\documentclass[
final,
   twoside,          
    openright, 
    titlepage,
    bibtotoc,
    BCOR13mm, 
    pointlessnumbers, 
DIV15, 
a4paper,12pt,parskip]{scrreprt}

\usepackage[T1]{fontenc} 

\usepackage{fourier}
\usepackage{cite}
\usepackage{amsmath}
\usepackage[thinspace,squaren]{SIunits}
\usepackage{amsfonts}
\usepackage{amssymb}
\usepackage[ngerman]{babel}
\usepackage{graphicx}
\usepackage[utf8]{inputenc}
\usepackage{fancyhdr}
\usepackage{float}
\usepackage{floatflt}
\usepackage{rotating}
\usepackage{dcolumn}

\usepackage{pstricks}

\usepackage{geometry}
\usepackage[off]{auto-pst-pdf}

\usepackage{psfrag}
\usepackage{textfit}
\usepackage{picins}
\usepackage{pdfpages}
\usepackage{pdflscape}

\usepackage[font=small,labelfont=bf]{caption}
\usepackage[font=small]{subfig}
\usepackage{xspace} 
\usepackage{lscape}

\usepackage{listliketab}
\usepackage{wrapfig}
\usepackage[section]{placeins}

\usepackage{tocvsec2}
\usepackage{ae}
\usepackage{array}
\usepackage{multirow}
\usepackage{booktabs}
\usepackage{paralist}
\usepackage{tabularx}
\usepackage{framed}
\usepackage{ltxtable}
\usepackage{longtable}
\usepackage{footnote}

\newcolumntype{v}[1]{>{\raggedleft\hspace{0pt}}p{#1}}
\newcolumntype{z}[1]{>{\centering\hspace{0pt}}p{#1}}
\newcolumntype{V}{>{\raggedleft\arraybackslash}X}
\newcolumntype{Q}{>{\raggedright\arraybackslash}X}
\newcolumntype{Z}{>{\centering\arraybackslash}X}
\newcolumntype{T}[1]{>{\raggedleft\hspace{0pt}}p{#1}}

\newcolumntype{R}[1]{>{\raggedleft\hspace{0pt}}p{#1}}
\newcolumntype{C}[1]{>{\centering\hspace{0pt}}p{#1}}
\newcolumntype{L}[1]{>{\raggedright\hspace{0pt}}p{#1}}

\renewcommand*\descriptionlabel[1]{\hspace\labelsep\normalfont #1}

\begin{document}
\setcounter{secnumdepth}{3} 
\setcounter{tocdepth}{2} 
\newcommand{\scanr}{Olympus-Scan$^{\mbox{\scriptsize{R}}}$\xspace}
\newcommand{\abb}{Abb.\xspace}
\newcommand{\kap}{Abschnitt~}
\newcommand{\tab}{Tabelle\xspace}
\newcommand{\formel}{Formel\xspace}
\newcommand{\matlab}{MATLAB$^\circledR$\xspace}
\newcommand{\blob}{\glqq Blob\grqq\xspace }
\newcommand{\blobs}{\glqq Blobs\grqq\xspace}
\newcommand{\lcfuenfzig}{LC$_{50}$"~Wert\xspace}
\newcommand{\hpf}{[hpf]}
\newcommand{\pmr}{\emph{PMR}\xspace}
\newcommand{\hts}{Hoch\-durch\-satz-Unter\-such\-ung\xspace}
\newcommand{\htsen}{Hoch\-durch\-satz-Unter\-such\-ungen\xspace}
\newcommand{\Nutzsig}{Nutz\-signal\xspace}
\newcommand{\Nutzsigs}{Nutz\-signals\xspace}
\newcommand{\Nutzsige}{Nutz\-signale\xspace}
\newcommand{\Nutzsigen}{Nutz\-signalen\xspace}
\newcommand{\NutzsigInf}{Nutz\-signal\-in\-for\-ma\-tion\xspace}
\newcommand{\NutzsigInfen}{Nutz\-signal\-in\-for\-ma\-tionen\xspace}
\newcommand{\NutzMat}{Nutz\-signal\-infor\-mations-Matrix\xspace}

\newcommand{\FischInfZeit}{Fisch-In\-forma\-tions\-zeit\-reihen\xspace}
\newcommand{\FischInfMerk}{Fisch-In\-forma\-tions\-merk\-mal\xspace}
\newcommand{\FischInfMerke}{Fisch-In\-forma\-tions\-merk\-male\xspace}
\newcommand{\FischInfMerken}{Fisch-In\-forma\-tions\-merk\-malen\xspace}
\newcommand{\Versuchspara}{Versuchsparameter\xspace}


\newcommand{\dhe}{d.h.\xspace}
\newcommand{\Dh}{D.h.\xspace}
\newcommand{\eV}{e.V.\xspace}
\newcommand{\zB}{z.B.\xspace}
\newcommand{\ZB}{Z.B.\xspace}
\newcommand{\oae}{o.ä.\xspace}
\newcommand{\uU}{u.U.\xspace}
\newcommand{\ua}{u.a.\xspace}
\newcommand{\ia}{i.A.\xspace}
\newcommand{\og}{o.g.\xspace}
\newcommand{\bzw}{bzw.\xspace}
\newcommand{\usw}{usw.\xspace}
\newcommand{\vgl}{vgl.\xspace}
\newcommand{\etc}{etc.\xspace}
\newcommand{\evtl}{evtl.\xspace}
\newcommand{\ca}{ca.\xspace}
\newcommand{\sog}{sog.\xspace}
\newcommand{\insbes}{insbes.\xspace}
\newcommand{\zT}{z.T.\xspace}
\newcommand{\ggf}{ggf.\xspace}
\newcommand{\inkl}{inkl.\xspace}

\sloppypar
\pagenumbering{Roman}

\begin{titlepage}
\pagestyle{empty}

\begin{center}
\centering


\vspace*{.3cm}


\LARGE \textbf{Konzept f\"{u}r Bildanalysen in Hochdurchsatz-Systemen am Beispiel des Zebrab\"{a}rblings}\\
\vspace{3cm}
\large{
Zur Erlangung des akademischen Grades \\[1ex]
\textbf{Doktor der Ingenieurwissenschaften}\\[1ex]
der Fakult\"{a}t f\"{u}r Maschinenbau\\
Karlsruher Institut f\"{u}r Technologie (KIT)}\\
\vspace{3cm}

\renewcommand{\baselinestretch}{1.2}


\normalsize
genehmigte \\
\textbf{Dissertation\\
}von\\[2ex]
Dipl.~Ing.~R\"{u}diger~Alshut\\
geboren am 14. Dezember 1980 in Mannheim

\end{center}

\vfill

\begin{tabbing}

Tag der m\"{u}ndlichen Pr\"{u}fung: \=\hspace{5cm} 12.Juli.2016\\[2ex]
Hauptreferent:  \> \hspace{5cm} Prof. Dr.-Ing. habil. G. Bretthauer\\
Korreferent: \> \hspace{5cm} Prof. Dr.-Ing. Ralf Mikut \\
Korreferent: \> \hspace{5cm} Prof. Dr. Uwe Str\"{a}hle\\

\hspace{6cm}

\kill


\end{tabbing}

\vfill


\newpage
\par\vspace*{\fill}
\includegraphics{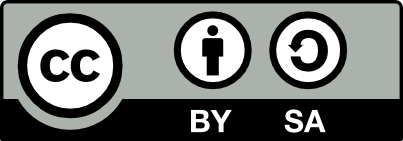}
\footnotesize
Dieses Werk ist lizenziert unter einer Creative Commons Namensnennung –\\
Weitergabe unter gleichen Bedingungen 3.0 Deutschland Lizenz\\
(CC BY-SA 3.0 DE): http://creativecommons.org/licenses/by-sa/3.0/de/

\cleardoublepage

\begin{titlepage}
\renewcommand{\baselinestretch}{1.5}
{\LARGE \textbf{Danksagung} \\ }

Die vorliegende Dissertation entstand in der Zeit von November 2008 bis Mai 2013 w\"{a}hrend meiner T\"{a}tigkeit am Institut f\"{u}r Angewandte Informatik des Karlsruher Instituts f\"{u}r Technologie. Im R\"{u}ckblick auf die Entstehungszeit dieser Arbeit habe ich reichlich Grund zur Dankbarkeit.

Herr Prof. Georg Bretthauer hat mir durch die Anstellung diese Arbeit erm\"{o}glicht und mich durch Diskussionen und wertvolle Hinweise beim Verfassen unterst\"{u}tzt. Frau Professor Jivka Ovtcharova hat das Korreferat \"{u}bernommen. Herr Prof. Uwe Str\"{a}hle erm\"{o}glichte durch seine Ideen und Zielsetzungen das Entstehen der Methoden und deren Anwendung f\"{u}r die Biologie. In besonderem Ma{\ss}e bin ich Herrn Prof. Ralf Mikut dankbar. In seiner Arbeitsgruppe und unter seiner Leitung wurden die hier vorgestellten Ergebnisse erarbeitet. Er war f\"{u}r mich sowohl ein gro{\ss}artiger Mentor als auch Ansprechpartner auf professioneller aber auch pers\"{o}nlicher Ebene. Keines unserer Treffen verlie{\ss} ich nicht ein wenig kl\"{u}ger und motivierter als zuvor.

Herrn Dr. Markus Reischl danke ich f\"{u}r die wertvolle fachliche Unterst\"{u}tzung und die unterhaltsamen Diskussionen, die einen erheblichen Teil zum Gelingen dieser Arbeit beigetragen haben. Sein Wissen und seine Fachkenntnis haben mich manches Mal verbl\"{u}fft und mir oft weitergeholfen. Auch seinen Humor und seinen sportlichen Ehrgeiz m\"{o}chte ich nicht missen. Herr Dr. Patrick Waibel hat mit mir, in jener Zeit, durch das gemeinsame B\"{u}ro mehr Zeit verbracht als jeder andere Mensch und ist seit den ersten Tagen am Institut sowohl in fachlicher als auch in pers\"{o}nlicher Hinsicht wichtig f\"{u}r mich gewesen. F\"{u}r die weiterhin anhaltende Freundschaft bin ich sehr dankbar. Herrn Dr. J\"{o}rg Matthes danke ich f\"{u}r die fachliche und freundschaftliche Unterst\"{u}tzung und die vielen Diskussionen w\"{a}hrend meiner Anstellung.

Auch die Kollegen und Studenten am Institut f\"{u}r angewandte Informatik waren stets eine gro{\ss}e Hilfe bei diversen Problemstellungen. Namentlich erw\"{a}hnen m\"{o}chte ich Frau Dr. Jasmin Lampert, Frau Daniela Sanchez, Herrn Prof. Lutz Gr\"{o}ll, Herrn Dr. Alexander Pfriem, Herrn Prof. Christian Pylatiuk.

F\"{u}r die finanzielle Unterst\"{u}tzung durch das Auslandsstipendium m\"{o}chte ich dem Karlsruhe House of Young Scientists (KHYS) danken, welches die Zusammenarbeit mit den Kollegen der Harvard Medical School erm\"{o}glicht hat. Die kurze aber intensive Kollaboration war au{\ss}erordentlich gewinnbringend. Von den Kollaborationspartnern gilt mein Dank besonders PhD Randall Peterson, in dessen Lab die Forschung statt fand, PhD Anjali Nath und PhD Xiang Li, die mich sofort kollegial in das Umfeld eingef\"{u}hrt haben und vor allem PhD David Kokel, der mich nicht nur in die Arbeiten integriert hat, sondern mir gleich vom ersten Tag an ein zu Hause in der Fremde geben hat.

Schlussendlich gilt ein gro{\ss}er Dank meiner Familie: Meiner Frau Christine f\"{u}r das flei{\ss}ige Korrekturlesen, meinen Eltern sowie meiner Schwester Marion f\"{u}r ihre Unterst\"{u}tzung in Worten und Taten.

%
%
\end{titlepage}

\cleardoublepage
\end{titlepage}

\tableofcontents
\clearpage
\pagenumbering{arabic}

\chapter{Einleitung} \label{chap:Einleitung}
\section{Bedeutung der Arbeit}
Bildbasierte Hochdurchsatz-Systeme erm\"{o}glichen das systematische massenweise Testen und Einordnen der Aktivit\"{a}t chemischer Substanzen in diversen Bereichen der Biologie, Toxikologie, Pharmakologie und Genetik. Machbarkeitsstudien zellbasierter Screens mit niedrigen St\"{u}ckzahlen entwickelten sich in den letzten Jahren schnell zu robusten und umfassenden Industriestandards mit teilweise immensen St\"{u}ckzahlen von Einzelversuchen sowie einer Gr\"{o}{\ss}enordnung bis in die Hunderttausende. Allerdings sind
Schlussfolgerungen (\zB Auswirkungen auf den Menschen und die Umwelt) aus den bisher fast ausschlie{\ss}lich zellbasierten Untersuchungen schwierig, Erkenntnisse m\"{u}ssen meist durch langwierige Tierversuche \"{u}berpr\"{u}ft und schlie{\ss}lich durch diese oft widerlegt werden \cite{Zon05}. Wird das Potenzial des Hochdurchsatzes durch die Ausweitung der Analyse auf einen vollst\"{a}ndigen Modellorganismus, wie \zB die Eier des Zebrab\"{a}rblings, ausgeweitet, so sind folgende \"{o}konomische sowie ethische Ziele erreichbar\cite{Baker11,Yang09,Zon10}:

\begin{itemize}
  \item eine h\"{o}here Effektivit\"{a}t durch geringere Fehlprognosen,
  \item die Reduktion von Tierversuchen zur Best\"{a}tigung von Ergebnissen,
  \item die Erh\"{o}hung des Durchsatzes bereits bestehender manueller Untersuchungen an Gesamtorganismen wie dem Zebrab\"{a}rbling sowie
  \item die Einsparung von Kosten, Personal und Zeit f\"{u}r die Versuchsdurchf\"{u}hrung.
\end{itemize}

Eine geeignete \hts erfordert eine auf die speziellen Anforderungen abgestimmte Versuchsauslegung, die schnelle und zuverl\"{a}ssige Erfassung und Analyse der Bildinhalte sowie eine klare zusammenfassende Darstellung der Ergebnisse. Herausfordernd ist, dass sich der auszuwertende Effekt von Versuch zu Versuch stark  unterscheidet und durch die hohe Versuchsanzahl die Qualit\"{a}t der aufgezeichneten Daten stark schwankt. Daher beschr\"{a}nken sich bisherige L\"{o}sungen auf die (Teil)Automatisierung von einfachen, informationsarmen oder die manuelle Auswertung von komplizierteren, informationsreichen Untersuchungen. In der vorliegenden Arbeit werden ein neues Konzept sowie neue Verfahren zur Gewinnung von Informationen aus  \htsen an Modellorganismen entwickelt und am Beispiel des Zebrab\"{a}rblings erprobt, was f\"{u}r eine schnelle automatisierte Erforschung biologischer Zusammenh\"{a}nge und der Analyse von Einfl\"{u}ssen innerhalb von Organismen sowie zur Entdeckung neuer Wirkstoffe eine Schl\"{u}sseltechnologie darstellt.

\section{Darstellung des Entwicklungsstandes}

    \subsection{Biologische Grundlagen}\label{subsec:Grundlagen_Bio}

        \subsubsection{\"{U}bersicht \hts} \label{subsubsec:\"{U}bersichtHochdurschsatz}
        Als \htsen werden Batch-Tests oder Siebtests bezeichnet, welchen in der Biologie gew\"{o}hnlich genetische, pharmakologische oder toxikologische Fragestellungen zu Grunde liegen und die eine gro{\ss}e Anzahl an Einzeluntersuchungen beinhalten. \htsen sind besonders in der Pharmakologie weit verbreitet und erm\"{o}glichen das schnelle Durchf\"{u}hren von Millionen von Tests. Mittels solcher Tests lassen sich beispielsweise rasch Antik\"{o}rper, (bio)aktive Substanzen oder Gene, die bestimmte biologische Wirkungspfade beeinflussen, identifizieren. Sie erm\"{o}glichen, in kurzer Zeit eine gro{\ss}e Anzahl von Proben in weiten Teilen automatisiert zu testen. In der Toxikologie wird aus den Ergebnissen der Tests zum einen das Gefahrenpotenzial von Substanzen bestimmt, zum anderen das Verst\"{a}ndnis biochemischer Prozesse verbessert. In der Pharmakologie dient die Identifikation bioaktiver Substanzen zudem als Ausgangspunkt zur Erforschung neuer Wirkstoffe, die potenziell als Basis neuartiger Medikamente in Frage kommen, oder wird zur Erforschung biochemischer Prozesse im Organismus herangezogen \cite{Bleicher03,Hueser06,Patton01}.

\begin{figure}[htb]
        \centering
\includegraphics[page=1]{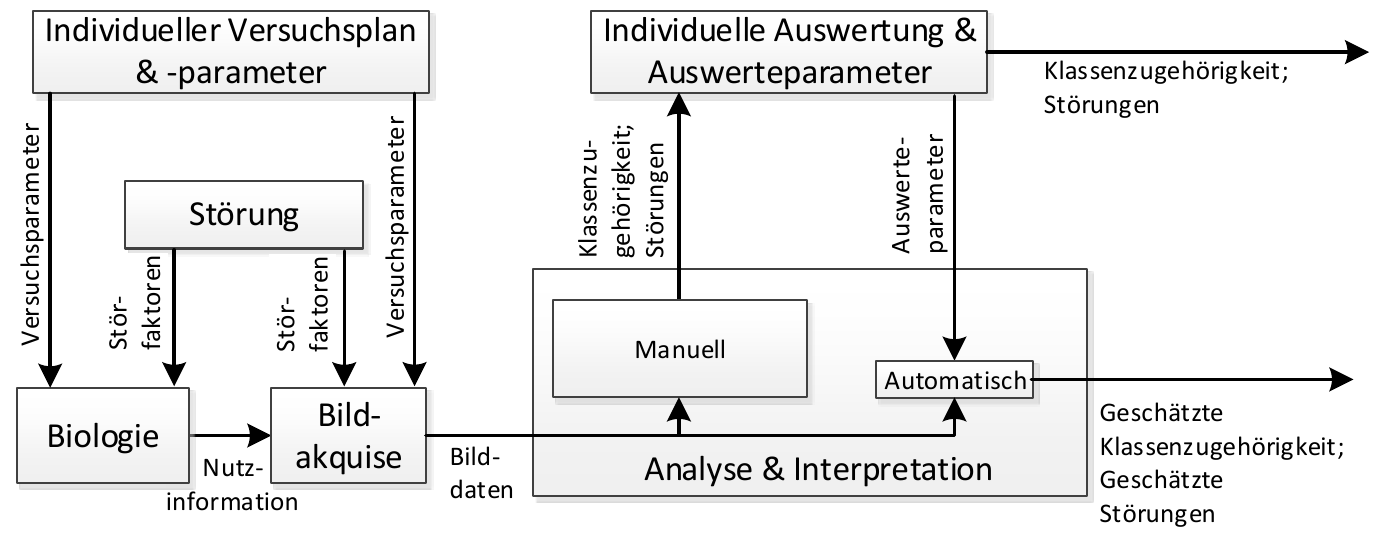}

\caption[Bisherige  Vorgehensweise bei der Versuchsauslegung bildbasierter \htsen.]{Bisherige  Vorgehensweise bei der Versuchsauslegung bildbasierter \htsen. Die Gr\"{o}{\ss}e der Bl\"{o}cke "`Manuell"' und "`Automatisch"' lehnt sich an die Verbreitung der Verfahren an (Analyse und Interpretation zurzeit weitgehend manuell). }
                 \label{fig:Neues_Konzept_Prinzip_1}
\end{figure}

Die \abb \ref{fig:Neues_Konzept_Prinzip_1} zeigt den derzeit \"{u}blichen Ablauf der Versuchsauslegung f\"{u}r \htsen. Auf die Parameter der Bl\"{o}cke "`Biologie"', "`Bildakquise"' und "`Ana\-lyse~\&~Inter\-pre\-ta\-tion"' wird hierbei Einfluss genommen. Es wird die Art der Bildakquise, die Auswertemethode, Versuchsparameter sowie die Anzahl der durchzuf\"{u}hrenden Einzeluntersuchungen manuell festgelegt. Die Auslegung der Untersuchungen wird in der Forschung gem\"{a}{\ss} der Erfahrung des Experimentators durchgef\"{u}hrt und basiert nicht auf systematischen Entscheidungskriterien.


        Unabh\"{a}ngig von Fragestellung, Ziel und Umfang der \hts wird ein biologisches Signal ben\"{o}tigt, welches sich bei der Datenerfassung beobachten und quantifizieren l\"{a}sst. Unter dem Begriff \emph{Biosignal} wird meist eine elektrische Spannung als Ergebnis biologischer Aktivit\"{a}ten verstanden \cite{Semmlow04} und als Nutzsignal in den folgenden Schritten weiter verarbeitet. Nutzsignale sind oft mittels Elektroden gewonnene (\"{A}nderungen von) Spannungen wie etwa die Elektrokardiografie (EKG), die Elektroenzephalografie (EEG), die Elektromyografie (EMG) oder das Elektroretinogramm (ERG). Generalisiert betrachtet umfasst der Begriff jedoch auch alle beobachtbaren und quantifizierbaren nicht-elektrischen Signale aus \zB der Mechanik, Akustik, Chemie oder Optik. F\"{u}r die in der vorliegenden Arbeit betrachteten bildbasierenden Verfahren sind vor allem optische Signale von Interesse, welche durch einen biologischen Effekt in der untersuchten Probe hervorgerufen werden. Die Effekte m\"{u}ssen mit geeigneten Mitteln reproduzierbar erzeugt und sichtbar gemacht werden, beispielsweise durch geeignete Pr\"{a}paration oder Stimulation. Im Anschluss werden die Signale quantifiziert (siehe hierzu Abschnitt \ref{subsec:Bildakquise+Bildverarbeitung}), um R\"{u}ckschl\"{u}sse ziehen zu k\"{o}nnen.

        Bei \htsen hat sich die Anwendung \sog Mikrotiterplatten etabliert \cite{Maruyama88}. Hierbei handelt es sich um kosteng\"{u}nstige Kunststoffplatten, in welche gitterartig angeordnete Vertiefungen (\sog \emph{Wells}) eingebracht sind. Typische Platten besitzen 96, oder Vielfache von 96 (\zB 384, 1536, 3456) solcher Vertiefungen und sind kommer\-ziell erh\"{a}ltlich. Je nach durchzuf\"{u}hrendem Versuch gibt es unterschiedliche Ausf\"{u}hrungen bzgl. der Farbe, beispielsweise mit Seitenw\"{a}nden aus schwarzem Plastik, oder bzgl. der Form, \zB mit runden oder flachen B\"{o}den der N\"{a}pfchen. Die Mikroskopie ist bei der Mehrheit der Untersuchungen invers, dabei wird das Bild "`von unten"' in Durchlicht-Aufnahmetechnik akquiriert. Die N\"{a}pfchen werden in den meisten F\"{a}llen mit einer Tr\"{a}gerfl\"{u}ssigkeit, beispielsweise einer w\"{a}ssrigen L\"{o}sung von \emph{Dimethylsulf\-oxid} (DMSO) und, je nach Versuch, einer oder mehreren zu untersuchenden Chemikalien gef\"{u}llt. W\"{a}hrend die Tr\"{a}gerfl\"{u}ssigkeit in allen Versuchen gleich bleibt, wird die gel\"{o}ste Chemikalie variiert. \"{U}blicherweise werden zu Kontrollzwecken einige N\"{a}pfchen pro Platte mit lediglich der Tr\"{a}gerfl\"{u}ssigkeit (\sog \emph{Negativkontrollen}) gef\"{u}llt und alle Versuche in mehrfacher Ausf\"{u}hrung durchgef\"{u}hrt (\sog \emph{Replika}). Zus\"{a}tzlich werden N\"{a}pfchen mit einem Reaktanten exponiert, dessen Wirkung auf die Organismen bekannt ist (\sog \emph{Positivkontrollen}). Die genannten Ma{\ss}nahmen dienen bei der Auswertung zur \"{U}berpr\"{u}fung, ob w\"{a}hrend des Versuchs Ungereimtheiten auftreten und das Testverfahren auch tats\"{a}chlich funktioniert. Bei Auff\"{a}lligkeiten muss die entsprechende Platte oder sogar die gesamte Versuchsreihe ausgeschlossen oder wiederholt werden \cite{Hueser06}. Die genauen Anweisungen, welche zur Versuchsdurchf\"{u}hrung notwendig sind, werden als "`Versuchsprotokoll"' bezeichnet.

\begin{figure}[htbp]
        \centering
        \includegraphics[width=.85\linewidth]{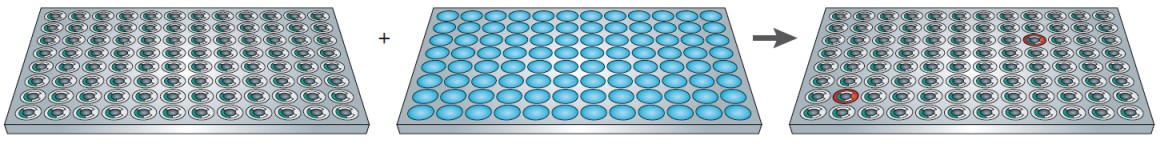}

        \caption[Einsatz von Mikrotiterplatten im Hochdurchsatz]{Mikrotiterplatten organisieren im Hochdurchsatz sowohl die Einzeluntersuchungen, als auch die chemischen Bibliotheken. Im obigen Beispiel aus \cite{Lieschke07} werden auf die in einer 96 Well-Plate vorbereiteten Larven (links) Substanzen aus einer Bibliothek (Mitte) geben. Im rechten Bild ist das exemplarische Ergebnis dargestellt. Hier wurden zwei Wells markiert, welche sich von den anderen unterscheiden, \zB durch einen abweichenden Entwicklungsprozess der Larven.}
        \label{fig:Mikrotiterplate_Chemische_Bibliothek}
        \end{figure}

        Mikrotiterplatten werden nicht nur bei der Versuchsdurchf\"{u}hrung angewandt, sondern auch zur Lagerung und Sortierung von chemischen Komponenten verwendet. In einem solchen Fall lagert in jedem N\"{a}pfchen eine Substanz, von der eine detaillierte Dokumentation existiert. Derart bef\"{u}llte Platten sind als Bibliotheken auch kommerziell erh\"{a}ltlich und werden als "`Stock-plates"' (engl.) bezeichnet (vgl. \abb \ref{fig:Mikrotiterplate_Chemische_Bibliothek}). Die Stock-Plates werden oft gefroren gelagert und nicht direkt in der \hts verwendet, sondern eine definierte Menge der zu untersuchenden Substanzen wird entnommen und der \hts zugef\"{u}hrt. Meist sind Wirkstoff-Familien auf einer solchen Stock-Plate zusammengefasst, und daher wird die gesamte Stock-Plate auf diese Weise dupliziert und untersucht. In einem typischen Versuch wird in das Gemisch aus Chemikalie und Tr\"{a}gerfl\"{u}ssigkeit eine biologische Einheit gegeben.
        Bei einer biologischen Einheit handelt es sich \zB um eine Zelle, Zellkultur oder einen biologischen Modellorganismus wie etwa einen tierischen Embryo. Nach einer Inkubationszeit, in welcher der Wirkstoff mit der biologischen Einheit interagieren kann, werden Messungen mittels diverser Detektoren (vgl. \kap \ref{sec:Beschreibung_Parameter}) automatisiert oder manuell von den Einheiten durchgef\"{u}hrt \cite{Schreiber07}. In bildbasierenden \htsen erfolgt die Generierung der Messwerte meist durch visuelle Auswertung der Versuche \cite{Carpenter07}. Die Messwerte werden auf Auff\"{a}lligkeiten, \sog \emph{hits} (engl. f\"{u}r Treffer), untersucht. Die M\"{o}glichkeit, einzelne Zellen \zB durch fluoreszierende Proteine zu markieren, hat die Robustheit und den Durchsatz der Versuche, zumindest bei zellbasierten Versuchen, in den vergangenen Jahren stark gesteigert \cite{Wittmann12}. Diese einfach gehaltenen Untersuchungen, bei denen lediglich die Anwesenheit oder Menge des fluoreszierenden Proteins ausgewertet wird, sind selbst ohne fortgeschrittene Bildverarbeitung und Segmentierungstechnik m\"{o}glich. Kombiniert mit Automatisierungstechnik lassen sich bei derartigen Untersuchungen sehr hohe Zahlen an Einzelversuchen pro Tag erreichen. Der Begriff \emph{Hoch}durchsatz ist indes nicht genau definiert, wird aber weithin bei einem Durchsatz zwischen 10\,000 und 100\,000 Proben pro Tag verwendet \cite{Woelcke01}.

        Manuell durchgef\"{u}hrte Experimente sind meist der Ausgangspunkt, aus dem ein Hochdurchsatz-Experiment abgeleitet wird. Die Versuchsprotokolle der manuellen Untersuchungen im Labor m\"{u}ssen angepasst werden, damit sie sich f\"{u}r den Hochdurchsatz eignen \cite{Inglese07}. Wichtige Bedingungen sind:
        \begin{itemize}
          \item Gute Reproduzierbarkeit und Stabilit\"{a}t des \Nutzsigs zwischen einzelnen N\"{a}pfchen und Platten. Dies gilt sowohl bez\"{u}glich des Reaktanten, der biologischen Einheit sowie der verwendeten Ger\"{a}te (z.B. bei Verwendung mehrerer Mikroskope etc.)
          \item Hohe Zuverl\"{a}ssigkeit der verwendeten Positiv"~ und Negativkontrollen
           \item Ausreichende Sensibilit\"{a}t des \Nutzsigs zur Identifikation von Substanzen mit schwachem Effekt auf die biologische Einheit
          \item Wirtschaftlichkeit der \hts (standardm\"{a}{\ss}ig gemessen in Kosten pro N\"{a}pfchen).
        \end{itemize}
        Aus den Abweichungen und Schwankungen zwischen den Platten und Kontrollen lassen sich Kontrollparameter berechnen, welche Auskunft \"{u}ber die Robustheit des Versuchs geben. Zur Validierung, inwieweit sich ein Protokoll f\"{u}r die \hts eignet, haben sich verschiedene Parameter etabliert (vgl. \tab \ref{tab:Validierungsparameter}).

        \begin{table} [htbp]
        \centering
        \footnotesize
        \renewcommand{\arraystretch}{2}
        \begin{tabularx}{\linewidth}{>{\raggedright\arraybackslash}l
         >{\raggedright\arraybackslash}l X}
        \toprule
        \textbf{Parameter }&\textbf{Formel} &\textbf{Beschreibung}\tabularnewline 
        \midrule
        Variationskoeffizient &$CV=\frac{\sigma}{\mu}\cdot 100$ & Parameter zur Quantifizierung der Messgenauigkeit relativ zum Mittelwert aller Messungen eines Versuchs in Prozent. Akzeptable Werte sind <15\%. \\

        Signal-Rausch-Verh\"{a}ltnis &$SNR=\frac{\mu_{\text{max}}-\mu_{\text{min}}}{\sigma_{\text{min}}}$ &  Parameter zur Quantifizierung der St\"{a}rke des \Nutzsigs\\

        Signal-Hintergrund-Verh\"{a}ltnis& $SHV= \frac{\mu_{\text{max}}}{\mu_{\text{min}}}$& Parameter, welcher in der Praxis mit Hilfe der Kontrollen ermittelt wird. Erwartungswert der Messwerte eines Versuchs (\Nutzsig) im Verh\"{a}ltnis zum Erwartungswert der Messwerte der Kontrollen (Hintergrund). Akzeptable Werte sind >2.  \\

        Signalfenster & $SF=\frac{\mu_{\text{max}}-\mu_{\text{min}}-3(\sigma_{\text{max}}+\sigma_{\text{min}})}{\sigma_{\text{max}}}$ & Parameter zur Quantifizierung der Signifikanz zwischen dem Maximal- und Minimalwert der Kontrollen. Akzeptable Werte sind >2. \\

        $Z'$-Faktor &$Z'=1-\frac{3(\sigma_{\text{max}}+\sigma_{\text{min}})}{|\mu_{\text{max}}-\mu_{\text{min}}|}$ & Alternative Repr\"{a}sentation des Signalfensters. Der Wert wird meist sowohl f\"{u}r die Kontrollen als auch f\"{u}r den Versuch selbst ermittelt. Akzeptable Werte sind >0.5. \\

        Minimum-Signifikanz-Verh\"{a}ltnis & $MSR=10^{\,2\,\sigma_d}$& Parameter zur Quantifizierung des minimalen Verh\"{a}ltnisses zwischen zwei Messungen, welches statistisch relevant ist (95\% Konfidenz).\\

        \bottomrule

        \end{tabularx}
        \caption[\"{U}bersicht ausgew\"{a}hlter Validierungsparameter f\"{u}r Hochdurchsatz-Untersuchungen~\cite{Inglese07}]{\"{U}bersicht ausgew\"{a}hlter Validierungsparameter f\"{u}r Hochdurchsatz-Untersuchungen~\cite{Inglese07} mit $\sigma$ Standardabweichung der Messwerte eines Versuchs; $\sigma_{\mathrm{max}/\mathrm{min}}$ Versuch mit gr\"{o}{\ss}ter/kleinster Standardabweichung; $\sigma_d$ Standardabweichung der ermittelten Wirksamkeiten in logarithmischer Skala (z.B. $LC_{50}$-Werte, vgl. Abschnitt \ref{subsec:Pr\"{a}s_EC50}); $\mu$ Erwartungswert der Messwerte eines Versuchs; $\mu_{\mathrm{max}/\mathrm{min}}$ Versuch einer Versuchsreihe mit gr\"{o}{\ss}tem/kleinstem Erwartungswert.}
        \label{tab:Validierungsparameter}
        \end{table}
        Das Signalfenster und der $Z$-Faktor dienen zur Absch\"{a}tzung zwischen dem Minimum und Maximum der Messungen und der Pr\"{a}zision der Werte innerhalb einer Platte und platten\"{u}bergreifend. Das Signalfenster bestimmt einen Parameter, welcher die Signifikanz zwischen dem maximalen und minimalen Messwert einer Versuchsreihe quantifiziert. Dieser ist jedoch nicht in gleichem Ma{\ss}e verl\"{a}sslich wie der $Z$-Faktor \cite{Iversen06}. Der $Z$-Faktor erm\"{o}glicht durch die einheitenlose Skala (von 0 bis 1) den Vergleich verschiedener Versuche und Versuchsdurchl\"{a}ufe miteinander. Hierbei werden auch Kontroll-N\"{a}pfchen ($Z'$) und Versuch-N\"{a}pfchen ($Z''$) miteinander verglichen. Niedrige Werte innerhalb der Kontrollen deuten beispielsweise auf eine Kontaminierung der N\"{a}pfchen oder ein anderes Problem in der Versuchsdurchf\"{u}hrung hin.
        Die Sensitivit\"{a}t der Untersuchung ist ein Messwert zur Bestimmung der Genauigkeit und Reproduzierbarkeit, welche sich mittels des Minimum-Signifikanz-Verh\"{a}ltnisses ($MSR$) quantifizieren l\"{a}sst\cite{AssayGuidanceManualCommitteeMaryland05,Eastwood06}.

        \subsubsection{Zebrab\"{a}rbling als Modellorganismus f\"{u}r den Hochdurchsatz}\label{subsubsec:Zebrab\"{a}rbling_Modellorg}
        Als biologische Einheiten kommen im Hochdurchsatz meist Modellorganismen zum Einsatz. Modellorganismen sind \zB ausgew\"{a}hlte Pflanzen, Tiere, Bakterien oder Pilze, die mit einfachen Methoden gez\"{u}chtet \bzw untersucht werden k\"{o}nnen und von besonderer Bedeutung f\"{u}r die biologische/biomedizinische Forschung sind.
        Sie zeichnen sich in der Regel durch eine kosteng\"{u}nstige sowie unkomplizierte Haltung \bzw Nachzucht aus und sind in vielf\"{a}ltiger Hinsicht sehr gut dokumentiert. Dar\"{u}ber hinaus geh\"{o}ren einige dieser Arten zu den ersten Spezies, deren komplettes Genom entschl\"{u}sselt werden konnte. Die Wahl des Modellorganismus h\"{a}ngt in der Regel von der biologischen Fragestellung ab. An Einzellern lassen sich beispielsweise gut zellbiologische Prozesse untersuchen. Mehrzellige Lebewesen hingegen werden vor allem f\"{u}r entwicklungsbiologische oder toxikologische Untersuchungen ben\"{o}tigt. Umfassend behandelt werden Modellorganismen in \cite{Mueller06c}.

        Bei der Forschung an Modellorganismen wird versucht, allgemein g\"{u}ltige und auf andere Organismen, insbesondere dem des Menschen, \"{u}bertragbare Erkenntnisse zu erhalten. Ein Grundproblem der etablierten Modellorganismen in der Toxikologie und Entwicklungsbiologie, wie z.B. der Fruchtfliege (\emph{Drosophila melanogster}), dem Fadenwurm (\emph{Caenorhabditis elegans}) oder dem
        Krallenfrosch (\emph{Xenopus laevis}), ist die relativ gro{\ss}e entwicklungsgeschichtliche Distanz zum Menschen \cite{Ingham97}. S\"{a}ugetiere, wie z.B. die Maus oder die Ratte, besitzen jedoch im Vergleich zu Insekten und Amphibien eine niedrigere Anzahl an Nachkommen und eine ungleich l\"{a}ngere Generationszeit. Zudem entwickeln sich die Nachkommen innerhalb des  Mutterleibes, wodurch toxikologische Untersuchungen bzw. Manipulationen am Embryo technisch aufw\"{a}ndig sind und sich nur schwer observieren lassen. S\"{a}ugetierorganismen sind dem  menschlichen Organismus am \"{a}hnlichsten, so dass an ihnen erzielte Forschungsergebnisse oft aussagekr\"{a}ftige Informationen \"{u}ber Verh\"{a}ltnisse beim Menschen liefern. Jedoch ist eine \hts aufgrund der oben geschilderten Nachteile und aufgrund der f\"{u}r die Analyse ben\"{o}tigten hohen Anzahl an Messdaten und damit an Individuen in der Praxis h\"{a}ufig nicht wirtschaftlich, technisch schwierig durchf\"{u}hrbar und ethisch bedenklich.

        Der Zebrab\"{a}rbling ist ein auf dem indischen Subkontinent beheimateter, ausgewachsen ca. 3 bis 5\,cm gro{\ss}er Karpfenfisch (\emph{Cyprinidae}) und hat in den wissenschaftlichen Laboratorien der Entwicklungsbiologen eine erstaunliche Karriere gemacht \cite{Eaton74}. Auch in der Industrie gewinnt der Zebrab\"{a}rbling an Bedeutung. Eine st\"{a}ndig wachsende Zahl von Laboren versucht, anhand des Fisches biologische Vorg\"{a}nge zu verstehen, die Ursachen menschlicher Erkrankungen aufzukl\"{a}ren und neue Medikamente zu entwickeln.

        Der Zebrab\"{a}rbling verbindet die Vorteile einer kurzen Generationszeit mit einer hohen Anzahl von Nachkommen und technisch leicht zug\"{a}nglichen Larven, die sich au{\ss}erhalb des Mutterleibes entwickeln. Als Vertreter der Wirbeltiere (\emph{Vertebraten}) besitzt er alle Organsysteme, die auch im Menschen vorkommen wie z.B. Auge, Hirn, Herz, Dr\"{u}sen etc. Sein Immunsystem l\"{a}sst sich im Gegensatz zu dem von Insekten gut mit dem des Menschen vergleichen. Des Weiteren wird davon ausgegangen, dass die \"{u}berwiegende Zahl aller  menschlichen Gene im Fisch nicht nur vorkommt, sondern auch sehr \"{a}hnliche oder sogar die gleichen Funktionen besitzen. Zudem sind die Zellen sowie die \"{a}u{\ss}ere Fruchth\"{u}lle um die Larven (das \sog Chorion)  transparent, wodurch eine einfache Beobachtung mit dem Lichtmikroskop m\"{o}glich ist. Die hohe Transparenz der Zellen erm\"{o}glicht eine genaue Beobachtung der sich bildenden Organe oder der Herzfrequenz und der Blutstr\"{o}mung am lebenden Embryo. W\"{a}hrend von den Zebrab\"{a}rblingslarven scharfe Detailaufnahmen die Identifizierung einzelner Zellen erm\"{o}glichen, stehen von anderen, lebenden, vertebralen Embryonen oder auch von lebenden menschlichen Embryonen nur weitaus schlechtere Bilder, wie z.B. unscharfe und kontrastarme Ultraschallaufnahmen, zur Verf\"{u}gung. Laut der EU  Direktive  \emph{2010/63/EU} fallen Versuche an fr\"{u}hen Tierembryonen  nicht unter das Tierschutzgesetz und m\"{u}ssen somit nicht als Tierversuche  angemeldet werden, was den administrativen Aufwand f\"{u}r Versuche mit hohen St\"{u}ckzahlen verringert. Der Zeitpunkt, ab dem die Embryonen unter das Tierschutzgesetz fallen, ist der Zeitpunkt, ab dem diese eigenst\"{a}ndig auf Futtersuche gehen. In \cite{Straehle11} wird der Zeitpunkt von 120 Stunden nach der Befruchtung [hpf] empfohlen. Weiterf\"{u}hrende Literatur zum Zebrab\"{a}rbling als Modellorganismus in der Forschung findet sich \zB in \cite{Langheinrich02,Metscher99,Sprague06,Traver03}.

        Die Zebrab\"{a}rblingslarve hat f\"{u}r \htsen gro{\ss}es Potenzial. Bereits 1996 wird von \cite{Eisen96} auf zahlreiche Literatur zum Potenzial des Zebrab\"{a}rblings hingewiesen, die Vorteile des Fisches in vielen Forschungsgebieten ausf\"{u}hrlich diskutiert und dieser als der Nachfolger der Fruchtfliege bezeichnet. An der Fruchtfliege als Modellorganismus gelangen bahnbrechende Erfolge bei der Identifizierung essentieller Gene in der Embryonalentwicklung. Diese Arbeit von Christiane N\"{u}sslein-Volhard wurde 1995 mit dem Nobelpreis f\"{u}r Medizin und Physiologie ausgezeichnet \cite{Nuesslein-Volhard80}. So f\"{o}rderte die Europ\"{a}ische Kommission seit Anfang 2004 bis Ende 2009 ein Gemeinschaftsprojekt von 15 europ\"{a}ischen Forschungseinrichtungen zur Forschung am Zebrab\"{a}rbling mit dem Namen \emph{ZF-MODELS - Zebrafish Models for Human Development and Disease} unter Leitung des T\"{u}binger Max-Planck-Instituts f\"{u}r Entwicklungsbiologie mit 12\,Millionen Euro. Ziel des Projektes war es, am Zebrab\"{a}rbling die Entwicklung von Wirbeltieren sowie menschliche Erbkrankheiten zu erforschen. W\"{a}hrend des Projektes wurden die Abl\"{a}ufe und Zusammenh\"{a}nge im Zebrab\"{a}rbling als Vertreter der Wirbeltiere untersucht. Die Ergebnisse waren derart vielversprechend, dass seit Juli 2010 das Folgeprojekt \emph{Zebrafish Regulomics for Human Health (ZF-HEALTH)} f\"{u}r einen Zeitraum von f\"{u}nfeinhalb Jahren mit 11,4\,Millionen Euro gef\"{o}rdert wird. Das ausgesprochene  Ziel des neuen Programmes ist \ua die Durchf\"{u}hrung von Hochdurchsatz-Ph\"{a}notypisierung mittels Verhaltens-Untersuchungen am Zebrab\"{a}rbling in 3D- und 4D-Bildaufnahmen.

\begin{figure}[h!tbp]

        \centering
        \begin{minipage} [b]{0.45\linewidth}
        \captionsetup{format=hang}
        \subfloat[\unit{1}{\hour}]{\includegraphics[width=0.3\textwidth]{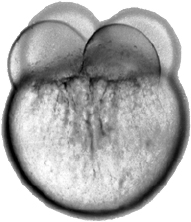}
        \label{fig:02a_Entwicklung_Zebrabaerbling_1h}} \hspace{.1cm}
         \subfloat[\unit{4}{\hour}]{\includegraphics[width=0.3\textwidth]{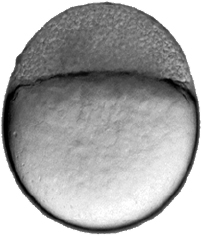} \label{fig:02a_Entwicklung_Zebrabaerbling_4h}} \hspace{.1cm}
         \subfloat[\unit{6}{\hour}]{\includegraphics[width=0.3\textwidth]{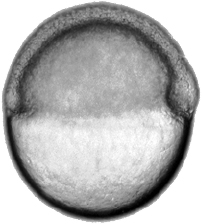}
        \label{fig:02a_Entwicklung_Zebrabaerbling_6h}}\\
        \vspace{.25cm}\\
        \subfloat[\unit{10}{\hour}]{\includegraphics[width=0.3\textwidth]{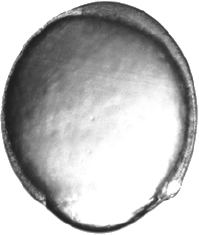} } \hspace{.1cm} \subfloat[\unit{12}{\hour}]{\includegraphics[width=0.3\textwidth]{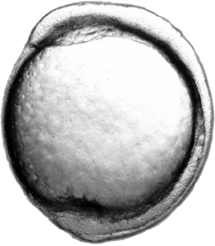} } \hspace{.1cm}
        \subfloat[\unit{18}{\hour}]{\includegraphics[width=0.3\textwidth]{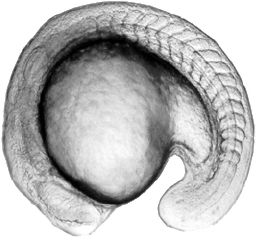}
        \label{fig:02a_Entwicklung_Zebrabaerbling_18h}}
        \vspace{.25cm}
        \end{minipage}
        \hfill
        \begin{minipage} [b]{0.45\linewidth}

        \subfloat[\unit{30}{\hour}]{\includegraphics[width=.98\textwidth]{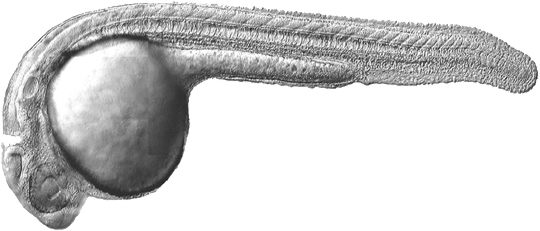}
        }\\
        \subfloat[\unit{42}{\hour}]{\includegraphics[width=.98\textwidth]{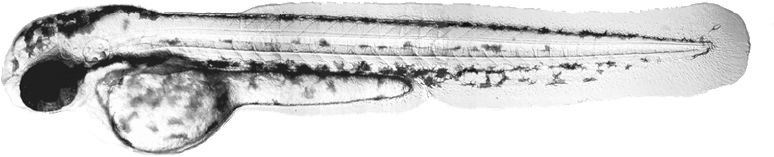}
        }\\
        \subfloat[\unit{48}{\hour}]{\includegraphics[width=.98\textwidth]{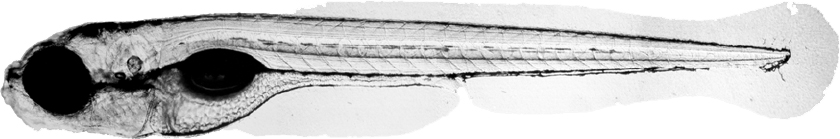}
        }
        \end{minipage}
        \caption[Ausgew\"{a}hlte Entwicklungsstadien der Zebrab\"{a}rblingslarven. ]{Ausgew\"{a}hlte Entwicklungsstadien der
        Zebrab\"{a}rblingslarven. Unter den Bildern ist jeweils das ungef\"{a}hre Alter in Stunden nach der Befruchtung angegeben. \cite{Westerfield93} }\label{fig:02a_Entwicklung_Zebrabaerbling}

        \end{figure}

        Verglichen mit anderen Wirbeltieren verl\"{a}uft die Entwicklung der Larve des Zebrab\"{a}rblings sehr schnell und fast jeder Entwicklungszeitpunkt l\"{a}sst sich verfolgen und in Bildern aufzeichnen. Die Eizelle, die sich oberhalb der Dotterkugel befindet, beginnt sich bereits Minuten nach dem Ablaichen und der Befruchtung zu teilen (\abb \ref{fig:02a_Entwicklung_Zebrabaerbling_1h}). Der beschriebene Vorgang wurde bereits mittels Bildverarbeitung gepr\"{u}ft und verfolgt von \cite{Keller08SC,Mikula11,Mikula08}. Diese Art von Auswertungen sind jedoch bisher aufgrund der gro{\ss}en Datenmengen nicht f\"{u}r den Hochdurchsatz geeignet.

        Aus den ersten Zellen bildet sich innerhalb von vier Stunden durch Zellteilungen eine Gruppe aus einigen tausend Zellen (\abb \ref{fig:02a_Entwicklung_Zebrabaerbling_4h}), welche in den folgenden sechs Stunden den Dotter umwachsen. Zehn Stunden nach der Befruchtung zeichnet sich allm\"{a}hlich der Kopf der Larve ab (\abb \ref{fig:02a_Entwicklung_Zebrabaerbling_6h}"~e) und nach 18 Stunden sind am Kopf die ovalen Vorl\"{a}ufer der Augen sowie im Rumpf die ersten Muskeln erkennbar. Kurze Zeit sp\"{a}ter beginnt der Schwanz sich zu strecken, die meisten Organe sind angelegt, das Herz f\"{a}ngt an zu schlagen und die nur knapp 24 Stunden alte Larve bewegt sich bereits im Chorion. W\"{a}hrend des dritten Tages nach der Befruchtung ist die Larve fertig entwickelt. Sie  schl\"{u}pft und schwimmt die ersten kurzen Strecken (\abb \ref{fig:02a_Entwicklung_Zebrabaerbling_18h}"~i). Nach f\"{u}nf Tagen schlie{\ss}lich  hat sie ihren Dottervorrat aufgebraucht und sucht selbstst\"{a}ndig nach Futter. Trotz der rasanten Entwicklung dauert es dann immerhin noch drei Monate, bis ein Zebrab\"{a}rbling geschlechtsreif wird. Eine genaue Beschreibung des  Entwicklungsprozesses findet sich in~\cite{Kimmel95}.
        Die beschriebenen Merkmale des Zebrab\"{a}rblings lassen sich lichtmikroskopisch leicht observieren. Alle Aufnahmen in \abb \ref{fig:02a_Entwicklung_Zebrabaerbling} wurden mit einem herk\"{o}mmlichen Lichtmikroskop aufgezeichnet. Es existieren bereits automatisierte Systeme zum Erkennen diverser Ph\"{a}notypen oder Endpunkte \cite{Spomer12,Gehrig09,Vogt09}.

        Zum Verst\"{a}ndnis eines weiteren Vorteils des Zebrab\"{a}rblings, dem Einsatz \sog Transgene, wird im Folgenden kurz auf das Gr\"{u}n Fluoreszierende Protein eingegangen. 1961 beschrieb und extrahierte erstmals Osamu Shimomura das Gr\"{u}n Fluoreszierende Protein (GFP) einer Tiefseequalle \cite{Shimomura05,Shimomura62}, welches bei Anregung mit blauem oder ultraviolettem Licht gr\"{u}n fluoresziert. Von unsch\"{a}tzbarem Wert ist das GFP, da es sich als Marker f\"{u}r die Genexpression nutzen l\"{a}sst. Eine mit Hilfe des Proteins markierte Zelle erf\"{u}llt die gleiche Funktion wie eine nicht markierte Zelle, mit dem Unterschied, dass die markierten Zellen nach Anregung mit einer passenden Wellenl\"{a}nge fluoreszieren \dhe, leuchten. Ein Organismus, in den ein solcher Marker eingebracht wurde, wird als \emph{Transgen} bezeichnet. Auch im Zebrab\"{a}rbling lassen sich auf die beschriebene Art Transgene generieren. Unmittelbar mit der Entstehung einer derart markierten Zelle im Zebrab\"{a}rbling lassen sich r\"{a}umliche und zeitliche Verteilung und Bewegungen der Zelle studieren. F\"{u}r die Entdeckung und Extraktion des GFP wurde im Jahr 2008 der Nobelpreis f\"{u}r Chemie an Osamu Shimomura, Martin Chalfie und Roger Tsien verliehen. Es existieren neben dem GFP bereits Varianten in anderen Farben (\zB CFP (cyan) oder YFP (yellow)), so dass sich auch mehrere Zelltypen des Zebrab\"{a}rblings gleichzeitig und unabh\"{a}ngig beobachten lassen\cite{Prasher95,Veith05}. Die M\"{o}glichkeit, die genannte Technik des Markierens von Zellen im Zebrab\"{a}rbling anzuwenden \cite{Lin00}, er\"{o}ffnet f\"{u}r den Modellorganismus alle M\"{o}glichkeiten, die die moderne Fluoreszenzmikroskopie bietet~ \cite{Spector06}. Hervorzuheben sind hier Aufnahmetechniken, die es erm\"{o}glichen, Zellbewegungen \"{u}ber die Zeit und/oder in 3D zu akquirieren.
        Die erzielbare Aufl\"{o}sung reicht von makroskopischen Bildern einfachster Aufbauten \cite{Blackburn11} \"{u}ber die bereits als Standard geltende Konfokal-Mikroskopie \cite{Pawley08}, die spezialisierte Hoch\-durch\-satz-Mikros\-kopie \cite{Liebel03} bis hin zur Nanoskopie, welche sich der Fluoreszenz bedient, um die physikalische Beugungsgrenze zu \"{u}berwinden \cite{Aquino11,Hell94}. Die einfachste Variante ist das Akquirieren des Fluoreszenzkanals mittels eines herk\"{o}mmlichen Lichtmikroskops, welches mit einem Fluoreszenzfilter ausgestattet ist. Beim Standard f\"{u}r die 3D-Bildakquise, der Konfokal-Mikroskopie, wird nur ein sehr kleiner Teil des Pr\"{a}parats zu jedem Zeitpunkt zur Fluoreszenz angeregt. Dies erm\"{o}glicht das Abrastern der Proben und so das Rekonstruieren von 3D-Modellen im Rechner. Das Abrastern ben\"{o}tigt jedoch Zeit und somit ist das Akquirieren von zeitlichen Ver\"{a}nderungen schwierig  \bzw die zeitliche Aufl\"{o}sung oftmals zu begrenzt. Eine weitere wichtige Mikroskopie-Technik, die sich der Fluoreszenz bedient und beim Zebrab\"{a}rbling Anwendung findet, ist die Single Plane Illumination Microscopy (SPIM)\cite{Huisken04}. Hierbei ist das Prinzip \"{a}hnlich wie bei der Konfokal-Mikroskopie. Allerdings wird nicht lediglich ein Punkt zur Fluoreszenz angeregt, sondern ein scharf definiertes Lichtvolumen. Durch das hohe Aspektverh\"{a}ltnis des Volumens wird auch von einem Lichtblatt gesprochen. Diese Variante erm\"{o}glicht eine wesentlich h\"{o}here Akquisegeschwindigkeit bei einem guten Signal-Rausch-Verh\"{a}ltnis.



        \subsubsection{Relevante Forschungsgebiete f\"{u}r eine bildbasierte \hts im Zebrab\"{a}rbling}\label{subsubsec:Relevante Forschungsgebiete}

         Der Vorteil, den eine \hts im gesamten Organismus gegen\"{u}ber der Zelle hat, ist, dass Verkn\"{u}pfungen und Gene in den Organismen oft redundant vorkommen oder erst durch das Zusammenspiel aller biochemischen Prozesse in der Wirkkette der Einfluss von Chemikalien aussagekr\"{a}ftig erforscht werden kann. Der Zebrab\"{a}rbling vereint die beschriebenen M\"{o}glichkeiten der Bildakquise mit den oben dargelegten Vorteilen in Haltung, gentechnischer Relevanz, kurzen Generationszyklen und klaren Mikroskopieaufnahmen. Somit erscheint das Tier als optimal geeignet f\"{u}r neue Hochdurchsatz-Untersuchungen mit bisher nicht erreichter Aussagekraft \cite{Baker11,Lieschke07,Love04,Starkuviene07,Yang09,Zon10}. Des Weiteren existieren bereits Firmen, welche spezielle Hoch\-durchsatz-Unter\-suchungen auf Auftragsbasis zur Entwicklung von Medikamenten am Zebrab\"{a}rbling durchf\"{u}hren. Eine Liste von kommerziellen Anbietern von \htsen findet sich in \cite{Blow2009}. Bez\"{u}glich der Einsatzgebiete f\"{u}r \htsen lassen sich drei weitl\"{a}ufige und biologisch relevante Bereiche unterscheiden:
        \begin{description}
          \item [\textbf{Entwicklungsbiologische oder genetische Forschung:}] Hier wird das Verst\"{a}ndnis \"{u}ber Funktionen und Eigenschaften von Genen und Mutationen erforscht. Die entwicklungsbiologische Forschung oder Ontogenese (aus dem Griechischen f\"{u}r \emph{Wesen} \emph{Geburt} \emph{Entstehung}) besch\"{a}ftigt sich mit den Vorg\"{a}ngen des Wachstums und der Entwicklung einzelner Organismen. Sie hat ihren Ursprung in der Embryologie. Im Hochdurchsatz lassen sich die genetische Kontrolle von Zellwachstum, Zelldifferenzierung und Zellspezialisierung in verschiedenen Zelltypen und Organen erforschen. Hierbei werden diverse Techniken wie \emph{Forward Genetics} und \emph{Reverse Genetics} angewandt \cite{lawson2011}. Damit ist es beispielsweise m\"{o}glich, gezielt die Expression von Genen auszul\"{o}sen oder zu verhindern. Durch die Techniken lassen sich Einblicke in die Funktion und Wirkungsweise der Gene gewinnen. Das Potenzial von \htsen im Zebrab\"{a}rbling wir u.a. in \cite{VandenBulck11,Patton01,Rinkwitz11} aufgezeigt.

          \item [\textbf{Toxikologische Forschung:}] Hier wird der Einfluss von Chemikalien auf den Organismus erforscht und Substanzen bez\"{u}glich der St\"{a}rke ihrer Toxizit\"{a}t untersucht. Es existieren auch bereits Regularien, die den \sog \emph{Fisch Embryo Test} (FET) beschreiben und festlegen, wie eine \hts f\"{u}r toxikologische Versuche und Frischwasser-Untersuchungen durchgef\"{u}hrt werden soll \cite{Braunbeck98,ISO07,Mayer86,Nagel02,OECD06a,OECD06}. Der Test soll dazu dienen, eine Vielzahl von Tierversuchen an ausgewachsenen Fischen zu vermeiden. Die Toxizit\"{a}ts-Tests m\"{u}ssen laut den gesetzlichen Vorschriften, wie sie in REACH\footnote{Verordnung (EG) Nr. 1907/2006 des Europ\"{a}ischen Parlaments und des Rates vom 18. Dezember 2006 zur Registrierung, Bewertung, Zulassung und Beschr\"{a}nkung chemischer Stoffe (REACH)}
               gefordert sind, f\"{u}r nahezu jede Chemikalie auf dem Markt durchgef\"{u}hrt werden \cite{Lahl06,Williams09}. Substanzen, vor allem Kunststoffe, Verbundstoffe und Polymere, die eine Jahresproduktion von einer Tonne erreichen, m\"{u}ssen nach dem neuen EU Gesetz \"{o}kotoxikologisch getestet werden. Unter das Gesetz fallen seit 2007 ca. 30.000 Substanzen. Die Toxizit\"{a}ts-Tests mittels Tierversuchen durchzuf\"{u}hren, w\"{u}rde bis zu 8 Mio. EUR pro Substanz kosten und jeweils bis zu 5 Jahre dauern \cite{Perkel12}. Daher investierte die EU seit 1986 \"{u}ber 300 Mio. EUR in alternative Tests, welche die Anzahl an Tierversuchen reduzieren k\"{o}nnen \cite{Ackerman04}. Ein weiterer (\"{o}ko)toxikologischer Anwendungsbereich ist die Bestimmung der Toxizit\"{a}t von Sedimentgestein \cite{Keiter10}. Hierf\"{u}r wird der \emph{Dantox}-Test (von \emph{danio rerio}, der lateinischen Bezeichnung des Zebrab\"{a}rblings) in einem BMBF-Projekt gef\"{o}rdert\footnote{DanTox - Entwicklung und Anwendung eines Verfahrens zur Ermittlung spezifischer Toxizit\"{a}t und molekularer Wirkungsmechanismen sedimentgebundener Umweltschadstoffe mit dem Zebrab\"{a}rbling (Danio rerio)}.  Die Aussagekraft und Anwendbarkeit von \htsen im Zebrab\"{a}rbling wurde allgemein best\"{a}tigt, jedoch wird diskutiert, inwieweit die Membranfunktion der Frucht"~ oder Eih\"{u}lle um die Zebrab\"{a}rblingslarve die Ergebnisse gegen\"{u}ber den herk\"{o}mmlichen Testverfahren verf\"{a}lscht \cite{Braunbeck05,Henn11,Lammer09a,Lammer09,Vaughan10,Yang09}.
              Das Potenzial der \hts im Zebrab\"{a}rbling auf diesem Gebiet wird ausf\"{u}hrlich beschrieben in \cite{VandenBulck11,Eimon09,Hill05, McGrath08,Selderslaghs12,Stegeman10,Straehle11,Yang09}. Limitierungen sind dem Verfahren allerdings durch die Bindung an das Wasser gesetzt. So k\"{o}nnen \zB nicht wasserl\"{o}sliche Substanzen oder Gase mit Hilfe des Fisch Tests oder Fisch Embryo Tests nur unzureichend getestet werden.

          \item [\textbf{Pharmakologische Forschung:}] Hier werden Chemikalien und kleine Molek\"{u}le bez\"{u}glich ihrer Eignung als Medikamente gepr\"{u}ft. Das Ziel ist es, Wirkstoffe zur Behandlung bisher unheilbarer Krankheiten oder Alternativen zu bereits etablierten Wirkstoffen zu finden \cite{Broach96,Hueser06,Gad05}. Bei pharmakologischen \htsen
              \begin{itemize}

              \item wird die Wirkung und das Wirkungsspektrum qualitativ und quantitativ ermittelt,
              \item wird versucht, den Angriffspunkt sowie den Wirkungsmechanismus zu kl\"{a}ren,
              \item der Einfluss auf verschiedene Organfunktionen festgestellt,
              \item die lokale und allgemeine Vertr\"{a}glichkeit gepr\"{u}ft und
              \item auf toxische Effekte geachtet.
              \end{itemize}

              Das Potenzial f\"{u}r \htsen im Zebrab\"{a}rbling auf diesem Gebiet wird ausf\"{u}hrlich beschrieben in \cite{Best08,Best08a,Bowman10,Carradice08,Yee05}. Eine \"{U}bersicht, welche detektierbaren oder vergleichbaren Auswirkungen menschliche Krankheiten \bzw Krankheitserreger auf den  Zebrab\"{a}rbling haben sowie eine Tabelle wichtiger Forschungsarbeiten bietet \cite{Lieschke07}.

        \end{description}

    \subsection{Grundlagen zur Bildakquise \& -verarbeitung f\"{u}r den Hochdurchsatz} \label{subsec:Bildakquise+Bildverarbeitung}
    Um eine \hts durchzuf\"{u}hren, m\"{u}ssen, wie bereits in Abschnitt~\ref{subsubsec:\"{U}bersichtHochdurschsatz} beschrieben ist, Biosignale erfasst werden. Dabei ist ein solches Signal immer notwendige Voraussetzung, unerheblich, auf welchem der Anwendungsgebiete aus Abschnitt~\ref{subsubsec:Relevante Forschungsgebiete} die \hts durchgef\"{u}hrt werden soll. F\"{u}r die unterschiedlichen Signale bietet der Markt spezielle Detektoren an. Die Dimension und der Aufwand der Akquise m\"{u}ssen dem Signal angepasst werden, um ein f\"{u}r den Hochdurchsatz geeignetes Sig\-nal-Rausch-Ver\-h\"{a}lt\-nis zu erreichen \cite{Colowick06}. Wird die Auswahl auf optische Signale beschr\"{a}nkt, so quantifizieren alle Detektoren die Menge an einfallendem Licht. Das auf den Sensor projizierte Signal ist zun\"{a}chst nichts weiter als eine zweidimensionale, zeitabh\"{a}ngige, kontinuierliche Verteilung von Lichtenergie. Die Lichtenergie kann ihren Ursprung in einer Lichtquelle oder in einer fluoreszierenden Zelle haben. Der Detektor hat die Aufgabe, drei wesentliche Schritte zu erf\"{u}llen \cite{Jahne05}:
    \begin{itemize}
      \item Eine r\"{a}umliche Abtastung der kontinuierlichen Lichtverteilung,
      \item eine zeitliche Abtastung der daraus resultierenden Funktion und
      \item eine Quantifizierung zur digitalen Darstellbarkeit und Verarbeitung.
    \end{itemize}

     Ein typisches Sensorelement ist eine Photodiode, deren Messung in der digitalen Repr\"{a}sentation als Pixel bezeichnet wird. Je nach Anordnung der Sensorelemente, geschickter Auswahl von Beleuchtungstechniken, zeitlichen Aufnahmen \usw l\"{a}sst sich die Dimension, also die zeitliche \bzw r\"{a}umliche Abtastung des akquirierten Datensatzes, beeinflussen. Die r\"{a}umliche Abtastung (\emph{spatial sampling}) erfolgt in der Regel durch die Geometrie des Detektors, die zeitliche Abtastung (\emph{temporal sampling}) geschieht durch Steuerung der Zeit, \"{u}ber die die Messung der Lichtmenge durch die einzelnen Sensorelemente erfolgt. Werden zu mehreren Zeitpunkten  Bilder akquiriert, entsteht eine Bildsequenz. Wie sich mehrdimensionale Repr\"{a}sentationen von Zebrab\"{a}rblingslarven mittels solcher Detektoren akquirieren lassen, wird im Verlauf der vorliegenden Arbeit aufgezeigt.

    Wird lediglich ein Sensorelement oder eine Reihe solcher verwendet, lassen sich eindimensionale Signale akquirieren. In \cite{Biometrica08,Pardo-Martin10,Pfriem11} werden solche Detektoren beispielsweise eingesetzt, um die Menge an abgeschattetem Licht oder die St\"{a}rke an Fluoreszenz zu quantifizieren. Eine kontinuierliche Aufnahme von Messwerten ist mittels fast aller Detektoren m\"{o}glich und ergibt eine Zeitreihe. Wird eine Zebrab\"{a}rblingslarve definiert an einem Sensor vorbei gef\"{u}hrt, so l\"{a}sst sich ein Profil \"{u}ber der Zeit aufzeichnen. Ein besonders hoher Durchsatz l\"{a}sst sich erzielen, wenn Zebrab\"{a}rblingslarven in einem Schlauch fortw\"{a}hrend gepr\"{u}ft werden. Die Information \"{u}ber den Ort der Quelle des Signals geht hierbei verloren oder ist durch die Anbringung des Sensors fix.

    \begin{figure}[!htbp]

        \centering
        \includegraphics[width=.8\linewidth]{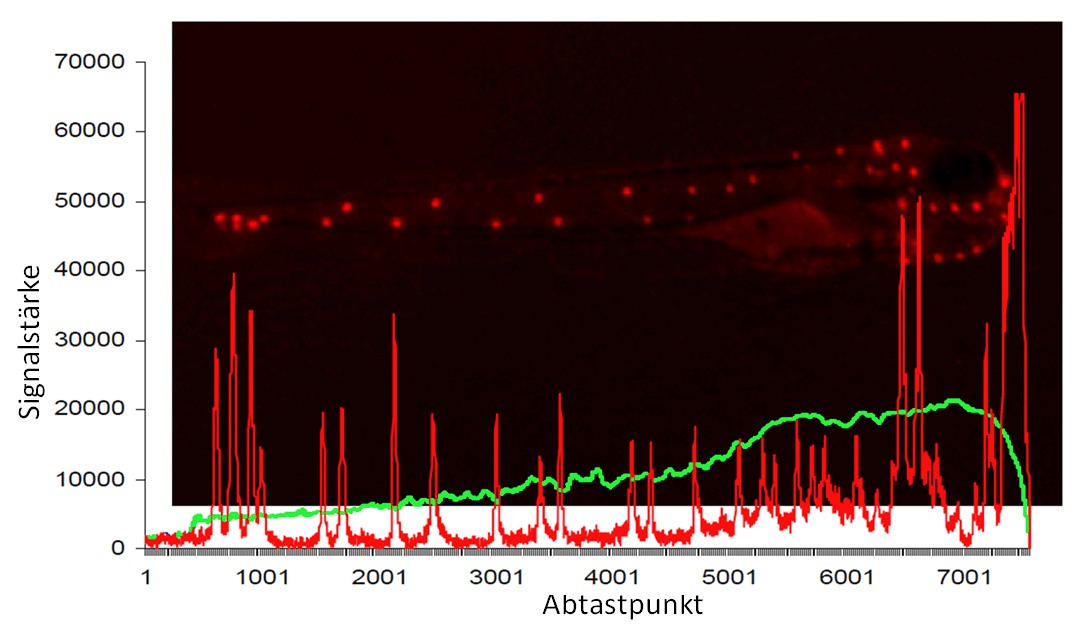}

        \caption[1D-Profil und 2D-Bildaufnahme einer Zebrab\"{a}rblingslarve im Vergleich]{Darstellung eines Zebrab\"{a}rblings mittels 1D-Profilen im Vergleich zu einer 2D-Bildaufnahme  \cite{Biometrica08}. Die gr\"{u}ne Linie zeigt die Menge an abgeschattetem Licht, w\"{a}hrend die rote Linie die Menge des Fluoreszenzsignals an der jeweiligen X-Position akkumuliert. Die jeweiligen Signale wurden getrennt aufgezeichnet und dann zur Visualisierung \"{u}berlagert.}
        \label{fig:2D_3D_Dektektor}

    \end{figure}
    \abb \ref{fig:2D_3D_Dektektor} zeigt eine \"{U}berlagerung einer zweidimensionalen Fluoreszenzaufnahme und Profile eindimensionaler Detektoren im Vergleich. Der erste Detektor (gr\"{u}ne Linie) quantifiziert die Menge an abgeschattetem Licht, \dhe je h\"{o}her der Wert, desto weniger Licht durchdringt die Larve. Der zweite Detektor (rote Linie) quantifiziert die Menge an von der Larve fluoreszierend ausgesendetem Signal, welches aufgrund eines anregenden Lasers ausgesendet wird. Der Vorteil der Methode liegt im erreichbaren Durchsatz und im g\"{u}nstigen Preis, w\"{a}hrend auf die Information der zweiten Ortskoordinate und eine bessere \"{o}rtliche Aufl\"{o}sung verzichtet werden muss.

    Eine typische Kamera, wie sie auch zur Akquise von Zebrab\"{a}rblingen verwendet wird, enth\"{a}lt eine regelm\"{a}{\ss}ige und rechtwinklige Anordnung solcher Sensoren \bzw Sensorelemente. Das aufgezeichnete Bild ist somit eine Matrix $\mathbf{I}$ mit den ganzzahligen Koordinaten $i_x = 1\dots I_x$; $i_y = 1 \dots I_y$ von Pixelwerten $I(i_x,i_y)$ innerhalb der Farbtiefe $\phi$:
    \begin{equation}
    \mathbf{I} \in \mathbb{N}^{I_x\times I_y} \quad \text{mit} \quad i_x,i_y\in\mathbb{N} \text{ und } {I}(i_x,i_y)< \phi
    \label{eqn:2D_Bild}
    \end{equation}
    F\"{u}r die Farbtiefe $\phi$ werden \"{u}blicherweise positive ganze Zahlen im Bereich $[0\dots 2^{\phi}-1]$ benutzt. Ein typisches Bild kann beispielsweise $\phi= 8$ bit besitzen und die Intensit\"{a}tswerte $[0 \dots 255]$ annehmen. Kleine Werte stehen hierbei f\"{u}r dunkle und hohe Werte f\"{u}r helle Pixel.

\begin{figure}[!htbp]
            \begin{minipage}[b]{0.45\linewidth}
        \centering
        \includegraphics[width=\linewidth]{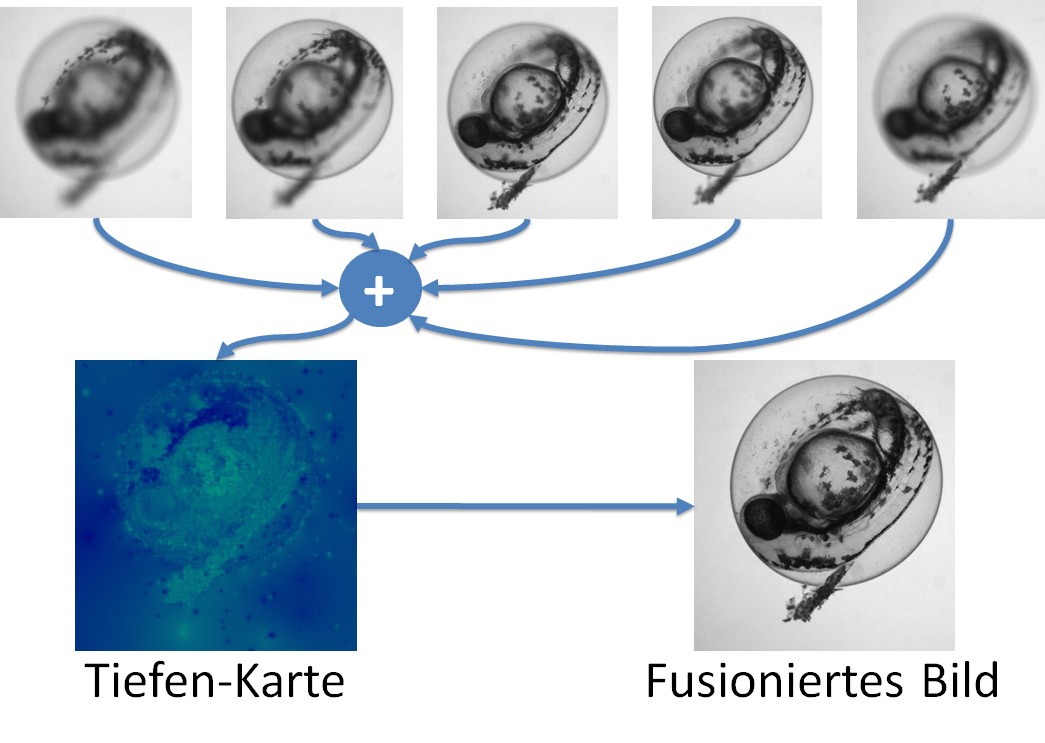}

        \caption[Mikroskopaufnahmen verschiedener Fokusebenen eines Zebrab\"{a}rblings]{Mikroskopaufnahmen verschiedener Fokusebenen eines Zebrab\"{a}rblings; aus der Wavelet-Transformation berechnete Tiefenkarte; fusioniertes Bild mit erweiterter Tiefensch\"{a}rfe}
        \label{fig:extendedfokus}

        \end{minipage}
        \hfill
        \begin{minipage}[b]{0.45\linewidth}
         \centering
        \includegraphics[width=\linewidth]{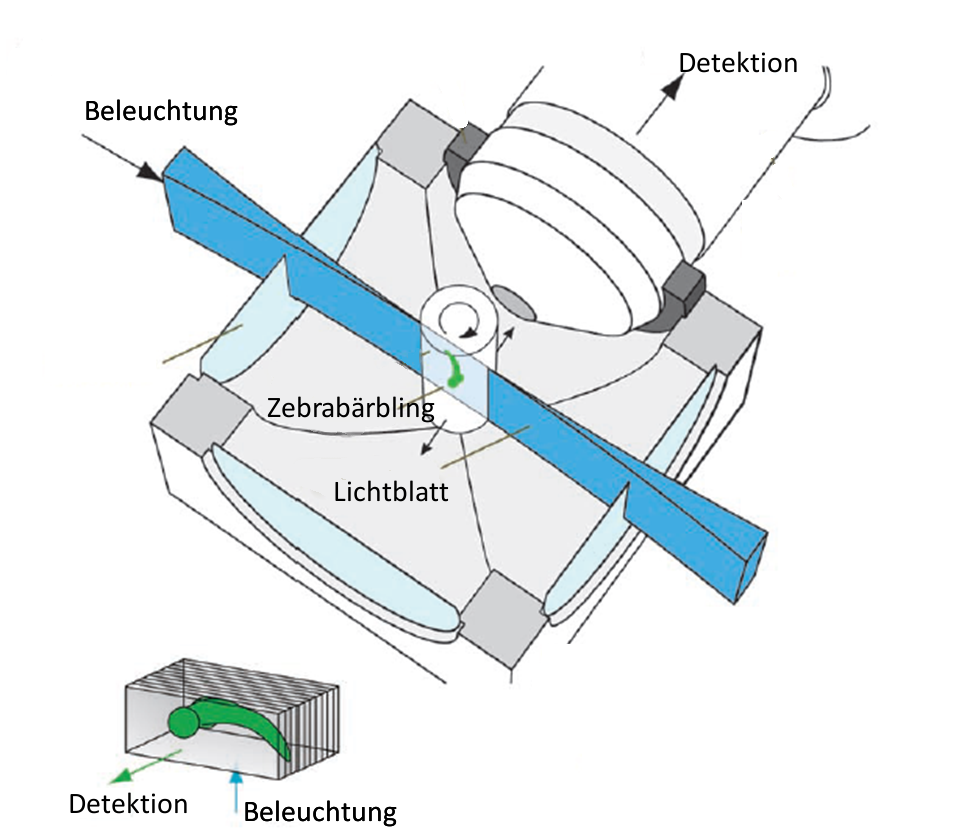}

        \caption[Prinzipskizze eines Lichtblatt-Mikroskops]{Prinzipskizze eines Lichtblatt-Mikroskops nach \cite{Huisken04} zur Akquise eines 4D-Datensatzes. Der Detektor  ist hier am Ende des Strahlenganges des Mikroskops angebracht.}
        \label{fig:SpimMikroskop}
        \end{minipage}
    \end{figure}

    Die 2D-Repr\"{a}sentation eines dreidimensionalen Zebrab\"{a}rblings ist je nach verwendetem Mikroskop, Objektiv, numerischer Apertur \etc meist nicht \"{u}ber die gesamte Tiefe des Modellorganismus scharf. Die optischen Limitationen beschr\"{a}nken die Fokusebene auf einen bestimmten Bereich, die \sog Tiefensch\"{a}rfe \cite{Russ02}. Eine dennoch scharfe Abbildung des gesamten Zebrab\"{a}rblings kann erfolgen, indem mehrere Bilder unterschiedlicher Fokusebenen fusioniert werden. Die Technik wird als \emph{Extended Focus} bezeichnet \cite{Forster04,Valdecasas01}. Hierbei wird \"{u}ber alle aufgenommenen Fokusebenen mittels einem Algorithmus (beispielsweise der Wavelet-Transformation) bestimmt, welche Pixel in welcher der Fokusebenen eine scharfe Abbildung des Objektes sind. Lediglich die scharf abgebildeten Pixel werden in einem einzigen resultierenden Bild mit \emph{erweitertem} Fokus gespeichert. Aus der r\"{a}umlichen Position des Pixels kann weiter eine dreidimensionale Repr\"{a}sentation rekonstruiert werden. Da keine Information vom Inneren des Volumens und von den verdeckten Fl\"{a}chen vorhanden ist, wird hier von einer 2,5D-Abbildung gesprochen (vgl. \abb \ref{fig:extendedfokus}).

     Vollst\"{a}ndige Volumenmodelle des Fisches lassen sich zerst\"{o}rungsfrei aus Schichtaufnahmen, wie sie beispielsweise mittels der Konfokal-Mikroskopie akquiriert werden k\"{o}nnen, erstellen. Die Schichtaufnahmen werden im Rechner zu einem vollst\"{a}ndigen 3D-Modell verbunden. Der Detektor ist auch hier ein herk\"{o}mmlicher 2D-Detektor. Entscheidend f\"{u}r die Erm\"{o}glichung der dreidimensionalen Akquise ist das Vorhandensein und die definierte Anregung von fluoreszierenden Zellen im Zebrab\"{a}rbling sowie die sp\"{a}tere Zuordnung der \"{o}rtlichen Information. Werden die Techniken der dreidimensionalen Fluoreszenzmikroskopie mit einer zeitlichen Aufnahme kombiniert wie beispielsweise bei der \emph{SPIM}-Methode (\vgl \abb \ref{fig:SpimMikroskop}), so lassen sich vierdimensionale Datens\"{a}tze aufzeichnen \cite{Huisken04,Keller08SC}.

    \subsection{Existierende L\"{o}sungsans\"{a}tze}
    Trotz den in Wirtschaft und Forschung weit verbreiteten, meist zellbasierten \htsen existieren nur wenige Publikationen, die, basierend auf Zebrab\"{a}rblingen, \htsen realisieren. Aus der Erfahrung der zellbasierten  Untersuchungen ist bekannt, dass sowohl die Durchf\"{u}hrung der \hts als auch deren Auslegung  strukturiertes Vorgehen verlangt. Die wichtigsten Arbeiten und deren Vorgehensweise werden im Folgenden kurz zusammengefasst.

    \subsubsection{Existierende bildbasierte \htsen im Zebrab\"{a}rbling}
    In der Literatur sind verschiedene Ans\"{a}tze zu finden, die das gro{\ss}e Potenzial des Zebrab\"{a}rblings f\"{u}r den Hochdurchsatz tats\"{a}chlich nutzen. Alle unterscheiden sich in der Art der Umsetzung, wie etwa dem Automatisierungsgrad oder der verwendeten Software. Es zeichnen sich bisher weder standardisierte Vorgehensweisen ab, noch hat sich eine Software etabliert. Zwar existiert in der Literatur eine freie, auf die Verwendung mit Zebrab\"{a}rblingen spezialisierte Bildverarbeitungs-Software. Die Software befindet sich allerdings in einem sehr fr\"{u}hen Entwicklungsstadium und wurde seit 2009 nicht weiter aktualisiert \cite{Liu07c,Liu07d,Liu08b}. Alle ver\"{o}ffentlichten, bildbasierten Siebtest-Untersuchungen gr\"{o}{\ss}eren Umfangs verfolgen ein eigens erstelltes Konzept mit einer eigens erstellten Bildverarbeitungsroutine oder ben\"{o}tigen eine kommerzielle und spezialisierte Plattform\cite{cachat11,green12}. Tabelle \ref{tab:Hochdurchsatz-Untersuchungen} bietet eine Auswahl an \htsen, eine weitere \"{U}bersicht findet sich in \cite{Lessman11}.

    \begin{table}
        \centering
       \footnotesize
        \begin{tabularx}{\linewidth}{
         >{\raggedright\arraybackslash}p{.7cm}
         >{\raggedright\arraybackslash}p{1.8cm} 
         >{\raggedleft\arraybackslash}p{1.4cm}
         >{\raggedleft\arraybackslash}p{1.4cm}
         >{\raggedright\arraybackslash}X
         >{\raggedleft\arraybackslash}p{2.5cm}
         >{\raggedleft\arraybackslash}p{.8cm}
         }
        \toprule
        \textbf{Jahr}& \textbf{Art des Screens}& \textbf{Anz. Chem.}&\textbf{Anz. Fische}& \textbf{Methode}& \textbf{Dim. der Bildakquise} & \textbf{Ref.}   \tabularnewline 
        \midrule
            2012 &  Mutant-Identifizierung & & >1000 & Bildauswertung und -akquise automatische Station: Tecan Infinite M1000 & 2D + Fluoreszenz & \cite{walker12}\\
            2012  &      Medikamenten-Untersuchung &     3   & 3000 &  Automatisches Mikroskop, Hellfeld + Fluoreszenz, automatisches Sortierverfahren, manuelle Bildverarbeitung (\emph{ImageJ})    &   1D+2D + Zoom + Fluoreszenz &     \cite{Carvalho11}\\
            2011 & Mutant-Identifizierung & 5 & ~ 1000 & Automatisches Mikroskop, Hellfeld, Skriptbasierte Bildverarbeitung & 2D & \cite{liu12}\\
            2011  &      Transgenetische Reporter  &     3   & &  Automatisches Mikroskop, Hellfeld + Fluoreszenz, Skriptbasierte Bildverarbeitung (\emph{Labview})    &   2D + Zoom + Fluoreszenz &     \cite{Peravali11}  \\



            2010     &      Verhaltens-Untersuchung  &    14\,000 & 70\,000  &   Automatisches Mikroskop, Kamera, \emph{Metamorph}-Control Software   &   2D + Zeit & \cite{Kokel10} \\

            2010     &      Verhaltens-Untersuchung  &    5\,648   & &   Megapixel Objektiv, Infrarot Kamera, \emph{Viewpoint} Tracking Software   &   2D + Zeit & \cite{Rihel10} \\

            2010  &      Medikamenten-Untersuchung  &     34   & 1\,700 &   Automatisches Mikroskop, Hellfeld + Fluoreszenz, Skriptbasierte Bildverarbeitung (\emph{Labview})    & 2,5D + Fluoreszenz  &  \cite{dAlencon10}\\

            2010  &      Entwicklung einer Hochdurchsatz-Methode  &       & 450 &   speziell angefertigte Hochdurchsatz-L\"{o}sung, mehrere Detektoren, keine automatisierte Bildauswertung & 1D + 2D + Fluoreszenz & \cite{Pardo-Martin10} \\

            2009  &      Verhaltens-Untersuchung  &     3   & &  Infrarot Kamera, \emph{Noldus} Tracking Software&   2D + Zeit &\cite{Irons10}\\

            2009  &      Mutant-Identifizierung  &     115   & &  Manuelles Mikroskop, Skriptbasierte Bildverarbeitung    & 2D &  \cite{Cao09} \\

            2009  &      Identifikation von Genkombinationen  &        &   17\,730   &   Automatisiertes Mikroskop, Hellfeld + Fluoreszenz, Skriptbasierte Bildverarbeitung
           (\emph{MATLAB-Software}) &  2,5D + Fluoreszenz &     \cite{Gehrig09} \\
           \bottomrule
        \end{tabularx}
        \caption{Ausgew\"{a}hlte Zebrab\"{a}rblings-Untersuchungen im Hochdurchsatz oder mit Hochdurchsatz-Potenzial mit Anzahl an Fischen oder Substanzen (falls in der Ver\"{o}ffentlichung angegeben). Eine weitere ausf\"{u}hrliche \"{U}bersicht und Gegen\"{u}berstellung bietet \cite{Mikut13}. }
        \label{tab:Hochdurchsatz-Untersuchungen}
        \end{table}

In den Untersuchungen von \cite{Gehrig09} werden sowohl Hellfeld-Aufnahmen in unterschiedlichen Fokusebenen als auch Fluoreszenzsignale, nach manueller Vorarbeit, automatisiert akquiriert und die St\"{a}rke des Fluoreszenzsignals mittels spezialisierten Algorithmen den Regionen im Fisch zugeordnet und miteinander verglichen. Da es sich beim Ziel der Untersuchung um die Identifikation von Genkombinationen handelt, kommen keine Chemikalien w\"{a}hrend des Screenings zum Einsatz. Die hochspezialisierte Bildverarbeitung leistet sowohl die Rotation der Fische als auch die vollautomatische Erkennung von Regionen sowie die Messung und Zuordnung von Fluoreszenzsignalen zu einzelnen Bereichen. Schlie{\ss}lich erfolgt die Auswertung und Visualisierung der Ergebnisse in \sog \emph{Fingerprints}, bei denen der St\"{a}rke des Signals eine Helligkeit und der Region im Fisch jeweils eine Farbe und ein Teil eines Rechtecks zugeordnet werden. Dies erm\"{o}glicht das schnelle Vergleichen und Ordnen der Ergebnisse sowie das Ableiten von Schlussfolgerungen. Die gesamte Programmierung, Bild"~ und Datenauswertung erfolgt mit der Software \emph{MATLAB}\footnote{The MathWorks, Inc.}. Es sei angemerkt, dass in \cite{kokel12} vorgeschlagen wird, solche \emph{Fingerprints} mittels moderner \emph{cloudbasierter} Techniken zu vernetzen.

In \cite{Cao09} f\"{u}hren die Autoren die \hts zwar vollst\"{a}ndig manuell durch, zeigen jedoch nur das Hochdurchsatz-Potenzial ihrer Methode auf, ohne derartige Versuche durchzuf\"{u}hren. Ziel ist es, einen mutierten Zebrab\"{a}rbling zu identifizieren, dessen Ph\"{a}notyp ein gekr\"{u}mmter R\"{u}cken ist. Die Autoren stellen einen sehr einfachen Bildverarbeitungsalgorithmus vor, der in der Lage ist, den R\"{u}cken der Tiere zu finden und auf Kr\"{u}mmung zu pr\"{u}fen. Bei zu dicht beieinander auftretenden Fischen versagt die Methode jedoch. Die Autoren vergleichen die Ergebnisse ihrer automatischen mit einer manuellen Analyse, kommen zu \"{u}bereinstimmenden Ergebnissen und schlie{\ss}en so auf Hochdurchsatz-\emph{F\"{a}higkeit} des Verfahrens. Zur Steuerung der Mikroskope kommt die Software \emph{Metamorph}\footnote{Molecular Devices, LLC, USA} zum Einsatz. Die Autoren machen keine Angabe, in welcher Software die Bildverarbeitung realisiert wurde.

In \cite{Irons10} wird die zur\"{u}ckgelegte Distanz von Zebrab\"{a}rblingslarven sowohl bei Helligkeit als auch im Dunkeln gemessen. Die Vorbereitung der Larven erfolgt manuell, w\"{a}hrend die Bildakquise innerhalb einer kommerziell erh\"{a}ltlichen gekapselten Box der Firma \emph{Noldus\footnote{Noldus Information Technology}} ohne Verwendung eines Mikroskops erfolgt. Die Bilder werden lediglich mittels einer Industriekamera akquiriert. Die Bildverarbeitung erfolgt durch Verwendung einer Tracking-Software der Firma \emph{Noldus}.

Das Ziel der Autoren in \cite{Pardo-Martin10} ist die Durchf\"{u}hrung einer Machbarkeitsstudie, um eine neue Automatisierungstechnik zur Analyse von Zebrab\"{a}rblingen zur Verf\"{u}gung zu stellen. Die Autoren stellen einen komplexen Apparat vor, der den Fisch innerhalb eines Schlauchs in einer Glas-Kapillaren transportiert, dort rotiert und w\"{a}hrenddessen Bilder akquiriert. Durch die Akquise von allen Seiten wird das Problem der undefinierten Lage der Tiere gel\"{o}st, welches bei herk\"{o}mmlicher Mikroskopie unumg\"{a}nglich ist. Eine belastbare  Anwendung der Konstruktion im Hochdurchsatz wurde bis zum heutigen Zeitpunkt jedoch noch nicht vorgestellt.

In \cite{dAlencon10} bedienen sich die Autoren eines automatisierten Mikroskops, wie es in \cite{Liebel03} vorgestellt ist. Es werden automatisiert mehrere Fluoreszenzkan\"{a}le akquiriert und es wird gezeigt, dass die Auswertung mittels eines automatisierten Skriptes erfolgen kann. Anders als in der Arbeit von \cite{Gehrig09} wird hier nicht der gesamte Fisch in Regionen aufgeteilt, sondern es werden lediglich Regionen von Interesse im Fisch (\sog Neuromasten) erkannt und die Menge an Fluoreszenz gemessen. Ein Vergleich der automatisierten mit der manuellen Auswertung ergibt \"{a}hnliche Ergebnisse. Da die Bildverarbeitung nicht in der Lage ist, jede der manuell erkennbaren Regionen zu identifizieren, ist es notwendig, die Menge an untersuchten Fischen zu erh\"{o}hen, um zu gleichen Ergebnissen von manueller und automatischer Analyse zu kommen. Die Steuerung der Mikroskope sowie die Bildverarbeitung wurde mit Hilfe der Software \emph{Labview\footnote{National Instruments Corporation}} realisiert.

In \cite{Rihel10} werden Schlaf"~ und Aktivit\"{a}tsphasen der Fische klassifiziert und \sog Verhaltens\-profile erstellt. Die Profile werden f\"{u}r \"{u}ber 5\,000 Substanzen ermittelt. Das Beobachten und Tracken der Fische erfolgt mittels einer kommerziell erh\"{a}ltlichen Box der Firma \emph{Viewpoint\footnote{Viewpoint S.A. - Viewpoint Life Sciences, Inc.}}, welche wie in \cite{Irons10} auf ein Mikroskop verzichtet, lediglich Hellfeld-Aufnahmen durchf\"{u}hrt und eine Tracking-Software nutzt. Das Verhaltensprofil der mit Chemikalien exponierten Fische wird mit dem der unbehandelten Fische verglichen. Abweichungen k\"{o}nnen mittels eines Farbcodes visualisiert und gruppiert werden. Mit der vorgestellten Methode lassen sich \"{a}hnlich wirkende Substanzen schnell zusammenfassen.

In \cite{Kokel10} werden Fische mittels eines Stimulus zu einer Bewegung animiert. W\"{a}hrend des Experiments wird die Bewegung der Fische durch ein automatisches Mikroskop der Firma Nikon und einer Kamera aufgezeichnet. Die Synchronisierung der Hardware sowie die Bildauswertung erfolgt mittels der Software \emph{Metamorph}\footnote{Molecular Devices, LLC, USA}. Es werden \"{u}ber 14\,000 Substanzen gepr\"{u}ft und vergleichbar zu \cite{Rihel10} wird eine Gruppierung der Substanzen vorgenommen, die auf einem Vergleich der Auswirkungen der Substanzen basiert.


In \cite{Peravali11} wird eine Bildanalyse bereits w\"{a}hrend der Bildakquise eingebunden. Dies hat den Vorteil, dass nicht relevante Strukturen oder Proben aussortiert werden k\"{o}nnen, wohingegen Bereiche im Fisch von Interesse in hoher Aufl\"{o}sung akquiriert werden. Die Mikroskopsteuerung und Bildverarbeitung erfolgt vollst\"{a}ndig mittels einer skriptbasierten speziellen Kombination aus \emph{XML}-Dateien\footnote{XML = Extensible Markup Language (engl. f\"{u}r "`erweiterbare Auszeichnungssprache"')} und \emph{Labview}-Skripten\footnote{National Instruments Corporation}.

In \cite{liu12} werden Larven durch ein automatisches System vorbereitet und mit Nanopartikeln beaufschlagt. Aus einer Reihe von Merkmalen, die aus den Bildern extrahiert werden, lassen sich drei verschiedene Endpunkte automatisch identifizieren. Das System wird an ca. 1000 Embryonen demonstriert. Die Erfolgsquote liegt bei \"{u}ber 90\%\cite{Broach96,Hueser06,Gad05}.

In \cite{Carvalho11} werden Larven automatisiert durch Injektion mit Tuberkulosebakterien infiziert. Es wird ein technisch realisierbarer Durchsatz von 2000 Larven pro Stunde angef\"{u}hrt. Das kommerzielle \emph{COPAS} System wird zur Analyse der Daten verwendet.

In \cite{walker12} findet ein kommerzielles hochspezialisiertes Auswertesystem f\"{u}r Mikrotiterplatten (Tecan Infinite M1000) Anwendung. Es wird ein theoretisch sehr hoher Durchsatz von \"{u}ber 50000 Einheiten pro Tag angegeben. Das System ist allerdings beschr\"{a}nkt auf Auswertungen in Fluoreszenzaufnahmen.

Die vorgestellten \htsen am Zebrab\"{a}rbling sind bislang oftmals noch auf manuelle Schritte bei der Versuchsvorbereitung angewiesen. Es existieren jedoch bereits automatische robotergest\"{u}tzte Verfahren, deren Ziel es ist, eine vollst\"{a}ndige Automatisierung der Vorbereitung der Untersuchungen zu erm\"{o}glichen. Um DNA, RNA oder Proteine in Zebrab\"{a}rblingseier einzubringen, wird beispielsweise die Mikroinjizierung angewandt \cite{Xu99}. Der Vorbereitungsschritt wurde bereits mehrfach durch automatisierte Systeme gel\"{o}st \cite{Hogg08,Lu07,Wang07,Xie11}. Ebenso existieren L\"{o}sungsans\"{a}tze zur automatischen Best\"{u}ckung der Mikrotiterplatten sowie der automatischen Entfernung des Chorions, welches die Zebrab\"{a}rblingseier umgibt \cite{Mandrell12,Zhang11}.

  Einige der Ver\"{o}ffentlichungen besch\"{a}ftigen sich speziell mit der Analyse des Herzschlags im Zebrab\"{a}rbling. Anders als die in Tabelle \ref{tab:Hochdurchsatz-Untersuchungen} aufgef\"{u}hrten Publikationen beschr\"{a}nken sich die Untersuchungen auf die methodische Durchf\"{u}hrbarkeit etwa durch  Extraktion des Signals der roten Blutk\"{o}rperchen aus Hellfeld-Aufnahmen \cite{bhat09} oder die Detektion des Herzschlags ohne ein Markieren des Herzens \bzw der Zellen \cite{bhat12}. Sie erfordern zumeist ein exaktes Positionieren der Larven \cite{Chan09, Ohn11}. Allen ist gemein, dass bisher keine systematische Analyse in gr\"{o}{\ss}erem Umfang durchgef\"{u}hrt wurde, was meist auf die komplizierten manuellen Pr\"{a}parationsschritte zur\"{u}ckgef\"{u}hrt werden kann. Somit war ein Einordnen in die Tabelle \ref{tab:Hochdurchsatz-Untersuchungen} nicht m\"{o}glich.

Es existieren eine ganze Reihe weiterer Ans\"{a}tze, von denen hier nur auf eine Auswahl der wichtigsten eingegangen werden konnte. Dabei wurden vor allem die j\"{u}ngeren Ver\"{o}ffentlichungen aufgef\"{u}hrt und solche, die eine hohe St\"{u}ckzahl \bzw einen hohen Durchsatz erzielen.



    \subsection{Offene Probleme}
Zusammenfassend ergibt sich aus den vorangegangenen Abschnitten die Erkenntnis, dass der Zebrab\"{a}rbling nach \"{u}bereinstimmender Meinung vieler Autoren erhebliches Potenzial f\"{u}r den Hochdurchsatz besitzt. Offensichtlich ist jedoch, dass aufgrund einer Reihe offener Probleme bisher nur ein geringer bis mittlerer Durchsatz erzielt werden konnte. Die wesentlichen technischen Probleme, weshalb \htsen im Zebrab\"{a}rbling bisher scheitern, sind:

\begin{enumerate}
  \item \emph{Entwurf der Hochdurchsatz-Prozesskette:}

  Bisher existiert kein systematisches Entwurfskonzept einer Hochdurchsatz-Prozesskette. Alle publizierten Ans\"{a}tze versuchen lediglich, manuelle Schritte teilweise oder ganz zu ersetzen. Die technischen, logistischen und informationstechnischen Voraussetzungen sind bei einer Anzahl von mehreren tausend Versuchen pro Tag jedoch komplex, daher bedarf es einer strukturierten Planung.

  \item \emph{Auswahl der geeigneten Art der Datenakquise:}

  Die Wahl der Art der Datenakquise aller in der Literatur vorhandenen L\"{o}sungsans\"{a}tze basiert im Wesentlichen auf der Erfahrung und dem Equipment, welches dem Entwickler zur Verf\"{u}gung steht. Entscheidende Vor"~ und Nachteile der einzelnen Methoden werden nicht oder nur unzureichend ber\"{u}cksichtigt \bzw ausgesch\"{o}pft. Dies f\"{u}hrt zu ineffizienter Datenakquise, oft unzureichenden Daten oder zu Datenvolumina, die ohne entsprechendes Fachwissen nicht zu bew\"{a}ltigen sind.

  \item \emph{Entwicklung und Auswahl geeigneter Bildverarbeitungs"~ und Datenanalysealgorithmen: }

    Ein systematischer \"{U}berblick \"{u}ber existierende Methoden und deren Anwendbarkeit f\"{u}r die biologischen \htsen mit Zebrab\"{a}rblingen existiert nicht. \"{A}hnlich der Auswahl der Datenakquise-Methode werden auch die Bild\-ver\-ar\-bei\-tungs-Meth\-oden basierend auf Vorlieben der Entwickler und dessen Kenntnisstand ausgew\"{a}hlt. Gleiches gilt f\"{u}r die gesamte Datenverarbeitung einschlie{\ss}lich Normierungs"~, Klassifikations"~ und Visualisierungsstrategien.

  \item \emph{Es existiert keine systematische Abstimmung der Datenakquise und Datenverarbeitung:}

      Die Bildverarbeitung muss oft auf unzureichend und nicht auf die Bed\"{u}rfnisse der automatisierten Auswertung abgestimmte Bilddaten zur\"{u}ckgreifen. Die beiden zuvor genannten Punkte sind hochgradig voneinander abh\"{a}ngig, werden bei der Versuchsauslegung jedoch kaum aufeinander abgestimmt. Es existieren keine Untersuchungen bez\"{u}glich der Auswirkung der Auswahl bestimmter Akquise-Methoden auf die Auswertung und umgekehrt.

\end{enumerate}

\section{Ziele und Aufgaben}

\htsen stellen ein leistungsstarkes Mittel zum Erlangen neuer Erkenntnisse in den Wissensgebieten der Genetik, Toxikologie oder Pharmazie dar. L\"{o}sungen f\"{u}r \htsen existieren bereits, beschr\"{a}nken sich jedoch entweder auf Versuche mit Zellen oder k\"{o}nnen, wenn sie einen gesamten Modellorganismus verwenden (wie \zB den Zebrab\"{a}rbling), den gew\"{u}nschten hohen Durchsatz nicht erreichen. Ein System zu finden, welches die Vorteile der \hts mit denen eines gesamten Modellorganismus verbindet, ist demnach von beachtlichem Wert. Ziel der vorliegenden Arbeit ist es daher, ein Konzept zu entwickeln, welches den systematischen Entwurf einer bildbasierten \hts auf Basis des Zebrab\"{a}rblings im Hinblick auf die Automatisierbarkeit erm\"{o}glicht.
Hierf\"{u}r sind die folgenden wissenschaftlichen Teilziele notwendig:

\begin{description}
  \item [1. Herleitung eines Konzeptes]\hfill \\
  Ein Konzept zur Versuchsauslegung ist zu entwickeln. Dieses muss universell einsetzbar und an die jeweilige biologische Fragestellung und den gew\"{u}nschten Durchsatz adaptierbar sein.

  \item [2. Ableitung von Auswertungskriterien]\hfill \\
  Zur Beurteilung der Durchf\"{u}hrbarkeit und zum Entwurf einer \hts sind Kriterien zu finden, die in der Lage sind, bekannte und neue   L\"{o}sungsans\"{a}tze zu analysieren und zu bewerten.

  \item [3. Entwicklung neuer Datenverarbeitungsverfahren]\hfill \\
  Neue Daten- und Bildverarbeitungsverfahren sind im Hinblick auf die Anforderungen der \hts abzuleiten.

  \item [4. Anpassungsf\"{a}hige Implementierung]\hfill \\
  Um die einfache und benutzerfreundliche Anwendbarkeit der Verfahren zu sichern, ist eine grafische Benutzeroberfl\"{a}che zu entwickeln, die die erarbeiteten L\"{o}sungen in sich vereint und den direkten Zugriff und die Visualisierung der Ergebnisse erm\"{o}glicht.

  \item [5. Experimentelle Erprobung am Zebrab\"{a}rbling]\hfill \\
  Im Hinblick auf die Besonderheiten, die sich bei der Verwendung des Zebrab\"{a}rblings  bei \htsen ergeben, ist das entwickelte Verfahren zu       konkretisieren. Insbesondere biologische, technische Assistenten m\"{u}ssen ohne Expertenwissen im Bereich der Datenverarbeitung in der Lage sein, Versuchsdaten einzulesen und auszuwerten.

  \item [6. Ableitung einer Aussage \"{u}ber die Leistungsf\"{a}higkeit]\hfill \\
  Zum Nachweis der Funktionalit\"{a}t sind Fallstudien durchzuf\"{u}hren. Die Ergebnisse geben Auskunft \"{u}ber den erreichbaren Durchsatz und die erreichbare Qualit\"{a}t des neu entwickelten Verfahrens.
\end{description}

In Kapitel \ref{chap:Neues_Konzept} werden zun\"{a}chst die Anforderungen an das zu entwickelnde Verfahren definiert und das neue Konzept zur Versuchsauslegung vorgestellt. Mit Hilfe des Konzeptes ist es dann m\"{o}glich, die Versuchsauslegung zu konkretisieren und die Versuchsparameter anforderungsgem\"{a}{\ss} zu w\"{a}hlen. Die zur technischen Umsetzung des Konzeptes neu entwickelten Methoden werden anschlie{\ss}end in Form von Modulen in Kapitel \ref{sec:BV_Module} vorgestellt. Darauf aufbauend vereint Kapitel \ref{chap:Implementierung} die vorgestellten Methoden in einer grafischen Benutzeroberfl\"{a}che, welche Zugriff auf alle wichtigen Parameter der Datenverarbeitung bietet. Anwendung findet das erarbeitete Verfahren in Kapitel \ref{chap:Anwendung} anhand konkreter biologischer Problemstellungen, f\"{u}r welche eine hochdurchsatzf\"{a}hige L\"{o}sung vorgestellt wird. Eine Zusammenfassung der wesentlichen Ergebnisse der Arbeit sowie ein Ausblick \"{u}ber weitere m\"{o}gliche Untersuchungen sind Gegenstand von Kapitel \ref{chap:Zusammenfassung}.

\chapter{Neues Konzept zur Ver\-suchs\-auswertung von \htsen am Zebrab\"{a}rbling} \label{chap:Neues_Konzept}


\section{\"{U}bersicht}
In diesem Kapitel wird erstmals ein zielgerichtetes, systematisches Konzept zur Versuchsauswertung bei \htsen  vorgestellt.

Ein solches Konzept kann unter Praxisbedingungen f\"{u}r \htsen an Zebrab\"{a}rblingen nur dann erfolgreich sein, wenn es die Schritte
\begin{itemize}
  \item Definition der Versuchs"~ und Auswerteparameter und
  \item anforderungsgerechte Versuchsauslegung
\end{itemize}
beinhaltet.

In Abschnitt \ref{sec:Konzept_Versuchsauslegung} wird zun\"{a}chst das Konzept beschrieben. Darauf aufbauend werden die notwendigen Anforderungen bez\"{u}glich der Durchf\"{u}hrbarkeit, der Messbarkeit und der Auswertbarkeit von \htsen zusammengetragen (Abschnitt \ref{sec:HDU_Anforderungen}). In Abschnitt \ref{sec:Beschreibung_Parameter} werden Versuchs"~ und Auswerteparameter identifiziert und daraufhin mathematisch beschrieben (Abschnitt \ref{sec:Parameter_mathematisch}). Auf den Ergebnissen aufbauend wird abschlie{\ss}end die Vorgehensweise zur anforderungsgerechten Versuchsauslegung vorgestellt (Abschnitt~\ref{sec:Anforderungsgerechte_Anwendung}).

\section{Neues Konzept zur Versuchsauswertung}\label{sec:Konzept_Versuchsauslegung}
Der Auslegung bildbasierter \htsen kommt aufgrund der hohen Spezifizierung auf das \Nutzsig eine entscheidende Bedeutung f\"{u}r den Erfolg des Versuchs zu. Durch die Biologie muss festgelegt werden, welches das \Nutzsig, \dhe die Information von Interesse, im zu untersuchenden Informationstr\"{a}ger ist. Das \Nutzsig (z.B. eine sich bewegende Zebrab\"{a}rblingslarve) muss daraufhin mit geeigneten Mitteln (z.B. Vereinzelung der Larven) derart pr\"{a}pariert werden, dass die Bildakquise und deren Verfahren (z.B. Mikroskopie) das Nutzsignal in Bilddaten abbilden kann. Die Daten m\"{u}ssen schlie{\ss}lich innerhalb des Blockes \emph{Analyse und Interpretation }nach neuen Erkenntnissen bez\"{u}glich der Fragestellung des Versuchs ausgewertet werden. Die Auswertung erfolgt manuell oder durch eigens entwickelte \bzw entsprechend adaptierte Auswertealgorithmen. Dabei wirken sowohl auf die Biologie als auch auf die Bildakquise St\"{o}rungen, die Einfluss auf die objektive Realit\"{a}t oder deren Abbildung haben.

\begin{figure}[!htb]
        \centering
                 \captionsetup[subfloat]{
                 singlelinecheck=false,
                 justification=RaggedRight
                 }
         \subfloat[Heutige Auslegung
                 ]{\includegraphics[page=1]{Bilder/Diss_Zusammenhang}
                 \label{fig:Neues_Konzept_Prinzip_a}}
                 \\
                 \subfloat[Neue Auslegung                ]{\includegraphics[page=2]{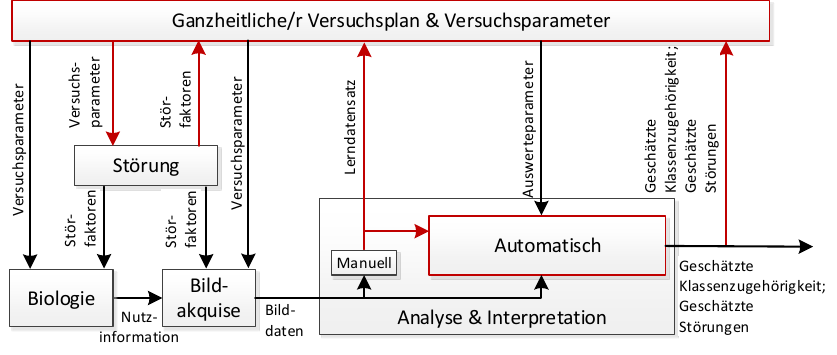}
                 \label{fig:Neues_Konzept_Prinzip_b}}

\caption[Bisherige und neue Versuchsauslegung f\"{u}r \htsen]{Bisherige Vorgehensweise (heutige Auslegung) und Sollzustand (neue Auslegung) f\"{u}r die Versuchsauslegung bildbasierter \htsen.}
\label{fig:Neues_Konzept_Prinzip}
\end{figure}

Die biologischen Schritte zur Bereitstellung der Nutzinformation ber\"{u}cksichtigen bisher lediglich, inwiefern die Information mittels Bildakquise erfasst werden kann, nicht ob sich diese zur (automatisierten) Auswertung eignet und ob M\"{o}glichkeiten zur Skalierung bestehen. \abb \ref{fig:Neues_Konzept_Prinzip_a} stellt die heutige Vorgehensweise bei der Auslegung dar. In Biologie und Bildakquise werden durch Versuchsparameter die Randbedingungen festgelegt. Auf den Versuch wirkt eine St\"{o}rung in Form von St\"{o}rfaktoren. In den biologischen Schritten wird eine \NutzsigInf (kurz: Nutzinformation) bereitgestellt, die in den Bilddaten abgebildet und durch Analyse und Interpretation einer Klasse zugeordnet wird. Die Klassenzuordnung geschieht bisher manuell. In neueren Arbeiten wird nun versucht, den manuellen Schritt durch eine Nachahmung der manuellen Arbeit zu ersetzen. Hierf\"{u}r sind Auswerteparameter notwendig. Das Ergebnis sind gesch\"{a}tzte Klassenzugeh\"{o}rigkeiten. Ebenso lassen sich die St\"{o}rungen \bzw deren Klassenzugeh\"{o}rigkeiten sch\"{a}tzen. Der Anteil an automatisch ausgef\"{u}hrten Auswertungen ist klein, was durch den deutlich kleineren Block im Schaubild visuell verdeutlicht wird. Die Auswahl der Versuchsparameter wie z.B. der Akquise-Methode beruht meist auf einer individuellen, nicht systematischen Entscheidungsgrundlage wie "`Verf\"{u}gbarkeit der Mikroskope"' oder "`pers\"{o}nliche Expertise des Laboranten"' \etc Die in der Literatur zu findenden \htsen f\"{u}hren die Versuchsauslegung oft willk\"{u}rlich, aus Erfahrungswerten oder anhand der vorhandenen Mikroskop"~ und Rechnerinfrastruktur durch \cite{Patton01,Sabaliauskas06,Tran07}. Ein methodisches und ganzheitliches Vorgehen ist jedoch unumg\"{a}nglich, um neuartige drei"~, vier"~ oder auch n-dimensionale (durch Ber\"{u}cksichtigung von Ver\"{a}nderungen im Raum und \"{u}ber die Zeit) Datens\"{a}tze effizient auswerten zu k\"{o}nnen. Des Weiteren basieren die Ergebnisse aller sich anschlie{\ss}enden Analyse"~ und Interpretationsschritte auf den derart akquirierten Bilddaten. Die Qualit\"{a}t der Bilddaten limitiert somit die erreichbare Qualit\"{a}t des Versuchs.
Die Auswertung erfolgt bisher in weiten Teilen manuell oder mittels (semi)automatischer Auswerteroutinen (vgl. Literatur\"{u}bersicht in \tab \ref{tab:Hochdurchsatz-Untersuchungen}). Die weitgehend individuelle und ungeordnete Versuchsauslegungsphilosophie hat den Nachteil, dass nachfolgende Verarbeitungsschritte nicht in die Entscheidungsfindung zur Parameterwahl mit einbezogen werden. Dies ist auch ein Grund, weshalb sich zur Zeit nur ein geringer Anteil der Analyse"~ und Interpretationsschritte automatisiert durchf\"{u}hren l\"{a}sst. Deshalb f\"{u}hrt die vorliegende Arbeit eine Analyse der Parameter durch, um eine Versuchsauslegung anforderungsgerecht durchf\"{u}hren zu k\"{o}nnen.

Das hier vorgeschlagene neuartige Konzept zur Versuchsauslegung der \hts bildet gem\"{a}{\ss} \abb \ref{fig:Neues_Konzept_Prinzip_b} ebenfalls die Nutzinformation mittels Bilddaten ab, jedoch werden alle Entscheidungen, Versuchs- sowie Auswerteparameter zielgerichtet auf die sp\"{a}tere automatisierte Auswertung hin getroffen. Der Prozess der Versuchsauslegung wird ganzheitlich vorgenommen und der Einfluss der jeweils getroffenen Entscheidung nicht nur auf den sich direkt anschlie{\ss}enden Schritt, sondern auf alle Bl\"{o}cke betrachtet. Die ganzheitliche Betrachtung schlie{\ss}t die Auswahl der Akquise-Methode sowie die Identifikation m\"{o}glichst vieler St\"{o}rgr\"{o}{\ss}en ein. Auf Basis der gesammelten Information werden Versuchs\-parameter gew\"{a}hlt. Des Weiteren wird der Anteil der automatisch vollzogenen Analyse und Interpretation stark erh\"{o}ht, was durch den gro{\ss}en Block verdeutlicht ist. Der manuelle Anteil ist deutlich kleiner und dient im optimalen Fall lediglich dazu, eine Wissensbasis f\"{u}r die automatische Auswertung zu schaffen. Der ausgewertete oder gelabelte (von engl. \emph{to label} = markieren, beschriften) Datensatz der Wissensbasis wird verwendet, einen Klassifikator anzulernen, mit dessen Hilfe die Klassenzugeh\"{o}rigkeit unbekannter Daten gesch\"{a}tzt wird.
Selbst einer vollst\"{a}ndigen Automatisierung der \hts muss indes zwingend eine manuelle Deutung der Ergebnisse folgen, da das Ziel, \bzw die Art der Treffer, welche durch die Untersuchung aufgezeigt werden sollen, im Vorfeld nicht oder nur unzureichend bekannt sind. Eine Ausnahme bilden selten durchgef\"{u}hrte \htsen, deren Ziel es ist, eine Hypothese zu pr\"{u}fen.

%


Da der bisherige Entwurf der \hts im Wesentlichen der Versuch ist, die manuellen Schritte der Bildauswertung automatisch nachzubilden, m\"{u}ssen nun f\"{u}r das neue Konzept die Anspr\"{u}che aus Biologie und Automatisierungstechnik aufeinander abgestimmt werden. F\"{u}r diese interdisziplin\"{a}re Aufgabe bedarf es einer Schnittstelle, mit deren Hilfe eine erfolgreiche Versuchsauslegung erfolgen kann. Die Schnittstelle sind die Einflussgr\"{o}{\ss}en auf die \"{u}bergeordneten Bl\"{o}cke "`Biologie"' und "`Bildakquise"' sowie "`Analyse und Interpretation"'. Die entsprechenden Parameter sind demnach die zu bestimmenden Gr\"{o}{\ss}en bei der Versuchsauslegung. Zwar muss die generelle Struktur der \hts individuell festgelegt werden, denn die in den Grundlagen der vorliegenden Arbeit bereits aufgezeigte gro{\ss}e Bandbreite an m\"{o}glichen biologischen, toxikologischen und genetischen Untersuchungen schlie{\ss}t eine universell anwendbare L\"{o}sung aus. Dennoch l\"{a}sst sich jede individuell generierte Struktur anhand der Versuchsparameter und Auswerteparameter auslegen und optimieren.

Versuchsparameter fassen hierbei alle Einflussgr\"{o}{\ss}en aus Bio\-logie und Bild\-akquise zusammen, w\"{a}hrend Auswerteparameter alle Einflussgr\"{o}{\ss}en beschreiben, die bei der Ana\-lyse~und~Inter\-preta\-tion festzulegen sind.
Die Literatur zeigt, dass eine gewissenhafte Auslegung der Parameter den Erfolg der Untersuchung zwar nicht garantiert, wohl aber wahrscheinlich macht und somit notwendige Voraussetzung f\"{u}r das Gelingen der \hts ist \cite{Goldsmith04,Patton01,Shariff10}. Eine mangelhaft ausgef\"{u}hrte Planung kann zum Scheitern der Untersuchung f\"{u}hren oder dazu zwingen, den gesamten Versuch wiederholen zu m\"{u}ssen. Da die Durchf\"{u}hrung der Untersuchung, je nach Anzahl und Aufwand der Einzelversuche, bis zu mehreren Monaten dauert, ist es sinnvoll, jeden Schritt sorgf\"{a}ltig in Vorversuchen zu pr\"{u}fen und zu optimieren.

Die Auswirkung einer Anpassung einzelner Parameter zeigt sich meistens erst nach dem Vollzug der vollst\"{a}ndigen Prozesskette, daher ist es notwendig, die Auswahl der Parameter iterativ vorzunehmen und in Vorversuchen die gesamte Kette mehrfach zu durchlaufen. Eine \"{A}nderung an einer beliebigen Stelle hat Auswirkungen auf viele, wenn nicht alle Teilprozesse. So kann beispielsweise eine ge\"{a}nderte Behandlung der Fischeier oder die Verwendung einer Mutation der Fischeier (\zB mit einem fluoreszierenden Marker) \"{A}nderungen bei der Wahl der  Mikroskopie"~ bzw. Akquise-Techniken erfordern. Eine solche \"{A}nderung f\"{u}hrt wiederum zu einer \"{A}nderung der Auswerteparameter der Analyse, deren Ergebnisse entsprechend in Interpretation und Evaluation neu dargestellt werden m\"{u}ssen.

Zu Beginn wird die Struktur der \hts geplant, auf deren Basis Vorversuche durchgef\"{u}hrt werden. Ziel ist es, schrittweise eine robuste Zuordnung des \Nutzsigs zu einer Klassen"~ oder Trefferzuweisung zu erhalten, die Anzahl der Einzelversuche zu h\"{o}heren Werten hin zu skalieren und gleichzeitig den Aufwand pro Einzelversuch so gering wie m\"{o}glich zu halten. Nach vollzogener Versuchsauslegung soll in einem Versuchsplan feststehen, welche Versuche mit welchen Behandlungen durchzuf\"{u}hren sind, mit welcher Methode die Nutzinformation akquiriert wird und auf welche Art die akquirierten Daten verarbeitet und ausgewertet werden.
Das hier vorgestellte Konzept zur Versuchsauslegung beschr\"{a}nkt sich allerdings auf die \emph{technische} Realisierbarkeit von \htsen und deren Optimierung. F\"{u}r weitere Details zur \emph{statistischen} Versuchsplanung, der Erstellung einer Rahmenvorschrift zur praktischen Durchf\"{u}hrung der Einzelversuche und der Minimierung und Optimierung der Anzahl der Versuche sei auf die Fachliteratur zur statistischen Versuchsplanung verwiesen \cite{Klein07,Montgomery05,Rasch07,Ruxton11}.

\begin{figure}[htbp]
        \centering
                \includegraphics[page=9,
                width=\linewidth
                ]{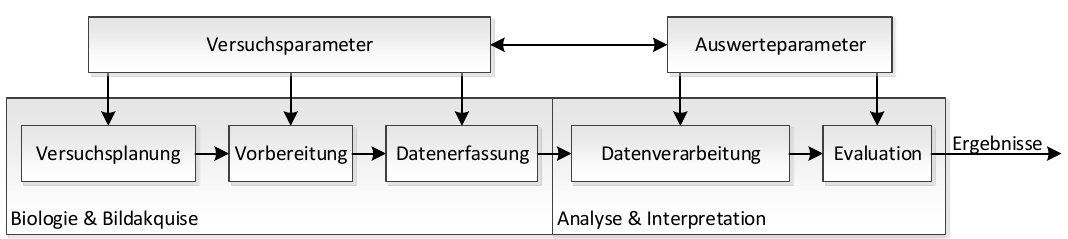}
\caption[Neues Konzept zur Optimierung der Versuchsauslegung]{Neues Konzept zur Optimierung der Versuchsauslegung durch Anpassung der Parameter der Prozesskette. }
\label{fig:Assay_Gestaltung}
\end{figure}

Eine bildbasierte \hts besteht \"{u}blicherweise aus zwei aufeinander folgenden Teilschritten (vgl. \abb \ref{fig:Assay_Gestaltung}): Der erste Teil, die "`Biologie \& Bildakquise"', setzt sich zusammen aus einer "`Versuchsplanung"', gefolgt von der "`Vorbereitung"' der Proben und der "`Datenerfassung"'. Der zweite Teil, die "`Analyse \& Interpretation"', besteht aus der "`Datenverarbeitung"' sowie deren "`Evaluation"'. Das vorgestellte Konzept ermittelt die Parameter aller in \abb \ref{fig:Assay_Gestaltung} dargestellten Bl\"{o}cke. Der Informations"~ \bzw Datenfluss verl\"{a}uft von links nach rechts, w\"{a}hrend die Parameter auf alle Bl\"{o}cke Einfluss nehmen und daher \"{u}bergeordnet sind. Wichtig hierbei ist die Einflussnahme und Ber\"{u}cksichtigung der Auswerte- und Versuchsparameter aufeinander, was durch den Doppelpfeil zwischen den Parameterbl\"{o}cken visuell verdeutlicht wird. 

Die Versuchsplanung im Block "`Biologie \& Bildakquise"' steht f\"{u}r die statistische Versuchsplanung. Das Vorgehen bei der statistischen Versuchsplanung sollte sich an der entsprechenden Fachliteratur orientieren und die Prinzipien der Wiederholung, Randomisierung und Blockbildung ber\"{u}cksichtigen \cite{Bandemer77,Klein07,Montgomery05,Rasch07,Ruxton11}. Ergebnis des Blockes "`Versuchsplanung"' ist ein statistisch optimaler Versuchsplan, welcher die Anzahl der Einzelversuche und deren jeweilige Behandlung festlegt und optimiert. F\"{u}r das vorgestellte Konzept sind die innerhalb der statistischen Versuchsplanung identifizierten St\"{o}rfaktoren von Bedeutung.
Der Block "`Vorbereitung"' bezieht sich auf alle Pr\"{a}parationen, die dazu beitragen, eine maximale Qualit\"{a}t der Datenakquise zu erzielen. Bei biologischen Versuchen besteht oftmals die M\"{o}glichkeit, eine Versuchseinheit derart vorzubereiten, dass die Nutzinformation besser erkennbar wird. Ein Beispiel ist die Bet\"{a}ubung und Ausrichtung der Zebrab\"{a}rblingslarven. Unterschiedlich aufw\"{a}ndige Pr\"{a}parationsschritte bieten so die M\"{o}glichkeit, die Bildauswertung zu vereinfachen. In manchen Versuchen ist das Ausrichten sogar zwingend erforderlich, um das \Nutzsig erfassen zu k\"{o}nnen. Werden die Larven bei der Pr\"{a}paration \zB durch speziell geformte Platten immer lateral ausgerichtet, k\"{o}nnen Bereiche im Zebrab\"{a}rbling leichter anhand von geometrischen Beziehungen gefunden werden \cite{Gehrig09}. Versuchsparameter sind somit alle gew\"{a}hlten Vorbereitungsschritte und St\"{o}rfaktoren, \zB die gew\"{a}hlte Position der Versuchseinheit in der Mikrotiterplatte.

Die Datenerfassung zeichnet mittels der gew\"{a}hlten Akquise-Methode einen Bildstrom auf. Je detaillierter der aufgenommene Bildstrom ist, desto mehr (Nutz)Informationen enth\"{a}lt er. Allerdings steigt die Gr\"{o}{\ss}e des Datensatzes. Der Anteil an "`unn\"{o}tigen"', d.h. Hintergrundinformationen, w\"{a}chst meist \"{u}berproportional, was die Gefahr birgt, dass die Daten allein aufgrund der Datenmenge schwer zu verarbeiten und zu archivieren sind. Auch der Aufwand f\"{u}r die Extraktion des \Nutzsigs steigt. Vor allem bei neuartigen mehrdimensionalen Mikroskopietechniken wie etwa der SPIM-Mikroskopie (vgl. Abschnitt \ref{subsec:Grundlagen_Bio}) fallen immens gro{\ss}e Datenmengen von mehreren Gigabyte (GB) oder sogar Terabyte (TB) pro Versuchseinheit an.
Mit Abschluss der Datenerfassung ist der \"{u}bergeordnete Block "`Biologie \& Datenakquise"' abgeschlossen und alle Versuchsparameter sind bestimmt. In Form des akquirierten Bildstroms steht nun eine Messung des \Nutzsigs zur Verf\"{u}gung. Dem Nutzsignal jeder Versuchseinheit einer Klasse \bzw einer Trefferfunktion zuzuordnen, ist Aufgabe des \"{u}bergeordneten Blocks "`Analyse \& Interpretation"', welcher sich wiederum weiter in die Schritte "`Datenverarbeitung"' und  "`Evaluation"' unterteilen l\"{a}sst.

Die Datenverarbeitung trennt das \Nutzsig von unwichtigen Informationen, \dhe von der Hintergrundinformation. Sie extrahiert Merkmale, welche schlie{\ss}lich der Klassifikation zugef\"{u}hrt werden. Der Prozess schlie{\ss}t die Bildverarbeitung mit ein und wird in der vorliegenden Arbeit detailliert betrachtet (\vgl Kapitel \ref{sec:BV_Module}). Das Ergebnis der Klassen"~ oder Trefferzuordnung wird in m\"{o}glichst \"{u}bersichtlicher Form dargestellt, sodass biologische Schl\"{u}sse gezogen werden k\"{o}nnen. Dies geschieht im Block "`Evaluation"'. Aus der automatischen Klassifikation muss der Biologe manuell eine Schlussfolgerung ziehen. Hierbei ist eine \"{u}bersichtliche Pr\"{a}sentation der Versuchsergebnisse hilfreich. Die anhand der Pr\"{a}sentation getroffene Deutung des Biologen soll die Fragestellung der \hts beantworten. In den Vorversuchen muss nach der ersten Evaluation anhand der vorl\"{a}ufigen Ergebnisse abgewogen werden, ob die Qualit\"{a}t der Ergebnisse ausreichend ist, um den Versuch auf die finale Gr\"{o}{\ss}e zu skalieren, \dhe ob mit dem gew\"{a}hlten Verfahren die gew\"{u}nschten Daten aussagekr\"{a}ftig abgebildet werden k\"{o}nnen und ob das gesuchte \Nutzsig in den Daten stark genug vorhanden ist. Ist dies der Fall, wird mit der tats\"{a}chlichen Durchf\"{u}hrung des vollst\"{a}ndigen Versuchs begonnen.

Mit dem dargestellten Konzept steht nun eine einheitliche Vorgehensweise und Schnittstelle f\"{u}r die Versuchsauswertung zur Verf\"{u}gung.

\section{Anforderungen an bildbasierte \htsen}\label{sec:HDU_Anforderungen}

Die \htsen m\"{u}ssen vielschichtigen Anforderungen gerecht werden. Die Anforderungen werden hier in strukturierter Form herausgearbeitet. Anforderungen sind je nach Versuchsvorgaben quantitativ (\zB Anzahl der zu untersuchenden Substanzen) oder qualitativ (\zB Sichtbarkeit des biologischen Effekts). W\"{a}hrend f\"{u}r quantitative Anforderungen klare Angaben gemacht werden k\"{o}nnen, lassen sich f\"{u}r qualitative Anforderungen keine genau messbaren Grenzwerte angeben, da \zB stets ein m\"{o}glichst gut sichtbarer biologischer Effekt oder eine geringe Fehlerrate anzustreben sind.

F\"{u}r \htsen lassen sich Ziele und Vorgaben allgemein bzw. in Abh\"{a}ngigkeit des konkreten biologischen Versuchs formulieren. Der Versuch spielt hierbei f\"{u}r die Anzahl \bzw Komplexit\"{a}t der Anforderungen die ma{\ss}gebende Rolle, daher sind nicht immer alle Anforderungen f\"{u}r den Einzelfall unbedingt erforderlich. Um den erreichbaren Durchsatz sicherzustellen, ist die Automatisierbarkeit aller notwendigen Versuchsschritte \"{u}bergeordnet. Zus\"{a}tzlich wurden in der vorliegenden Arbeit folgende Anforderungen f\"{u}r \htsen identifiziert:

%

\begin{enumerate}
\item Durchf\"{u}hrung
    \begin{itemize}

        \item Legitimit\"{a}t

        Die Untersuchung muss den gesetzlichen Vorschriften sowie den ethischen Prinzipien der jeweiligen Forschungseinrichtung entsprechen. Es sind beispielsweise Vorgaben zur Genmanipulation, dem Umgang mit toxikologischen Substanzen und Tierversuchen zu beachten.

         \item Logistische Machbarkeit

        Da es sich bei den Versuchen mit Modellorganismen um sich entwickelnde und damit ver\"{a}nderliche Proben handelt, m\"{u}ssen vergleichbare Daten auch zu vergleichbaren Entwicklungs-Zeitpunkten aufgenommen werden. Die Forderung ist von logistischer und biologischer Bedeutung, da f\"{u}r einen gew\"{u}nschten Durchsatz der entsprechende Nachschub an Fischeiern gegeben sein muss. Die meisten Arbeitsschritte m\"{u}ssen weitgehend parallel erfolgen. Da die Eier jedoch immer zu einem bestimmten Zeitpunkt, dem Sonnenaufgang, gelegt werden und somit die Entwicklung immer gleich startet, ist das Zeitfenster f\"{u}r die Bildakquise vergleichbarer Daten an einem Tag klein.
      \item Biologische Realisierbarkeit

            Die zu untersuchende Fragestellung muss sich am Zebrab\"{a}rbling \bzw dem gew\"{a}hlten Modellorganismus durchf\"{u}hren lassen.

      \item Reproduzierbarkeit

            Ein erneutes Durchf\"{u}hren identischer Versuche muss zum qualitativ gleichen Ergebnis f\"{u}hren.

    \item Finanzielle Realisierbarkeit

            Durch die hohe Anzahl an Einzeluntersuchungen entstehen hohe laufende Kosten f\"{u}r die Verbrauchsmaterialien und f\"{u}r die zu untersuchenden Substanzen. Hinzu kommen Kosten f\"{u}r \zT hochaufl\"{o}sende Mikroskope, robotergest\"{u}tzte Automatisierungstechniken und Computertechnik zur Bildverarbeitung sowie Archivierung. In der Summe \"{u}bersteigen solche Kosten schnell das Budget kleinerer Forschungseinrichtungen.

    \end{itemize}

\item Messung
  \begin{itemize}
    \item Eindeutige Pr\"{a}senz der biologisch relevanten Information

         Die grundlegende Forderung der eindeutigen Pr\"{a}senz ist bei vielen Versuchen nicht erf\"{u}llt. So k\"{o}nnen Biologen oftmals aus Erfahrung und anhand von kleinen Hinweisen Dinge (etwa biologische Gewebe, Ph\"{a}notypen \etc) in Organismen erkennen, welche f\"{u}r eine weniger ge\"{u}bte Person und auch f\"{u}r die automatische Datenverarbeitung in den Daten nicht zu identifizieren sind. In einem solchen Fall ist mit automatischen Verfahren keine robuste Auswertung m\"{o}glich.

    \item R\"{u}ckwirkungsfreiheit der Messung

        Die Akquise darf die Ergebnisse nicht wesentlich beeinflussen. Manche Akquise-Techniken erfordern beispielsweise sehr lichtstarke Beleuchtungen oder bei der konfokalen  Mikroskopie sogar Laser zur Akquise der Daten. Biologische Proben k\"{o}nnen \zB bei Verhaltensuntersuchungen auf Licht reagieren oder fluoreszierende Marker k\"{o}nnen durch starke Beleuchtung ausgeblichen werden. Daher muss vor Beginn der Untersuchung der Einfluss der Akquise-Methode auf das \Nutzsig ber\"{u}cksichtigt werden.

    \item Schnelligkeit der Messung

      Die Akquise muss in ad\"{a}quater Zeit erfolgen.
      Da aus den \og logistischen Einschr\"{a}nkungen der Durchsatz deutlich unter der theoretischen Kapazit\"{a}t automatischer Mikroskope liegt, muss gepr\"{u}ft werden, ob die Fragestellung auch in der gew\"{u}nschten Zeit realisierbar ist. Die Realisierbarkeit ist abh\"{a}ngig von den zur Verf\"{u}gung stehenden menschlichen und technischen Ressourcen.
  \end{itemize}

\item Auswertung
  \begin{itemize}
    \item Segmentierbarkeit des biologischen Effekts im Bild

           Der biologische Effekt muss nach der Messung klar in den aufgezeichneten Bildern oder dem aufgezeichneten Bildstrom vorhanden sein, sodass eine Segmentierung, \dhe Abgrenzung des biologischen Effekts von anderen Bildinhalten, m\"{o}glich ist.

        \item Quantifizierbarkeit des biologischen Effekts im Bild

        Der biologische Effekt kann beispielsweise ein Ph\"{a}notyp sein, welcher nur selten unter Tausenden von Untersuchungen vorkommt, oder eine Statistik, die aus einer hohen St\"{u}ckzahl an Bildern extrahiert wird. Das Nutzsignal muss sich durch Zahlenwerte, die aus den Bildern ermittelt werden, repr\"{a}sentieren lassen, \dhe die Bildauswertung muss die relevanten Informationen extrahieren und die Merkmalsextraktion der Bildauswertung muss so konzipiert sein, dass sich das biologische \Nutzsig aus den Merkmalen klassifizieren l\"{a}sst. Auch in der gro{\ss}en Anzahl an Einzeluntersuchungen m\"{u}ssen selten vorkommende Ereignisse erkennbar sein und auch erkannt werden.

    \item Niedrige Fehlerrate bei der Detektion des biologischen Effekts

            Die Bildauswertung muss robust sein, \dhe Fehlklassifikationen durch Bildfehler aus Akquise, Pr\"{a}paration, Robotik \etc sollen ebenso wie ein kritischer Abbruch der Bildverarbeitungsroutine vermieden werden.

    \item Schnelligkeit der Auswertung

          Die Bildauswertung muss in ad\"{a}quater Zeit erfolgen, da je nach biologischer Fragestellung \uU Echtzeitf\"{a}higkeit gefordert werden muss. Zumeist ist jedoch eine Rechenzeit von einigen Minuten pro Einzelexperiment vertretbar, solange das Gesamtexperiment in einem angemessenen Zeitraum (\"{u}blicherweise zwischen einigen Stunden und Wochen) auswertbar ist.

    \item Pr\"{a}sentierbarkeit der Auswertung

          Die Ergebnisse m\"{u}ssen nach der Durchf\"{u}hrung und Auswertung der \hts dargestellt werden. Die hohe Anzahl von Einzeluntersuchungen macht eine \"{u}bersichtliche Art und Weise der Darstellung notwendig. Die Klassifikationsergebnisse und die \ggf identifizierten \emph{Treffer} (selten vorkommende Ereignisse, auf welche die \hts ausgerichtet sein kann) m\"{u}ssen visualisiert werden und den Versuch in \"{u}bersichtlicher Weise darstellen, sodass von Biologen weitere Schl\"{u}sse gezogen werden k\"{o}nnen.

    \item Wissenschaftliche Archivierung

          Um die Reproduzierbarkeit zu sichern, m\"{u}ssen alle Versuche inkl. Versuchsparameter, akquirierten Bildern, verwendeter Software, deren Version und der Visualisierung archiviert werden. Die Archivierung schlie{\ss}t sowohl die Bildverarbeitungsroutine als auch den Zugriff auf alle Auswertungs"~ und Visualisierungsroutinen ein.
          Bei mehrdimensionalen Untersuchungen im Hochdurchsatz fallen sehr hohe Datenmengen an. Die Daten m\"{u}ssen bei den meisten biologischen Untersuchungen f\"{u}r mindestens 10 Jahre archiviert werden \cite{Stotzka11}. Die zur Verf\"{u}gung stehende Archivierungseinrichtung muss daher der Datenmenge und Dauer der Archivierung entsprechend ausgelegt sein.
  \end{itemize}
\end{enumerate}


\newpage

\section{Identifikation der Versuchs"~ und Auswerteparameter}\label{sec:Beschreibung_Parameter}


Zur konkreten Anwendung des im vorangegangenen Abschnitt eingef\"{u}hrten Konzeptes m\"{u}ssen in allen Prozessschritten (vgl. \abb \ref{fig:Assay_Gestaltung}) Entscheidungen getroffen werden, durch die Versuchs- und Auswerteparameter bestimmt werden. Die Versuchsparameter bilden zusammen mit der Anweisung zur praktischen Durchf\"{u}hrung aus der statistischen Versuchsplanung den Versuchsplan der \hts. Der folgende Abschnitt leistet einen Beitrag zur Entscheidungsfindung f\"{u}r die technische Durchf\"{u}hrung und zeigt wichtige Eigenschaften und gegenseitige Abh\"{a}ngigkeiten zwischen den einzelnen Versuchs\-parametern und Entscheidungen auf. Die gew\"{a}hlten Optionen haben unmittelbare Auswirkungen auf die Aussagen und den Durchsatz, der mittels einer \hts erzielt werden kann. Zum Teil stehen f\"{u}r ein beabsichtigtes Ergebnis bzw. eine bestimmte biologische Fragestellung und somit ein gew\"{u}nschtes \Nutzsig eine Reihe an Entscheidungen aus Gr\"{u}nden der Realisierbarkeit bereits fest. Als einf\"{u}hrendes Beispiel sei ein Zelltracking einer einzelnen Zelle angef\"{u}hrt. Eine solche Aufgabenstellung ist in einer Hellfeld-Aufnahme schwer realisierbar, da die \NutzsigInf (die Zelle von Interesse) schwer von benachbarten Zellen zu unterscheiden ist. Hier bietet sich das Markieren der Zellkerne durch fluoreszierende Proteine an, da sich diese, bei Anregung mit der entsprechenden Wellenl\"{a}nge separat, ohne Auftreten der nicht markierten Zellen, aufzeichnen lassen. Ein auf morphologischen Formen basierendes \Nutzsig ist dagegen nur mit Hellfeld-Aufnahmen abbildbar. Die beiden Beispiele zeigen die Notwendigkeit auf, die Akquise-Methode so zu w\"{a}hlen, dass das \Nutzsig abgebildet wird. Im Folgenden werden  Vor- und Nachteile der Entscheidungsm\"{o}glichkeiten diskutiert und anhand von Beispielen die Folgen verdeutlicht. Manche der Entscheidungen schlie{\ss}en sich im \"{U}brigen nicht aus, sondern es k\"{o}nnen auch zwei oder mehrere Entscheidungen gleichzeitig getroffen werden (im vorigen Beispiel das gleichzeitige Aufzeichnen von Hellfeld"~ und Fluoreszenzaufnahmen und die sp\"{a}tere Zuordnung zu biologischen Signalen). Ein solches Vorgehen ist f\"{u}r biologische Fragestellungen zum Teil sogar unmittelbar erforderlich und hat erheblichen Einfluss auf den f\"{u}r die Bildakquise bzw. Bildauswertung notwendigen Aufwand und die zu akquirierende  Datenmenge.

\begin{figure}[!htbp]
        \centering
        \includegraphics[page=7]{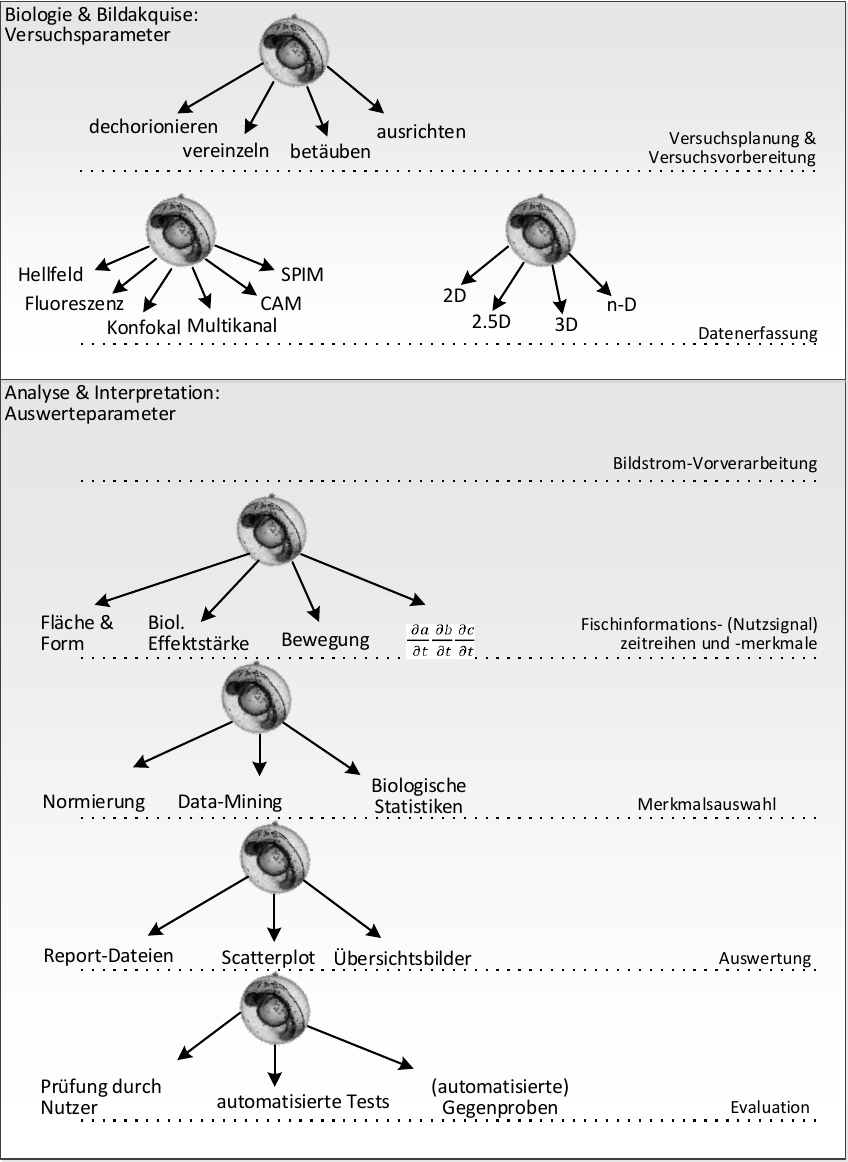}
        \caption{Entscheidungsm\"{o}glichkeiten und Ablauf einer bildbasierten \hts}
        \label{fig:Entscheidungsbaum}

\end{figure}

 In \abb \ref{fig:Entscheidungsbaum} ist eine \"{U}bersicht der wichtigsten Entscheidungsm\"{o}glichkeiten zur Bestimmung der Versuchsparameter dargestellt. Die \"{U}bersicht ist nach dem Schema des Konzeptes aus Abschnitt \ref{sec:Konzept_Versuchsauslegung} gegliedert und die bereits eingef\"{u}hrten, \"{u}bergeordneten Bl\"{o}cke "`Biologie \& Bildakquise"' sowie "`Analyse \& Interpretation"' lassen sich abgrenzen.

\subsection{Biologie und Bildakquise}\label{subsec:Bio_u_Bildakquise}
\begin{figure}[!htbp]
        \centering
                \includegraphics[width=3cm]{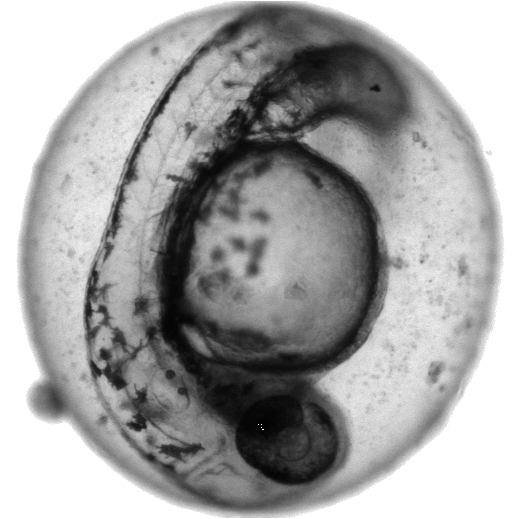} \hspace{2cm}
        \includegraphics[width=6cm]{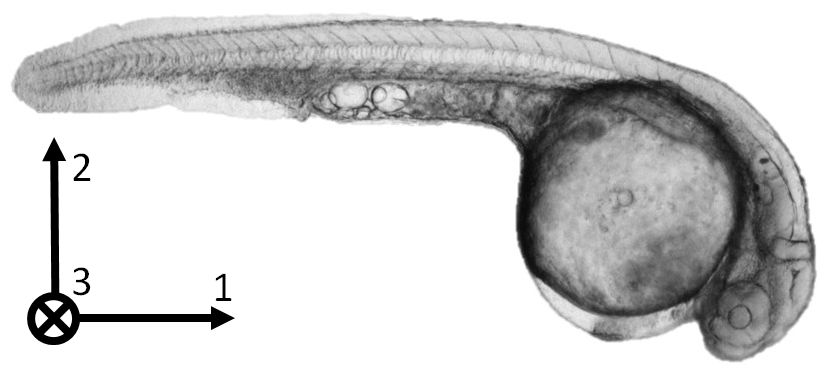}
        \caption[Larve eines Zebrab\"{a}rblings im Chorion und nach dem Dechorionieren]{Larve eines Zebrab\"{a}rblings im Chorion (links) und nach dem Dechorionieren (rechts), \dhe dem Entfernen der Fruchth\"{u}lle}
        \label{fig:dechorionieren_1}
    \end{figure}

Im Biologie"~ und Bildakquiseteil m\"{u}ssen die Versuchsparameter bestimmt werden.
F\"{u}r eine gute Qualit\"{a}t bei der Datenerfassung bietet es sich an, w\"{a}hrend der Versuchsvorbereitung die Larven zu pr\"{a}parieren. Der Aufwand f\"{u}r das Pr\"{a}parieren f\"{a}llt unterschiedlich hoch aus, ist jedoch in fast allen F\"{a}llen manuell durchzuf\"{u}hren, limitiert somit den Durchsatz und setzt ein entsprechendes Personal voraus. Die wichtigsten Pr\"{a}parationsschritte sind das Dechorionieren, das Vereinzeln, das Ausrichten und das Bet\"{a}uben der Larven. Das Dechorionieren ist das Befreien der Larven von der umschlie{\ss}enden Fruchth\"{u}lle (dem \sog Chorion). Dabei wird die Fruchth\"{u}lle entweder mit einer Pinzette manuell entfernt oder mittels einer Chemikalie aufgel\"{o}st. Bei Verwendung der Chemikalie lassen sich viele Larven gleichzeitig dechorionieren. \abb \ref{fig:dechorionieren_1} stellt einer Larve im Chorion eine dechorionierte Larve gegen\"{u}ber.

Der Vorteil der Pr\"{a}paration ist, dass sich die Larven nach dem Dechorionieren ausstrecken und hierdurch alle seitlich sichtbaren Bereiche akquirierbar sind. Bis zu einem Alter von 72\,hpf\footnote{hpf = \emph{h}ours \emph{p}ast \emph{f}ertilization = Stunden nach der Befruchtung} liegt die Larve zudem definiert entweder in lateraler oder dorsaler Lage\footnote{Zu einem sp\"{a}teren Zeitpunkt richtet sich die Larve auf, da sich die Schwimmblase entfaltet hat \cite{Westerfield93}.}. Durch die runde Form des Dottersacks kann die Larve allerdings leicht in eine beliebige Richtung gekippt sein, was zur Folge hat, dass die absolute H\"{o}he von Bereichen im Fisch variiert. Da es technisch nicht m\"{o}glich ist, die gesamte Larve, auch in der Tiefe, mit einer Aufnahme scharf abzubilden, m\"{u}ssen auch dechorionierte Larven in mehreren Fokusebenen akquiriert werden. F\"{u}r die Bildauswertung ist die definierte seitliche Position allerdings von gro{\ss}em Vorteil, da sich ein Koordinatensystem definieren l\"{a}sst und geometrische Beziehungen zum Auffinden von Bereichen im Fisch ableitbar sind \cite{Gehrig09}, \dhe die Larven m\"{u}ssen lediglich um die dritte Achse rotiert und \ggf gespiegelt werden, um eine vergleichbare Position zu erhalten (vgl. \abb \ref{fig:dechorionieren_1}). Den genannten Vorteilen steht der gro{\ss}e zeitliche Aufwand f\"{u}r die Entfernung des Chorions und f\"{u}r das Ausrichten auf der Seite gegen\"{u}ber. Dies gilt im Besonderen, wenn auf die Dechorionierung mittels Chemikalien aufgrund von m\"{o}glichen wechselseitigen Abh\"{a}ngigkeiten zwischen den Substanzen, die bei dem Versuch zum Einsatz kommen, verzichtet und somit das Dechorionieren f\"{u}r jede Larve einzeln und manuell durchgef\"{u}hrt werden muss. Zudem ist die Vereinzelung von dechorionierten Zebrab\"{a}rblingen schwieriger, da die Larven bereits "`stromlinienf\"{o}rmig"' sind und das exakte Pipettieren und vor allem dessen Automatisierung vor Probleme stellt.

Die Vorteile der Zebrab\"{a}rblingslarven im Chorion sind u.a. die leichtere Automatisierbarkeit sowie die kurze Pr\"{a}parationszeit. Bis eine robuste robotergest\"{u}tzte L\"{o}sung f\"{u}r die \og Pr\"{a}paration entwickelt ist, kann f\"{u}r \htsen lediglich die Larve im Chorion empfohlen werden, denn bei manueller Pr\"{a}paration l\"{a}sst sich der geforderte Durchsatz von mehreren tausend Einzelversuchen pro Tag nicht erreichen, was f\"{u}r die Bildverarbeitung jedoch den Nachteil, dass die Larven im Chorion frei beweglich sind. Somit ist deren Lage auf den akquirierten Bildern \"{a}u{\ss}erst unterschiedlich. Als Vorteil der Zebrab\"{a}rblingslarve als Modellorganismus wird zwar vor allem Transparenz angef\"{u}hrt, was jedoch nicht bedeutet, dass alle Seiten der Larve gleich gut auf dem Bild sichtbar sind. Selbst kontraststarke Bereiche wie das Auge sind, wenn sie verdeckt auftreten, zwar durch die Larve hindurch noch zu erahnen, die Konturen werden von den dar\"{u}ber liegenden K\"{o}rperteilen (z.B. Kopf oder Dottersack) jedoch stark verwischt oder abgeschw\"{a}cht. Je nach Lage der Larve sind somit verschiedene Details im Zebrab\"{a}rbling mit Chorion nicht oder nur schwer zu erkennen. Damit lassen sich hier lediglich von der Lage unabh\"{a}ngige Nutzsignale untersuchen.

Die letzte der hier aufgezeigten Pr\"{a}parationsm\"{o}glichkeiten ist das Bet\"{a}uben der  Zebrab\"{a}rblingslarven. Ab \ca 24\,hpf beginnen die Larven, sich spontan in ihrer Eih\"{u}lle zu bewegen. Die H\"{a}ufigkeit dieser spontanen Bewegungen schwankt w\"{a}hrend der Entwicklung der Larve. Im Mittel kann jedoch alle drei Sekunden eine Bewegung beobachtet werden, was zur Folge hat, dass, wenn die Bewegung w\"{a}hrend der Bildakquise auftritt, verwaschene Konturen in der Mikroskopaufnahme erscheinen. Solche Aufnahmen sind meist f\"{u}r die weitere Auswertung unbrauchbar. Bei Aufnahmetechniken wie etwa der Konfokal-Mikroskopie oder SPIM sind lange Belichtungszeiten pro Aufnahme ohne Bewegung erforderlich. Soll \"{u}ber eine l\"{a}ngere Zeit eine Zuordnung von Bereichen in der Larve m\"{o}glich sein, m\"{u}ssen solche Bewegungen unterbunden werden. Dies kann durch ein Bet\"{a}ubungsmittel, \zB Trikaine, erreicht werden, hat allerdings einen manuellen Pr\"{a}parationsschritt mit \og Nachteilen zur Folge. Auch kann ein Zusammenwirken des An\"{a}sthetikums mit den Substanzen, deren Einfluss bestimmt werden soll, auftreten. Daher muss bei der Versuchsauslegung gepr\"{u}ft werden, ob die Verwendung von Bet\"{a}ubungsmitteln vertretbar ist. Eine alternative M\"{o}glichkeit zum Demobilisieren der Larven ist es, die Larven kurz vor der Bildakquise mit Hilfe von kaltem Wasser zu k\"{u}hlen \cite{zhang1995}.

Nachdem nach \abb \ref{fig:Entscheidungsbaum} eventuelle manuelle Pr\"{a}parationsschritte gew\"{a}hlt wurden und die Larven in definierter Anzahl (\"{u}blicherweise 1-10) in Mikrotiterplatten platziert und \ggf mit Chemikalien exponiert sind, muss nun f\"{u}r das \Nutzsig eine geeignete Akquise-Technik  gew\"{a}hlt werden (\vgl \kap \ref{subsec:Bildakquise+Bildverarbeitung}). F\"{u}r jede Zebrab\"{a}rblingslarve wird nun eine Messung mittels Detektoren erfolgen. Unabh\"{a}ngig von den gew\"{a}hlten Akquise-Techniken wird bei den aufgezeichneten Daten immer von einem Bildstrom gesprochen. Je nach Protokoll der Aufzeichnung handelt es sich bei den Daten um ein einzelnes Bild, eine Bildsequenz, oder mehrere Bildsequenzen innerhalb einer l\"{a}ngeren Zeitspanne. Es m\"{u}ssen also, ebenfalls unabh\"{a}ngig von der gew\"{a}hlten Akquise-Technik und angepasst an das \Nutzsig, die Aufnahmezeitpunkte und deren Frequenz festgelegt werden. Daf\"{u}r muss gekl\"{a}rt werden, ob die \NutzsigInf nur selten auftritt, unter bestimmten Voraussetzungen erkennbar ist oder durch geeignete Mittel provoziert werden kann. Dies hat direkten Einfluss auf Anzahl der Wiederholungen sowie Dauer und Frequenz der Akquise pro Einzelversuch. Im einfachsten Fall ist der Zeitpunkt der Akquise bzw. das Zeitfenster f\"{u}r vergleichbare Ergebnisse gro{\ss} (mehrere Stunden bis Tage) und der Effekt immer sichtbar. Ist das nicht der Fall, \zB wenn ein Entwicklungsvorgang der Larven oder von Zellen untersucht wird, so sind Bildsequenzen vonn\"{o}ten, deren Abtastraten an die Entwicklungsgeschwindigkeit anzupassen sind. Die steigende Akquise-Zeit der Einzelversuche hat Leerlaufzeiten \bzw Wartezeiten der Mikroskope zur Folge. Solche Zeiten des Stillstandes schr\"{a}nken die erreichbare Anzahl an Einzeluntersuchungen pro Tag ein, da die Mikroskope l\"{a}nger durch die Bildakquise je Einzelversuch blockiert werden und durch die geforderte Vergleichbarkeit der Einzelversuche und die schnelle Entwicklung der Larven nur ein kleines Zeitfenster zur Akquise existiert. Die Auftrittsh\"{a}ufigkeit hat daher \"{u}berproportionalen Einfluss auf den erreichbaren Durchsatz.

Bei der Datenerfassung hat die Wahl der Akquise-Technik, wie \zB die Pr\"{a}paration, unterschiedliche, weitreichende Auswirkungen auf die Bildauswertung. Entsprechend \abb \ref{fig:Entscheidungsbaum} bieten sich verschiedene Optionen, welche die Modalit\"{a}t des Bildstroms bestimmen. Die Akquise-Methode ist dem Auftreten des  \Nutzsigs anzupassen. Es muss in jedem Fall sichergestellt werden, dass der Bildstrom das \Nutzsig und damit die \NutzsigInf abbildet.

Ist die auszuwertende Beobachtung beispielsweise in Hellfeld-Aufnahmen und in Gewebefl\"{a}chen der Larve zu finden, so ist eine einzelne Mikroskopaufnahme bereits ausreichend, um die weiteren Schritte erfolgreich abzuschlie{\ss}en. Ist die Aufnahme ein Grauwertbild, so bleibt die vom Detektor aufgezeichnete Matrix zweidimensional. F\"{u}r Farbbilder sind bereits drei Kan\"{a}le (f\"{u}r rot, gr\"{u}n und blau) aufzuzeichnen. Tritt der biologische Effekt lediglich unter Verwendung von Fluoreszenzmikroskopie in Erscheinung, so kann ebenfalls ein zweidimensionales Bild akquiriert werden. Oft ist jedoch die Zuordnung des fluoreszierenden Bereichs im Hellfeld-Kanal gew\"{u}nscht, was die Dimension der Akquise mindestens um einen Grad erh\"{o}ht. Zu Beginn der Akquise der \hts werden somit Modalit\"{a}t \bzw die Anzahl der Kan\"{a}le des Bildstroms festgelegt. Hierbei stehen die in der Einleitung (Abschnitt \ref{subsec:Bildakquise+Bildverarbeitung}) beschriebenen M\"{o}glichkeiten zur Verf\"{u}gung.

F\"{u}r eine \hts bieten sich in erster Linie Hellfeld-Aufnahmen an, da solche traditionell der Standard in der Mikroskopie und daher weit verbreitet sind. Des Weiteren steht eine gro{\ss}e Auswahl an (teil)automatisierten Mikroskopen zur Verf\"{u}gung. Sie k\"{o}nnen  zur oberfl\"{a}chlichen Abbildung fast aller Objekte verwendet werden. Es lassen sich der Zebrab\"{a}rbling als Ganzes oder auch Bereiche, wie \zB dessen einsehbare Organe akquirieren. Eine besondere Schwierigkeit ist das (automatische) Auffinden der Fokus\-ebene, welche das \Nutzsig optimal abbildet. Die optimale Sch\"{a}rfe der Zebrab\"{a}rblingslarven im Chorion ist unter mehreren Gesichtspunkten schwierig zu ermitteln. F\"{u}r jedes Mikroskop kann der scharf abgebildete Bereich, die sog. Sch\"{a}rfentiefe, aufgrund der physikalischen Gegebenheiten (Aufl\"{o}sung/Numerische Apertur) errechnet werden. Bei hinreichend ebenen Objekten wird daher der gesamte Bereich scharf abgebildet. Da das Chorion jedoch eine relativ gro{\ss}e Ausdehnung in der Tiefe hat, l\"{a}sst sich in jeder Fokus\-ebene nur ein kleiner Teil scharf darstellen. Zwar k\"{o}nnen z.B. immer die Mitte oder der obere Bereich scharf fokussiert werden, da die Lage der Larven jedoch variiert, ist immer ein anderer morphologischer Bereich der Larve scharf abgebildet. Um das \Nutzsig sp\"{a}ter vergleichen zu k\"{o}nnen, muss es jedoch auch m\"{o}glichst gleich akquiriert sein. Werden nun verschiedene z-Ebenen aufgenommen, sog. Stacks, muss dasjenige ausgew\"{a}hlt werden, welches die interessierende Region m\"{o}glichst scharf wiedergibt. Eine Vergr\"{o}{\ss}erung der Anzahl an Fokusebenen ergibt jedoch nur so lange Sinn, wie die Summe der Sch\"{a}rfentiefen in den Bildern nicht den Durchmesser des Chorions des Zebrab\"{a}rblings \"{u}berschreitet. Je nach Genauigkeit und Abgrenzbarkeit des \Nutzsigs in der Larve muss somit die Anzahl der Fokusebenen festgelegt werden.
Es entsteht eine \sog 2.5-dimensionale Aufnahme. Die Modalit\"{a}t ist hierbei Eins (\bzw Drei f\"{u}r Farbaufnahmen) und die Daten sind dreidimensional.
Werden Entwicklungsverl\"{a}ufe oder Bewegungsmuster der Zebrab\"{a}rblinge ausgewertet, m\"{u}ssen eine oder mehrere Bildsequenzen aufgezeichnet werden. F\"{u}r jeden weiteren Abtastzeitpunkt multipliziert sich die Gr\"{o}{\ss}e des Bildstroms.

Wird die Fluoreszenzmikroskopie angewandt, lassen sich f\"{u}r eine \hts die Intensit\"{a}tswerte eines mittels Markers im Zebrab\"{a}rbling gekennzeichneten Bereichs akquirieren. Bei der Fluoreszenzmikroskopie ist die Selektivit\"{a}t der Aufzeichnung gleichzeitig Vor"~ und Nachteil. Ist die \NutzsigInf markiert, wird bei der Fluoreszenzmikroskopie lediglich die \NutzsigInf akquiriert. Dies hat f\"{u}r die Bildverarbeitung den Vorteil, dass die Region von Interesse nicht vom Hintergrund getrennt werden muss und auch vorliegt, wenn dessen Quelle im Inneren des Zebrab\"{a}rblings liegen sollte. Der Nachteil jedoch ist, dass in den akquirierten Daten keinerlei Information \"{u}ber die Position der Daten relativ zum Zebrab\"{a}rbling enthalten ist. Sind beispielsweise nur wenige Zellen markiert, enth\"{a}lt der Bildstrom keine Informationen, ob die Zellen \zB aus dem Kopf oder dem R\"{u}cken stammen. Bildsequenzen lassen sich mit der Konfokal-Mikroskopie nur mit relativ niedrigen Abtastfrequenzen und langen Aufnahmezeiten realisieren. Daher ist eine Bet\"{a}ubung der Larven erforderlich. Um die Fluoreszenzkan\"{a}le dennoch einem Bereich im Zebrab\"{a}rbling zuordnen zu k\"{o}nnen, besteht die M\"{o}glichkeit, sowohl Fluoreszenz- als auch Hellfeld-Aufnahmen von der gleichen Larve zum m\"{o}glichst gleichen Zeitpunkt unter m\"{o}glichst gleichen Bedingungen zu akquirieren.

Erweitern lassen sich die Multikanalaufnahmen durch das Hinzuf\"{u}gen weiterer Fluoreszenzkan\"{a}le, da bei einer \hts mehrere Zellen unterschiedlich markiert sein k\"{o}nnen.  Dies k\"{o}nnen sowohl Autofluoreszenzkan\"{a}le als auch markierte Kan\"{a}le unterschiedlicher Emissionswellenl\"{a}ngen sein.
Je akquirierter Fluoreszenzwellenl\"{a}nge ist ein gesonderter Kanal aufzuzeichnen und f\"{u}r eine Hellfeld-Aufnahme ein weiterer (\bzw drei f\"{u}r Farbaufnahmen). Bei der Akquise solcher Multikanalaufnahmen entstehen somit schnell gro{\ss}e Datenmengen, da sich hier die Datenmenge zus\"{a}tzlich zu den Abtastzeitpunkten nochmals mit der Kanalanzahl multipliziert. Es sei darauf hingewiesen, dass sich Zellen biologisch oft nicht sofort markieren lassen. Sog. transgene Linien, die sich f\"{u}r diesen Zweck eignen, m\"{u}ssen erst gez\"{u}chtet werden. Das Z\"{u}chten f\"{u}hrt zu einer Wartezeit eines oder mehrerer voller Generationszyklen der Fische. Zudem muss beachtet werden, dass f\"{u}r eine \hts eine entsprechend gro{\ss}e Population der transgenen Linie vorhanden sein muss, um den Nachschub an Eiern pro Tag zu gew\"{a}hrleisten, was zu weiteren Wartezeiten f\"{u}hren kann.

Bei der 3. Dimension kann entweder die Hinzunahme der Zeit oder die geometrische Ausdehnung in Richtung der z-Achse gemeint sein. Zur Vermeidung von Missverst\"{a}ndnissen werden in der vorliegenden Arbeit nur r\"{a}umliche Bildstr\"{o}me von Zebrab\"{a}rblingslarven als dreidimensional bezeichnet, ansonsten wird von Abtastzeitpunkten gesprochen. Wie bei den 2.5"~dimensionalen Aufnahmen erweitert sich der Bildstrom bez\"{u}glich der $z$"~Achse und alle Pixel werden zu Voxeln. Allerdings ist bei solchen Daten im Gegensatz zu 2.5"~dimensionalen Bildstr\"{o}men eine Volumeninformation gegeben, da die verdeckt liegenden Bereiche im Fisch ebenfalls abgebildet werden k\"{o}nnen. Es steigen jedoch die Anforderungen an die Computer-Hardware sowohl durch den gr\"{o}{\ss}eren Speicherplatzbedarf als auch in der Auswertung, da die Volumenbilder meist als Ganzes in den Arbeitsspeicher geladen werden m\"{u}ssen.

Die gr\"{o}{\ss}ten Bildstr\"{o}me entstehen bei der Kombination von Multikanalaufnahmen mit dreidimensionalen Aufnahmen. Solche n-dimensionalen Datens\"{a}tze lassen sich mit dem \sog SPIM Mikroskop aufzeichnen (\vgl \kap \ref{subsec:Bildakquise+Bildverarbeitung}). F\"{u}r die \hts bleibt die Verarbeitung hoch-dimensionaler Bilddaten jedoch zum Zeitpunkt des Entstehens der vorliegenden Arbeit nur eine theoretische M\"{o}glichkeit, da sowohl die Datenakquise, die Rechenleistung und die Speicherkapazit\"{a}ten solche Datens\"{a}tze f\"{u}r den Hochdurchsatz ausschlie{\ss}en.

Einen Kompromiss bietet CAM (Computer Aided Microscopy) \cite{Peravali11}. Mittels der CAM-Technologie wird ein \"{U}bersichtsbild oder ein dreidimensionales Bild in niedriger Aufl\"{o}sung aufgenommen, live ausgewertet und lediglich der Bereich von Interesse akquiriert, beispielsweise mit Hilfe eines Zoomobjektives. Der in sehr hoher Aufl\"{o}sung und \ggf Multikanaltechnik akquirierte Bereich beschr\"{a}nkt sich hierbei m\"{o}glichst lediglich auf den Bereich im Zebrab\"{a}rbling, der die \NutzsigInf tr\"{a}gt.


\subsection{Analyse und Interpretation}

Bei der Analyse und Interpretation (vgl. weiterhin \abb \ref{fig:Entscheidungsbaum}) m\"{u}ssen nach der Bildstrom-Vorverarbeitung die Art des biologischen Effekts von Interesse (das \Nutzsig) und die darin enthaltene Information (die \NutzsigInf) bestimmt werden. Eine typische \NutzsigInf sind entweder eine charakteristische Form oder Fl\"{a}che eine Bereichs im Zebrab\"{a}rbling oder auch des gesamten Modellorganismus, die biologische Effektst\"{a}rke, die Bewegung oder die Ver\"{a}nderung einer der zuvor genannten \NutzsigInfen \"{u}ber die Zeit.

Die in der Segmentierung gefundenen Bereiche m\"{u}ssen bez\"{u}glich des Nutzsignals quantifiziert werden. Je nach Nutzsignal k\"{o}nnen entweder Merkmale, nach einer Bildvorverarbeitung wie \zB Verbesserungen des Kontrastes (vgl. \kap \ref{sec:BildstromVorverarbeitung}), direkt extrahiert werden oder es m\"{u}ssen erst Zeitreihen ermittelt und die Zeitreihen weiter zu Merkmalen reduziert werden. Hierbei handelt es sich also um Fischinformationszeitreihen und -merkmale oder einfacher Nutzinformationszeitreihen und -merkmale. In Momentaufnahmen wird oft ein Bild in einem Zwischenschritt durch die Segmentierung erzeugt. Solche Bilder enthalten nach M\"{o}glichkeit lediglich das Nutzsignal, dessen Information sich durch formbeschreibende Werte quantifizieren l\"{a}sst und damit Aufschluss \"{u}ber die charakteristischen Bereiche, \dhe der \NutzsigInf des Zebrab\"{a}rblings, gibt. Typische solcher Merkmale sind \zB die Gr\"{o}{\ss}e, die Rundheit, der Umfang des Bereichs oder die lange/kurze Halbachse einer umschlie{\ss}enden Ellipse. Die biologische Effektst\"{a}rke zeigt sich in der Signalst\"{a}rke im Bild. Das Merkmal der Effektst\"{a}rke wird vornehmlich bei Fluoreszenzaufnahmen eingesetzt. Die vom Detektor aufgezeichneten Lichtintensit\"{a}tswerte entsprechen hier der Signalst\"{a}rke eines bestimmten Bereichs und sind das direkte Ma{\ss} f\"{u}r die St\"{a}rke des biologischen Effekts. Bewegungen lassen sich gut in Bildsequenzen detektieren, von Interesse sind hierbei die H\"{a}ufigkeit einer Bewegung oder Bewegungsmuster. Quantifiziert werden Bewegungsmuster meist \"{u}ber die Ver\"{a}nderung von Pixelwerten \"{u}ber der Zeit. In Bildsequenzen besteht zus\"{a}tzlich die M\"{o}glichkeit, alle zuvor genannten Merkmale an unterschiedlichen Zeitpunkten zu bestimmen. Die so entstehenden Zeitreihen lassen eine Bewertung der Ver\"{a}nderung \"{u}ber die Zeit zu. Beispielsweise kann die Ver\"{a}nderung der Gr\"{o}{\ss}e des Dottersacks w\"{a}hrend der Entwicklung ein solches Ma{\ss} sein.

Alle extrahierten Merkmale repr\"{a}sentieren die \NutzsigInf und k\"{o}nnen nun weiter verarbeitet werden. Ziel ist es, die Merkmale der zugeh\"{o}rigen Ausgangsklasse zuzuweisen. Zu Beginn wird der Einfluss von St\"{o}rfaktoren mittels einer Normierung versucht zu beseitigen. Daraufhin werden m\"{o}glichst aussagekr\"{a}ftige Merkmale extrahiert und die trennungsst\"{a}rksten in einer Merkmalsauswahl identifiziert. F\"{u}r die Identifikation sowie f\"{u}r die sp\"{a}tere Klassifikation ist ein Lerndatensatz erforderlich. Hierf\"{u}r wird ein Teil der Daten manuell anhand der Bilddaten der entsprechenden Klasse zugeordnet. Mittels der gefundenen signifikanten Merkmale und des Lerndatensatzes wird daraufhin ein Klassifikator angelernt und die Klassenzugeh\"{o}rigkeit der restlichen, nicht manuell zugeordneten Daten gesch\"{a}tzt. Aus den Klassenzugeh\"{o}rigkeiten lassen sich dann in einer Auswertung (vgl. weiter \abb \ref{fig:Entscheidungsbaum}) charakteristische \bzw biologisch oder toxikologisch wichtige Kurven und Werte berechnen und interpretieren. Bei einer toxikologischen Untersuchung kann \zB eine Dosis-Effekt-Kurve dargestellt werden. Dabei wird die gesch\"{a}tzte Klassenzugeh\"{o}rigkeit jeder Einzeluntersuchung gegen\"{u}ber der Dosis einer Chemikalie aufgetragen. Aus dieser Kurve lassen sich dann charakteristische Werte wie \zB die h\"{o}chste Konzentration, bei der kein Effekt auftritt (NOEL; engl.  \textbf{N}o \textbf{O}bserved \textbf{E}ffect \textbf{L}evel), berechnen. Ein weiterer typischer Wert ist die Konzentration, bei der die H\"{a}lfte aller Versuchseinheiten nicht \"{u}berlebensf\"{a}hig ist (EC$_{50}$; engl. \textbf{E}ffect \textbf{C}oncentration 50).


Bei der Evaluation muss ein Biologe die pr\"{a}sentierten Daten entsprechend interpretieren. Die Deutung geschieht fast ausschlie{\ss}lich manuell. F\"{u}r bekannte Effekte l\"{a}sst sich jedoch eine Plausibilit\"{a}tspr\"{u}fung oder eine automatisierte Gegenprobe einf\"{u}hren, welche eine Sicherheit gegen\"{u}ber Fehlauswertungen bietet. Die Ergebnisse m\"{u}ssen schlie{\ss}lich archiviert werden, um zu einem sp\"{a}teren Zeitpunkt eine Reproduktion der Ergebnisse zu erm\"{o}glichen.

 \subsection{Einfluss der Versuchsparameter auf die Bildqualit\"{a}t}\label{subsec:Bildqualit\"{a}t}
Die f\"{u}r \htsen geforderten Parameter wie \zB m\"{o}glichst kurze Gesamtzeit der Versuchsdurchf\"{u}hrung haben zur Folge, dass die Bildakquise automatisiert durchgef\"{u}hrt werden muss. Hierf\"{u}r finden die im vorigen Abschnitt erw\"{a}hnten speziellen Hochdurchsatz-Mikroskope Anwendung.  Die schnelle Akquise f\"{u}hrt jedoch leicht zu Inhomogenit\"{a}ten in den Bildstr\"{o}men, da die Aufnahmen ohne menschliche \"{U}berwachung oder Korrektur durchgef\"{u}hrt werden. Solche Inhomogenit\"{a}ten beeinflussen die Bildqualit\"{a}t und -information. Die Bildverarbeitung und Klassifikation muss entsprechend abgestimmt werden \cite{Khan12smps,Stegmaier12a}. Zur Betrachtung der Auswirkungen auf die Bildauswertung lassen sich die Inhomogenit\"{a}ten in zwei Bereiche einteilen:
\begin{enumerate}
\item Inhomogenit\"{a}t der \emph{Qualit\"{a}t} durch Fehler bei der Mikroskopie und
\item Inhomogenit\"{a}t der \emph{Information} durch fehlerhafte Objekte.
\end{enumerate}
Die Qualit\"{a}t bezieht sich hierbei auf die G\"{u}te der Bilder bez\"{u}glich Sch\"{a}rfe, Beleuchtung und Reinheit, w\"{a}hrend mit Information der Inhalt der Bilder gemeint ist, also inwieweit die Akquise den biologischen Effekt von Interesse im Zebrab\"{a}rbling auf dem Bild wiedergibt.
Die Information ist die Forderung, dass der Bildstrom die \NutzsigInf enth\"{a}lt. Die Akquise-Methode muss also so gew\"{a}hlt werden, dass der biologische Effekt von Interesse im Bildstrom vorhanden ist.
\abb \ref{fig:Qualit\"{a}tsprobleme_Daten} zeigt Beispiele f\"{u}r Beeintr\"{a}chtigungen, die bei der Bildakquise im Hellfeld entstehen k\"{o}nnen. In \abb \ref{fig:Qualit\"{a}tsprobleme_Daten_a} und \abb \ref{fig:Qualit\"{a}tsprobleme_Daten_b} stellt sich der automatische Fokus des Mikroskops auf Schmutz in der Fl\"{u}ssigkeit oder auf der Linse scharf. Dadurch sind die Larven nur undeutlich zu erkennen. In \abb \ref{fig:Qualit\"{a}tsprobleme_Daten_c} und \abb \ref{fig:Qualit\"{a}tsprobleme_Daten_d} sind die Larven von Schmutz umgeben und teilweise \"{u}berdeckt. In \abb \ref{fig:Qualit\"{a}tsprobleme_Daten_e} und \abb \ref{fig:Qualit\"{a}tsprobleme_Daten_f} schlie{\ss}lich treten gro{\ss}e Fremdk\"{o}rper in den Bildern auf, die zum Teil sogar die
Kontur um die Eih\"{u}lle unterbrechen.
\begin{figure}[!htb]
        \centering
         \captionsetup[subfloat]{labelfont = {small ,bf},format = hang}
                \subfloat[\small Fehlerhafte Fokussierung
]{\includegraphics[width=0.29\linewidth]{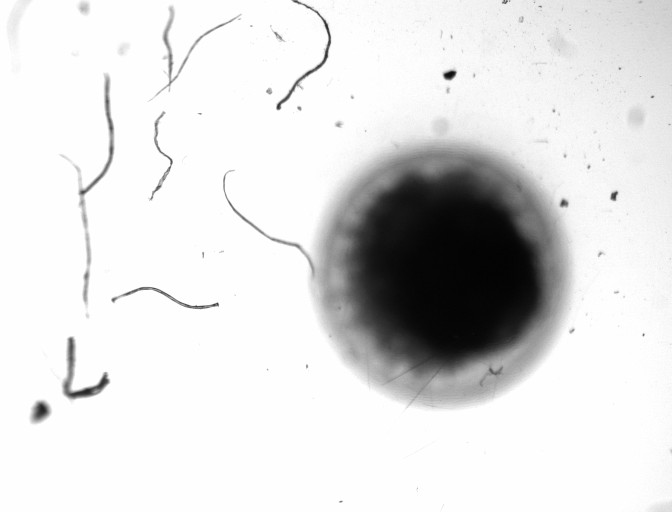}\label{fig:Qualit\"{a}tsprobleme_Daten_a}}
\hfill
                \subfloat[\small Fehlerhafte Fokussierung
                ]{\includegraphics[width=0.29\linewidth]{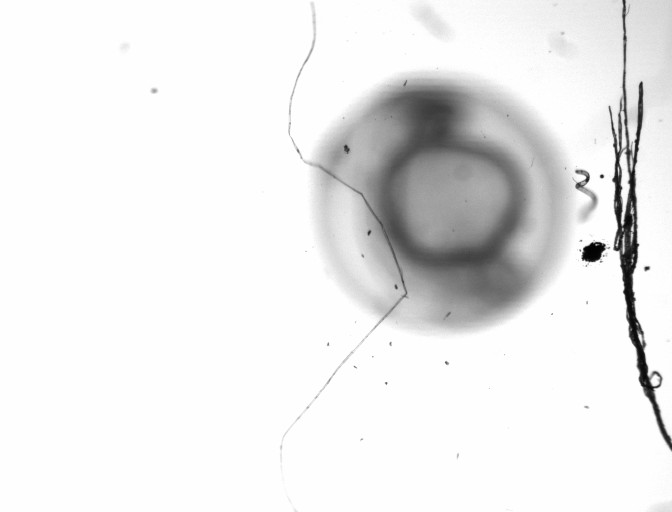}\label{fig:Qualit\"{a}tsprobleme_Daten_b}}
                \hfill
                \subfloat[\small Schmutzpartikel im Wasser
]{\includegraphics[width=0.29\linewidth]{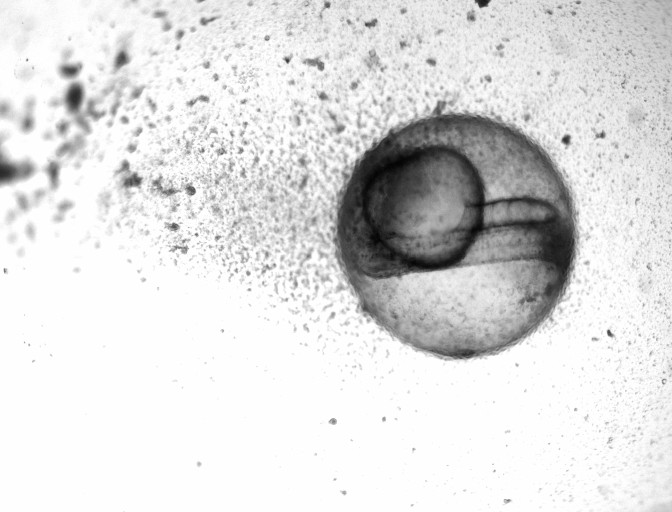}\label{fig:Qualit\"{a}tsprobleme_Daten_c}}
                \\
                \subfloat[\small Schmutzpartikel im Wasser
                ]{\includegraphics[width=0.29\linewidth]{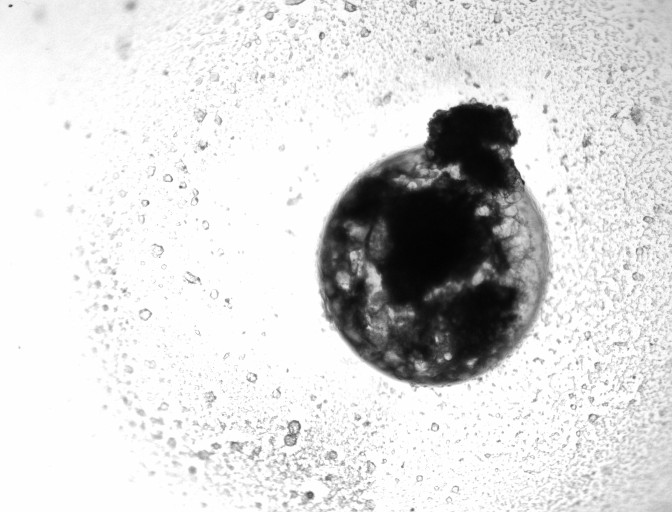}\label{fig:Qualit\"{a}tsprobleme_Daten_d}}
\hfill
                \subfloat[\small Fremdk\"{o}rper im Bild
]{\includegraphics[width=0.29\linewidth]{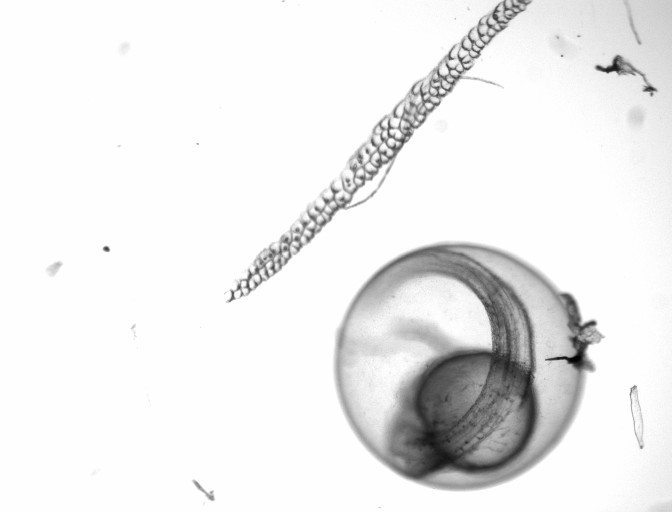}\label{fig:Qualit\"{a}tsprobleme_Daten_e}}
                \hfill
                \subfloat[\small Fremdk\"{o}rper im Bild
]{\includegraphics[width=0.29\linewidth]{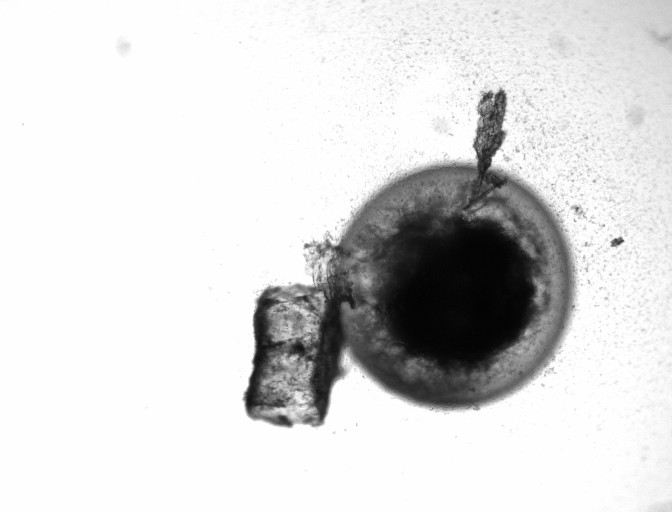}\label{fig:Qualit\"{a}tsprobleme_Daten_f}}
        \caption[Qualitative Inhomogenit\"{a}t des Bildstroms]{Qualitative Inhomogenit\"{a}t des Bildstroms. Zur Akquise wurde ein Hochdurchsatz-Mikroskop des Typs \scanr verwendet \cite{Alshut08}.}
        \label{fig:Qualit\"{a}tsprobleme_Daten}
\end{figure}

Die Qualit\"{a}t des Bildstroms l\"{a}sst sich ma{\ss}geblich durch Sorgfalt bei der Bildakquise und Pr\"{a}paration verbessern. So werden z.B. s\"{a}mtliche Verunreinigungen, die sich auf dem Objektiv, in der Luft oder im Wasser um die Larven befinden, mit aufgezeichnet und verdecken oder verschlechtern Informationen und damit die Qualit\"{a}t aller weiteren Schritte der Auswertung und k\"{o}nnen im schlimmsten Fall zu einer Fehlklassifikation  f\"{u}hren. Je konstanter die Umgebungsbedingungen wie z.B. die Raumbeleuchtung sind, desto konstanter ist auch die Qualit\"{a}t des Bildstroms. Ebenso kann der automatische Fokus des Mikroskops sich durch \"{u}berm\"{a}{\ss}ige Verschmutzungen der Probe falsch justieren, sodass irrelevante Bildteile scharf abgebildet werden. In den im Rahmen der vorliegenden Arbeit betrachteten Bildstr\"{o}men erwiesen sich Schmutzpartikel, Bildunsch\"{a}rfe und eine ungleichm\"{a}{\ss}ige Ausleuchtung als die am h\"{a}ufigsten auftretenden qualitativen Einschr\"{a}nkungen.

\begin{figure}[!htb]
        \centering
        \small
         \captionsetup[subfloat]{labelfont = {small ,bf}}
                \subfloat[\small Eih\"{u}lle ist mechanisch zerst\"{o}rt
                 ]{\includegraphics[width=0.29\linewidth]{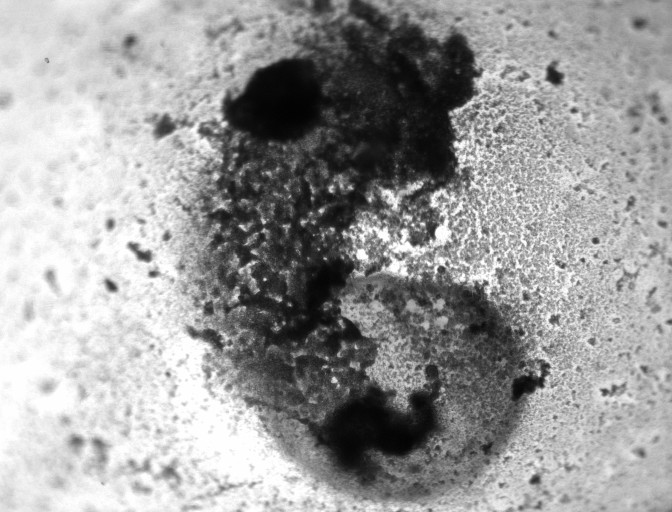}\label{fig:Informationsprobleme_Daten_a}}
                \hfill
                \subfloat[\small Weitere zerfallene Eih\"{u}lle im Bild
                ]{\includegraphics[width=0.29\linewidth]{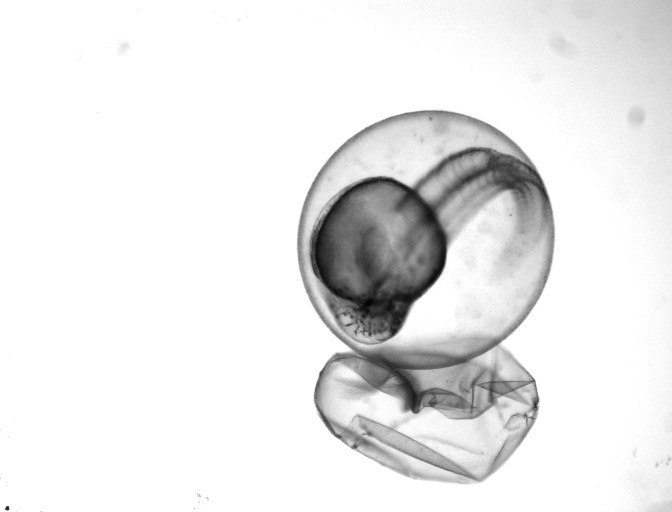}\label{fig:Informationsprobleme_Daten_b}}
                \hfill
                \subfloat[\small Eih\"{u}lle unvollst\"{a}ndig
]{\includegraphics[width=0.29\linewidth]{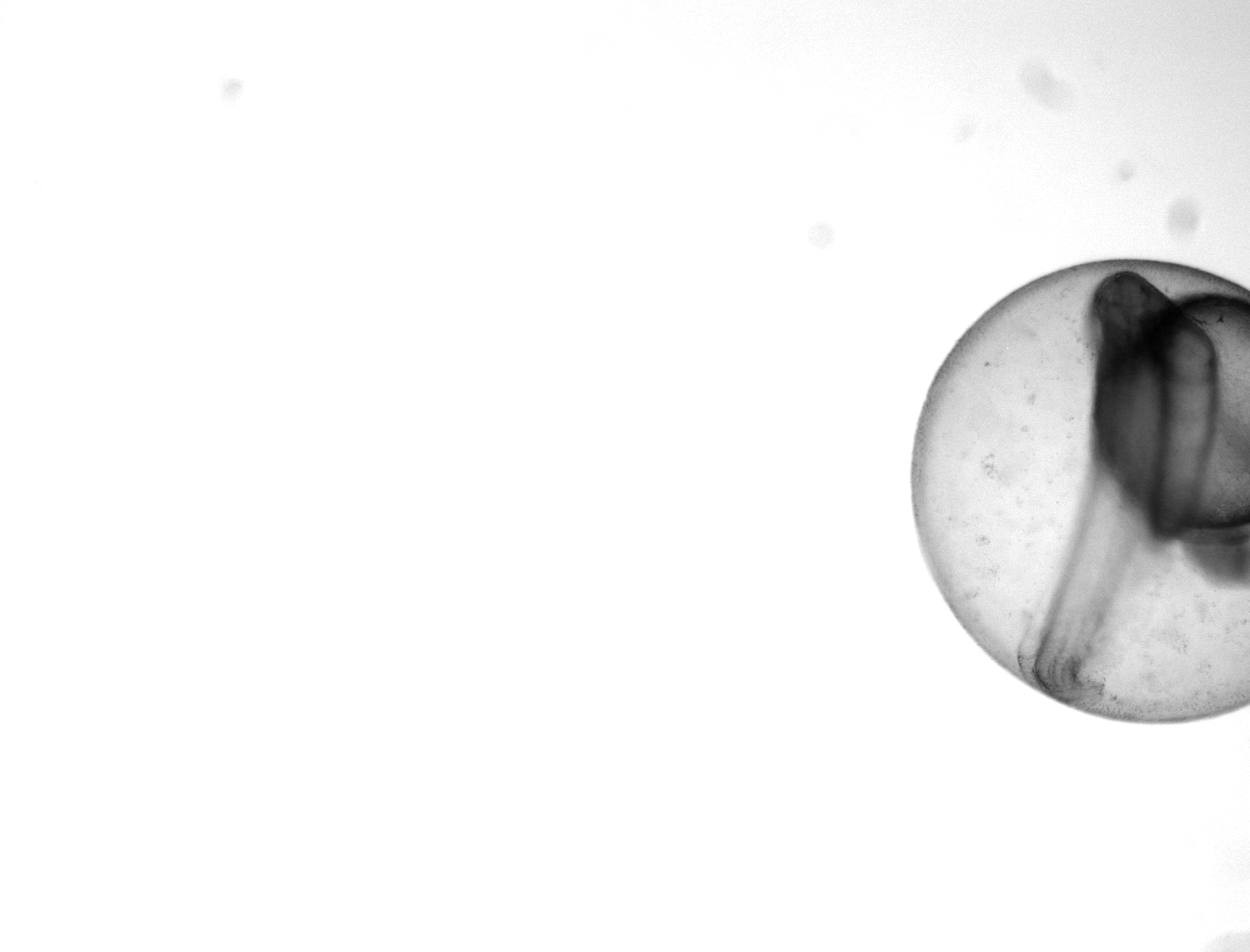}\label{fig:Informationsprobleme_Daten_c}}
                \\
                \subfloat[\small Zwei Larven abgebildet
                ]{\includegraphics[width=0.29\linewidth]{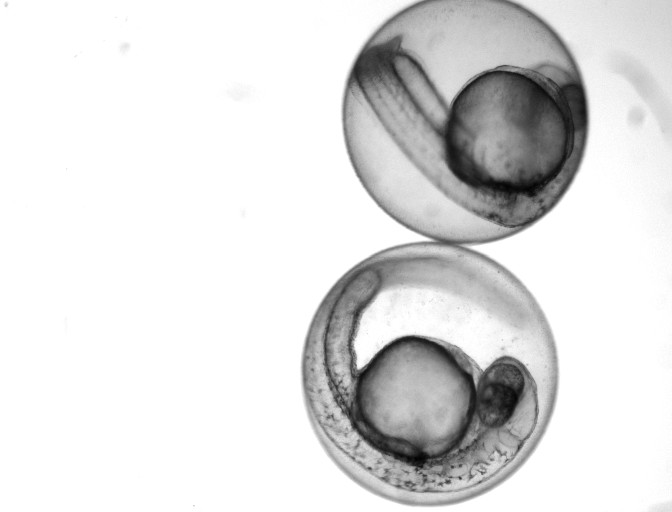}\label{fig:Informationsprobleme_Daten_d}}
                \hfill
                \subfloat[\small Ei ist unbefruchtet
]{\includegraphics[width=0.29\linewidth]{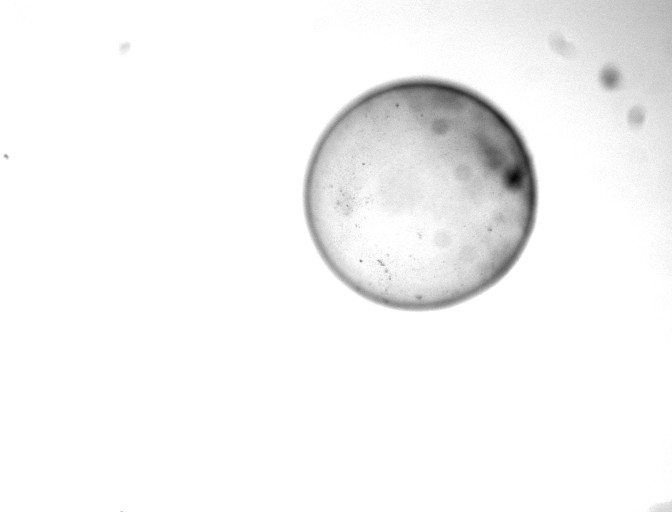}\label{fig:Informationsprobleme_Daten_e}}
                \hfill
                \subfloat[\small Leeres N\"{a}pfchen  abgelichtet
]{\includegraphics[width=0.29\linewidth]{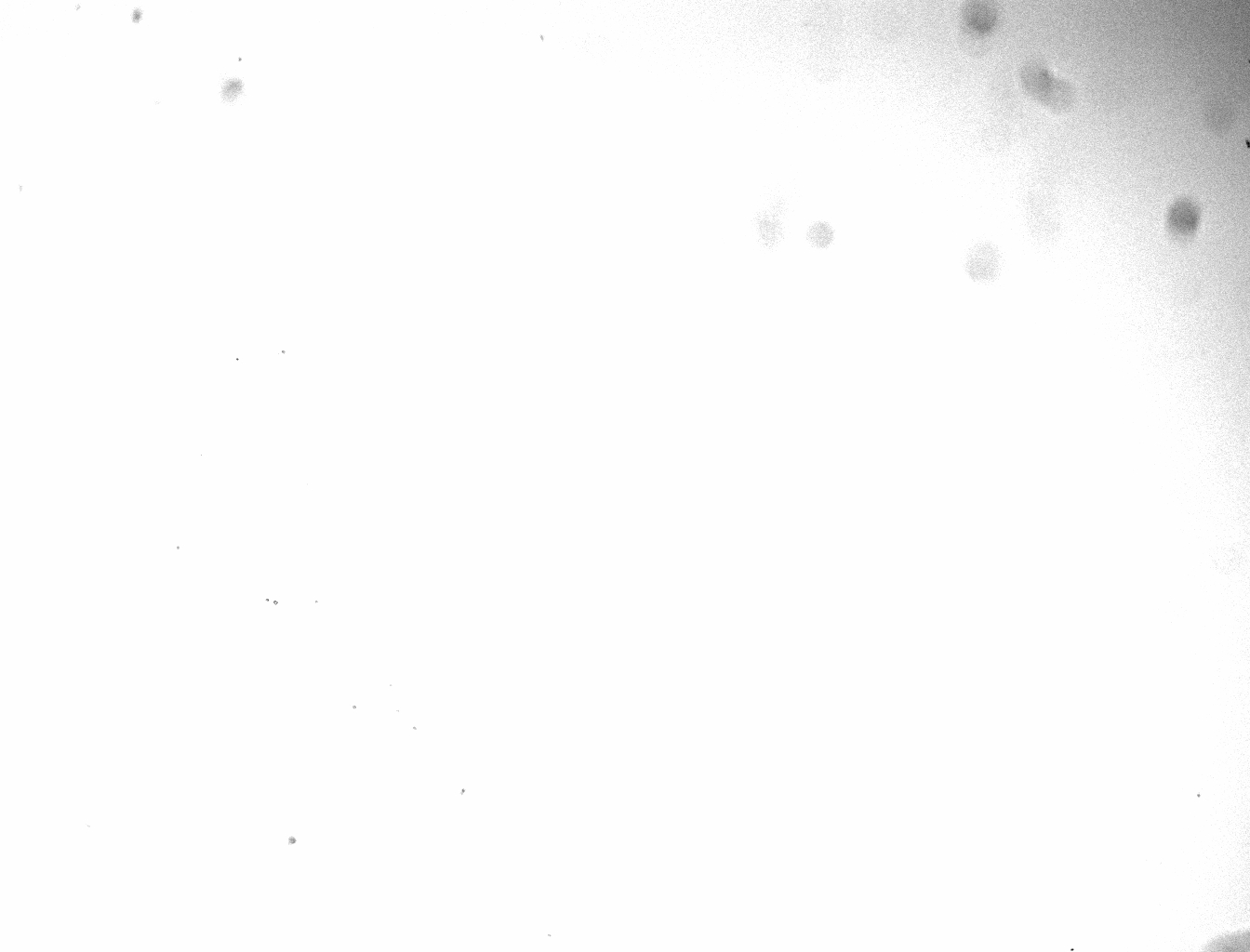}\label{fig:Informationsprobleme_Daten_f}}
        \caption{Informelle Inhomogenit\"{a}t der Datens\"{a}tze \cite{Alshut08}}
        \label{fig:Informationsprobleme_Daten}
\end{figure}

Oft sind die Aufnahmen zwar von guter Qualit\"{a}t bez\"{u}glich der Sch\"{a}rfe und der Beleuchtung, jedoch wird die \NutzsigInf vom Bildstrom nicht oder nicht komplett wiedergegeben. Larven sind \zB nicht vollst\"{a}ndig auf dem Bild abgelichtet oder
auf einem Bild sind mehrere Larven \bzw auch gar keine zu finden (vgl. \abb \ref{fig:Informationsprobleme_Daten}), obwohl im Versuchsprotokoll genau eine Larve gefordert war. In \abb \ref{fig:Informationsprobleme_Daten_a} ist das Zebrab\"{a}rblingsei durch mechanische Einwirkung bei der Vereinzelung mit der Pipette oder beim Transport zerst\"{o}rt worden. In \abb \ref{fig:Informationsprobleme_Daten_b} befindet sich neben der Eih\"{u}lle der lebenden Larve eine weitere, leere Eih\"{u}lle, die m\"{o}glicherweise von einer bereits geschl\"{u}pften Larve stammt.
\abb \ref{fig:Informationsprobleme_Daten_c} bildet nur ca. 1/3 der Larve ab, w\"{a}hrend in \abb \ref{fig:Informationsprobleme_Daten_d} zwei Larven gleichzeitig bei der Bilderfassung in dem N\"{a}pfchen der Mikrotiterplatte waren. In \abb \ref{fig:Informationsprobleme_Daten_e} ist das Zebrab\"{a}rblingsei nicht befruchtet, es erscheint auf dem Bild als "`leerer Ring"'. In \abb \ref{fig:Informationsprobleme_Daten_f} schlie{\ss}lich ist ein leeres N\"{a}pfchen abgelichtet. Die gezeigten Beispiele k\"{o}nnen nicht das gesamte Spektrum von Inhomogenit\"{a}ten abbilden, sollen jedoch die Vielf\"{a}ltigkeit der Fehlerquellen aufzeigen.

\begin{figure}[h!tbp]
\hspace{2cm}
   \subfloat[\small richtig \label{fig:Registrierung_a}
                 ]{

                  \includegraphics[height=4.7cm]{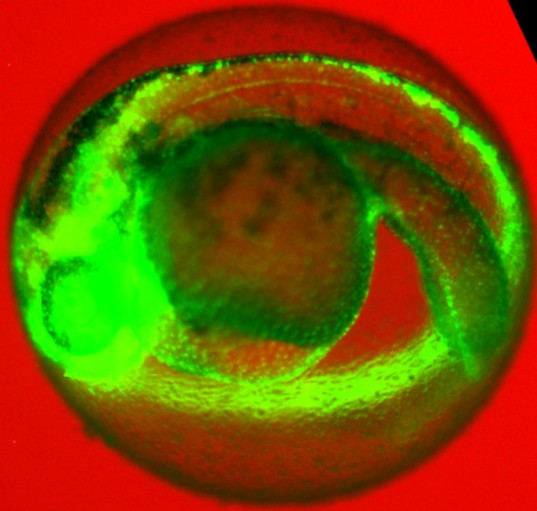}}
                \hspace{2cm}
                \subfloat[\small falsch \label{fig:Registrierung_b}
                ]{ \includegraphics[height=4.8cm]{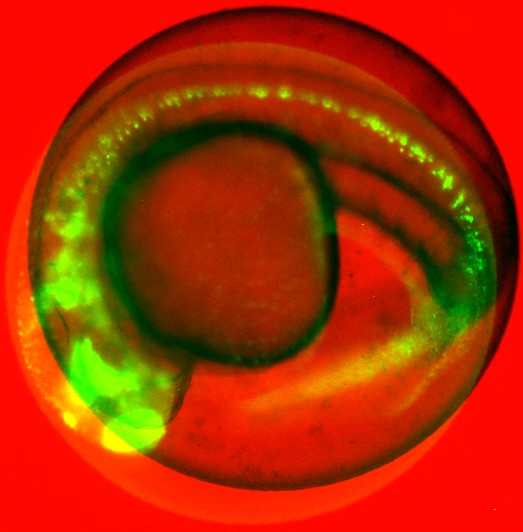}}
                \hfill
                \caption[Beispiele f\"{u}r die Registrierung von Multikanalaufnahmen.]{Beispiele f\"{u}r die Registrierung von Multikanalaufnahmen. \"{U}berlagerung der Grauwertbilder durch gr\"{u}n eingef\"{a}rbte Fluoreszenzaufnahmen auf rotem Hintergrund (Falschfarben-Darstellung)}\label{fig:Registrierung}
                \end{figure}

Bei Multikanalaufnahmen ist ein weiterer Aspekt die Problematik des zeitlichen Versatzes, welcher zwischen den Aufnahmen der einzelnen Kan\"{a}le entsteht. \ZB nimmt das Mikroskop \scanr bei mehreren Fokusebenen in einem ersten Durchlauf alle Fokusebenen der Reihe nach auf, beginnend mit den Hellfeld-Aufnahmen. Daraufhin wird das Durchlicht durch eine Blende verschlossen, ein Fluoreszenzfilter automatisch eingesetzt und die entsprechenden Fokusebenen f\"{u}r die Fluoreszenzkan\"{a}le akquiriert. Die Bildakquise erfolgt damit nicht simultan \cite{Liebel03}, was Abweichungen der Bilder zueinander zur Folge hat, da zum einen Toleranzen des Mikroskops den optischen Fluss zwischen beiden Aufnahmeserien ver\"{a}ndern und zum anderen  sich das Objekt entweder eigenst\"{a}ndig (z.B. durch Muskelzucken der Larven) oder durch St\"{o}rungen bewegt haben kann. Das hat zur Folge, dass gleiche Strukturen im Fisch nicht an der gleichen Stelle im Bild abgebildet werden. Kleine Toleranzen k\"{o}nnen oder m\"{u}ssen in einem solchen Fall durch eine Registrierung, \dhe eine Zuordnung der einzelnen Kan\"{a}le durch Translationen und Rotationen, ausgeglichen werden, w\"{a}hrend die Registrierung bei gro{\ss}en Toleranzen scheitert oder einen zu gro{\ss}en Aufwand erfordert. Solche F\"{a}lle m\"{u}ssen dann \ggf mittels geeigneter Validit\"{a}tspr\"{u}fungen erkannt und ausgeschlossen werden (\vgl Kapitel \ref{sec:BV_Module}).
Ein Beispiel f\"{u}r eine solche Abweichung ist in \abb \ref{fig:Registrierung} gegeben. W\"{a}hrend in der linken Abbildung \ref{fig:Registrierung_a} sich der Fluoreszenzkanal mit dem Hellfeldkanal deckt, sind in \abb \ref{fig:Registrierung_b} die Kan\"{a}le nicht deckungsgleich was deutlich am Kopfbereich des Fisches zu sehen ist.

\section{Mathematisches Modell der \hts}\label{sec:Parameter_mathematisch}


Bei der Betrachtung der \hts ist eine allgemein anwendbare Notation, welche sich auf alle Auspr\"{a}gungen der Untersuchung adaptieren l\"{a}sst, hilfreich. Daher wird im folgenden Abschnitt f\"{u}r alle identifizierten Parameter der \hts eine modellhafte Beschreibung eingef\"{u}hrt, welche sich universell auf \htsen anpassen l\"{a}sst und ein einheitliches Beschreibungsmodell darstellt.

\begin{figure}[htbp]
        \centering
        \includegraphics[page=1
        ]{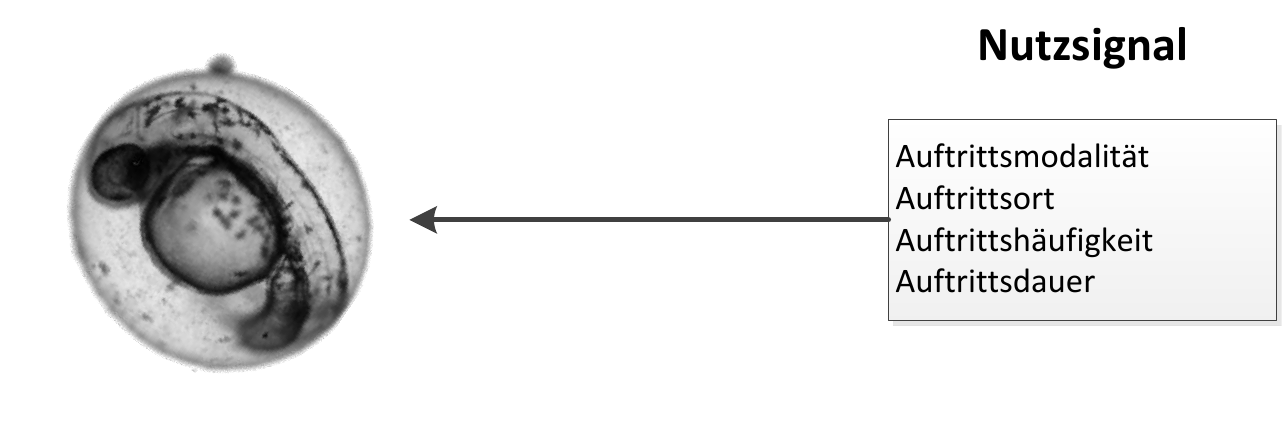}
\caption{Charakterisierung des \Nutzsigs}
\label{fig:Nutzsignaleigenschaften}
\end{figure}

Werden die identifizierten Parameter aus Sicht der Eingangsdaten der Bildverarbeitung beschrieben, so kann eine formale Betrachtung vorgenommen werden, aus der sich eine modellhafte Beschreibung ableiten l\"{a}sst. Der biologische Effekt und somit das \Nutzsig werden charakterisiert durch vier Parameter, die sich ausnahmslos auf die Art oder den Ort beziehen, an dem das \Nutzsig auftritt. Die Parameter sind Auftrittsmodalit\"{a}t, "~ort, "~h\"{a}ufigkeit und "~dauer (vgl. \abb \ref{fig:Nutzsignaleigenschaften})

Bei der Betrachtung der \hts muss die \textbf{Auftrittsmodalit\"{a}t} des Signals gekl\"{a}rt und in der Versuchsauslegung ber\"{u}cksichtigt werden. Der Begriff Auftrittsmodalit\"{a}t steht f\"{u}r die verschiedenen Kan\"{a}le, mit denen ein \Nutzsig aufgezeichnet werden kann, etwa mittels der Aufzeichnung von Grauwertbildern von einem Hellfeld-Mikroskop und \zB einem weiteren Fluoreszenz-Kanal. Die Auftrittsmodalit\"{a}t hat daher Einfluss auf die Dimension der Bildakquise. F\"{u}r den Parameter wird der Bezeichner $m_\Psi$ eingef\"{u}hrt.
Ein weiterer Parameter ist der \textbf{Auftrittsort} des \Nutzsigs. Der Auftritts\-ort ist gleichzeitig der Ort bzw. der Raum von Interesse (engl. \emph{(ROI)} Region of Interest) und durch Indizes definiert, welche die Lage des Raums innerhalb des Nutzsignals beschreiben. F\"{u}r den Auftrittsort wird $\mathbf{{u}}_\Psi$ eingef\"{u}hrt mit  $\mathbf{{u}}_\Psi=[x1_\Psi, x2_\Psi, y1_\Psi, y2_\Psi, z1_\Psi, z2_\Psi]^T$.
Die \textbf{Auftrittsh\"{a}ufigkeit} beschreibt, inwieweit das zu beobachtende \Nutzsig st\"{a}ndig sichtbar ist (\zB st\"{a}ndig oder nur zu einer bestimmten Wahrscheinlichkeit in einem von 1000 Zeitpunkten).  F\"{u}r die Auftrittsh\"{a}ufigkeit wird der Bezeichner $f_{\Psi}$ eingef\"{u}hrt.
W\"{a}hrend die Auftrittsh\"{a}ufigkeit des \Nutzsigs vornehmlich zu Wartezeiten und niedrigerem Durchsatz f\"{u}hrt, wirkt sich die \textbf{Auftrittsdauer} auf die L\"{a}nge der Bildsequenzen aus, sollte das \Nutzsig \"{u}ber die gesamte Auftrittsdauer beobachtet werden m\"{u}ssen.  F\"{u}r die gew\"{a}hlte Aufnahmedauer des \Nutzsigs wird der Bezeichner $t_{\Psi}$ eingef\"{u}hrt.

%


Eine konkrete bildbasierte \hts besteht aus einer Anzahl $N$ von divers  behandelten Zebrab\"{a}rblingslarven. Es wird vereinfachend davon ausgegangen, dass eine Larve Tr\"{a}ger genau eines biologischen Effekts und damit die Versuchseinheit ist. Die Larven existieren als reale Versuche ${\mathbf{RV}}$, \dhe in Form von kontinuierlichen Werten und deren digitaler Abbildung in einem Bildstrom ${\mathbf{BS}}$.
Die Anzahl an untersuchten Zebrab\"{a}rblingslarven setzt sich zusammen aus drei Klassen. Die Klasse der Positiv-Kontrollen bestehend aus $n_{C^+}$ Versuchseinheiten, die der Negativ-Kontrollen bestehend aus $n_{C^-}$ Versuchseinheiten und die Klasse der den Rahmenbedingungen ausgesetzten Proben bestehend aus $n_s$ Versuchseinheiten. Die Anzahl an Zebrab\"{a}rblingslarven in der \hts ist dann gegeben durch:
\begin{equation}\label{eqn:HDU_Umfang}
n=1 \dots N \quad \text{mit} \quad N = n_{C^+}+n_{C^-}+n_{s}, \quad  N \in \mathbb{N}^+.
\end{equation}


Die Versuchseinheit wird durch Gr\"{o}{\ss}en beschrieben, die die Versuchsbedingungen bestimmen. Die Gr\"{o}{\ss}en werden in Anlehnung an \cite{Rasch07} als Faktoren bezeichnet. Es erweist sich als zweckm\"{a}{\ss}ig, Plan"~ und St\"{o}rfaktoren zu unterscheiden. Die Planfaktoren beschreiben jene Gr\"{o}{\ss}en, deren Einfluss durch systematische Variation untersucht werden soll. In den St\"{o}rfaktoren werden alle anderen Gr\"{o}{\ss}en zusammengefasst, die ebenfalls einen Einfluss auf die Versuchsergebnisse haben. In der Literatur werden die St\"{o}rfaktoren weiterhin unterteilt in Konstant"~, Rest"~ und Blockfaktoren \cite{Rasch07}. Konstantfaktoren sind Faktoren, die ihren Wert w\"{a}hrend der Versuchsdurchf\"{u}hrung nicht \"{a}ndern, Blockfaktoren sind Faktoren, die sich zu Bl\"{o}cken oder Stufen zusammenfassen lassen und Restfaktoren sind die \"{u}brigen Faktoren. Von der zus\"{a}tzlichen Unterteilung wird in der vorliegenden Arbeit abgesehen, da die Unterteilung vornehmlich f\"{u}r die statistische Versuchsplanung von Interesse ist. Jeder Faktor besteht aus einer Menge von Klassen, den Ausgangsklassen.

\begin{figure}[htbp]
        \centering
        \includegraphics[page=5]{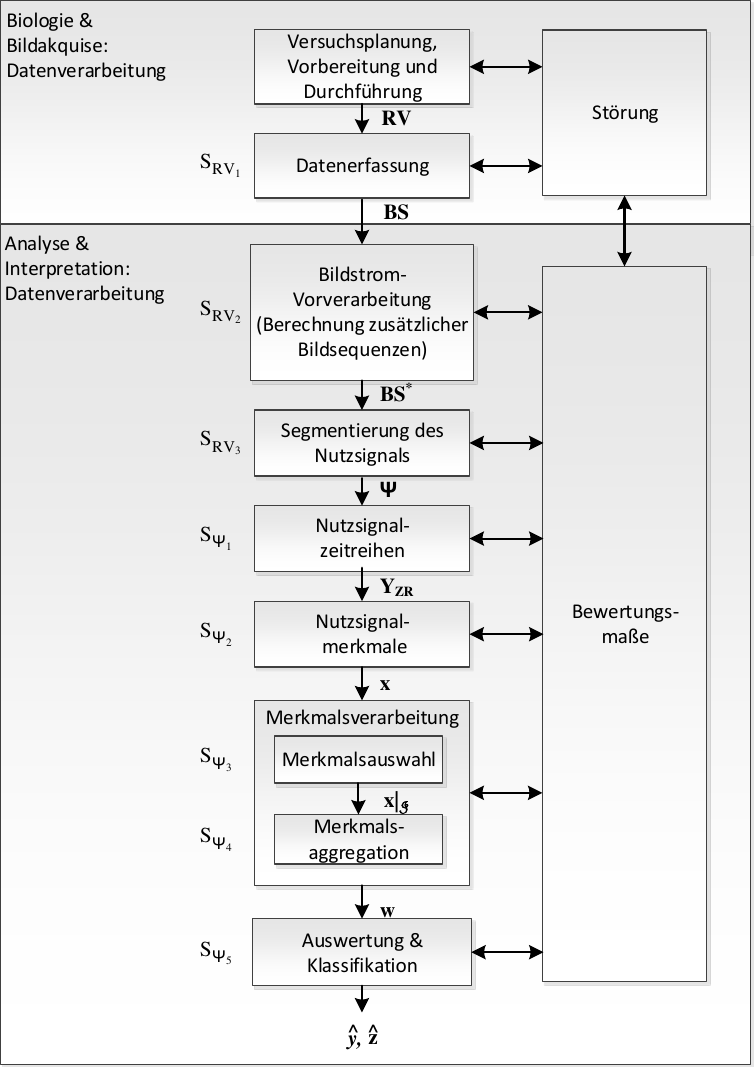}
\caption[Schema zum mathematischen Modell f\"{u}r bildbasierte \htsen]{Schema zum mathematischen Modell f\"{u}r bildbasierte \htsen aus Sicht der Datenverarbeitung}
\label{fig:Signalmodell}
\end{figure}

Jede \hts enth\"{a}lt eine Anzahl $i_{sy}=1 \dots s_y$ von Planfaktoren $y$ und eine Anzahl $i_{sz}=1 \dots s_z$ von St\"{o}rfaktoren $z$. Die Planfaktoren lassen sich anhand der Versuchseinheiten in einer ($N,s_y$)"--dimensionalen Matrix $\mathbf{Y}$ anordnen. Im Folgenden wird meist, zur einfacheren Notation, der Fall eines eindimensionalen Planfaktors ($s_y=1$) betrachtet. Analog hierzu werden die St\"{o}rfaktoren in einer ($N,s_z$)"--dimensionalen Matrix  $\mathbf{Z}$ notiert.
Die Anzahl an Klassen des $i_{sy}$-ten Planfaktors werden im Folgenden mit $m_{y,i_{sy}}$ und die Anzahl an Klassen des $i_{sz}$-ten St\"{o}rfaktors als $m_{z,i_{sz}}$ bezeichnet. F\"{u}r jede Versuchseinheit wird bez\"{u}glich jedes Faktors eine Klassenzuordnung getroffen.
Weiter werden alle Faktoren durch die noch zu definierenden Versuchsparameter $(\mathbf{p}_z)_n,(\mathbf{p}_y)_n$ bestimmt
\begin{equation}
\mathbf{Y}=
\begin{pmatrix}
y_1[1] & \cdots & y_{s_y}[1]  \\ \vdots & \ddots & \vdots \\ y_1[N]  & \cdots & y_{s_y}[N]
\end{pmatrix},\;
\mathbf{Z}=
\begin{pmatrix}
z_1[1] & \cdots & z_{s_z}[1]  \\ \vdots & \ddots & \vdots \\ z_1[N]  & \cdots & z_{s_z}[N]
\end{pmatrix}.
\end{equation}
Zum besseren Verst\"{a}ndnis sei als Beispiel eine \hts angef\"{u}hrt, in welcher Zebrab\"{a}rblingseier nach Koaguliertheit und Herzschlag untersucht werden sollen und welche an zwei Mikroskopen und an unterschiedlichen Tagen aufgenommen wurde.
In diesem Beispiel sind die Zebrab\"{a}rblingseier die Versuchseinheiten. Der Herzschlag sowie die Koaguliertheit sind die Planfaktoren, mit jeweils $m_{y,1}=m_{y,2}=3$ Klassen, $y_1[n]=\{1,2,3\}$, $y_2[n]=\{1,2,3\}$. Hierbei steht bei beiden Planfaktoren die Klasse $1$ f\"{u}r \emph{ja}, die Klasse $2$ f\"{u}r \emph{vielleicht} und Klasse $3$ f\"{u}r \emph{nein}. Die Mikroskope und Aufnahmetage sind St\"{o}rfaktoren mit jeweils einer Klasse f\"{u}r jedes Mikroskop und jeden Aufnahmetag. Die gew\"{a}hlten Parameter des Mikroskops wie Vergr\"{o}{\ss}erung, Beleuchtungsst\"{a}rke \etc sind die Versuchsparameter jeder Versuchseinheit $(\mathbf{p}_{z})_n,(\mathbf{p}_{y})_n$. Die Versuchsparameter sind f\"{u}r die Planfaktoren und St\"{o}rfaktoren normalerweise identisch ($\mathbf{p}_{z} =\mathbf{p}_{y}=\mathbf{p}_{zy}$).
Der gesamte Ablauf der \hts, vom mit $\mathbf{RV}$ bezeichneten realen Versuch bis zur Klassenzuweisung der jeweiligen Klassen aus $\mathbf{Y, Z}$ ist in dem Schema in \abb \ref{fig:Signalmodell} dargestellt.

F\"{u}r die wichtigsten Faktoren der verwendeten Zebrab\"{a}rblingslarven, n\"{a}mlich dem  Alter $t_{\mathrm{hpf}}$ und dem Faktor f\"{u}r die gesamte erforderliche Zeit f\"{u}r die Realisierung $t_{{RV}}$, werden Bezeichner eingef\"{u}hrt. Alle weiteren Faktoren werden in $\mathbf{z}_{{RV}}$ zusammengefasst. Die Realisierung jedes Versuchs $\mathbf{RV}$ ist Element einer Menge, deren Inhalt abh\"{a}ngig von den Randbedingungen des Versuchs ist und mit $\mathbb{RV}$ bezeichnet wird. Es l\"{a}sst sich nun formulieren:
\begin{eqnarray}\label{eqn:RV}
\begin{pmatrix}\mathbf{RV}_1\\\vdots\\\mathbf{RV}_N\end{pmatrix}&;&\mathbf{RV} \in \mathbb{RV}\\
    \mathbf{RV}_n&=&f(\mathbf{p}_{zy})_n\\
    \text{mit} \quad (\mathbf{p}_{zy})_n&=&(t_{\mathrm{hpf}},t_{{RV}},\mathbf{z}_{{RV}})^T_n.
\end{eqnarray}
Die notwendige Zeit f\"{u}r einen Versuch (ohne Parallelisierung der Arbeitsabl\"{a}ufe) $t_{{RV}}$ setzt sich zusammen aus der Summe der Dauer aller notwendigen Pr\"{a}parationsschritte $(t_p)_{i_p}$, mit der Anzahl an Pr\"{a}parationsschritten $n_p$ und der Dauer der Akquise $t_{aq}$
 \begin{equation}\label{eqn:RV_Dauer}
 t_{{RV}}=\sum\limits_{i_p=1}^{n_p}(t_p)_{i_p}\,+\,t_{aq}.
 \end{equation}
Die Zugeh\"{o}rigkeit der Versuche $\mathbf{RV}$ zu den jeweiligen Klassen $\mathbf{y}$ ist unbekannt und soll in der \hts durch einen Klassifikator gesch\"{a}tzt werden. Sch\"{a}tzungen werden mit einem Dach gekennzeichnet. Ebenso l\"{a}sst sich die Klassenzugeh\"{o}rigkeit der Restfaktoren aus $\mathbf{z}$ sch\"{a}tzen. Es ergeben sich die Abbildungen $S_y$ und $S_z$
 \begin{eqnarray}\label{eqn:RVtoY}
S_y(\mathbf{p}): \mathbf{RV} \mapsto \hat{\mathbf{y}}\\
S_z(\mathbf{p}):    \mathbf{RV} \mapsto \hat{\mathbf{z}}.
\end{eqnarray}
Die Zuordnung zu den Klassen von $\mathbf{y}$ ist Ziel der \hts und wird in Abh\"{a}ngigkeit des \Nutzsigs und des noch zu definierenden Parametervektors $\mathbf{p}$  erfolgen. Das \Nutzsig steht nicht unmittelbar zur Verf\"{u}gung, sondern muss aus dem Bild extrahiert werden. Die Extraktion setzt sich aus den Teilschritten Datenerfassung $S_{RV_1}$, Bildstrom-Vorverarbeitung $S_{RV_2}$ und Segmentierung $S_{RV_3}$ zusammen.

Das \Nutzsig wird mittels der Matrix $\boldsymbol{\Psi}$ dargestellt. F\"{u}r eine gute Fragestellung muss das \Nutzsig in jedem realen Versuch $\mathbf{RV}$ enthalten sein.
Der Bildstrom $\mathbf{BS}$ ist die digitale Repr\"{a}sentation  des realen Versuchs $\mathbf{RV}$. Er ist die Abtastung von $\mathbf{RV}$ bez\"{u}glich der Zeit und des Ortes und muss f\"{u}r eine erfolgreiche \hts das \Nutzsig abbilden. Daher muss f\"{u}r die Abtastung auch das Abtasttheorem \cite{Shannon98} eingehalten werden. Die Bildakquise $S_{RV_1}$ bildet somit $\mathbf{RV}$ derart auf $\mathbf{BS}$ ab, dass der biologische Effekt und damit das \Nutzsig $\boldsymbol{\Psi}$ aus dem Bildstrom extrahierbar sind:
\begin{equation}
S_{RV_1}(\mathbf{p}_{BS}):\, \mathbf{RV}\, \mapsto \, \mathbf{BS}.
\end{equation}
Der Bildstrom ist abh\"{a}ngig von den Parametern $\mathbf{p}_{BS}$. Dabei handelt es sich zum einen um die Art des Detektors und der aufgezeichneten Wellenl\"{a}nge, \dhe der Akquise-Modalit\"{a}t des Bildstroms $(m_{{BS}})_n$. Des Weiteren muss der Raum der Akquise als Parameter ber\"{u}cksichtigt werden und wird durch den Vektor $\mathbf{({u}_{{BS}}})_n=[x_{BS_1},x_{BS_2},y_{BS_1},y_{BS_2},z_{BS_1},z_{BS_2}]_n^T$ beschrieben. Schlie{\ss}lich m\"{u}ssen noch die Anzahl akquirierter Frames $(t_{{BS}})_n$ und die Aufnahmefrequenz (engl. Framerate) $(f_{{BS}})_n$ gew\"{a}hlt werden. Damit l\"{a}sst sich f\"{u}r den Bildstrom schreiben:
\begin{eqnarray}\label{eqn:Bildstromi}
    \mathbf{BS}_n&=&f(\mathbf{p}_{BS})_n\\
    \text{mit}\quad (\mathbf{p}_{BS})_n&=&(m_{{BS}};\mathbf{{u}_{{BS}}};t_{{BS}};f_{{BS}})^T_n.
\end{eqnarray}
Die wichtigste Forderung an die Bildakquise ist, dass der Bildstrom $\mathbf{BS}$ ein geeignetes \Nutzsig enth\"{a}lt. Das bedeutet, dass die Parameter des Bildstroms in Abh\"{a}ngigkeit der Parameter des \Nutzsigs gew\"{a}hlt werden. Die Parameter des \Nutzsigs $\mathbf{p}_{\Psi}$ lassen sich somit in \"{U}bereinstimmung mit
 \abb \ref{fig:Nutzsignaleigenschaften} analog zu den Parametern des Bildstroms formulieren
 \begin{eqnarray}\label{eqn:Psi_Parameter}
 \boldsymbol{\Psi}_n&=&f(\mathbf{p}_{\Psi})_n\\
 \text{mit} \quad (\mathbf{p}_{\Psi})_n&=&(m_{{\Psi}};\mathbf{{u}_{{\Psi}}};t_{{\Psi}};f_{{\Psi}})^T_n.
 \end{eqnarray}
Gut gew\"{a}hlte Akquise-Parameter des Bildstroms bilden also m\"{o}glichst ausschlie{\ss}lich das \Nutzsig ab. 

Der mit genannten Parametern akquirierte Bildstrom $\mathbf{BS}$ setzt sich aus Einzelbildern $\mathbf{I}_E$ zusammen, welche jeweils durch eine Matrix von Intensit\"{a}tswerten beschrieben werden:
\begin{eqnarray}
  \left(\mathbf{I}_{E}[i_f,i_w]\right)_{n} \quad \in \mathbb{N}^{\;I_x \times I_y \times I_z },\\
       i_w, i_f \in \mathbb{N}.\nonumber
\end{eqnarray}
Hierbei ist $i_f= 1 \dots r_F$ der Index aller Abtastungen und $i_w=1\dots r_w$ der Index der Wiederholungen. Weiter ist $I_x$ die Aufl\"{o}sung des Bildes in $x$- und $I_y$ die Aufl\"{o}sung des Bildes in $y$-Richtung, $i_z=1\dots I_z$ der Index f\"{u}r Schichtaufnahmen \bzw Fokusebenen und $i_c=1\dots I_c$, der Index der Modalit\"{a}ten. Der gesamte Bildstrom l\"{a}sst sich somit schreiben als:
\begin{eqnarray}\label{eq:Bildstrom}
  \mathbf{BS}_n=
    \begin{pmatrix}
  \mathbf{I}_{E}[\;\;1,\;\;1] & \cdots & \mathbf{I}_{E}[\;\;1\;,r_w] \\
 \vdots&\ddots&\vdots\\
  \mathbf{I}_{E}[r_F,1] & \cdots & \mathbf{I}_{E}[r_F,r_w] \\
  \end{pmatrix}_{i_c,n},\\
    \mathbf{BS}_n \quad \in \mathbb{N}^{\;I_x \times I_y \times I_z \times I_c \times r_F \times r_w}, \nonumber\\
i_c,r_F,r_w\in \mathbb{N}.\nonumber
\end{eqnarray}

\begin{figure}[!htbp]
\centering
        \centering
       \includegraphics[page=1,width=\linewidth]{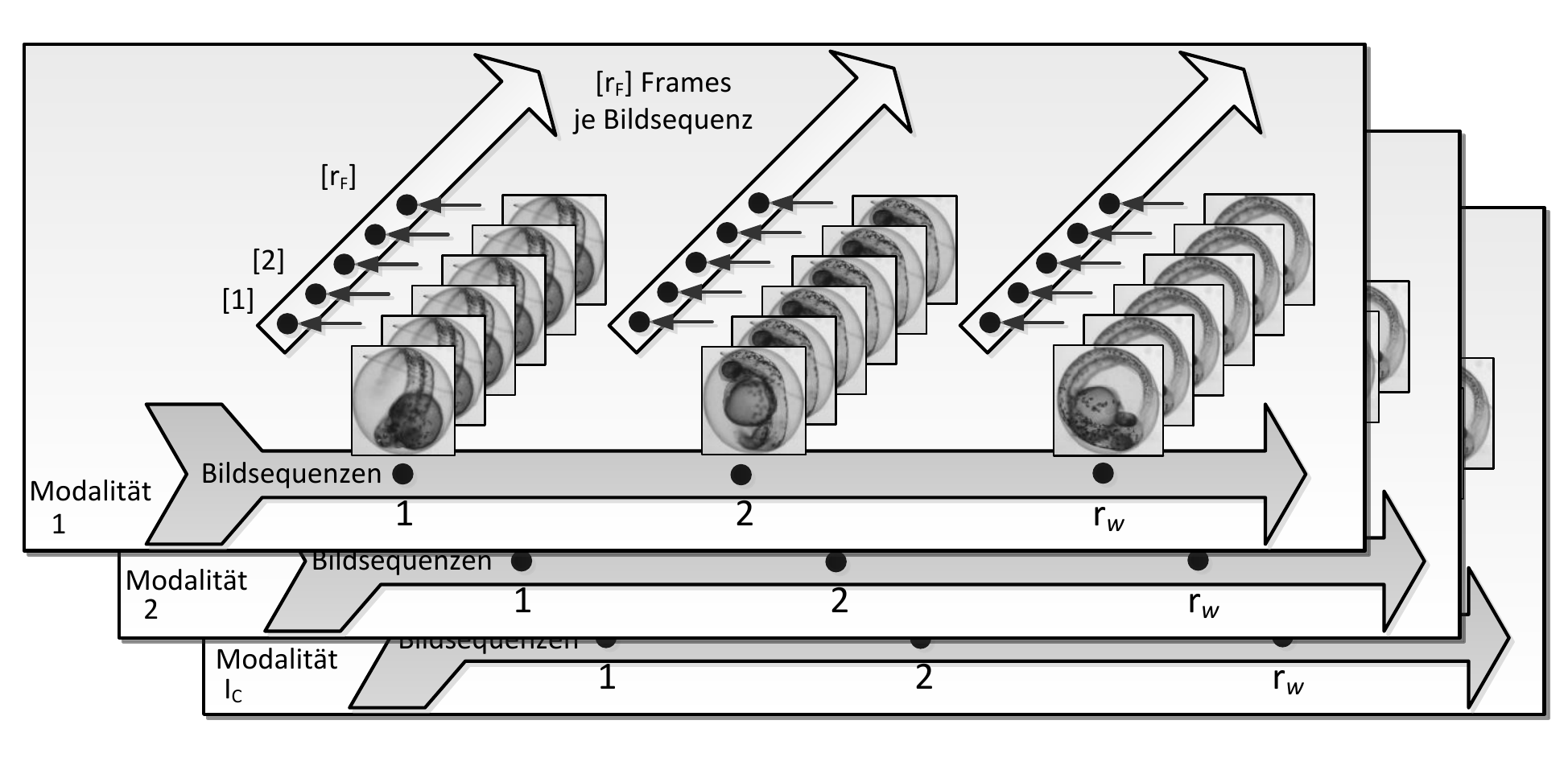}

        \caption{Grafische Veranschaulichung eines Elementes des Bildstroms $\mathbf{BS}_n$ \cite{Alshut11AT} 
        }
        \label{fig:Bildstrom}
\end{figure}

Es ergeben sich die Bild-Rohdaten des Bildstroms $\mathbf{BS}$ mit insgesamt $B$ Bildern.
\begin{equation}
  B=I_c \cdot I_z \cdot r_w \cdot r_F.
\end{equation}
Der Einfluss der Bildstromparameter ist anschaulich in \abb \ref{fig:Bildstrom} dargestellt.

In vielen F\"{a}llen ist eine Vorverarbeitung des Bildstroms sinnvoll oder notwendig. In einer Bildvorverarbeitung werden \zB Extended-Fokus-Bilder berechnet (vgl. Abschnitt~\ref{subsec:Bildakquise+Bildverarbeitung}) oder es werden Punktoperationen wie Histogrammanpassungen oder Korrekturen des Gammawertes vorgenommen. Ebenso fallen alle linearen und nichtlinearen Bildfilter-Operatoren in den genannten Bereich. Die Bildvorverarbeitung wird hier allgemein beschrieben als eine Abbildung des Bildstroms $\mathbf{BS}$ auf einen f\"{u}r die weitere Verarbeitung verbesserten Bildstrom $\mathbf{BS}^*$, welche von einem oder mehreren Filterparametern $\mathbf{p}_{f_n}$ abh\"{a}ngt. Die Parameter $\mathbf{p}_{f_n}$ sind \zB die gew\"{a}hlten Filtermasken, Schwellenwerte \usw Der Aufbau von $\mathbf{BS}^*$ ist hierbei analog zu Formel (\ref{eq:Bildstrom})
\begin{equation}\label{eqn:Filterung}
S_{RV_2}(\mathbf{p}_{f_n}):\mathbf{BS} \mapsto \mathbf{BS}^*.
\end{equation}
Im Allgemeinen ist eine Dimensionsreduktion der Daten beabsichtigt. In speziellen F\"{a}llen steigt die Dimension jedoch auch an, \zB wenn es notwendig ist, neue Bilder zu berechnen. Bildfusionen hingegen f\"{u}hren immer zu einer Dimensionsreduktion.

Die Bildverarbeitung hat nun die Aufgabe, das \Nutzsig $\boldsymbol{\Psi}$ in jedem Bild aus dem Bildstrom $\mathbf{BS}$ zu extrahieren oder zu segmentieren. Die Segmentierung $S_{RV_3}$ bildet somit $\mathbf{BS}^*$ auf $\boldsymbol{\Psi}$ ab
\begin{eqnarray}\label{eq:BSaufPSI}
S_{RV_3}(\mathbf{p}_\psi):\, \mathbf{BS^*}\, \mapsto \, \boldsymbol{\Psi} =
\begin{pmatrix}
   \mathbf{I}_{E_\Psi}[\;\;\;\;1\;,\,1] & \cdots & \mathbf{I}_{E_\Psi}[\;\;\;\;1\;,r_{w{\Psi}}]\\
 \vdots&\ddots&\vdots\\
  \mathbf{I}_{E_\Psi}[r_{F{\Psi}},1] & \cdots & \mathbf{I}_{E_\Psi}[r_{F{\Psi}},r_{w{\Psi}}]\\
\end{pmatrix}_{i_{c\Psi},n},\\
  \boldsymbol{\Psi}_n \quad \in \mathbb{N}^{\;I_{x\Psi} \times I_{y\Psi} \times I_{z\Psi} \times I_{c\Psi} \times r_{F\Psi} \times r_{w\Psi}}.
\end{eqnarray}
Hierin sind $\mathbf{I}_{E_\Psi}$ Ausschnitte jedes Einzelbildes $\mathbf{I}_E$ des originalen Bildstrom, welche nun m\"{o}glichst nur das Nutzsignal enthalten
\begin{eqnarray}
  \left(\mathbf{I}_{E_\Psi}[i_f,i_w]\right)_{n} \quad \in \mathbb{N}^{\;\;I_{x\Psi} \times I_{y\Psi} \times I_{z\Psi} },\\
      i_{w\Psi}, i_{f\Psi} \in \mathbb{N}.\nonumber
\end{eqnarray}
Der segmentierte Bildstrom setzt sich analog zum Bildstrom zusammen. Die Dimension von $\boldsymbol{\Psi}$ \"{a}ndert sich bei der Segmentierung. Da auf die Region von Interesse Bezug genommen wird, wird die Dimension gegen\"{u}ber $\mathbf{BS}^*$ meist geringer. Lediglich die Modalit\"{a}t $I_{c\Psi}$ kann durch das Errechnen bzw. Extrahieren von informationstragenden neuen Bildern, wie beispielsweise Differenzbilder, auch gr\"{o}{\ss}er werden.
Als Klassifikatoren werden Systeme bezeichnet, die in der Lage sind, Eingangsgr\"{o}{\ss}en verschiedenen Klassen zuzuordnen. Die extrahierte \NutzsigInf wird nun in Anlehnung an \cite{Mikut08,Reischl04a,Quinlan86} einer Klasse zugewiesen. Der Ablauf der Klassifikation setzt sich aus den Teilschritten Merkmalsextraktion $S_{\Psi_1}$ und $S_{\Psi_2}$, Merkmalsauswahl $S_{\Psi_3}$, Merkmalsaggregation $S_{\Psi_4}$ und der Entscheidungsfindung mittels einer Entscheidungsregel $S_{\Psi_5}$ zusammen (vgl. \abb \ref{fig:Signalmodell} unten). F\"{u}r jeden Teilschritt sind Parameter zu w\"{a}hlen, f\"{u}r die die Bezeichner $\mathbf{p}_{S\Psi}=[\mathbf{p}_{S{\Psi_1}},\mathbf{p}_{S{\Psi_2}},\mathcal{I},\mathbf{p}_{S{\Psi_4}},\mathbf{p}_{S{\Psi_5}}]^T$  eingef\"{u}hrt werden und auf die im Folgenden n\"{a}her eingegangen wird.

Es soll nun nach \abb \ref{fig:Signalmodell} mittels der \NutzsigInf $\boldsymbol{\Psi}$ jeder Versuchseinheit eine Klasse der Planfaktoren aus $\mathbf{Y}$ zugewiesen werden. Analog l\"{a}sst sich auch die Klassenzugeh\"{o}rigkeit zu einer der St\"{o}rgr\"{o}{\ss}en aus $\mathbf{Z}$ sch\"{a}tzen:
\begin{eqnarray}
S_{\psi_y}:\ \boldsymbol{\Psi} \mapsto \mathbf{\hat{y}}\\
S_{\psi_z}:\ \boldsymbol{\Psi} \mapsto \mathbf{\hat{z}}
\end{eqnarray}
Da eine direkte L\"{o}sung des Problems anhand der Bilddaten nicht m\"{o}glich ist, wird die \NutzsigInf durch Ersatzgr\"{o}{\ss}en, die Merkmale, charakterisiert. Die Aufgabe ist die Berechnung niederdimensionaler und informationstragender, \dhe das \Nutzsig  $\boldsymbol{\Psi}$ beschreibender, Merkmale. Die wichtigsten Merkmale basieren auf Pixeln, Kanten, Texturen, Regionen, Objekten und Szenen (vgl. \cite{Lehmann02,Mikut08}).

Die Merkmalsextraktion transformiert die Daten der \NutzsigInf f\"{u}r jeden Abtastzeitpunkt in einen $m_1$-dimensionalen Merkmalsraum in Abh\"{a}ngigkeit der Parameter $\mathbf{p}_{S\Psi_1}$. F\"{u}r jede Zebrab\"{a}rblingslarve enth\"{a}lt die Matrix daher Merkmalswerte $y_{ZR}$ zu jedem Abtastzeitpunkt und somit eine Zeitreihe $\mathbf{Y}_{ZR}$ f\"{u}r jedes Merkmal. Diese Zeitreihen werden f\"{u}r die \hts \FischInfZeit genannt.
\begin{eqnarray}
    S_{\Psi_1}(\mathbf{p}_{S\Psi_1}):\ \boldsymbol{\Psi} \mapsto \mathbf{Y}_{ZR}=
     \begin{pmatrix}
  {y}_{ZR}[\;\;1,\;\;1] & \cdots &{y}_{ZR}[\;\;1\;,r_w]\\
  \vdots & \ddots & \vdots \\
  {y}_{ZR}[r_F,1] & \cdots & {y}_{ZR}[r_F,r_w]\\
  \end{pmatrix}_{m_1},\\
    \mathbf{Y}_{ZR} \quad \in \mathbb{R}^{\;r_F \times r_w \times m_1 }.\nonumber
    \label{eq:Merkex}
\end{eqnarray}
Die \FischInfZeit werden durch eine Dimensionsreduktion wiederum mittels Merkmalen beschrieben. Es entsteht schlie{\ss}lich ein $m_2$-dimensionaler Merkmalsraum in Abh\"{a}ngigkeit der Parameter $\mathbf{p}_{S\Psi_2}$. Solche Merkmale werden f\"{u}r die \hts \FischInfMerke genannt
\begin{eqnarray}
    S_{\Psi_2}(\mathbf{p}_{S\Psi_2}): \mathbf{Y}_{ZR} \mapsto
    \mathbf{x}^T=\left(x_1, \dots, x_{m_2}\right),\\
    \mathbf{x}\in \mathbb{R}^{1\times m_2}
    \label{eq:Merkex1}
\end{eqnarray}
Die extrahierten Merkmale sind h\"{a}ufig redundant oder enthalten \zT keine Information, die sich f\"{u}r eine Klassifikation eignet. Daher wird eine Merkmalsauswahl angewandt, die sich eines Kriteriums bedient, um die aussagekr\"{a}ftigsten Merkmale zu identifizieren. Eine \"{U}bersicht zu solchen Kriterien ist in \cite{Ahrens74,Mikut01} zu finden. Der Bezeichner $\mathcal{I}$ steht f\"{u}r die Indexmenge der ausgew\"{a}hlten $m_3$ Merkmale und es ergibt sich die Abbildung
\begin{eqnarray}\label{eq:Abbildung_S3}
    S_{\Psi_3}:\ \mathbf{x} \mapsto \mathbf{x}|_\mathcal{I}=
\left(x_1, \dots, x_{m_3}\right),\\
\mathbf{x}|_\mathcal{I} \in \mathbb{R}^{1 \times {m_3}},\; {m_2}\geq {m_3}.\nonumber
    \label{eq:Merkex2}
\end{eqnarray}
Die Merkmalsaggregation dient zur Reduktion der Dimension des Merkmalsraumes. Hierdurch werden zudem Klassifikationen, die auf Merkmalsr\"{a}umen mit $m_3>3$ basieren, grafisch darstellbar. Es existieren verschiedene Verfahren unterschiedlicher Arbeitsweise, von denen die bekanntesten die Hauptkomponentenanalyse (HKA) und die lineare Diskriminanzanalyse (LDA) sind. Eine \"{U}bersicht findet sich in \cite{Carreira-Perpinan97,Jain00,Scholkopf96,Reischl04a}. Beide Verfahren liefern als Ergebnis einen $m_d$-dimensionalen Merkmalsraum, welcher eine Linearkombination aller $m_3$ Merkmale ist und von dem Parameter $\mathbf{p}_{S\Psi_4}$ der gew\"{a}hlten Aggregation abh\"{a}ngt.
\begin{eqnarray}\label{eq:Abbildung_S4}
    S_{\Psi_4}(\mathbf{p}_{S\Psi_4}): \mathbf{x}|_\mathcal{I} \mapsto \mathbf{w}\\
\mathbf{w} \in \mathbb{R}^{\, 1 \times m_d}\nonumber
, \quad m_3 \geq m_d.
    \label{eq:Merkagg}
\end{eqnarray}
Den abschlie{\ss}enden Schritt des Klassifikatorentwurfs stellt die Konstruktion der Entscheidungsregel dar, die den entstandenen Merkmalssatz einer der Klassen von $\mathbf{Y}$ \bzw $\mathbf{Z}$, in Abh\"{a}ngigkeit der Klassifikationsparameter $\mathbf{p}_{S\Psi_5}$, zuweist. Da es sich hierbei um eine Sch\"{a}tzung handelt, wird die Zuordnung $\mathbf{\hat y}$ \bzw  $\mathbf{\hat z}$ bezeichnet
\begin{eqnarray}\label{eq:Abbildung_S5}
    S_{{\Psi_5y}}(\mathbf{p}_{S\Psi_5}):&  \mathbf{w} \mapsto \mathbf{\hat{y}},&\\
        S_{\Psi_5z}(\mathbf{p}_{S\Psi_5}):&  \mathbf{w} \mapsto  \mathbf{\hat{z}}.&
    \label{eq:Entscheidung}i
\end{eqnarray}
Nun sind alle Versuchsparameter des Versuchsplans der \hts bestimmt und werden in dem Parametervektor $\mathbf{p}$ zusammengefasst
\begin{equation}
  \mathbf{p}=(\mathbf{p}_{y};\mathbf{p}_{z};\mathbf{p}_{BS};\mathbf{p}_{\Psi};\mathbf{p}_{fn};\mathbf{p}_{S\Psi})^T.
\end{equation}

Die gesamte Abbildung ergibt dann:
\begin{eqnarray}
\mathbf{RV}\xrightarrow{S_y}\mathbf{\hat{y}} \mathop{\widehat{=}} \mathbf{RV} \xrightarrow{S_{RV_1}} \mathbf{BS} \xrightarrow{S_{RV_2}} \mathbf{BS^*} \xrightarrow{S_{RV_3}} \boldsymbol{\Psi} \xrightarrow{S_{\Psi_1}} \mathbf{Y}_{ZR} \xrightarrow{S_{\Psi_2}} \mathbf{x} \xrightarrow{S_{\Psi_3}} \mathbf{x}|\mathcal{I} \xrightarrow{S_{\Psi_4}} \mathbf{w}\xrightarrow{S_{\Psi_5}}\mathbf{\hat{y}}\,\\
\mathbf{RV}\xrightarrow{S_z}\mathbf{\hat{z}} \mathop{\widehat{=}}
\underbrace{\mathbf{RV} \xrightarrow{S_{RV_1}} \mathbf{BS} \xrightarrow{S_{RV_2}} \mathbf{BS^*} \xrightarrow{S_{RV_3}}}_
{\text{Bildverarbeitung}}
 \boldsymbol{\Psi}
\underbrace{
 \xrightarrow{S_{\Psi_1}} \mathbf{Y}_{ZR} \xrightarrow{S_{\Psi_2}} \mathbf{x} \xrightarrow{S_{\Psi_3}} \mathbf{x}|\mathcal{I} \xrightarrow{S_{\Psi_4}} \mathbf{w}\xrightarrow{S_{\Psi_5}}\mathbf{\hat{z}}}
 _{\text{Merkmalsextraktion/Klassifikation}}.
\end{eqnarray}


%

\section{Anforderungsgerechte Versuchsauslegung}\label{sec:Anforderungsgerechte_Anwendung}

\begin{figure}[htbp]
        \centering
        \includegraphics[page=4]{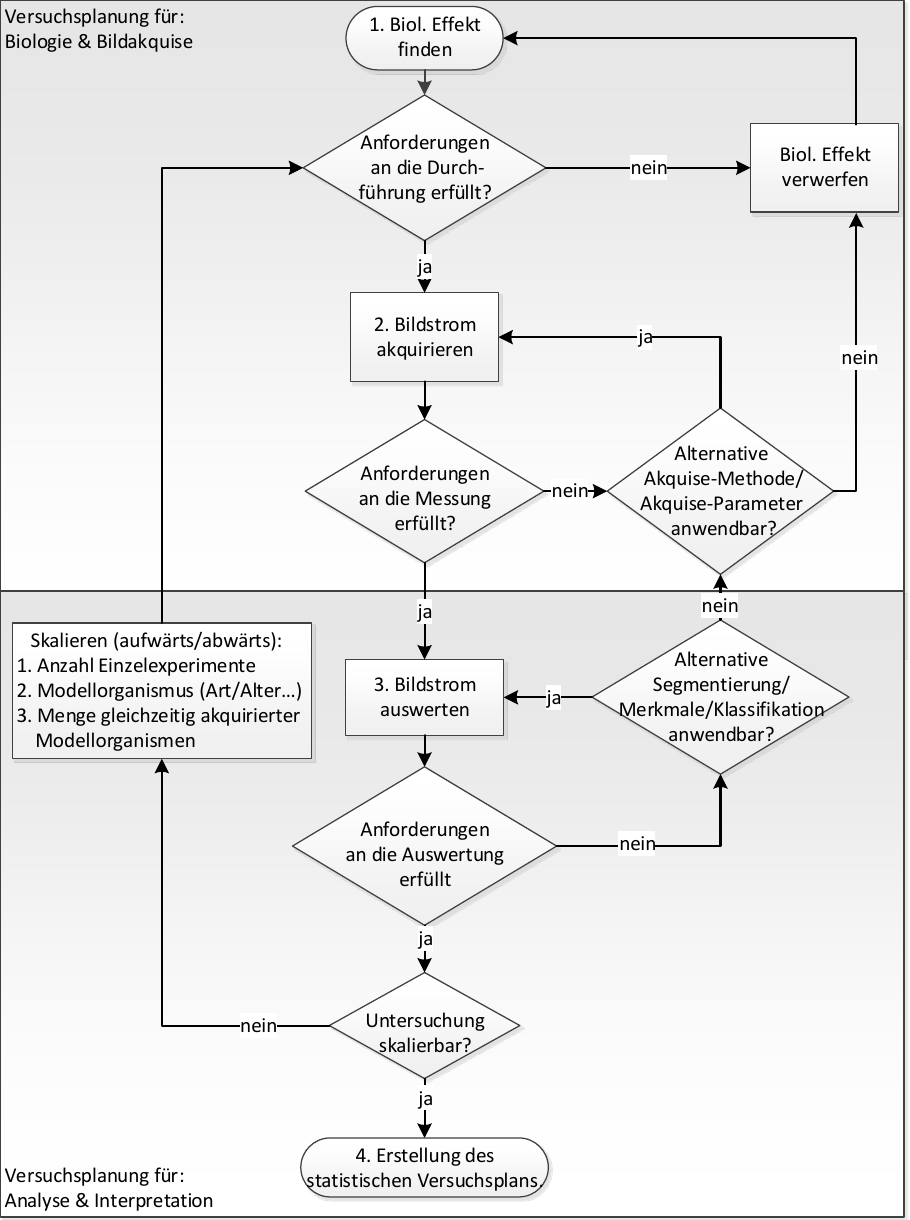}
\caption[Flussdiagramm f\"{u}r die anforderungsgerechte Strukturierung]{Flussdiagramm f\"{u}r die anforderungsgerechte Strukturierung des Versuchsplans von \htsen}
\label{fig:Ablaufdiagramm_Screendesign}
\end{figure}

Nach der Formulierung der Anforderungen in Abschnitt \ref{sec:HDU_Anforderungen} und dem Aufzeigen der zur Verf\"{u}gung stehenden Versuchs- und Auswerteparameter besteht eine schwierige Aufgabe darin, diejenigen Parameter auszuw\"{a}hlen, welche die Anforderungen auch erf\"{u}llen. Daher wird, um die Struktur der \hts den Anforderungen  entsprechend zu erstellen, im folgenden Abschnitt ein Flussdiagramm zur erfolgreichen Gestaltung eines Hochdurchsatzversuchs vorgestellt, welches in \abb \ref{fig:Ablaufdiagramm_Screendesign} dargestellt ist. Beim Durchlaufen des Diagramms werden Schritt f\"{u}r Schritt Parameter festgelegt und jeweils gepr\"{u}ft, ob die Auslegung den Anforderungen gerecht wird.

In diesem Flussdiagramm werden vier Versuchspl\"{a}ne erstellt, die zur erfolgreichen Strukturierung erforderlich sind:
 \begin{enumerate}
  \item Versuchsplan f\"{u}r die Identifikation des biologischen Effekts, \dhe Ermittlung des Nutzsignals,
  \item Versuchsplan f\"{u}r die Akquise des Nutzsignals, \dhe Durchf\"{u}hrung der Messung,
  \item Versuchsplan f\"{u}r die Extraktion und Klassifikation der Merkmale, \dhe Auswertung des Bildstroms,
  \item Versuchsplan f\"{u}r die Durchf\"{u}hrung der \hts, \dhe die Erstellung des statistischen Versuchsplans.
\end{enumerate}

Innerhalb der ersten drei Schritte werden, f\"{u}r gew\"{a}hlte Parameter, die in Abschnitt \ref{sec:HDU_Anforderungen} formulierten Anforderungen gepr\"{u}ft, bevor der 4. Schritt, die Durchf\"{u}hrung der \hts, vollzogen werden kann. Dabei k\"{o}nnen nicht alle Anforderungen in der Reihenfolge, in der sie in den Kategorien beschrieben sind, gepr\"{u}ft werden. Beispielsweise ist die Pr\"{u}fung auf Skalierbarkeit erst m\"{o}glich, wenn alle anderen Parameter bereits gew\"{a}hlt sind, da jeder Abschnitt sich f\"{u}r die hohe St\"{u}ckzahl an Einzelversuchen eignen muss.
 Die zu pr\"{u}fenden Kategorien sind nach Abschnitt \ref{sec:HDU_Anforderungen}:
 \begin{itemize}
  \item Anforderungen an die Durchf\"{u}hrung,
  \item Anforderungen an die Messung,
  \item Anforderungen an die Auswertung
  \item und Skalierbarkeit der \hts.
\end{itemize}

Der 1. Schritt, das Finden des biologischen Effekts (des \Nutzsigs), ist vollst\"{a}ndig biologisch motiviert. Die gesetzlichen Vorschriften m\"{u}ssen beachtet werden, eine permanente Pr\"{u}fung bei der Auswahl neuer Verfahren ist vorzusehen. Die biologische Realisierbarkeit zeigt sich im Vorversuch, \dhe der Biologe muss verst\"{a}ndlicherweise in der Lage sein, den Effekt, der im Hochdurchsatz untersucht werden soll, beispielsweise durch eine Stimulation manuell zu erzeugen. Entscheidend f\"{u}r die technische Durchf\"{u}hrbarkeit der \hts ist, ob das \Nutzsig reproduziert werden kann. Die Reproduzierbarkeit kann durch mehrfaches Wiederholen des biologischen Versuchs gepr\"{u}ft werden. Ist dies nicht gegeben, so ist mit dem jeweiligen Effekt keine \hts m\"{o}glich und ein anderer, besser geeigneter Effekt muss gew\"{a}hlt werden. Der Schritt pr\"{u}ft die Parameter der beiden Arbeitsschritte  "`Versuchsplanung"' und "`Vorbereitung"' in \abb~\ref{fig:Assay_Gestaltung}.

Wurde ein reproduzierbarer biologischer Effekt gefunden, wird versucht, das \Nutzsig zu messen, \dhe mittels einer der erw\"{a}hnten Methoden zu akquirieren. Um den Anforderungen \emph{Eindeutige Pr\"{a}senz der biologisch relevanten Information} und \emph{R\"{u}ckwirkungsfreiheit der Messung} aus Abschnitt \ref{sec:HDU_Anforderungen} zu entsprechen, muss die gew\"{a}hlte Methode dabei:
 \begin{itemize}
   \item das \Nutzsig robust abbilden,
   \item ein gutes \Nutzsig-Rausch-Verh\"{a}ltnis aufweisen,
   \item die Probe wenig beeinflussen,
   \item dem Abtasttheorem gerecht werden.
 \end{itemize}

Die Pr\"{u}fung kann weitgehend durch einfaches Ansehen der Bilder erfolgen. Sind die auszuwertenden Bereiche beispielsweise ein Gewebe in der Larve, so ist auf einen guten Kontrast dieser Bereiche zu achten. Beim \"{U}berpr\"{u}fen der gew\"{a}hlten Abtastfrequenz und bei mit niedriger Frequenz auftretenden \Nutzsigen muss deren Frequenz zuvor mittels entsprechender Messeinrichtungen ermittelt werden. Eine m\"{o}glichst vollst\"{a}ndige Repr\"{a}sentation des \Nutzsigs im akquirierten Bildstrom muss angestrebt werden. Die gro{\ss}e Bandbreite an m\"{o}glichen biologischen Effekten und damit \Nutzsigen  l\"{a}sst eine allgemein g\"{u}ltige Pr\"{u}fmethode nicht zu. Die Erfahrung zeigt jedoch, dass sich Daten qualitativ f\"{u}r eine automatische Auswertung eignen, wenn es einem unge\"{u}bten menschlichen Betrachter m\"{o}glich ist, das \Nutzsig im Bildstrom leicht zu identifizieren und \ggf zu markieren. Andernfalls muss eine alternative Methode zur Bildakquise angewandt werden oder der biologische Effekt muss verworfen werden (vgl. \abb \ref{fig:Ablaufdiagramm_Screendesign}). Die Akquise findet sich in \abb~\ref{fig:Assay_Gestaltung}  im Block "`Mikroskopie"'. Das Ergebnis der Akquise ist der Bildstrom.

Ist das \Nutzsig robust im Bildstrom enthalten, besteht der dritte Schritt darin, Methoden der Bildverarbeitung anzuwenden und die Anforderungen an die Auswertung zu pr\"{u}fen. Hier zeigt sich, ob die Pr\"{a}senz des \Nutzsigs mit der Akquise-Methode \bzw den gew\"{a}hlten Akquise-Parametern des 2. Schritts ausreichend ist. Es ist zu pr\"{u}fen, ob:
\begin{itemize}
    \item das \Nutzsig durch die Segmentierung von anderen Informationen  im Datensatz getrennt werden kann,
  \item die Segmentierung ausreichend robust gegen\"{u}ber Helligkeit, Reflexionen, Lage des Objekts \ua ist,
  \item der Berechnungsaufwand vertretbar ist.
\end{itemize}

Nach der Extraktion von Merkmalen aus dem Bildstrom sind die Merkmale ebenfalls auf das Erreichen der Anforderungen aus Abschnitt \ref{sec:HDU_Anforderungen} zu pr\"{u}fen.  Merkmale m\"{u}ssen zwar, je nach \Nutzsig der \hts, unterschiedlichen Bedingungen gerecht werden, dennoch lassen sich aus der Praxis allgemein g\"{u}ltige Pr\"{u}fungen formulieren:
\begin{enumerate}
  \item Vom \Nutzsig unabh\"{a}ngige Gr\"{o}{\ss}en (wie etwa die Beleuchtung) d\"{u}rfen die Merkmale nicht oder nur unwesentlich beeinflussen.
  \item Der Berechnungsaufwand soll m\"{o}glichst gering sein.
  \item Die multivariate Trenng\"{u}te muss gew\"{a}hrleistet sein, \dhe die Kombination der Merkmale muss eine gute Unterscheidung der Klassen des \Nutzsigs erlauben.
  \item Die Auspr\"{a}gungen der Merkmale m\"{u}ssen m\"{o}glichst robust sein und daher eine geringe Empfindlichkeit gegen\"{u}ber Rauschen, Messfehlern oder zeitvarianten \"{A}nderungen aufweisen.
  \item Das Ergebnis muss sich \"{u}bersichtlich pr\"{a}sentieren lassen.

\end{enumerate}
Sollte einer der genannten Punkte nicht erf\"{u}llt sein, muss, wie in Abbildung \ref{fig:Ablaufdiagramm_Screendesign} ablesbar, gepr\"{u}ft werden, ob mittels einer anderen Akquise-Methode aussagekr\"{a}ftige Merkmale extrahiert werden k\"{o}nnen. Falls nicht, so muss der biologische Effekt verworfen werden.

Nach erfolgreicher Merkmalsextraktion ist das \Nutzsig durch Zahlenwerte abgebildet und es muss abschlie{\ss}end gepr\"{u}ft werden, ob der Umfang der \hts mit der ausgearbeiteten L\"{o}sung durchgef\"{u}hrt werden kann. Auf der Basis der bis zum jetzigen Zeitpunkt verf\"{u}gbaren Informationen kann eine Hochrechnung f\"{u}r die wichtigsten Rahmenbedingungen f\"{u}r den Versuch erfolgen. Zu pr\"{u}fende Werte sind die Akquise-Dauer, der notwendige Speicherplatz, die notwendige Rechenkapazit\"{a}t und zur Verf\"{u}gung stehende Computer-Hardware, die zur Verf\"{u}gung stehenden Modellorganismen, der Durchsatz der Mikroskop-Plattformen, die Arbeitskraft f\"{u}r die manuell durchzuf\"{u}hrenden Schritte sowie die Pr\"{a}sentierbarkeit der Ergebnisse. Die berechneten Umf\"{a}nge m\"{u}ssen mit den zur Verf\"{u}gung stehenden Mitteln verglichen werden. Ist die Untersuchung nicht durchf\"{u}hrbar, m\"{u}ssen Parameter skaliert werden. Eine M\"{o}glichkeit bietet \zB das Anpassen der Anzahl von Einzelexperimenten. Bez\"{u}glich des Modellorganismus kann auf einen gr\"{o}{\ss}eren oder kleineren Organismus ausgewichen werden. Beim Zebrab\"{a}rbling l\"{a}sst sich \zB die Gr\"{o}{\ss}e durch das Alter skalieren, \dhe durch das Verwenden j\"{u}ngerer oder \"{a}lterer Fische \bzw Larven. Auch die Anzahl gleichzeitig akquirierter Modellorganismen kann variiert werden. Ebenso ist das Verwenden von mehreren Mikroskopen zur Erh\"{o}hung des Durchsatzes m\"{o}glich.
%

Wichtig bei der Skalierung ist, dass sich bei Anwendung einer der aufgezeigten L\"{o}sungsvorschl\"{a}ge (oder auch anderer) eine Ver\"{a}nderung auf jeden Schritt der Untersuchung auswirkt, wie im Konzept in Abschnitt \ref{sec:Konzept_Versuchsauslegung} erarbeitet. Daher muss nach jeder Iteration das \emph{gesamte} Flussdiagramm, mit den genannten Pr\"{u}fungen, nochmals durchlaufen werden. Erst wenn die \hts in der gew\"{u}nschten Weise ohne weitere Anpassungen durchf\"{u}hrbar ist, darf mit der tats\"{a}chlichen Ausf\"{u}hrung des Experiments begonnen werden. Ein typischer Fehler ist es, w\"{a}hrend der Durchf\"{u}hrung den Prozess, \zB durch Parallelisierung, ohne erneute Pr\"{u}fung zu beschleunigen. Ein solches Vorgehen f\"{u}hrt oftmals zu St\"{o}rfaktoren, deren Einfluss sich im Nachhinein nur schwer beseitigen l\"{a}sst oder gar zum Scheitern der \hts f\"{u}hrt.

\section{Bewertung}

Das Ziel des in Kapitel \ref{chap:Neues_Konzept} vorgestellten neuen Konzeptes ist es, den Erfolg einer automatisierten Auswertung von \htsen sicherzustellen, was durch eine gesamtheitliche Betrachtung des Versuchs bei der Auswahl der Versuchs- und Auswerteparameter erreicht wird. Hierf\"{u}r wurde ein breites Spektrum solcher Parameter identifiziert, eine allgemein anwendbare formale Notation eingef\"{u}hrt und hieraus wurden m\"{o}glichst allgemein formulierte Anforderungen f\"{u}r eine erfolgreiche Versuchsauslegung abgeleitet. Mittels des abschlie{\ss}end vorgestellten Flussdiagramms wird eine strukturierte Vorgehensweise zur Umsetzung des Konzeptes unter Einhaltung der Anforderungen f\"{u}r die Praxis geliefert. Im folgenden Kapitel wird ein Modulkatalog vorgestellt, der f\"{u}r die Umsetzung des vorgeschlagenen Konzeptes die notwendigen Werkzeuge enth\"{a}lt.

\chapter{Modulkatalog f\"{u}r die Auswertung und Pr\"{a}sentation von \htsen am Zebrab\"{a}rbling}\label{sec:BV_Module}
\section{Einf\"{u}hrung}
Das in der Kapitel \ref{chap:Einleitung} aufgezeigte breit gef\"{a}cherte Anwendungsfeld der \htsen wird \"{u}ber Jahre hinweg neue \Nutzsige am Zebrab\"{a}rbling hervorbringen, f\"{u}r die eine bildbasierte \hts eine geeignete und schnelle Analysemethode ist \cite{MacRae12}. In allen F\"{a}llen wird nach der Bildakquise ein Bildstrom vorliegen, der analysiert und interpretiert werden muss. Jedes weitere \Nutzsig ist eine neue Herausforderung an jeden Teilbereich der Auswertung, was eine universell anwendbare L\"{o}sung f\"{u}r alle bekannten wie zuk\"{u}nftigen Fragestellungen ausschlie{\ss}t. Die Analyse der in der Literatur zu findenden \htsen sowie die Erfahrung aus den f\"{u}r die vorliegende Arbeit durchgef\"{u}hrten Untersuchungen (vgl. Kapitel \ref{chap:Anwendung}) zeigen jedoch, dass bei \htsen von biologischen Datens\"{a}tzen verschiedene Methoden sowie M\"{o}glichkeiten der Vereinfachung aus Sicht der Bildverarbeitung wiederholt vorkommen. Der hohe Durchsatz erm\"{o}glicht es, auf aufw\"{a}ndige Auswertungen schlechter oder schwieriger Daten von Einzelversuchen zu verzichten, solche Daten zu verwerfen und mit der Auswertung der \"{u}brigen Einzelversuche fortzufahren. Eine kurze Auswertedauer und ein gutes, \"{u}bersichtliches Datenmanagement sowie die nachvollziehbare Darstellung des L\"{o}sungsweges sind von gr\"{o}{\ss}erer Bedeutung als die perfekte Analyse des Einzelversuches. Durch die massenhafte Pr\"{a}paration und Akquise ist zudem eine hohe Schwankung der Qualit\"{a}t der Daten unumg\"{a}nglich. Die Schwankung resultiert \ua  aus unterschiedlichen F\"{u}llh\"{o}hen der N\"{a}pfchen mit Fl\"{u}ssigkeit, schwankender Beleuchtung, schwankendem Alter der Larven und Bewegungen der Larven w\"{a}hrend der Aufnahme (\vgl \kap \ref{subsec:Bildqualit\"{a}t}).

\begin{figure}[htbp]
\centering
        \centering
               \includegraphics[page=6,
              width=\linewidth
               ]{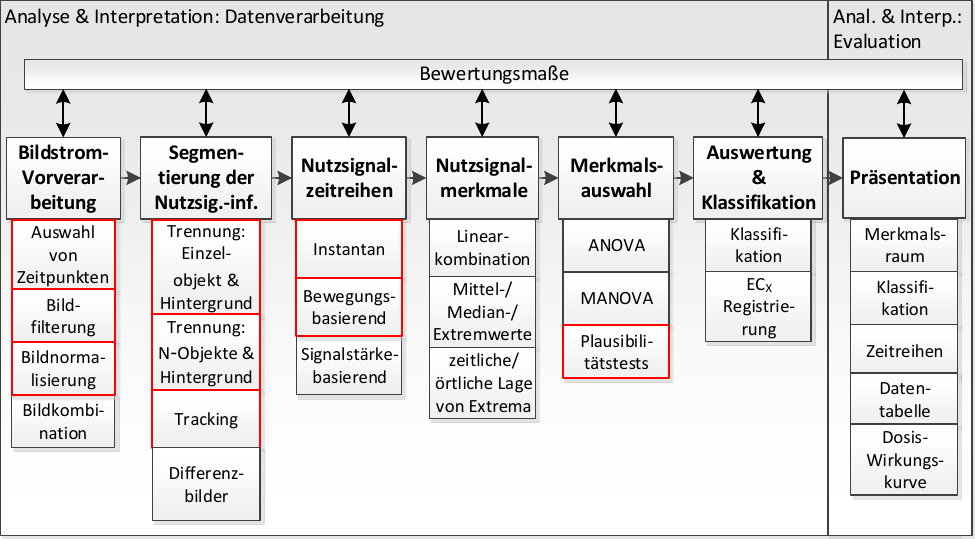}

        \caption[Modulkatalog]{Modulkatalog mit Kategorien und den \"{u}bergeordneten Bewertungsma{\ss}en. In der vorliegenden Arbeit neu entwickelte Module sind rot gekennzeichnet.}
        \label{fig:Prozess_Uebersicht}
\end{figure}

Trotz der genannten Schwierigkeiten ist es f\"{u}r jede \hts m\"{o}glich, die Auswertungsschritte, die am Bildstrom vollzogen werden, in wenige Kategorien zusammenzufassen. F\"{u}r jede der Kategorien schl\"{a}gt die vorliegende Arbeit eine Reihe von L\"{o}sungen vor. Die L\"{o}sungen bestehen f\"{u}r jede Kategorie aus zu Modulen zusammengefassten Methoden, welche in den folgenden Abschnitten vorgestellt werden. Alle Module einer Kategorie haben gemeinsam, dass sie die gleichen Ein"~ und Ausgangsgr\"{o}{\ss}en besitzen. F\"{u}r die jeweils vorliegende biologische Fragestellung m\"{u}ssen aus jeder Kategorie ein oder mehrere passende Module gew\"{a}hlt und spezifiziert werden. Die vorkommenden Datenverarbeitungsmethoden wurden entweder in der vorliegenden Arbeit entwickelt, wie \zB die neuartige Normalisierung f\"{u}r \htsen (\kap \ref{subsec:Modulkatalog_Normalisierung}), die Trennung von Einzelobjekt (Fischei) vom Hintergrund (\kap \ref{subsec:Segmentierung}) und das neue Verfahren zum Tracking des Chorions (\kap \ref{subsec:Tracking}) oder es handelt sich um bereits bekannte Methoden, die f\"{u}r die \hts modifiziert wurden. In \abb~ \ref{fig:Prozess_Uebersicht} ist der Modulkatalog visualisiert und die neu entwickelten Module sind rot hervorgehoben.  Das vorliegende Kapitel stellt alle Module jeder Kategorie zu einem Modulkatalog zusammen. Der Vorteil des Kataloges ist, dass es mit dessen Hilfe m\"{o}glich wird, einen Gro{\ss}teil der bekannten als auch zuk\"{u}nftigen Problemstellungen von bildbasierten \htsen ohne gro{\ss}en Entwicklungsaufwand zu l\"{o}sen. Lediglich die Auswahl geeigneter Module und die Anpassung der jeweiligen Auswerteparameter auf die vorliegende Fragestellung sind durch den Anwender durchzuf\"{u}hren.

Der Ablauf der Datenverarbeitung, die Analyse und Interpretation des im Bildstrom enthaltenen \Nutzsigs, l\"{a}sst sich gem\"{a}{\ss} \abb \ref{fig:Prozess_Uebersicht} in sieben Kategorien aufspalten, auf die in den folgenden Abschnitten eingegangen wird. Die Entwurfsphase der Auswertung verl\"{a}uft nicht schematisch, sondern ist ein iterativer Prozess, wie in \abb \ref{fig:Ablaufdiagramm_Screendesign} vorgestellt wurde. Dabei werden anhand der Zwischenergebnisse Module, Auswerteparameter und Bewertungsma{\ss}e variiert. Die Bewertungsma{\ss}e sind \zB die G\"{u}te der Klassifikationen oder die visuelle Kontrolle der Ergebnisse der Segmentierung. Sie nehmen Einfluss auf die genannte Variation und basieren auf den Zwischen"~ und Endergebnissen der Bl\"{o}cke der Modulkategorien. Sie helfen bei der Findung der am besten geeigneten Struktur und Auswerteparameter sowie bei der Einsch\"{a}tzung der Qualit\"{a}t der L\"{o}sung \cite{Mikut08}.

Zum besseren Verst\"{a}ndnis der Module werden einige durch konkrete Beispiele veranschaulicht. Bei der Vorstellung der Module wird zudem bei Beispielen immer versucht, die Beispiele und Datens\"{a}tze m\"{o}glichst einfach zu halten. Bildverarbeitungsmodule werden beispielsweise bevorzugt anhand von einzelnen Grauwertbildern entsprechend Formel~(\ref{eqn:2D_Bild})  beschrieben, auch wenn die Module in der Praxis auf den wesentlich komplexeren Bildstrom nach Formel~(\ref{eq:Bildstrom}) angewandt werden. Sollte die Anwendung auf einen Datensatz h\"{o}herer Dimension nicht m\"{o}glich sein, wird an der entsprechenden Stelle darauf hingewiesen.

\section{Bildstrom-Vorverarbeitung}\label{sec:BildstromVorverarbeitung}
Notwendige Voraussetzung f\"{u}r eine erfolgreiche Auswertung ist ein Bildstrom, dessen Bilddaten ein konsistentes und ausreichend starkes \Nutzsig enthalten. Durch den Einfluss der vielen bereits in Abschnitt \ref{sec:Beschreibung_Parameter} diskutierten St\"{o}rungen ist dies in realen Problemstellungen oft nicht unmittelbar gegeben. Des Weiteren m\"{u}ssen die Rohdaten bei Schichtaufnahmen erst zu einem verarbeitbaren Datensatz zusammengesetzt werden. Die Bild-Vorverarbeitung hat zum Ziel, den Bildstrom f\"{u}r die weitere Verarbeitung aufzubereiten \bzw zu verbessern.

Eingang der Modulkategorie \emph{Bildstrom-Vorverarbeitung} sind die Rohdaten, \dhe der unverarbeitete Bildstrom $\mathbf{BS}$, wie er bei der Akquise aufgezeichnet wurde. Als Ausgang steht nach erfolgreicher Verarbeitung ein verbesserter Bildstrom $\mathbf{BS^*}$ zur Verf\"{u}gung, der sich f\"{u}r die Segmentierung eignet und dessen Dimension so gro{\ss} wie n\"{o}tig, jedoch so gering wie m\"{o}glich ist.

\subsection{Auswahl von Zeitpunkten}\label{subsec:Auswahl_Zeitpkt}
Der erste Unterpunkt der Bildstrom-Vorverarbeitung in \abb \ref{fig:Prozess_Uebersicht}, die Auswahl von Zeitpunkten, steht f\"{u}r das Ausw\"{a}hlen nur bestimmter Aufnahmen zu bestimmten Zeitpunkten aus dem Bildstrom. Er stellt eine Dimensionsreduktion des Bildstroms $\mathbf{BS}$ dar. Hierbei wird ein gro{\ss}er Bildstrom \zB in mehrere Teile unterteilt oder auf der Basis von Vorwissen oder durch Sichtpr\"{u}fung auf einen wesentlichen Teil beschr\"{a}nkt. Die Methode kommt vor allem dann zum Einsatz, wenn der Bildstrom zur Auswertung mehrerer \Nutzsige akquiriert wurde und der auszuwertende Planfaktor im Bildstrom redundant vorhanden ist. Ein Beispiel sind Bewegungssequenzen, die zur Detektion von Herzschl\"{a}gen aufgezeichnet wurden. Mit einem solchen Datensatz lassen sich neben den Bewegungen auch instantane \Nutzsige wie \zB die entwickelte Gr\"{o}{\ss}e bestimmen. F\"{u}r solche \Nutzsige ist aus der Sequenz bereits ein einziges Bild ausreichend, daher k\"{o}nnen die anderen zeitlichen Daten bei der Analyse ausgeschlossen werden. Ein Auswerten aller Bilder des Bildstroms w\"{u}rde wenig neue Information zur Klasseneinteilung beitragen. In der Praxis zeigt sich zudem, dass manche Detektoren f\"{u}r die Erfassung von Bildern eine kurze Zeit ben\"{o}tigen, um die Betriebstemperatur zu erreichen, was zur Folge hat, dass die ersten Aufnahmen im Bildstrom \zT verf\"{a}lscht dargestellt werden. Daher ist es ratsam, bei der Auswahl nicht intuitiv die erste(n) Aufnahme(n), sondern zeitlich sp\"{a}ter aufgezeichnete Bilder zu w\"{a}hlen, oder, wenn das Nutzsignal es zul\"{a}sst, die Auswahl der Bilder nach einer festgesetzten Vorschrift zu vollziehen. Die Auswahl der Bilder erfolgt somit zumeist nach einer manuell festgelegten Liste oder nach einer automatischen Auswahlvorschrift, wie \zB der Ermittlung der sch\"{a}rfsten Schichtaufnahme mittels Wavelet-Transformation \cite{kautsky02}.

In jedem Fall wird eine Indexmenge $\mathcal{I}_{BV}$ bestimmt, die aus dem Bildstrom nur bestimmte Bilder ausw\"{a}hlt. Das "`Weglassen"' der restlichen Bilder ist hier eine einfache M\"{o}glichkeit zur Verringerung von Dimension und Gr\"{o}{\ss}e des Datensatzes. Es sei jedoch darauf hingewiesen, dass der Schritt der Bildauswahl bei nicht sorgf\"{a}ltiger Auswahl die Messgenauigkeit verringert. Die Auswahl muss daher immer mithilfe von Vorversuchen und Bewertungsma{\ss}en \"{u}berpr\"{u}ft werden.

\subsection{Neues Verfahren zur Bildfilterung und Normalisierung f\"{u}r inhomogene Datens\"{a}tze}\label{subsec:Modulkatalog_Normalisierung}
Einige Einfl\"{u}sse von St\"{o}rfaktoren bei \htsen lassen sich mit vergleichsweise wenig A"~priori"~Wissen durch Methoden der Bildvorverarbeitung beseitigen oder abschw\"{a}chen. Die Bildfilterung bezeichnet lineare und nichtlineare Punktoperationen wie Kontrasterh\"{o}hung, Registrierung, Korrektur von Beleuchtungseffekten, Rauschunterdr\"{u}ckung und Reduzierung von Artefakten \cite{burger06,Gonzalez08}. Auch Operationen zur Transformation, \"{A}nderung der Farbtiefe oder die Skalierung \zB mittels \sog Bild"~ oder Gau{\ss}-Pyramiden fallen darunter \cite{burt1983}. Sie transformieren den Bildstrom wie in Formel (\ref{eqn:Filterung}) allgemein beschrieben. Das Ergebnis der genannten Methoden sind Bilder, welche sich im Vergleich zu den Rohdaten besser f\"{u}r alle nachfolgenden Schritte der Auswertung des \Nutzsigs eignen.

Fast immer ist eine Bildnormalisierung notwendig, die als eine besondere Form von Filterung betrachtet werden kann. Gem\"{a}{\ss} \abb \ref{fig:Neues_Konzept_Prinzip_a}  wirken St\"{o}rgr\"{o}{\ss}en direkt auf Biologie und Bildakquise. Solche St\"{o}rungen sind \zB das Aufzeichnen der Bilddaten zu unterschiedlichen Tageszeiten und Temperaturen, durch unterschiedliche Laboranten (\zB mit Unterschieden in der Vorgehensweise bei nicht genau spezifizierten Versuchsprotokollen), mit ge\"{a}nderter Beleuchtung \usw Die Absolutwerte der f\"{u}r die Auswertung und Klassifikation aus dem Bildstrom extrahierten Merkmalszeitreihen (vgl. \kap \ref{subsec:FischInf_Zeitreihe}) h\"{a}ngen somit nicht, wie gew\"{u}nscht, lediglich von der Nutzinformation ab, sondern ebenso von den genannten St\"{o}rgr\"{o}{\ss}en. Data-Mining Methoden zur Merkmalsauswahl sind somit mit hoher Wahrscheinlichkeit in der Lage, Merkmale zu identifizieren, die auf signifikante Unterschiede im \Nutzsig hindeuten. Die Ursache dieser Signifikanz ist jedoch nicht zwingend im \Nutzsig begr\"{u}ndet, was w\"{u}nschenswert ist, sondern oftmals durch die St\"{o}rgr\"{o}{\ss}en verursacht. Wird auf eine Normierung g\"{a}nzlich verzichtet, k\"{o}nnen Einfl\"{u}sse der St\"{o}rgr\"{o}{\ss}en ein Auswerten der Daten verschlechtern oder scheitern lassen.

\begin{figure}[htb]
\centering

    \begin{minipage}[b]{0.48\linewidth}
        \centering
                \includegraphics[width=.7\linewidth]{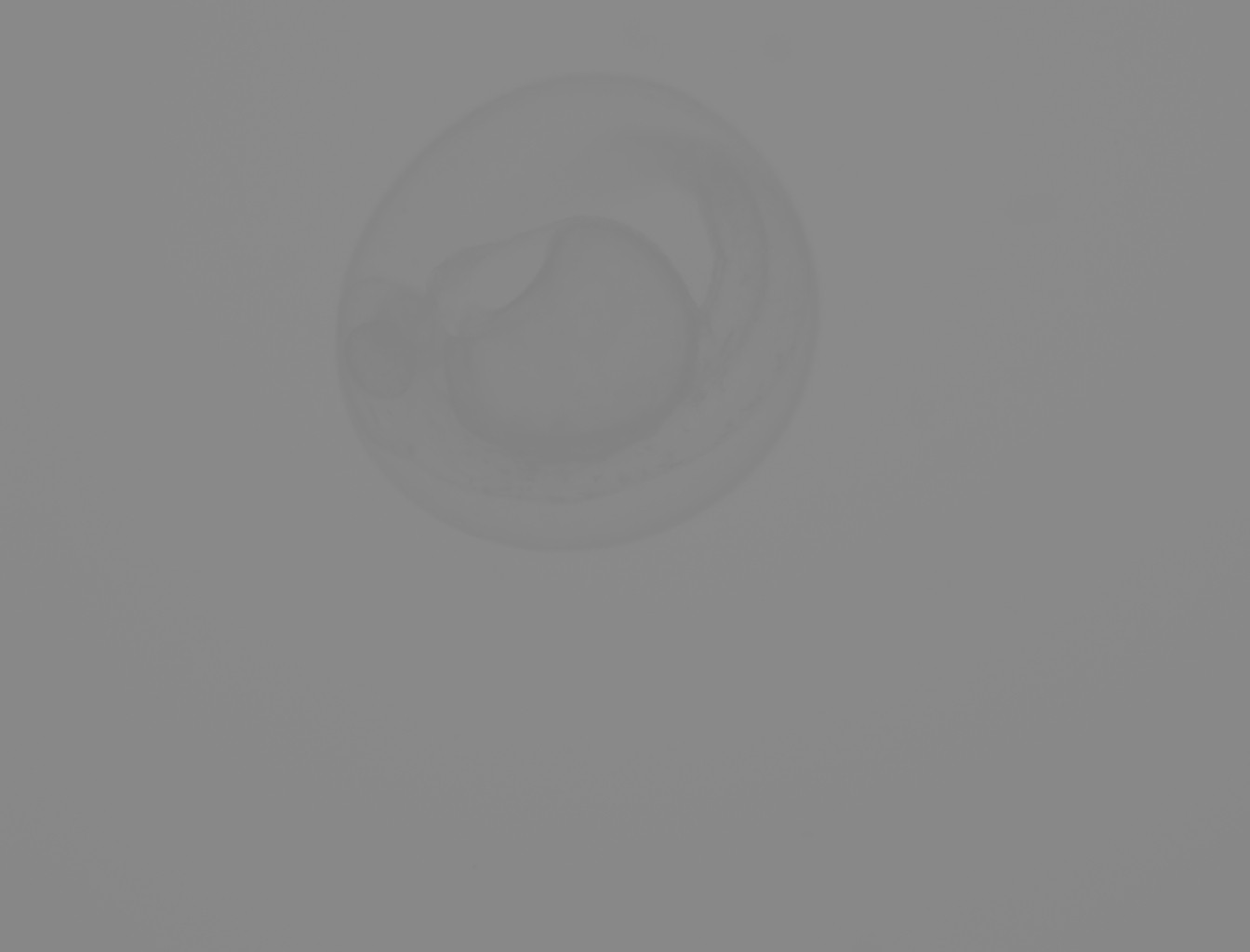}
                                \vspace{0.1cm}\\
        \includegraphics[width=\linewidth]{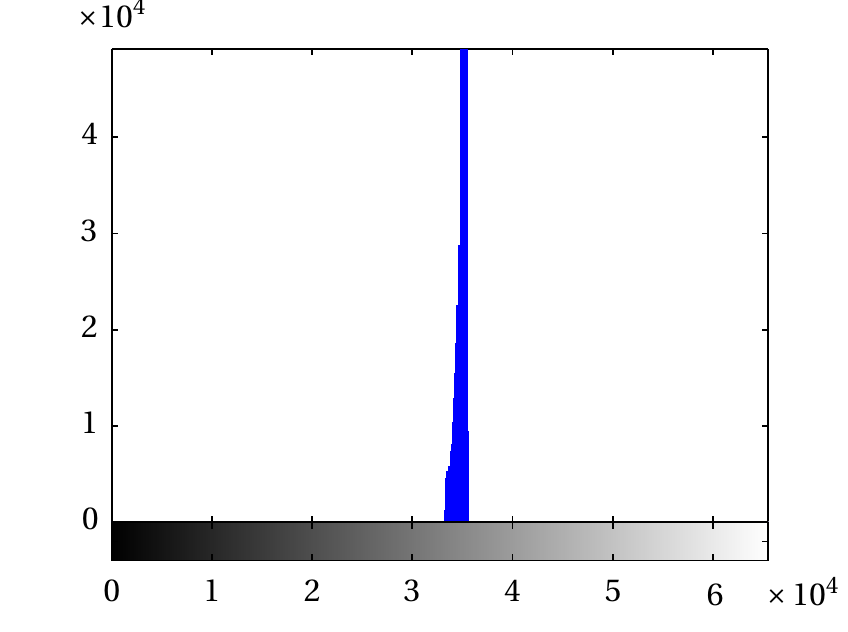}
        \caption[Bild und Histogramm der Rohdaten]{Bild und Histogramm der Rohdaten des \scanr-Hochdurchsatzmikroskops\\}
        \label{fig:Histogram_original}
    \end{minipage}
\hfill
    \begin{minipage}[b]{0.48\linewidth}
        \centering
                \includegraphics[width=.7\linewidth]{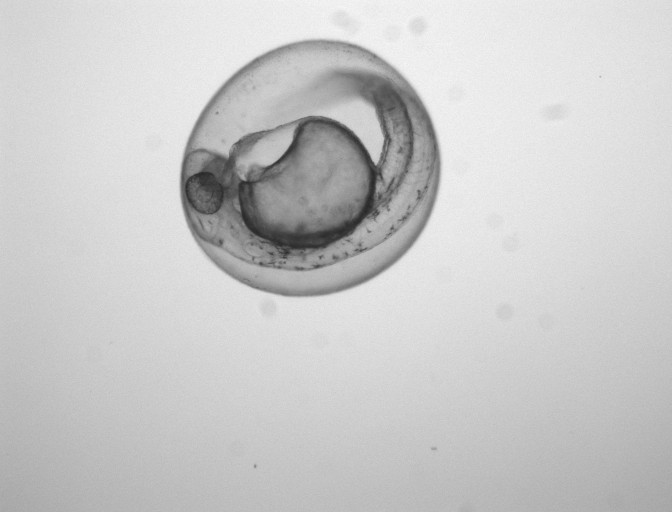}
                \vspace{0.1cm}\\
        \includegraphics[width=\linewidth]{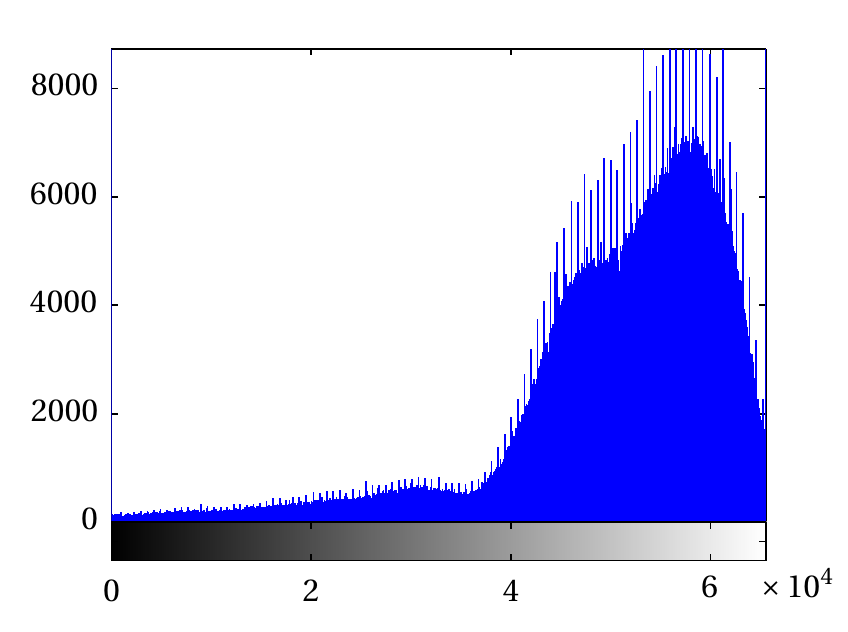}
        \caption[Bild und Histogramm der normalisierten Rohdaten]{Bild und Histogramm der normalisierten Rohdaten (gleiche Rohdaten wie \abb \ref{fig:Histogram_original}) gem\"{a}{\ss} Formel (\ref{eq:saturation}) mit den Parametern $I_{\text{hoch}}  $=98\% und $I_{\text{tief}} $=2\%.}
        \label{fig:Histogram_neu}
        \end{minipage}
\end{figure}

Als Beispiel sei eine typische Aufnahme des Hochdurchsatzmikroskops  \scanr angef\"{u}hrt. Die Intensit\"{a}tswerte konzentrieren sich lediglich auf einen kleinen Wertebereich (vgl. \abb \ref{fig:Histogram_original}), was zur Folge hat, dass eine Vielzahl \"{u}blicher Algorithmen wie etwa adaptive Schwellenwertverfahren \cite{niblack85,sauvola97} oder Regionen-Wachstum-Verfahren \cite{adams94} aufgrund des schlechten relativen Kontrastes scheitern. In einem solchen Fall wird die Verarbeitungskette aufgrund von Validit\"{a}tspr\"{u}fungen abgebrochen, oder, sollten diese fehlen, werden sogar g\"{a}nzlich falsche Werte zur Klassifikation herangezogen. Derartig falsche Merkmalswerte f\"{u}hren im ung\"{u}nstigsten Fall zu systematischen Fehlern und Fehlklassifikationen.

Die St\"{o}rgr\"{o}{\ss}en sind hier \ua das nicht vollst\"{a}ndige Ausnutzen des zur Verf\"{u}gung stehenden Wertebereichs, was zu einer schlechten Bilddynamik f\"{u}hrt sowie abweichende Mikroskop-Einstellungen, das Rauschen des Detektors \etc Ein Unterdr\"{u}cken des Einflusses der St\"{o}rgr\"{o}{\ss}en kann durch eine Korrektur der Intensit\"{a}tswerte erfolgen. Hierbei wird jeder Pixel mittels des Parameters $q$ zentriert und durch einen die Verteilung beschreibenden Parameter $v$ dividiert. F\"{u}r ein zweidimensionales Bild gilt:
\begin{equation}\label{eq:normalisation}
I_{i_xi_y}^* = \frac{I_{i_xi_y}-q}{v}.
\end{equation}
F\"{u}r die Parameter $q$ und $v$ eignen sich \zB statistische Parameter wie Mittelwert oder Varianz der Intensit\"{a}tswerte
\begin{equation}
q_1=\frac{1}{I_x I_y}\sum\limits_{i_x=1}^{I_x}\sum\limits_{i_y=1}^{I_y}I_{i_xi_y}, \quad v_1 = \sqrt{\frac{1}{I_x I_y-1}\sum\limits_{i_x=1}^{I_x}\sum\limits_{i_y=1}^{I_y} (I_{i_xi_y}-q)^2}.
\label{standarddev}
\end{equation}
oder auch Quantile und Extremwerte
\begin{equation}
q_2=\min_{i_x = 1 \dots I_x}\left(\min_{i_y = 1 \dots I_y}(I_{i_xi_y})\right), \quad v_2=\max_{i_x = 1 \dots I_x}\left(\max_{i_y = 1 \dots I_y}(I_{i_xi_y})\right) - \min_{i_x = 1 \dots I_x}\left(\min_{i_y = 1 \dots I_y}(I_{i_xi_y})\right).
\label{eq:maxminnorm}
\end{equation}
Durch das Einf\"{u}hren der Forderung einer prozentualen Mindestanzahl an ges\"{a}ttigten Pixeln l\"{a}sst sich die Robustheit der Normierung mittels Gleichung (\ref{eq:saturation})  deutlich verbessern. Sinnvolle Werte f\"{u}r  $I_\text{tief}$  sind 2 Prozent und $I_\text{hoch}$ 98 Prozent:
\begin{equation}
I_{i_xi_y}^* = \left\{
\begin{array}{ll}
0                                                           &\text{f\"{u}r } I_{i_xi_y}\leq I_\text{tief}\\
\frac{I_{i_xi_y}- I_\text{tief}}{I_\text{hoch}-I_\text{tief}} &\text{f\"{u}r } I_\text{tief} < I_{i_xi_y}<I_\text{hoch}\\
1                                                           &\text{f\"{u}r } I_{i_xi_y}\geq I_\text{hoch}\\
\end{array}
\right.
\label{eq:saturation}
\end{equation}

Das vorgestellte Verfahren wurde nach Formel (\ref{eq:saturation}) auf das Bild in \abb \ref{fig:Histogram_original} angewandt. Das Ergebnis ist in \abb \ref{fig:Histogram_neu} dargestellt. Im Originalbild wird lediglich ein kleiner Bereich der m\"{o}glichen Grauwerte ausgenutzt, wie im Histogramm zu erkennen ist (\abb \ref{fig:Histogram_original}, unten). Auch ist die Larve im Bild fast nicht zu erkennen. Nach Anwendung des Verfahrens sind die Grauwerte \"{u}ber den gesamten Wertebereich verteilt und die Kontraste gut ausgebildet. Ohne weitere Validit\"{a}tspr\"{u}fung birgt die Methode allerdings die Gefahr, dass lediglich das Bildrauschen auf den gesamten Wertebereich gestreckt wird. Der genannte Fall tritt \zB ein, wenn bei der Pr\"{a}paration versehentlich ein N\"{a}pfchen nicht bef\"{u}llt wurde. Solche Bilder lieferten in der weiteren Verarbeitung scheinbar sinnvolle Werte und k\"{o}nnen somit die Versuchsergebnisse stark verf\"{a}lschen. Abhilfe schaffen hier Validit\"{a}tspr\"{u}fungen, die \"{u}ber den gesamten Bildinhalt Auskunft liefern und fehlerhafte Bilder aussortieren. Eine einfache und robuste Pr\"{u}fung ist beispielsweise der Mittelwert aller Intensit\"{a}tswerte. F\"{u}r ein leeres Well (\emph{Well} = N\"{a}pfchen in einer Kunststoffplatte \vgl \kap \ref{subsubsec:\"{U}bersichtHochdurschsatz})  f\"{a}llt dieser aufgrund der fehlenden Abschattung durch das Pr\"{a}parat deutlich h\"{o}her aus als mit Pr\"{a}parat.

\begin{figure}[htb]
    \begin{minipage}[t]{0.48\linewidth}
        \centering
    \includegraphics[width=\linewidth]{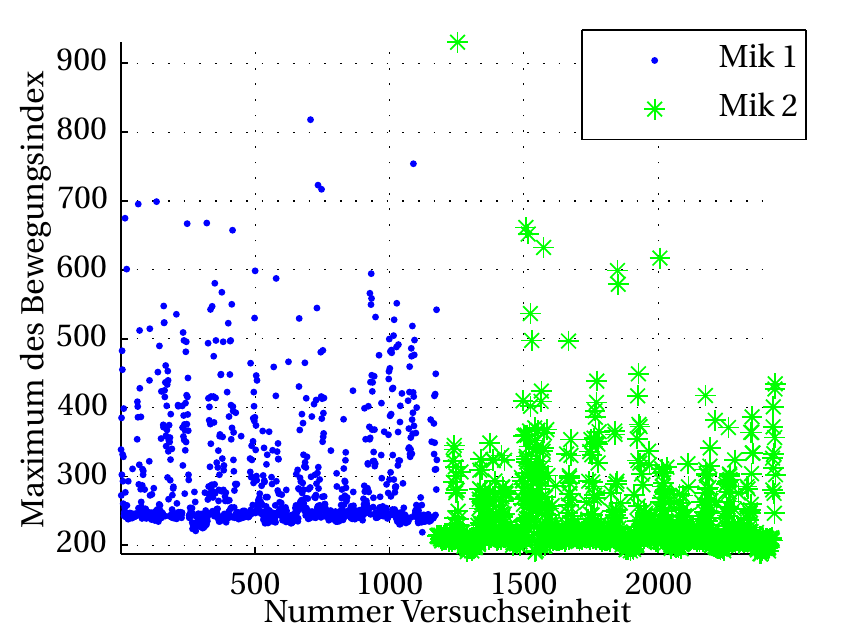}
    \caption[Verteilung der Merkmalswerte in einer \hts]{Verteilung der Merkmalswerte in einer \hts zur Ermittlung der akuten Toxikologie an zwei baugleichen Mikroskopen vom Typ \scanr}
    \label{fig:Norm_Tox_vor_Normalisierung}
        \end{minipage}
        \hfill
\centering
    \begin{minipage}[t]{0.48\linewidth}
        \centering
                \includegraphics[width=\linewidth]{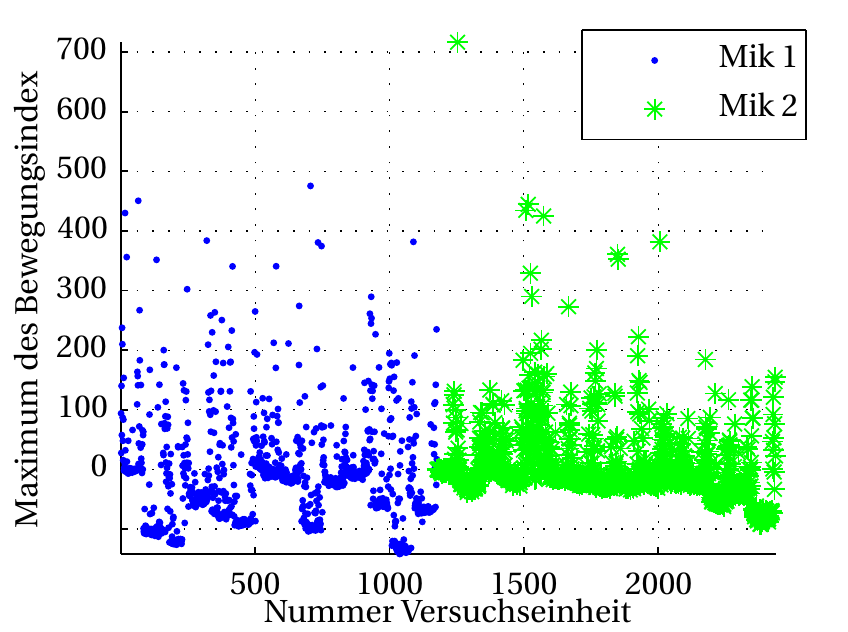}

        \caption[Scatterplot der Merkmalswerte der Kontrollen]{\label{fig:Norm_Tox_Kontrollen} Scatterplot der mittels der Kontrollen aus \abb \ref{fig:Norm_Tox_vor_Normalisierung} normierten Merkmalswerte \\}

    \end{minipage}
\end{figure}

\begin{figure} [htb]
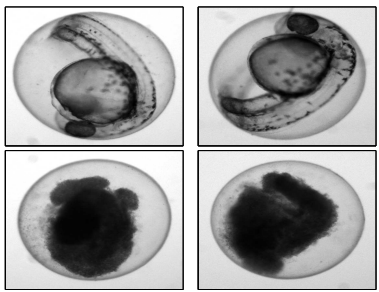
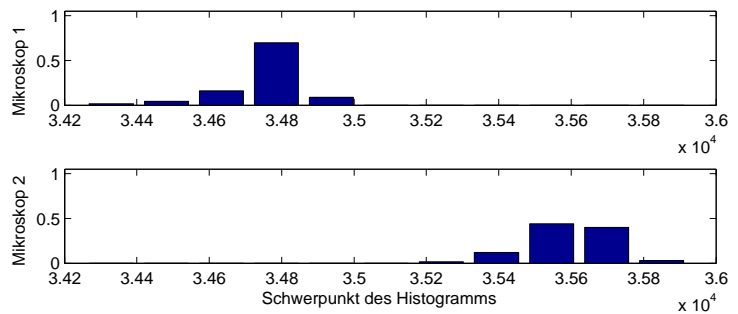

\begin{minipage}{.15\textwidth}
    \includegraphics[width=1\columnwidth]{Bilder/02_Neues_Konzept/Normierung/sys1_alive(a1).eps}
    \includegraphics[width=1\columnwidth]{Bilder/02_Neues_Konzept/Normierung/sys1_dead(e1).eps}
\end{minipage}
\begin{minipage}{.15\textwidth}
    \includegraphics[width=1\columnwidth]{Bilder/02_Neues_Konzept/Normierung/sys2_alive(a1).eps}
    \includegraphics[width=1\columnwidth]{Bilder/02_Neues_Konzept/Normierung/sys2_dead(e1).eps}
\end{minipage}
\hfill
\begin{minipage}{.6\textwidth}
    \begin{center}
    \includegraphics[width=1\columnwidth]{Bilder/02_Neues_Konzept/Normierung/histmikr.eps}
    \end{center}
\end{minipage}
\caption[Mikroskopbilder mit zugeh\"{o}rigen Histogrammen eines Merkmals]{\label{fig:hist}a (links): Mikroskopbilder mit verschiedenen Planfaktoren (\emph{lebendig} in Zeile~1, \emph{koaguliert} in Zeile~2) und St\"{o}rfaktoren (\emph{Mikroskop 1} in Spalte 1, \emph{Mikroskop 2} in Spalte 2).\\ b (rechts): Schwerpunkt des Histogramms aller Bilder zweier Mikroskope des in Abbildung \ref{fig:Norm_Tox_vor_Normalisierung} gezeigten Beispiels \cite{Reischl10}. Obwohl die Aufnahmen in \ref{fig:hist}a, spaltenweise verglichen, optisch gleich wirken \"{u}berlappen die Histogramme des dargestellten Merkmals nicht.}
\end{figure}

Die Methode korrigiert jedoch keine Faktoren, die sich direkt auf den Versuch beziehen, wie \zB ein unterschiedliches Alter der Fische oder einen unterschiedlichen Mikroskoptyp. Als Beispiel sei hier die \hts zur Ermittlung der akuten Toxikologie gew\"{a}hlt \cite{OECD06}. In \abb \ref{fig:Norm_Tox_vor_Normalisierung} ist ein Scatterplot einer \hts von 2436 Versuchseinheiten (hier jeweils ein Zebrab\"{a}rbling) aufgetragen, welche an zwei unterschiedlichen Mikroskopen aufgezeichnet wurde. Ein typisches Merkmal zur Klassifikation der Bewegung wurde ausgew\"{a}hlt (\vgl hierzu \kap \ref{subsec:Merkmale_Bewegung}). Der Einfluss der beiden Mikroskope ist deutlich, durch einen Offset der Merkmalswerte zueinander zu erkennen. Solche Einfl\"{u}sse, wie sie in diesem Beispiel bereits zwischen zwei Mikroskopen des gleichen Typs entstehen, sind zwischen unterschiedlichen Laboren ungleich gr\"{o}{\ss}er.

Um Versuche auch zwischen unterschiedlichen Laboren, Mikroskopen und Tagen vergleichbar zu machen, muss der Bildstrom durch auf den Versuch bezogene Parameter normalisiert werden. Hier ist meist das Normalisieren mittels der Negativ- oder Positiv-Kontrollen erfolgreich \cite{Malo06}. Im gezeigten Beispiel scheitert die Methode jedoch ebenfalls, da auch die Merkmalswerte in den Positiv-Kontrollen schwanken. Kontrollen sollten zwar ein vergleichbares Verhalten zeigen, was hier jedoch nicht der Fall ist, wie \abb \ref{fig:Norm_Tox_Kontrollen} zeigt. In der Abbildung wurden alle Merkmalswerte mittels der Kontrollen normiert. Insbesondere bei den Merkmalswerten von Mikroskop 1 (blau dargestellt) streuen nach der Normierung die Merkmalswerte st\"{a}rker als zuvor. Beispielsweise haben die ersten 84 Versuchseinheiten (96 Versuche auf einer Platte davon 12 Kontrollen zur Normalisierung) nach der Normalisierung Merkmalswerte welche minimal etwa bei 0 liegen. Die n\"{a}chste Platte (Nummer Versuchseinheit 84 bis 168) zeigt Minimalwerte unter -100.
%
Klarer wird die unterschiedliche Verteilung der Merkmalswerte durch den Einfluss der Mikroskope in Abbildung \ref{fig:hist}, in der diese anhand eines Histogramms veranschaulicht wird. Die hier dargestellten Larven sind sich in beiden Aufnahmen (Reihenweise verglichen) sehr \"{a}hnlich, die Merkmalswerte (hier Schwerpunkt des Histogramms, $x_1$ vgl. Abschnitt \ref{subsec:Module_Instantane_Merkmale}) zeigen jedoch erhebliche Unterschiede. In einem konsistenten und normalisierten Versuch sollte sich wenig Unterschied bei der Merkmalsverteilung der gleichen Planfaktoren von vergleichbaren Versuchseinheiten zeigen, selbst wenn sie von unterschiedlichen Mikroskopen stammen. Bisherige L\"{o}sungen m\"{u}ssen einen solchen Datensatz verwerfen.

Deshalb wurde im Rahmen der vorliegenden Arbeit eine neue Methode erarbeitet, welche in der Lage ist, ein solches Problem zu l\"{o}sen und auch inhomogene Datens\"{a}tze vergleichbar macht \cite{Reischl10}. 
Ein Klassifikator ist umso besser, je exakter der Klassifikator in der Lage ist, die Faktoren $\hat{Y}$ oder $\hat{Z}$ der wahren Klasse $Y$ oder $Z$ zuzuordnen. Es l\"{a}sst sich die Erfolgsrate f\"{u}r Planfaktoren $\xi_y$ und St\"{o}rfaktoren $\xi_z$ f\"{u}r jede Versuchseinheit $n$ \"{u}ber alle Ausgangsklassen $s_y$ angeben durch:

\begin{equation}
   \xi_{y,n} = \frac{1}{s_y}\sum_{i_{sy}=1}^{s_y} \xi_{y,i_{s_y},n}
   \quad
   \text{mit}
   \quad
   \xi_{y,i_{sy},n} =
       \begin{cases}
       1 &  \hat{y}_{i_{sy},n} = y_{i_{sy},n}\\
       0 & \text{sonst}\\
   \end{cases},
\end{equation}
\begin{equation}
    \xi_{z,n} = \frac{1}{s_z}\sum_{i_{sz}=1}^{s_z} \xi_{z,i_{s_z},n}
   \quad
   \text{mit}
   \quad
   \xi_{z,i_{sy},n} =
       \begin{cases}
       1 &  \hat{z}_{i_{sz},n} = z_{i_{sz},n}\\
       0 & \text{sonst}\\
   \end{cases}.
\end{equation}
Die mittlere Klassifikationsg\"{u}te $\overline{\xi_{y}}, \overline{\xi_{z}}$ wird bestimmt durch:
\begin{equation}
   \overline{\xi_{y}} = \frac{1}{N}\sum_{n=1}^N \xi_{y,n}, \quad  \overline{\xi_{z}} = \frac{1}{N}\sum_{n=1}^N \xi_{z,n}.
   \end{equation}
Im Falle einer perfekten Klassifikation werden alle Planfaktoren korrekt der zugeh\"{o}rigen Ausgangsklasse zugeordnet, w\"{a}hrend sich die Klassen aller anderen Faktoren nicht trennen lassen und daher gleichverteilt nach dem Zufallsprinzip auf die Ausgangsklassen aufgeteilt werden:
\begin{equation}
    \overline{\xi_{y,\text{ziel}}}=1, \quad \overline{\xi_{z,\text{ziel}}}\approx \frac{1}{s_z} \sum_{i_{sz}=1}^{s_z} \frac{1}{m_{y,sz}}.
\end{equation}
Die letzte Formel setzt allerdings voraus, dass die Merkmale von den St\"{o}rfaktoren unabh\"{a}ngig sind, was in realen Systemen nicht der Fall ist. Daher m\"{u}ssen alle Parameter in $\mathbf{p}$ so gew\"{a}hlt werden, dass der Einfluss der St\"{o}rfaktoren klein wird. Dieses Problem kann mittels Filter"~ oder Wrapperans\"{a}tzen gel\"{o}st werden \cite{Kohavi97,Hall99}. Filter bestimmen die Versuchsparameter ohne Kenntnis der wahren oder gesch\"{a}tzten Klassenzugeh\"{o}rigkeit
\begin{equation}
    \mathbf{p}^* = f(\mathbf{p)}. 
\end{equation}
Wrapper hingegen ber\"{u}cksichtigen die Klassifikationsergebnisse bei der Parameteradaption:
\begin{equation}
    \mathbf{p}^{**} = f(\mathbf{p},  \mathbf{Y}, \mathbf{Z}, \hat{\mathbf{Y}}, \hat{\mathbf{Z}}).
\end{equation}
Wrapper liefern \"{u}blicherweise bessere Ergebnisse auf Kosten eines deutlich h\"{o}heren Rechenaufwands.

Die Optimierung sucht nach einer Parameterkombination, welche einen Kompromiss zwischen $\overline{\xi_{y}}-\overline{\xi_{y,\text{ziel}}}$ und $\overline{\xi_{z}}-\overline{\xi_{z,\text{ziel}}}$ sucht.

Ein m\"{o}gliches Kriterium zur Optimierung ist
\begin{equation}
    \min_{\mathbf{p}}(O)\quad\text{mit} \quad O= \alpha\left|\overline{\xi_{y}}-\overline{\xi_{y,\text{ziel}}}\right|+(1-\alpha) \left|\overline{\xi_{z}}-\overline{\xi_{z,\text{ziel}}}\right|, \quad \alpha \in [0.5, 1].
    \label{eq:optim}
\end{equation}

Der Parameter $\alpha$ wichtet die Klassifikationsergebnisse zwischen Plan"~ und St\"{o}rfaktoren und muss f\"{u}r sinnvolle Ergebnisse mit signifikantem Einfluss der Planfaktoren gr\"{o}{\ss}er als 0.5 gew\"{a}hlt werden  (Bei Werten < 0.5 ist der Einfluss der St\"{o}rfaktoren auf das Optimierungskriterium gr\"{o}{\ss}er als der Einfluss der Planfaktoren). F\"{u}r den Fall $\alpha=1$ wird die Optimierung zu einem gew\"{o}hnlichen Wrapper \cite{Reischl03}. Es lassen sich L\"{o}sungsans\"{a}tze sowohl mit Wrappern als auch Filtern realisieren oder auch durch eine Kombination beider. Eine Kombination kann etwa die Optimierung eines normalisierten Bildstroms $BS^*$ mittels eines Wrappers sein.

F\"{u}r den Fall, dass alle Merkmalswerte zur gleichen Klasse der St\"{o}rfaktoren geh\"{o}ren, liefern konventionelle Klassifikatoren in einer Kreuzvalidierung gute Ergebnisse, auch f\"{u}r Testdaten~\cite{Efron95,Kohavi95}. Geh\"{o}ren die Testdaten jedoch einer anderen Klasse der St\"{o}rfaktoren an, welche bei der Kreuzvalidierung nicht vertreten war, kann das tats\"{a}chliche Ergebnis deutlich schlechter ausfallen. Daher ist die Kreuzvalidierung kein geeignetes Mittel, die Validit\"{a}t der Klassifikation zu \"{u}berpr\"{u}fen. Die besten Ergebnisse lassen sich erzielen, wenn in den Trainings- und Testdaten alle Klassen der St\"{o}r"~ und Planfaktoren vertreten sind.

\begin{figure} [htbp]
    \begin{center}
        \includegraphics[width=0.8\columnwidth]{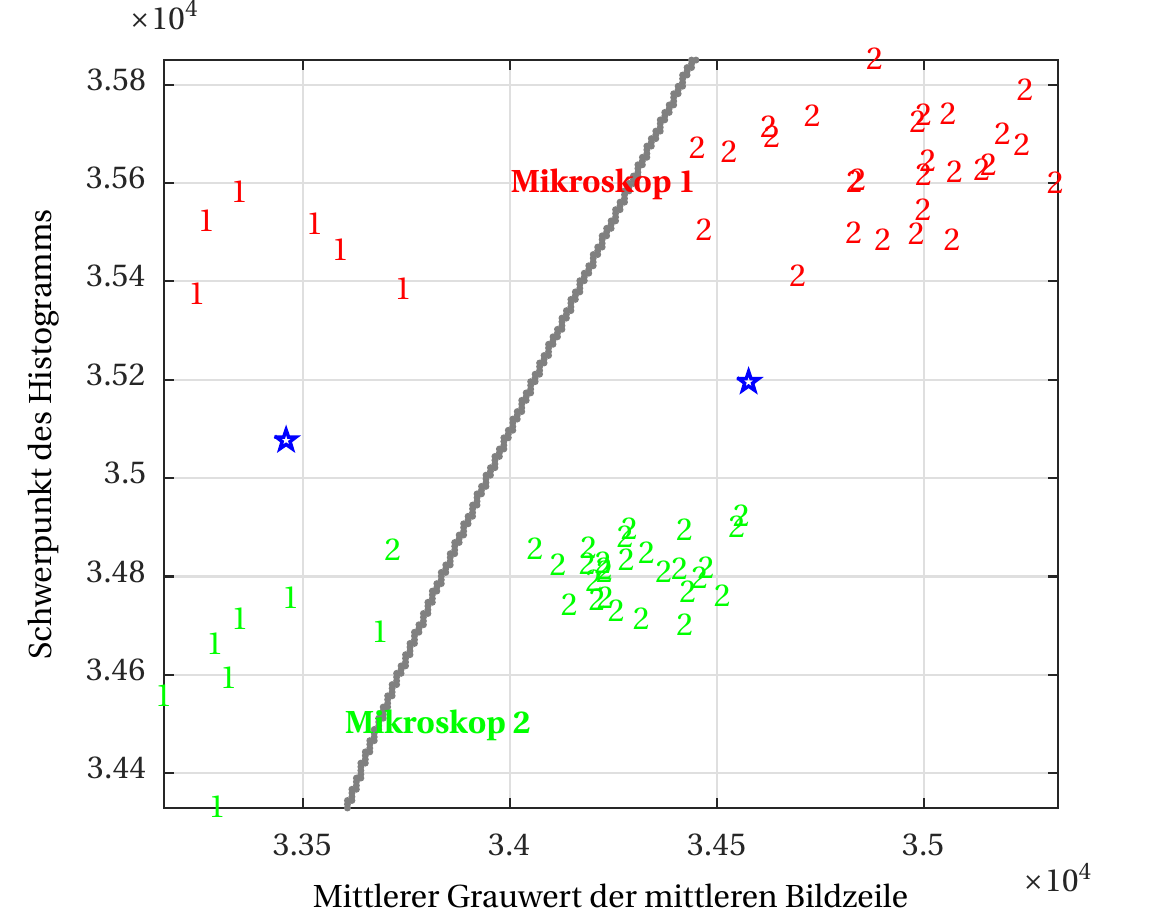}
    \end{center}

\vspace{-2em}
\caption[Merkmalsraum und Diskriminanzfunktion f\"{u}r zwei Platten]{\label{fig:resultwrapper1}Merkmalsraum und Diskriminanzfunktion f\"{u}r zwei Platten aus Abbildung \ref{fig:Norm_Tox_vor_Normalisierung} vor Normalisierung \cite{Reischl10}. }
    \begin{center}
        \includegraphics[width=0.8\columnwidth]{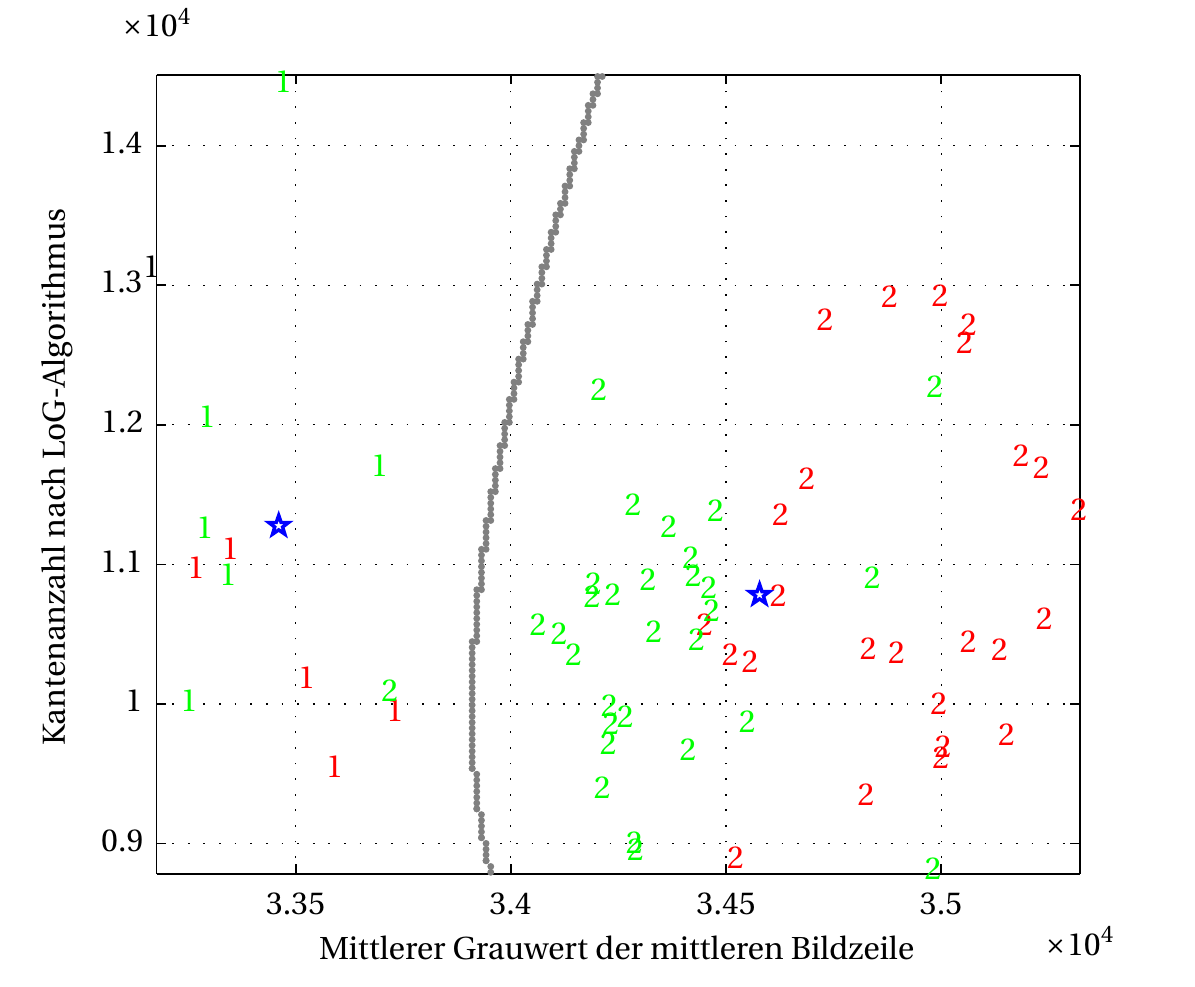}
   \end{center}
\vspace{-2em}
\caption[Merkmalsraum und Diskriminanzfunktion f\"{u}r zwei Platten]{\label{fig:resultwrapper2}Merkmalsraum und Diskriminanzfunktion f\"{u}r zwei Platten aus Abbildung \ref{fig:Norm_Tox_vor_Normalisierung} nach der Normalisierung mittels des Wrapperansatzes \cite{Reischl10}.}
\end{figure}

Den Erfolg der Methode verdeutlichen die Abbildungen \ref{fig:resultwrapper1} und \ref{fig:resultwrapper2} am Beispiel der Merkmalswerte aus Abbildung \ref{fig:hist}.
Die Datenpunkte in Abbildung \ref{fig:resultwrapper1} welche mit $1$ gekennzeichnet sind geh\"{o}ren der Klasse \emph{koaguliert} an, Datenpunkte mit $2$ geh\"{o}ren der Klasse \emph{lebendig} an. Datenpunkte von Mikroskop 1 sind rot, Datenpunkte von Mikroskop 2 sind gr\"{u}n dargestellt. Der Einfluss des St\"{o}rfaktors \emph{Mikroskop} ist deutlich zu erkennen. Obwohl die Merkmalswerte beider Mikroskope Versuchseinheiten mit \"{a}hnlichem Status abbilden, \"{u}berschneidet sich die Streuung der Merkmalswerte nicht. In Abbildung \ref{fig:resultwrapper2} schlie{\ss}lich sind die gleichen, jedoch mittels des Wrapperansatzes normalisierten Daten dargestellt.
Ebenfalls geh\"{o}ren die mit $1$ gekennzeichneten Datenpunkte der Klasse \emph{koaguliert} an und die mit $2$ gekennzeichneten Datenpunkte der Klasse \emph{lebendig} an. Datenpunkte von Mikroskop 1 sind wie zuvor rot und Datenpunkte von Mikroskop 2 gr\"{u}n gekennzeichnet. Nach der Normalisierung ist eine Unterscheidung der Mikroskope, wie gewollt, nicht mehr unmittelbar m\"{o}glich. Es l\"{a}sst sich lediglich der Planfaktor \emph{Koaguliertheit} klar abgrenzen. Damit sind die zuvor sowohl durch die Plan- als auch durch die St\"{o}rfaktoren trennbaren Merkmalsr\"{a}ume nach der Normalisierung nur noch durch die Planfaktoren unterscheidbar und somit deutlich robuster f\"{u}r eine \hts.

\subsection{Bildkombination}
Der letzte Modulblock in der Kategorie Bildstrom-Vorverarbeitung sind bildkombinierende Methoden. Solche kommen zum Einsatz, wenn bei der Bildakquise mehrere Fokusebenen oder Schichtaufnahmen akquiriert wurden, was vor allem bei \Nutzsigen der Fall ist, bei denen bestimmte Bereiche im Fisch unterschieden werden sollen. Die Bildkombination erstellt aus mehreren zweidimensionalen Bildern kombinierte Bilder und reduziert somit die Dimension. Bei \htsen kommt vor allem die Kombination von Farbkan\"{a}len zu einem Grauwertkanal zum Einsatz, da Farbkan\"{a}le in bestimmten F\"{a}llen keine zus\"{a}tzliche Information zur \NutzsigInf liefern. Eine ebenfalls wichtige Methode ist der Extended Focus \cite{Pieper83}. Die Verfahren hierzu basieren auf der Annahme, dass alle zu fusionierenden Bilder den gleichen Inhalt zeigen, die Bildinformation in einer der Aufnahmen jedoch weniger gest\"{o}rt aufgezeichnet wurde als in anderen. Hier wird mittels eines Bewertungsma{\ss}es in jeder Fokusaufnahme versucht, die beste Repr\"{a}sentation der Szene auszuw\"{a}hlen und in das Ergebnisbild einzuf\"{u}gen. Es lassen sich damit deutliche Steigerungen der Sch\"{a}rfentiefe und damit des auswertbaren Bereichs innerhalb des Zebrab\"{a}rblings erzielen. Eine ausf\"{u}hrliche Beschreibung der Methoden findet sich u.a. in \cite{burger06,Jahne05,Steinbrecher93}.

\section{Segmentierung der Nutzinformation}\label{subsec:Segmentierung}

Um das biologische \Nutzsig einer \hts quantifizieren zu k\"{o}nnen, muss es von allen anderen Daten im Bildstrom separiert werden.  Alle hier vorgestellten Module haben als Eingang den vorverarbeiteten Bildstrom $\mathbf{BS^*}$ und als Ausgang Merkmalsbilder $\boldsymbol{\Psi}$. In die Kategorie \emph{Segmentierung der Nutzinformation} geh\"{o}ren die folgenden drei Module. \begin{enumerate}
         \item Je Bildstrom ist genau eine Versuchseinheit zu jedem Zeitpunkt aufgezeichnet (Trennung Einzelobjekt und Hintergrund).
         \item Im Bildstrom sind mehrere Versuchseinheiten zu jedem Zeitpunkt aufgezeichnet (Trennung mehrere Objekte und Hintergrund).
         \item Zus\"{a}tzlich zu 1. oder 2. ist die Versuchseinheit \"{u}ber einen bestimmten Zeitraum aufgezeichnet (Tracking).
       \end{enumerate}
Je Versuchseinheit werden Merkmalsbilder errechnet. Merkmalsbilder sind aus den Rohdaten abgeleitete Bilder wie \zB Bin\"{a}rbilder (nach Anwendung eines Schwellenwertes) oder Differenzbilder von Bildsequenzen. Die Separation der Information im vorverarbeiteten Bildstrom $\mathbf{BS^*}$ erfolgt durch die \sog Segmentierung. Die Segmentierung ist das Gruppieren zusammengeh\"{o}riger Bereiche. Im Folgenden werden die Bildregionen, welche die Information des biologischen Effekts enthalten, als Vordergrund und alle anderen Regionen als Hintergrund bezeichnet. Die Segmentierung trennt somit den Vordergrund vom Hintergrund und erm\"{o}glicht so die weitere Auswertung der im Vordergrund enthaltenen \NutzsigInf. Alle Methoden bewirken eine Reduktion des vorverarbeiteten Bildstroms $\mathbf{BS^*}$ auf die \NutzsigInf. Dabei unterscheidet sich das Vorgehen stark f\"{u}r die F\"{a}lle, ob mehrere Versuchseinheiten im Bild vorhanden sind \bzw sein k\"{o}nnen oder ob in der Versuchsauslegung festgelegt ist, dass lediglich eine Versuchseinheit, \zB der Zebrab\"{a}rbling (\inkl Chorion), im Bildstrom vorliegt.

\subsection{Neue Methode zur Trennung von Einzelobjekt und Hintergrund}\label{Trennung von Einzelobjekt und Hintergrund}
Zur Trennung der \NutzsigInf einer einzelnen Versuchseinheit vom Hintergrund sind bei akzeptabler Bildqualit\"{a}t bereits Schwellenwertverfahren in Kombination mit morphologischen Operatoren erfolgreich \cite{Alshut08}. Adaptive Schwellenwertverfahren wie die Methode von Otsu \cite{Otsu75,Otsu79} oder Fuzzy-basierende Schwellenwertverfahren \cite{Hathaway87,Huang95} bieten hier jedoch deutlich robustere Ergebnisse \cite{Alshut09,Alshut10}. Der Vorteil der genannten Verfahren ist die kurze Rechenzeit, was gerade bei Roboteranwendungen wie \zB dem automatischen Pipettieren von Eiern wichtig ist \cite{Pfriem11}. Ein Schwellenwertverfahren trennt jedoch ohne jegliche Pr\"{u}fung Vorder"~ von Hintergrundpixeln. Daher muss f\"{u}r jedes gefundene Objekt nachtr\"{a}glich gepr\"{u}ft werden, ob es sich um eine Versuchseinheit oder ein Artefakt, \zB Schmutz o.\"{a}., handelt. Die besten Ergebnisse lassen sich unter Einbezug von Vorwissen, beispielsweise der Gr\"{o}{\ss}e und Kontur der \NutzsigInf der \hts, erzielen. Solche Informationen unterscheiden sich stark f\"{u}r die beiden F\"{a}lle:
\begin{enumerate}
  \item Die Larve befindet sich im Chorion.
  \item Die Larve befindet sich nicht mehr im Chorion.
\end{enumerate}
In der vorliegenden Arbeit wurden neue Methoden f\"{u}r den ersten Fall entwickelt, da f\"{u}r den Hochdurchsatz das Dechorionieren, wie in \kap \ref{subsec:Bio_u_Bildakquise} beschrieben, deutliche Nachteile hat. Unter Einbezug von Vorwissen, wie zum Beispiel der Kontur des Zebrab\"{a}rblings, kann jedoch auch f\"{u}r den zweiten Punkt eine Segmentierung erfolgreich sein \cite{Gehrig09}.

Im ersten Fall (der Zebrab\"{a}rbling befindet sich im Chorion), l\"{a}sst sich das zu findende Objekt gut durch die Gr\"{o}{\ss}e und Form charakterisieren. Daher wird ein Schwellenwert mittels Intervallschachtelung so lange angepasst, bis das gr\"{o}{\ss}te Objekt den Suchkriterien der \hts gerecht wird. Bei der Suche nach dem Ei des Zebrab\"{a}rblings ist das Ziel der Suche ein einzelnes Objekt, welches ein Mindestma{\ss} f\"{u}r Rundheit und eine Gr\"{o}{\ss}e innerhalb eines zuvor festgelegten Toleranzbereichs besitzt.
Als Ma{\ss}einheit empfiehlt es sich, sich von der Bildaufl\"{o}sung unabh\"{a}ngige Ma{\ss}e, wie \zB die wahre Gr\"{o}{\ss}e in mm,  zu w\"{a}hlen, da sich die Algorithmen somit auch auf andere Mikroskope, Kameras und Vergr\"{o}{\ss}erungen \"{u}bertragen lassen. Werden nach einer maximalen Anzahl an Iterationen die \og Kriterien nicht erreicht, wird die Suche abgebrochen und die Versuchseinheit als fehlerhaft markiert. Das Verfahren setzt voraus, dass die Mehrzahl der Versuche korrekt pr\"{a}pariert ist, \dhe genau ein Ei enth\"{a}lt, da solche Versuchseinheiten sonst als fehlerhaft detektiert, aus der weiteren Verarbeitung ausgeschlossen werden und somit wertlos sind. Die Validit\"{a}tspr\"{u}fung geschieht hier durch Bestimmung von instantanen Merkmalen des gr\"{o}{\ss}ten Objektes. Typische Werte f\"{u}r die Ei-Gr\"{o}{\ss}e und das Verh\"{a}ltnis der Durchmesser einer umschlie{\ss}enden Ellipse werden zu Beginn f\"{u}r typische Beispiele ermittelt und dann mit den Werten jeder Versuchseinheit verglichen. Alle Bilder au{\ss}erhalb eines Toleranzbereichs von 20\% des ermittelten Wertes werden verworfen. F\"{u}r n\"{a}here Ausf\"{u}hrungen zu Merkmalen und deren Extraktion sei auch auf Abschnitt \ref{subsec:FischInf_Zeitreihe} verwiesen. Schlie{\ss}lich wird das Bild auf den Bereich, der das gr\"{o}{\ss}te Objekt enth\"{a}lt, zugeschnitten. Der Zuschnitt erfolgt durch Anpassung der Indizes des Bildstroms (\vgl Formel (\ref{eq:BSaufPSI})).

\subsection{Neue Methode zur Trennung von mehreren Objekten und Hintergrund }\label{Trennung mehrere Objekte_Hintergrund}
F\"{u}r die Trennung mehrerer Eier von Zebrab\"{a}rblingen vom Hintergrund kann ebenfalls auf eine Kette von Schwellenwerten und morphologischen Operatoren zur\"{u}ckgegriffen werden.  Je mehr Versuchseinheiten pro Bild aufgezeichnet werden, desto gr\"{o}{\ss}er ist jedoch die Wahrscheinlichkeit, dass sich die Versuchseinheiten bei der Aufnahme ber\"{u}hren und nach der Anwendung eines Schwellenwertes im Bin\"{a}rbild als ein einziges gro{\ss}es Objekt erscheinen. Wenn es nicht m\"{o}glich ist, solche Cluster zu verwerfen, m\"{u}ssen diese anschlie{\ss}end durch zus\"{a}tzliche Bildverarbeitungsschritte getrennt werden, \zB mittels Regionen-Wachstumsverfahren oder Levelset-Algorithmen \cite{Cremers07,Pavlidis90,Ziou98}. Da es sich hierbei um aufw\"{a}ndigere  Verfahren handelt, w\"{a}chst der Rechenaufwand deutlich und macht solche Verfahren f\"{u}r \htsen weniger geeignet. Durch Ausnutzung von spezifischem Vorwissen l\"{a}sst sich jedoch eine robuste Segmentierung finden, welche den besten Kompromiss zwischen Rechenaufwand und Robustheit liefert. Bei der Segmentierung mehrerer Zebrab\"{a}rblingseier l\"{a}sst sich eine Konstante im Bild ausnutzen, die in jeder Bildsituation und in jeder Position sowie bei sich bewegenden Larven vorkommt: Die Au{\ss}enkante des Chorions. Die Au{\ss}enkante bildet zudem auch den Abschluss des zu untersuchenden Objektes zu Nachbarobjekten und dem Hintergrund. Das Chorion kann vereinfacht durch einen Kreis beschrieben werden, welcher, wenn dieser einen akzeptablen Kontrast zum Hintergrund aufweist, durch den weitgehend bekannten Radius eine robuste Detektion zul\"{a}sst. Zudem ver\"{a}ndert sich der Durchmesser des Chorions w\"{a}hrend der Entwicklung der Larve nur unwesentlich \cite{Westerfield93}. Damit l\"{a}sst sich die Trennung von mehreren Versuchseinheiten zumindest f\"{u}r die initiale Detektion der Eier auf eine Kreissuche reduzieren. F\"{u}r die Kreissuche existieren in der Literatur bereits diverse Ans\"{a}tze \cite{Carolyn75,Hough62}. Die Hough-Kreisdetektion kann erfolgreich zur Detektion des Chorions angewandt werden.
Hierf\"{u}r wird zu Beginn ein Kantenbild beispielsweise mittels des Canny-Filters errechnet \cite{Canny83}. Jedem Pixel im Kantenbild werden daraufhin alle potenziellen Mittelpunkte des Chorions zugeordnet und jeweils in einer Akkumulationsmatrix markiert. Mittels der Maxima der so entstandenen Matrix l\"{a}sst sich daraufhin die Position der Zebrab\"{a}rblingseier ermitteln. Limitierungen hat die beschriebene Methode bei niedrigem Kontrast des Chorions zum Hintergrund. In einem solchen Fall wird das Chorion nicht vollst\"{a}ndig als Kante erkannt und das Maximum in der Akkumulationsmatrix ist weniger ausgepr\"{a}gt. Bei sehr schlechtem Kontrast f\"{u}hrt dies dazu, dass der R\"{u}cken der Zebrab\"{a}rblinge st\"{a}rker im Kantenbild repr\"{a}sentiert ist als der Rand des Chorions, was zu einer fehlerhaften Kreisdetektion f\"{u}hrt. Abbildung \ref{fig:PMR_Konzeptumsetzung_Figures_Kantenbild} in Kapitel \ref{chap:Anwendung} zeigt ein Beispiel der Anwendung der Methode.
Allerdings ist der Rechenaufwand zum Erstellen der Akkumulationsmatrix gr\"{o}{\ss}er als bei den erw\"{a}hnten reinen Schwellenwertverfahren. Im Gegensatz zu jenen ist die Kreisdetektion jedoch sehr robust und das Ergebnis l\"{a}sst sich anhand einfacher Validit\"{a}tskriterien wie \zB dem mittleren Grauwert des Kreisinhaltes \"{u}berpr\"{u}fen.

\subsection{Neues Verfahren zum Tracking des Chorions}\label{subsec:Tracking}
Liegen Bildsequenzen von mehreren Zebrab\"{a}rblingseiern vor, ist es notwendig, die Position des \Nutzsigs an jedem Zeitpunkt zu bestimmen und einander zuzuordnen. Die in Abschnitt \ref{Trennung mehrere Objekte_Hintergrund} vorgestellte Methode zur Detektion des Chorions l\"{a}sst sich zwar auf jedes Frame eines Bildstroms anwenden, allerdings bedeutet dies einen hohen Rechenaufwand. G\"{u}nstiger ist es, das Tracking mittels Kreuzkorrelation hinzu zu kombinieren \cite{Jahne05}.  Tabelle~\ref{tab:Rechenaufwand_Tracking} stellt den Rechenaufwand der beiden Methoden gegen\"{u}ber.
\begin{table}[!htbp]
  \begin{tabularx}{\linewidth}{L{3cm}VV}
\toprule
\textbf{Methode}&\textbf{Rechenaufwand [s]} & \textbf{Rechenaufwand f\"{u}r typische Sequenz [s]}\\
\midrule
  Kreisdetektion &$ 0.131284 $&$ 131.284$ \\
Korrelation  & $0.001889$&$ 15.112$  \\
\bottomrule
\end{tabularx}
\caption[Gegen\"{u}berstellung des Rechenaufwands von Korrelation und
Kreisdetektion ]{Gegen\"{u}berstellung des Rechenaufwands von Korrelation und
Kreisdetektion auf einem handels\"{u}blichen Desktop-PC f\"{u}r eine typische Sequenzl\"{a}nge von 1000 Frames und 8 Versuchseinheiten}\label{tab:Rechenaufwand_Tracking}
\end{table}

\begin{figure}[htb]
\centering
        \includegraphics[width=\linewidth]{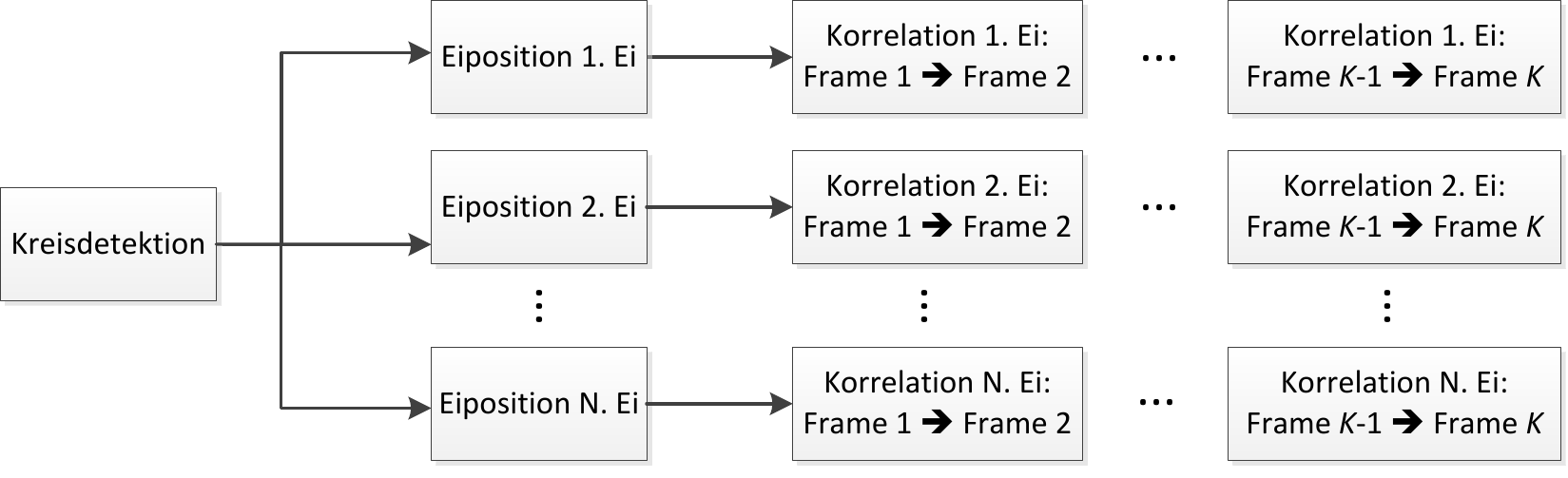}
        \caption[Ablaufschema des Trackings bei einer \hts]{Ablaufschema des Trackings bei einer \hts mit noch im Chorion befindlichen Zebrab\"{a}rblingslarven. $K$ bezeichnet die Anzahl an Frames und $N$ die Anzahl an Eiern in der untersuchten Sequenz.}
        \label{fig:03aFischtracking}
\end{figure}

 Die Kreisdetektion ben\"{o}tigt gegen\"{u}ber der Korrelation zweier Bilder signifikant mehr Rechenzeit. Ein besserer Ansatz ist es daher, die in einer initialen Kreisdetektion gewonnene Information der Eiposition auszunutzen und das Chorion mittels Korrelation zu tracken. Allerdings muss die Korrelation f\"{u}r jedes zu trackende Ei durchgef\"{u}hrt werden, w\"{a}hrend der Aufwand der Kreisdetektion von der Eianzahl unabh\"{a}ngig ist. Dennoch ergibt sich f\"{u}r eine typische Sequenz mit etwa 1000 Frames und 8 Eiern ein deutlicher Geschwindigkeitsvorteil gegen\"{u}ber der ausschlie{\ss}lichen Kreisdetektion. Bei der Korrelation wird der Suchraum des Ortes im darauf folgenden Frame (das \sog Zielbild) auf einen kleinen Bereich um die bekannte Eiposition verringert. Die Einschr\"{a}nkung des Suchraums kann getroffen werden, da sich das Ei zwischen zwei Frames nur um eine endliche Pixel\-anzahl verschiebt und zudem die Driftbewegungen der Eier verh\"{a}ltnism\"{a}{\ss}ig langsam bez\"{u}glich der Abtastrate sind.
 Problematisch hierbei sind jedoch kleine Abweichungen jeder gefundenen Position in jedem getrackten Frame, was zu einem "`Wegdriften"' der aktuellen Position von der tats\"{a}chlichen Eiposition f\"{u}hren kann. Das Problem des Wegdriftens wurde gel\"{o}st, indem als Suchbild beim Tracking nicht das gefundene Ei, sondern ein einfacher Kreis mit dem Radius der Eih\"{u}lle verwendet wurde. Da im Suchbild nur die Eih\"{u}lle einen Kreis darstellt und im Suchraum nur ein Ei vorhanden sein kann, ist ein Wegdriften des Suchfeldes bei ausreichender Bildqualit\"{a}t ausgeschlossen. Mit dem zu Beginn aus der Hough-Kreisdetektion ermittelten Eidurchmesser wird das Suchbild mit lediglich einem Kreis erstellt. Dieses Bild wird im Suchraum mittels Korrelation so positioniert, dass es das Chorion \bzw das Kantenbild des Chorions m\"{o}glichst gut abdeckt. Hierf\"{u}r wird eine \sog G\"{u}te-Matrix erstellt, die einen Wert f\"{u}r die \"{U}bereinstimmung des Suchbildes im Zielbild enth\"{a}lt. Das Maximum in der G\"{u}te-Matrix stellt die wahrscheinlichste neue Position des Chorions dar und wird als Startpunkt f\"{u}r die Suche im n\"{a}chsten Frame verwendet. Der Vorgang wird wiederholt, bis die gesamte Bildsequenz abgearbeitet ist und in jedem Frame die Position des Chorions bestimmt wurde. \abb \ref{fig:03aFischtracking} veranschaulicht den Ablauf des Trackings.

Als Empfehlung zur Bestimmung der Parameter erweist es sich als zweckm\"{a}{\ss}ig, den mittleren Eidurchmesser in den ersten Frames der ersten Bildsequenzen einer \hts manuell zu ermitteln. Dieser kann dann mit einer Toleranz von 10\% zur weiteren automatischen Kreissuche angewandt werden. F\"{u}r den Suchraum wurden um das Suchbild problemspezifisch definierte Entfernungen (Standardparameter: vier Pixel) in $X$- und $Y$-Richtung hinzugef\"{u}gt. Dies erweist sich f\"{u}r die Zebrab\"{a}rblinge bei einer Bildwiederholungsfrequenz von 30\,fps als ausreichend.

\subsection{Differenzbilder}
Differenzbilder sind eine einfache M\"{o}glichkeit die Ver\"{a}nderung der Pixel\-werte und damit die Bewegung der Zebrab\"{a}rblingslarve in einem Bild darzustellen. Hierzu wird bei einer Bildsequenz das zeitlich sp\"{a}ter aufgezeichnete Bild vom vorhergehenden abgezogen. Um negative Pixel\-werte zu vermeiden, wird der Absolutwert gebildet. Werden alle Bilder des Bildstroms $\mathbf{BS}$ nach dem Zeitpunkt der Akquise geordnet und wird sich zur einfacheren Notation hierbei auf lediglich eine Modalit\"{a}t $I_c=1$ und eine Schichtaufnahme $I_z=1$ beschr\"{a}nkt, so ergibt sich eine Zeitreihe ${I}_{i_x i_y}[k]$ mit $k=1\dots K$ Abtastzeitpunkten. F\"{u}r die Differenzbilder l\"{a}sst sich dann schreiben:
 \begin{equation}\label{eqn:Differenzbilder}
\boldsymbol{\Psi}_d[k]=|\mathbf{I}[k]- \mathbf{I}[k-1]| \quad \text{mit } k=2\dots K
  \end{equation}
Die entstandenen Differenz- oder Merkmalsbilder bilden die Basis f\"{u}r eine Vielzahl von Einzelmerkmalen, deren gemeinsames Ziel es ist, die Menge an Bewegung im gesamten Bild oder in Ausschnitten zu quantifizieren. Ein einzelnes Differenzbild aus einer solchen Zeitreihe wird im Folgenden mit $\mathbf{I}_d$ bezeichnet.

\section{Nutzsignalzeitreihen und -merkmale}\label{subsec:FischInf_Zeitreihe}

Drei Typen von \Nutzsigen werden unterschieden:
\begin{enumerate}
  \item \Nutzsige, die sich auf das Aussehen des Zebrab\"{a}rblings beziehen,
  \item \Nutzsige, die sich auf die Bewegung des Zebrab\"{a}rblings beziehen und
  \item \Nutzsige, die sich auf die Signalst\"{a}rke von \zB fluoreszierenden Markern beziehen.
\end{enumerate}
Ziel der Merkmale ist es, die \Nutzsige zu quantifizieren und somit vergleichbar zu machen. F\"{u}r jeden Typ der \Nutzsige wurden mehrere Merkmale entwickelt. Zwar ist ein einziges, signifikantes Merkmal zur Klassifikation oft bereits ausreichend, doch kann durch die Ber\"{u}cksichtigung mehrerer Merkmale eine niedrigere Fehlerrate erzielt werden. Gerade bei der starken Beeintr\"{a}chtigung der \Nutzsige durch St\"{o}rfaktoren, wie es bei \htsen der Fall ist, ist die Extraktion von mehreren Merkmalen von Vorteil. Generell hat es sich f\"{u}r Data-Mining Methoden bei \htsen bew\"{a}hrt, m\"{o}glichst viele Merkmale zur Auswertung heranzuziehen und signifikante Merkmale sp\"{a}ter ausfindig zu machen. In den folgenden Abschnitten  wird ein spezifischer, beschreibender Merkmalssatz aufgebaut, auf dessen Basis daraufhin eine Bewertung erfolgen kann. Damit ergeben sich als Eingangsgr\"{o}{\ss}en der Modulkategorie von der Segmentierung bereitgestellten \Nutzsige $\boldsymbol{\Psi}$ und als Ausgangsgr\"{o}{\ss}en Zeitreihen $\mathbf{Y}_{ZR}$ oder Merkmale $\mathbf{X}$ (vgl. Formel (\ref{eq:Merkex2}) und \abb \ref{fig:Signalmodell}).

\subsection{Instantane Merkmale} \label{subsec:Module_Instantane_Merkmale}
F\"{u}r den ersten Typ Merkmale, solche, die sich auf das Aussehen des Zebrab\"{a}rblings beziehen, sind  Einzelbilder zur Merkmalsextraktion ausreichend. Sie werden entweder durch Auswahl von Zeitpunkten, nach den Methoden aus Abschnitt \ref{subsec:Auswahl_Zeitpkt} aus dem Bildstrom entnommen, oder es wurden bereits bei der Bildakquise lediglich Einzelbilder je Versuchseinheit akquiriert. Die Merkmale beurteilen ohne Ber\"{u}cksichtigung einer zeitlichen Ver\"{a}nderung den Zustand zum jeweiligen Aufnahmezeitpunkt. Zur Merkmalsextraktion wird ein einzelnes, vorverarbeitetes und segmentiertes Bild herangezogen.

Ein histogramm-basierendes Merkmal ist der Schwerpunkt des Histogramms $H$ \"{u}ber die Pixel\-werte im Bild~$\boldsymbol{I}$ und berechnet sich f\"{u}r ein Einzelbild gem\"{a}{\ss}:
\begin{equation}\label{eqn:Merkmal_1}
x_1=\frac{\sum_{i_h=0}^{\phi}\limits i_h \cdot H(i_h) }{\sum_{i_h=0}^{\phi}\limits H(i_h) }.\\
\end{equation}
wobei f\"{u}r das Histogramm  $H$  gilt:\\
\begin{equation}
H(i_h)=\text{card}\{(i_x,i_y)|I(i_x,i_y)=i_h\}\quad \text{mit } i_h=1\dots \phi, i_h \in \mathbb{N} .
\end{equation}
Der Operator \emph{$\text{card}$} bezeichnet die Kardinalit\"{a}t oder M\"{a}chtigkeit, welche f\"{u}r endliche Mengen gleich der Anzahl der Elemente einer bestimmten Menge ist.

Ein \"{a}hnliches, grauwert-basiertes Merkmal ist der Mittelwert \"{u}ber alle Pixel im segmentierten Bild  $\mathbf{I}$. Es hat sich hierbei bew\"{a}hrt, Hintergrundpixel zuvor auf den Wert $0$ zu setzen und bei der Berechnung auszuschlie{\ss}en:
\begin{equation}
x_2=\frac{\sum_{i_x=1}^{I_x}\limits \sum_{i_y=1}^{I_y}\limits I(i_x,i_y)}{I_x \cdot I_y - \,
\operatorname{card}\{\,(i_x,i_y)\, |\, I(i_x,i_y)=0\}}.
\label{eqn:03Merkmal_2}
\end{equation}
Beide Werte geben einen allgemeinen Hinweis, ob der Inhalt des segmentierten Bildes vergleichbar ist. Eine \"{a}hnlich entwickelte Zebrab\"{a}rblingslarve hat bei gleichen Akquise-Parametern auch eine vergleichbare Grauwertverteilung und somit einen vergleichbaren Schwerpunkt des \og Histogramms \bzw einen \"{a}hnlichen Mittelwert der Grauwerte.

Steht etwa, wie bei den Eiern des Zebrab\"{a}rblings, das Vorwissen zur Verf\"{u}gung, dass sich die relevante Information in der Mitte des segmentierten Bildes (\dhe im Zentrum des Chorions) befindet, lassen sich weitere Merkmale wie der "`mittlere Grauwert der mittleren Bildzeile"' und die "`Schwankung der Grauwerte entlang der mittleren Bildzeile"' berechnen. Beide Merkmale sind ein guter Indikator f\"{u}r den Entwicklungsstand der Zebrab\"{a}rblinge. Die Merkmale lassen sich weiter verbessern, wenn sie auf das Zentrum des segmentierten Bereichs einschr\"{a}nken werden, indem lediglich die mittleren 30\% der mittleren Bildzeile ausgewertet werden:
\begin{equation}
x_3=\frac{\sum_{i_x= n_1 }^{n_2} \limits I\left(i_x,\left(\frac{I_y}{2}\right)\right)}{n_2 - n_1 +1}
\label{eqn:03Merkmal_3}
\end{equation}
mit gerundeten Werten $n_1= [0.3 \cdot I_x]$ und $n_2= [0.7 \cdot I_x]$.

Die Schwankung um die mittlere Bildzeile l\"{a}sst sich nach Korrektur um die mittleren Grauwerte $c$ berechnen durch
\begin{equation}
\delta (i_x) = \left|I\left((i_x,\left(\frac{I_y}{2}\right)\right)-c\right|.
\end{equation}
Nach Aufsummieren ergibt sich das Merkmal ebenfalls \"{u}ber den mittleren Bereich der mittleren Bildzeile
\begin{equation}
x_4=\sum_{i_x=n_1}^{n_2} \limits \delta(i_x).
\end{equation}
Die Anzahl an Kanten in Kantenbildern ist ein guter Indikator f\"{u}r die Anzahl an ausgepr\"{a}gten Details im Zebrab\"{a}rbling und somit f\"{u}r die Vergleichbarkeit des Entwicklungsstandes. Die Algorithmen zur Kantendetektion "`Canny"' $I_{\text{Canny}}$ \cite{Canny83} und "`Laplacian of Gaussian (LoG)"' $I_{\text{LoG}}$ \cite{Huertas86} wurden angewandt und ausgewertet:
\begin{align}\label{eqn:03Merkmal_6}
x_{5}=\operatorname{card}\{(i_x,i_y)\,|\,&I_{\text{Canny}}(i_x,i_y)=1\}, \\
x_6=\operatorname{card}\{(i_x,i_y)\,|\,&I_{\text{LoG}}(i_x,i_y)\;\;=1\}.
\end{align}
Es lassen sich weitere Merkmale ermitteln, indem das Aussehen der Objekte im Bild untersucht wird. Beispielsweise l\"{a}sst sich die "`Kugeligkeit"' oder "`Rundheit"' des Bildinhalts des segmentierten Bildstroms bestimmen. Daf\"{u}r wird nach Anwendung des ISO-Data Algorithmus \cite{Hathaway87} die l\"{a}ngste $M_{\text{max}}$ mit der k\"{u}rzesten Halbachse $M_{\text{min}}$ der umschlie{\ss}enden Ellipse verglichen. ($\mu_{11},\mu_{22},\mu_{12}$) sind hierf\"{u}r die notwendigen zentralen Momente des gefilterten Bildes (\vgl  \cite{Haralick92}).
\begin{eqnarray}
M_{\text{max}}= 2 \cdot \sqrt{2} \cdot \sqrt{\mu_{11} + \mu_{22} + \sqrt{(\mu_{11} - \mu_{22})^2 + 4 \cdot \mu_{12}^2}}.\\
M_{\text{min}}= 2 \cdot \sqrt{2} \cdot \sqrt{\mu_{11} + \mu_{22} - \sqrt{(\mu_{11} - \mu_{22})^2 + 4 \cdot \mu_{12}^2}}.
\end{eqnarray}
\begin{equation}
x_7=M_{\text{max}} - M_{\text{min}}.
\end{equation}
F\"{u}r einen Validit\"{a}tstest wird die Gr\"{o}{\ss}e jedes Bildes als Merkmal extrahiert. So lassen sich bei der Auswertung deutlich gr\"{o}{\ss}ere oder kleinere Bildinhalte ausschlie{\ss}en.
\begin{equation}
x_8= I_x \cdot I_y.
\end{equation}

\subsection{Auf Bewegungen bezogene Merkmale}\label{subsec:Merkmale_Bewegung}
Alle Merkmale, die sich auf die Bewegung der Zebrab\"{a}rblinge beziehen, wurden aus Differenzbildern extrahiert. Die Merkmale unterscheiden sich vornehmlich dadurch, dass sie unterschiedlichen Wert auf charakteristische Ph\"{a}nomene der Bewegungen der Larven, wie etwa lokal oder global auftretende \"{A}nderungen, legen.

Ein geeignetes Merkmal ist die Standardabweichung in den Differenzbildern nach Formel~(\ref{eqn:Differenzbilder}):
  \begin{equation}
x_9[k]=\frac{1}{I_x I_y -1}\sum\limits_{i_x=1}^{I_x}\sum\limits_{i_y=1}^{I_y}\left(I_{d,i_x i_y}[k]-\overline I_{d}[k]\right)^2
  \end{equation}
  mit
  \begin{equation}
\overline I_ {d}[k]=\frac{1}{I_x I_y} \sum\limits_{i_x=1}^{I_x}\sum\limits_{i_y=1}^{I_y}I_{d,i_x i_y}[k].
  \end{equation}
Das Merkmal quantifiziert die Variation der Pixel im Differenzbild. Bei Bewegung steigt die Standardabweichung an. Eine Variante ist es, das Merkmal \"{u}ber die Anzahl der Pixel und die Farbtiefe $ \phi$ (Die Farbtiefe wird \"{u}blicherweise bei der Bildakquise festgelegt) zu normalisieren:
\begin{equation}
x_{10}[k]= \frac{1}{I_x I_y \; \phi} \sum\limits_{i_x=1}^{I_x}\sum \limits_{i_y=1}^{I_y}I_{d,i_x i_y}[k]
  \end{equation}
Das Merkmal $x_{10}$ gibt Auskunft \"{u}ber die mittlere relative \"{A}nderung der Pixel.

Um Differenzen innerhalb dunkler und heller Bildbereiche direkt vergleichbar zu machen, wird der Mittelwert \"{u}ber die Intensit\"{a}tswerte der Pixel normalisiert. Um Divisionen durch Null zu vermeiden, wird eine Konstante $x_{\text{off}}$ im Nenner addiert:
  \begin{equation}
x_{11}[k]=\frac{1}{I_xI_y}\sum\limits_{i_x=1}^{I_x}\sum\limits_{i_y=1}^{I_y}
\frac{I_{d,i_x i_y}[k]}{I_{\text{max},i_x i_y}[k]+x_{\text{off}}}
  \end{equation}
  mit
    \begin{equation}
I_{\text{max},i_x i_y}[k]=\text{max}(I_{i_x i_y}[k],I_{i_x i_y}[k-1]).
  \end{equation}
Stark bewegte Regionen im Bild lassen sich quantisieren, wenn die Anzahl relevanter Pixel\-\"{a}nderungen im Differenzbild bestimmt wird. Dazu werden Pixel gez\"{a}hlt, die Differenzen gr\"{o}{\ss}er als ein dynamischer Schwellenwert $I_{\text{dyn}}$ aufweisen:
\begin{equation}\label{eq:Merkmal12}
    x_{12}[k]=\underset{i_x,i_y}{\text{card}}(I_{d,i_xi_y}[k]|I_{d,i_xi_y}[k] > I_{ \text{dyn}}[k])
\end{equation}
Der dynamische Schwellenwert wird so bestimmt, dass er Differenzwerte \"{u}bertrifft, die aus Bildrauschen resultieren. Bildrauschen bewirkt eine Abweichung der gemessenen Pixel\-werte, ohne tats\"{a}chliche \"{A}nderung des aufgezeichneten Objektes. Wird ein Pixel eines mit Rauschen \"{u}berlagerten Bildes mit $I_d^{*}$  bezeichnet, so liefern der Mittelwert und die Standardabweichung (hier durch das dreifache der Standardabweichung 99,7\% aller Pixelwerte ber\"{u}cksichtigt) dieses Pixels $I_{dyn}$:
\begin{equation}
I_{dyn}[k]= \overline I^{*}_{d}[k]+3s^*_d[k]
  \end{equation}
Durch Bildrauschen entstehen Ausrei{\ss}er im Differenzbild mit Werten oberhalb eines mittels des c-Quantils \"{u}ber den Pixelwerten des Bildes gebildeten Schwellenwertes $I_{\text{quant}_c}$ oder unterhalb eines mittels des (1-c)-Quantils \"{u}ber den Pixelwerten des Bildes gebildeten Schwellenwertes $I_{\text{quant}_{1-c}}$. Ihre Pixel\-werte lassen sich bestimmen durch:
\begin{equation}
\{(i_x,i_y)\}=\{(i_x,i_y)|I_{\text{quant}_c}<I_{ d,i_x i_y}<I_{\text{quant}_{1-c}}\}, \quad 0\le c < 0.5 .
  \end{equation}
c muss hierbei kleiner sein als der Anteil der maximal im Bild auftretenden Bewegung. Ein Wert von c=0.4 hat sich in der Praxis bew\"{a}hrt. Damit ergibt sich
\begin{equation}
s^*_d[k]=\frac{1}{(1-c)I_xI_y-1}\sum\limits_{(i_x, i_y)|c<I_{d,i_x i_y}<1-c}(I_{d,i_x i_y}[k]-\overline{I}_{\,d,i_x i_y}^{*}[k])^2
  \end{equation}
  mit%
\begin{equation}
\overline{I}_{\,d,i_x i_y}^{*}[k]=\frac{1}{(1-c)I_xI_y-1}\sum\limits_{(i_x, i_y)|c<I_{d,i_x i_y}<1-c}I_{d,i_x i_y}[k]
  \end{equation}
Das Merkmal $x_{13}[k]$ bezieht sich auf die Anzahl relevanter, relativer Pixel\-\"{a}nderungen im Differenzbild und berechnet sich wie $x_{12}$ in Formel (\ref{eq:Merkmal12}), allerdings mit normalisierten Differenzwerten
$
\frac{I_{d,i_x i_y}[k]}{I_{x}I{y}}
$
anstelle von $I_{d,i_x i_y}[k]$.

\begin{equation}\label{eq:Merkmal13}
    x_{13}[k]=\text{card}\left((i_x,i_y)\left|\frac{I_{d,i_x i_y}[k]}{I_{x}I_{y}} > I_{ \text{dyn}}[k]\right)\right.
\end{equation}

Die maximale \"{A}nderung im Differenzbild  ist ein Ma{\ss} f\"{u}r die st\"{a}rkste Bewegung der Zebrab\"{a}rblingslarve:
\begin{equation}
x_{14}[k]=\underset{i_x,i_y}{\text{max}}(I_{\,d,i_x i_y}[k])
  \end{equation}%
Eine Variante des Merkmals $x_{14}$ ist es, das Merkmalsbild zuvor mit einem 3x3 Gau{\ss}filter (Maximum im gegl\"{a}tteten Differenzbild) zu falten. Dies reduziert den Einfluss einzelner Ausrei{\ss}er im Differenzbild. Es hat sich bew\"{a}hrt f\"{u}r $\sigma$ 0.5 zu w\"{a}hlen:
 \begin{equation}
h(i_x,i_y)=\exp -\frac{i_x^2+i_y^2}{2\sigma ^2}:
  \end{equation}%
 \begin{equation}
x_{15}[k]=\underset{I_x,I_y}{\text{max}}(I_{\,d,\text{smoothed},i_x i_y}[k])
  \end{equation}%
  mit
\begin{equation}
  I_{d,\text{smoothed},i_x i_y}[k]=I_{\,d,i_x i_y}[k]\ast h(i_x,i_y).
\end{equation}
wobei $\ast$ f\"{u}r den Faltungsoperator und $h(i_x,i_y)$ f\"{u}r die Filtermaske steht.

Die Summe aller \"{A}nderungen im Differenzbild ist ein Ma{\ss} f\"{u}r die Menge an Bewegung im Bild.
Verschiedene Einfl\"{u}sse bei der Bildakquise k\"{o}nnen dazu f\"{u}hren, dass im Bild ein Beleuchtungs"~ \bzw Helligkeitsverlauf auftritt, was die Bilddynamik beeinflusst. Da die Absolutwerte der Differenzbilder von der Dynamik in den Rohdaten begrenzt wird, entstehen so auch Abweichungen in der Messung. Um die Vergleichbarkeit der Merkmale und deren absoluten Werte sicherzustellen, ist eine Normierung sinnvoll. Daher wurden die Merkmalswerte  mittels der Standardabweichung \"{u}ber alle Pixel\-werte normalisiert. Somit berechnet sich das Merkmal nach:
  \begin{equation}
x_{16}[k]=\frac{1}{x_{9}[k]}\sum\limits_{i_x=1}^{I_x}\sum\limits_{i_y=1}^{I_y}
I_{d,i_x i_y}[k]
  \end{equation}
Wichtig bei einer solchen Normalisierung ist jedoch ein vergleichbarer Bildinhalt. Ist in den Bildern beispielsweise eine variierende Anzahl an Eiern abgebildet, so scheitert die Methode.

Die einfachste Form zur Ermittlung der Bewegung im Differenzbild ist die Summe aller \"{A}nderungen, ohne jegliche Normalisierung nach:
  \begin{equation}
x_{17}[k]=\sum\limits_{i_x=1}^{I_x}\sum\limits_{i_y=1}^{I_y}
I_{d,i_x i_y}[k]
  \end{equation}

\subsection{Auf Signalst\"{a}rke bezogene Merkmale}

Merkmale zur Bestimmung der Signalst\"{a}rke finden Anwendung, wenn die St\"{a}rke des auszuwertenden \Nutzsigs durch den optischen Detektor erfasst wurde. Bei \htsen mit Zebrab\"{a}rblingen ist dies vor allem bei mittels Fluoreszenzmikroskopie generierten Bilddaten der Fall. Solche Merkmale gestatten einen R\"{u}ckschluss auf die Menge an fluoreszent leuchtenden Markern. Die Berechnung erfolgt meist mittels Addition aller Pixel\-werte im segmentierten Bildstrom oder es wird auf das arithmetische Mittel \bzw den Median \"{u}ber die Pixel\-werte zur\"{u}ckgegriffen. Bei der Auswertung von auf Basis fluoreszierend leuchtender Marker akquirierter Bilder ist zur richtigen Deutung jedoch eine Ber\"{u}cksichtigung der Messeinrichtung n\"{o}tig. Eine weitere Schwierigkeit sind fluoreszente Strahlungen des Hintergrundes sowie ein oft hohes Rauschen im aufgezeichneten Bild. Eine genauere Betrachtung sowie eine Checkliste zu den Besonderheiten der Quantifizierung von Fluoreszenzbildern finden sich in \cite{waters09}.


\section{Merkmalsauswahl}

Bei der Merkmalsauswahl werden die extrahierten Merkmale f\"{u}r die Klassifikation vorbereitet bzw. ausgew\"{a}hlt. Eingangsgr\"{o}{\ss}en der Modulkategorie sind alle $m_2$ extrahierten Merkmale $\mathbf{X}$. Die Ausgangsgr\"{o}{\ss}e ist eine reduzierte Anzahl an Merkmalen $m_3$ bzw. $m_d$.
Zu Beginn werden Merkmalswerte, welche von fehlerhaften Bilddaten wie leeren N\"{a}pfchen, zu vielen Versuchseinheiten, schlechter Beleuchtung \etc stammen, in einem Validit\"{a}tstest ausgeschlossen. Hierbei kann auf einfache Schwellenwerte zur\"{u}ckgegriffen werden, die im Allgemeinen aus Vorwissen stammen. Wurde beispielsweise die Eigr\"{o}{\ss}e als Merkmal extrahiert, kann ein Bereich sinnvoller Werte bestimmt werden. Merkmale au{\ss}erhalb des bestimmten Bereichs sollten verworfen werden. Sind solche Pixel ausgeschlossen, kann eine Auswahl der besten Merkmale erfolgen, denn die extrahierten Merkmale sind \zT redundant und es kommt im konkreten Anwendungsfall auf die wirkende St\"{o}rung an, welches der Merkmale eine trennstarke Information enth\"{a}lt. Je mehr redundante Merkmale zur Verf\"{u}gung stehen, desto schwieriger wird die Entscheidungsfindung. Eine zu gro{\ss}e Anzahl an hierbei ber\"{u}cksichtigten Merkmalen kann sogar zu einer Fehlklassifikation f\"{u}hren \cite{Cristianni00}. Wird die Auswertung in Echtzeit zur Verf\"{u}gung gestellt und auf die Akquise zur\"{u}ckgekoppelt wie \zB bei Roboteranwendungen, ist es schon aus Gr\"{u}nden der Rechenzeit notwendig, die Berechnung auf notwendige Merkmale zu limitieren. 

Zur automatischen Merkmalsauswahl kommen verschiedene bekannte Methoden zum Einsatz. Beispiele sind die univariate Varianzanalyse (ANalysis Of VAriance -- ANOVA) und die multivariate Varianzanalyse  (Multivariate ANalysis Of VAriance -- MANOVA) \cite{Ahrens74}. Die Verfahren vergleichen die Innerklassenvarianzen mit den Zwischenklassenvarianzen der Merkmale und geben so Aufschluss dar\"{u}ber, welches Merkmal oder welche Merkmalskombination am besten zur Unterscheidung der gew\"{a}hlten Klassen geeignet ist. Mit Hilfe der ANOVA wird das Merkmal identifiziert, dessen Relevanz zur Klassentrennung den h\"{o}chsten Wert erreicht. Daraufhin wird das identifizierte Merkmal, mit Hilfe der MANOVA, mit allen verbliebenen Merkmalen kombiniert, die Relevanz der Merkmalskombination ermittelt und die beste Zweier-Merkmalskombination bestimmt. Der Schritt entspricht der Formel (\ref{eq:Abbildung_S3}) und Formel (\ref{eq:Abbildung_S4}). Es hat sich in der Praxis bei \htsen bew\"{a}hrt, den Merkmalsraum auf eine solche Zweier-Merkmalskombination zu reduzieren und zur Klassifikation heranzuziehen. Eine solche Zweier-Merkmalskombination ist ein guter Kompromiss zwischen Interpretierbarkeit, Rechenzeit und G\"{u}te der Klassifikation \cite{Ahrens74,Mikut08}.

\section{Auswertung und Klassifikation}

Nach Auswahl geeigneter Module zur Extraktion von informationstragenden Merkmalen sind mit der formulierten Fragestellung aus Abschnitt \ref{sec:Anforderungsgerechte_Anwendung} alle notwendigen Voraussetzungen f\"{u}r eine erfolgreiche L\"{o}sung eines Klassifikations"~ oder Regressionsproblems in \htsen am Zebrab\"{a}rbling gegeben. Zur Klassenzuordnung kann zumeist auf bekannte Verfahren der Klassifikation zur\"{u}ckgegriffen werden\cite{Mikut08}. Entscheidend f\"{u}r den Erfolg ist die Bereitstellung eines informationstragenden Merkmalssatzes $\mathbf{W}$. Wie bei einer solchen Art von Problemstellung \"{u}blich, muss ein Kompromiss zwischen einer niedrigen Fehlerrate und einer hohen Interpretierbarkeit der Ergebnisse, sowie einem m\"{o}glichst geringen Rechenaufwand gefunden werden. Die Eingangsgr\"{o}{\ss}en bei dieser Modulkategorie sind die vorverarbeiteten Merkmale $\mathbf {W}$ und die Ausgangsgr\"{o}{\ss}en sind die Klassenzuweisungen der Planfaktoren $\mathbf{\hat{Y}}$ bzw. der St\"{o}rfaktoren~$\mathbf{\hat{Z}}$ (vgl. auch Formel~(\ref{eq:Abbildung_S5})).

\subsection{Klassifikation}

Zur Durchf\"{u}hrung der Klassifikation unbekannter Daten muss der Klassifikator zu Beginn mittels eines Lerndatensatzes angelernt werden. Die Frage nach der Anzahl an Versuchseinheiten, die hierf\"{u}r notwendig ist, kann nicht pauschal beantwortet werden und h\"{a}ngt von den Eigenschaften und St\"{o}rfaktoren der \hts ab \cite{Mikut08}. In den meisten F\"{a}llen sind jedoch 20 bis 50 Versuchseinheiten einer Klasse ausreichend. F\"{u}r das Labeln, also das manuelle Zuordnen einer Versuchseinheit der Lerndaten zu einer Klasse, wurde eine komfortable Softwarel\"{o}sung entwickelt und in das Softwarepaket Gait-CAD implementiert (vgl. \kap \ref{subsec:Gait-CAD}), welcher dem Anwender ein zuf\"{a}lliges Beispiel aus dem Datensatz zeigt und die M\"{o}glichkeit bereitstellt, per Mausklick einer der vorhandenen Ausgangsklassen zuzuweisen. So kann auf schnelle Weise f\"{u}r eine \hts am Zebrab\"{a}rbling der Lerndatensatz kreiert werden.

Zur Klassifikation kommen vornehmlich Bayes-Klassifikatoren zum Einsatz, deren Aufgabe es ist, die Zugeh\"{o}rigkeit der Versuchseinheiten auf Basis der extrahierten, reellwertigen Merkmalswerte $\mathbf{W}$ einer der Klassen aus $\mathbf{Y}$ oder $\mathbf{Z}$ diskret zuzuordnen. Mittels des Lerndatensatzes wird ein hierf\"{u}r funktioneller Zusammenhang bestimmt. Der Zusammenhang wird als Diskriminanzfunktion bezeichnet und ist entweder eine Entscheidungsfunktion, die den Merkmalswerten eine Klasse explizit zuordnet, oder besteht aus Distanzen zu den Klassen, welche die Zugeh\"{o}rigkeit implizit ausdr\"{u}cken. Es wird eine Normalverteilung der Merkmale f\"{u}r jede Ausgangsklasse angenommen. Die Diskriminanzfunktion wird mittels der Mahalanobis-Distanz und klassenspezifischen Kovarianzmatrizen bestimmt \cite{mahalanobis1936}. Im Anschluss lassen sich mittels Validierungsverfahren die G\"{u}te \"{u}ber unbekannte Datentupel absch\"{a}tzen und die Auswirkungen zuf\"{a}lliger St\"{o}reinfl\"{u}sse ermitteln \cite{Mikut08}.

\subsection{Regression der Dosis-Wirkungs-Kurve} 
\begin{figure}[htb]
    \centering
    \includegraphics[width=0.45\linewidth]{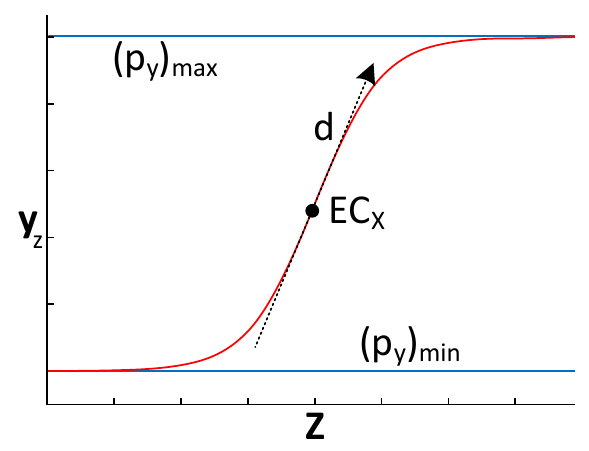}
    \caption[Sigmoidkurve der Dosis-Wirkungs-Kurve]{Sigmoidkurve der Dosis-Wirkungs-Kurve mit den Parametern EC$_{X}$, $(p_y)_{\min}$, $(p_y)_{\max}$ und $d$ \cite{cleland1963}.}
    \label{fig:EC50_Registr}
\end{figure}

\label{subsec:Pr\"{a}s_EC50}
Oft ist es das Ziel, den Effekt des in der \hts systematisch variierten St\"{o}rfaktors  $\mathbf{Z}$ auf einen oder mehrere Planfaktoren $\mathbf{Y}$ zu bestimmen. Bei Toxinen werden hier \"{u}blicherweise Verd\"{u}nnungsreihen angelegt. Es kann auch der Effekt anderer St\"{o}rfaktoren, wie \zB Spannungen oder Temperaturen, be\-obachtet werden. Jede Versuchsreihe wird mit einer unterschiedlichen Konzentration (oder Dosis) ausgef\"{u}hrt. Voraussetzung ist, dass f\"{u}r jede gew\"{a}hlte Stufe eine ausreichende Anzahl an Versuchseinheiten pr\"{a}pariert wurde. Die Klassifikation erfolgt dabei nacheinander f\"{u}r jede der gew\"{a}hlten Konzentrationsstufen und es wird die Klassenzugeh\"{o}rigkeit aller Versuchseinheiten $\hat{\mathbf{Y}}$ gesch\"{a}tzt. Wird die beobachtete Wirkung durch den variierten St\"{o}rfaktor stimuliert, so wird in den meisten F\"{a}llen eine kontinuierliche Zunahme des Effekts erwartet. Die Konzentration wird von einem niedrigen Effekt-Niveau $(p_y)_{\min}$ zu einem hohen Effekt-Niveau $(p_y)_{\max}$ verlaufen (\vgl \abb \ref{fig:EC50_Registr}). Die Dosis-Wirkungs-Kurve folgt typischerweise einer Sigmoidfunktion, welche mittels Regression an die Messwerte angepasst wird. Das entstehende Optimierungsproblem ist parameternichtlinear und wird numerisch gel\"{o}st.

\begin{equation}
  \mathbf{y}_{z_i}=(p_y)_{\min} +\frac{(p_y)_{\max} - (p_y)_{\min}}{1+(\frac{\mathbf{z}_{c_i}}{EC_{X}})^d}
\end{equation}
Hier ist ${\mathbf{y}_z}$ der Wert bez\"{u}glich des untersuchten Effekts (meist in Prozent), $(p_y)_{\min},(p_y)_{\max}$ der niedrigste \bzw h\"{o}chste ermittelte Wert des Effekts, der Vektor $\mathbf{z}_c$ enth\"{a}lt die Konzentrationen und $d$ ist der Absolutwert der Steigung der Sigmoidfunktion \cite{cleland1963}. $EC_{X}$ bezeichnet die Wendestelle der Kurve, welcher einer Konzentration/Dosis zugeordnet wird. Der Parameter $d$ wird als Hill-Koeffizient bezeichnet. F\"{u}r den Fall, dass steigende Konzentrationen die Wirkung hemmen, wird der Parameter negativ und es ergibt sich ein abfallender Verlauf der Kurve in \abb \ref{fig:EC50_Registr}. Die Wirksamkeit auf den gesuchten Effekt l\"{a}sst sich am $EC_{X}$-Wert ablesen. Entscheidend f\"{u}r den Erfolg der Methode ist jedoch, dass sich der betrachtete Effekt bei der Variation des Planfaktors auch einstellt. Sollte sich der Effekt nicht einstellen, scheitert die Regression und es muss \ggf eine Anpassung des Versuchsplanes erfolgen.

%
%
%

\section{Pr\"{a}sentation}
Entscheidend f\"{u}r den Erfolg der \hts ist es, die Pr\"{a}sentation des Ergebnisses derart zu gestalten, dass die eingangs gestellte biologische Frage m\"{o}glichst direkt beantwortet wird. Ist das nicht unmittelbar m\"{o}glich, so muss versucht werden, m\"{o}glichst exakte Hinweise zu liefern. Biologische Wirkzusammenh\"{a}nge sowie die Komplexit\"{a}t der Auswertungskette machen es notwendig, nicht nur das Endergebnis, sondern auch den L\"{o}sungsweg transparent darzustellen. Hierzu ist es wichtig, auch Zwischenergebnisse jedes Moduls nachvollzieh"~ und interpretierbar darstellen zu k\"{o}nnen. Eingangsgr\"{o}{\ss}en der Modulkategorie \emph{Pr\"{a}sentation} sind die Klassifikationsergebnisse $\hat{\mathbf{Y}}$ bzw. $\hat{\mathbf{Z}}$. Ausgangsgr\"{o}{\ss}en sind die aufbereiteten Daten in diverser Form (vgl. Tabelle \ref{tab:Pr\"{a}sentationstechniken}) sowie die Archivierung der Untersuchungsergebnisse.

Ist die Fragestellung beispielsweise die Suche nach der Effekt-Konzentration 50 (EC$_{50}$), bei der ein gesuchter Effekt bei 50\% der Versuchseinheiten auftritt, so ist das Ergebnis der \hts klar formulierbar. Entweder die Regression der Messwerte konvergieren bez\"{u}glich der Dosis-Wirkungs-Kurve und die gesuchte Konzentration EC$_{50}$ l\"{a}sst sich ermitteln oder die gew\"{a}hlten Konzentrationsreihen  m\"{u}ssen angepasst werden.  Ein m\"{o}glichst genauer Hinweis f\"{u}r eine solche Anpassung kann durch die Darstellung der Klassifikationsergebnisse in Form eines Histogramms gegen\"{u}ber der Konzentration erfolgen.
Andere Fragestellungen k\"{o}nnen nicht direkt beantwortet werden, sondern es kann lediglich ein Hinweis durch die Daten gegeben werden, der von Experten gedeutet werden muss. Um einen guten \"{U}berblick \"{u}ber alle Daten zu erhalten, sind automatisch generierte Reportdateien ein geeignetes Mittel. Solche fassen f\"{u}r jede Larve oder jede Platte wichtige Ergebnisse \"{u}bersichtlich zusammen.

Die beste Pr\"{a}sentationstechnik ist daher abh\"{a}ngig von der durchzuf\"{u}hrenden Analyse zu w\"{a}hlen. In Tabelle \ref{tab:Pr\"{a}sentationstechniken} ist eine Auswahl solcher Techniken \"{u}ber bestimmten Problemstellungen aufgelistet, die sich bei \htsen bew\"{a}hrt haben. Die Aufz\"{a}hlung ist keinesfalls ersch\"{o}pfend. Eine etwas allgemeinere Aufz\"{a}hlung der genannten und weiterer Pr\"{a}sentationstechniken findet sich in \cite{Mikut08}. Die Tabelle ordnet einer Art der Analyse jeweils eine m\"{o}gliche Pr\"{a}sentationstechnik zu. In der Bildverarbeitung sind beispielsweise Merkmalsbilder, d.h. Bilder, die typische Bildinhalte in Form von Merkmalen repr\"{a}sentieren und aus den Rohdaten (Rohbildern) berechnet werden, n\"{u}tzlich (vgl. \kap \ref{subsec:Segmentierung}). Auch eine \"{U}berlagerung solcher Merkmalsbilder mit den Rohdaten ist oftmals sehr aussagekr\"{a}ftig (vgl. \zB \abb \ref{fig:PMR_Konzeptumsetzung_Figures_Hough_Ergebniss}). Ein weiteres Beispiel ist die Pr\"{a}sentation des Ergebnisses aus dem Anlernen eines Klassifikators. Das Ergebnis kann z.B. eine Diskriminanzfunktion sein. Hier eignet sich ein Scatterplot \"{u}ber den Lerndaten mit der \"{U}berlagerung der Diskriminanzfunktion. So kann bereits schon durch Anschauen des Ergebnisses eine Aussage \"{u}ber die G\"{u}te des Klassifikators getroffen werden (Ein Beispiel findet sich in \abb~\ref{fig:resultwrapper2}).
\begin{table}[h!tbp]
\centering
\begin{tabularx}{\linewidth}{
L{5cm}X
}
\toprule
 \textbf{Analyse}& \textbf{Pr\"{a}sentationstechnik} \\
\midrule
Bildverarbeitungsanalyse & Merkmalsbilder \bzw Sequenzen von Merkmalsbildern. \"{U}berlagerung von Rohdaten mit Merkmalsbildern\\[1ex]
Diskriminanzfunktion & Scatterplot \"{u}ber Lerndaten mit Einblendung der Trennfl\"{a}chen\\[1ex]
Ergebnisinterpretation &  Sequenzen und Bilder des Bildstroms \evtl mit Einblendung der trennst\"{a}rksten Merkmale\\[1ex]
Falschfarbenbilder & Darstellung eines Intensit\"{a}ts- oder Grauwertbilds durch Zuordnung einer Farbe meist als \"{U}berlagerung eines anderen Bildes (vgl. \abb \ref{fig:Registrierung}) \\[1ex]
Gesamt\"{u}bersicht & Reportdatei mit einer Auswahl der hier genannten Pr\"{a}sentationstechniken bezugnehmend auf Plan- und St\"{o}rfaktoren\\[1ex]
Heatmaps & Darstellung der Werte einer Matrix in Form eines Bildes mittels einer Farbkodierung (vgl. \abb \ref{fig:PMR_Konzeptumsetzung_Figures_Akkumulationsmatrix})\\[1ex]
Klassifikationsergebnisse & Klassenspezifische Histogramme, Ergebnistabellen, Konfusionsmatrix, \"{U}berlagerung von Bildern und Sequenzen des Bildstroms mit den Klassifikationsergebnissen, Scatterplot mit Klassenzuweisung und Diskriminanzfunktion\\[1ex]
Merkmalsselektion& Sortierte Merkmalslisten mit Merkmalsrelevanzen \\[1ex]
Merkmalsverteilung & Boxplot, Histogramm eines Merkmals\\[1ex]
Merkmalsverteilung (paarweise) & Scatterplot, Falschfarbenbilder, Heatmaps\\ \\[-1ex]
Rohdatenanalyse & Sequenzen und Bilder der Rohdaten sowie deren Grauwertverteilung, \zB in Form von Histogrammen \\[1ex]
Zeitreihen & 2D-Plot, Heatmap, \evtl mit Einblendung von Klassifikationsergebnissen wie \zB Bewegungsphasen und "~ereignissen \\[1ex]
\bottomrule
\end{tabularx}
\caption{Ausgew\"{a}hlte Pr\"{a}sentationstechniken f\"{u}r \htsen}\label{tab:Pr\"{a}sentationstechniken}
\end{table}

\section{Bewertung}
Der in Kapitel \ref{sec:BV_Module} vorgestellte Modulkatalog bietet eine Auswahl an neuen (wie \zB \kap \ref{subsec:Modulkatalog_Normalisierung} und \kap \ref{Trennung von Einzelobjekt und Hintergrund}) sowie bew\"{a}hrten Methoden (\zB \kap \ref{subsec:Auswahl_Zeitpkt}) f\"{u}r die Bildanalyse in Hochdurchsatzsystemen am Beispiel des Zebrab\"{a}rblings. Je nach den Gegebenheiten des zu untersuchenden Nutzsignals und akquirierten Bildstroms ist es nun f\"{u}r eine Vielzahl von Versuchen m\"{o}glich, passende Methoden auszuw\"{a}hlen und zu einer leistungsf\"{a}higen Auswertungskette  zusammenzuf\"{u}gen. Der Katalog deckt dabei alle Verarbeitungsschritte von den Rohdaten bis hin zur Pr\"{a}sentation und Archivierung ab. Die in \abb \ref{fig:Prozess_Uebersicht} \"{u}bersichtlich nach Reihenfolge der Abarbeitung aufgelisteten Methoden sind vorzugsweise nach dem bereits in Abschnitt \ref{sec:Anforderungsgerechte_Anwendung} vorgestellten Ablaufdiagramm auszuw\"{a}hlen und im Hinblick auf die Erf\"{u}llung der in Abschnitt \ref{sec:HDU_Anforderungen} eingef\"{u}hrten Anforderungen im Zusammenspiel mit den anderen Versuchsparametern zu pr\"{u}fen. Der modulare Aufbau erm\"{o}glicht zudem das leichte Erg\"{a}nzen weiterer, an bisher unbekannte Problemstellungen angepasste Methoden. Die folgenden  Kapitel zeigen  die exemplarische Implementierung und Anwendung des erarbeiteten Konzeptes und des Modulkataloges.

\chapter{Implementierung und Skalierung}\label{chap:Implementierung}
\section{\"{U}bersicht}

Ziel der Implementierung f\"{u}r \htsen ist es, die vorgestellten Module lauff\"{a}hig umzusetzen. Die Implementierung soll m\"{o}glichst einfach und anwenderfreundlich sein und somit allen Beteiligten der interdisziplin\"{a}ren Projekte zur Verf\"{u}gung stehen. Des Weiteren soll Einfluss auf die Parameter bestehen und sowohl die Bilddaten als auch die zugeh\"{o}rigen Repr\"{a}sentationen, Faktoren und Klassen m\"{u}ssen einsehbar und anschaulich pr\"{a}sentierbar sein. Ein Filtern der Daten auf Basis der Faktoren ist ebenso unverzichtbar wie die direkte Ausf\"{u}hrbarkeit der Bildverarbeitungsalgorithmen. Zum Beispiel kann es von Interesse sein, alle Bildsequenzen einzusehen, die mittels eines bestimmten Mikroskops aufgenommen wurden oder alle Merkmalswerte eines bestimmten Merkmales, welches an einem bestimmten Tag aufgenommen wurde, zu plotten. Die Bilddaten werden meist auf einem zentralen, \"{u}ber eine Netzwerkverbindung angeschlossenen Server gespeichert und sollten f\"{u}r eine schnelle, parallele Berechnung auf mehrere Computer (\sog Clients) verteilbar ausgef\"{u}hrt werden k\"{o}nnen. Eine Parallelisierung wiederum erfordert eine zus\"{a}tzliche M\"{o}glichkeit, die Teilergebnisse zu fusionieren. Der schnelle direkte Zugriff auf alle Methoden, Daten und Klassenzugeh\"{o}rigkeiten ist wichtig f\"{u}r ein gutes Verst\"{a}ndnis des Datensatzes und zur Kontrolle auf Inkonsistenzen. So kann \zB das Aufl\"{o}sen von Daten nach Positionen in der Mikrotiterplatte neue St\"{o}rfaktoren aufzeigen. Aufgrund der interdisziplin\"{a}ren Problemstellung ist es zudem wichtig, dass die Auswertung sowie eine Anpassung der Parameter auf neue Datens\"{a}tze weitgehend auch ohne Expertenwissen auf dem Gebiet der Informatik (\zB von einem Biologen) ausf\"{u}hrbar sind. Die Datentypen sind hierbei sehr heterogen, und es sind jeweils Bilddaten, Merkmale (und Zeitreihen) und Metadaten zuzuordnen und neu generierte Daten zu speichern. Die Bilddaten m\"{u}ssen prozessiert werden, es muss bekannt sein, mit welchem Algorithmus, welche Zwischenergebnisse erstellt wurden und wie die Einzelversuche miteinander zu vergleichen sind. Des Weiteren muss die Berechnung schnell erfolgen und die Ergebnisse m\"{u}ssen pr\"{a}sentierbar sein. Ein weiterer Aspekt ist die Archivierung. Dabei m\"{u}ssen nicht nur die Versuchsdaten und Ergebnisse reproduzierbar abgelegt werden, sondern auch die verwendete Version der Algorithmen und die gew\"{a}hlten Parameter archiviert werden.

Um die genannten Aufgaben zu erf\"{u}llen, wurde f\"{u}r die vorliegende Arbeit eine neue grafische Oberfl\"{a}che (GUI) erstellt und die bereits bestehende MATLAB Toolbox Gait-CAD erweitert. Skriptbasierte L\"{o}sungen wurden verworfen, da solche den Nachteil haben, dass der Zugriff auf Parameter und Optionen oft kryptisch ist und eine lange Einarbeitungszeit erfordert. In einer GUI kann jeder Parameter mit entsprechenden Hinweisen versehen werden. Auch ist eine GUI oft selbsterkl\"{a}rend oder die Auswirkung einzelner Optionen kann durch den Nutzer leicht in Erfahrung gebracht werden. Zudem m\"{u}ssen die Ergebnisse im Merkmalsraum oder die Merkmalsbilder ohnehin in einer grafischen Ausgabe dargestellt werden.

Zusammenfassend kann f\"{u}r die Implementierung gefordert werden, dass sie bezugnehmend auf \abb \ref{fig:Neues_Konzept_Prinzip_b} und die Kategorien des Modulkatalogs gem\"{a}{\ss} \abb \ref{fig:Prozess_Uebersicht}

\begin{enumerate}
  \item die Einzelversuche verkn\"{u}pft mit
  \begin{itemize}
    \item den zugeh\"{o}rigen Faktoren,
    \item dem Bildstrom,
    \item den Zwischenergebnissen,
    \item der Repr\"{a}sentation des Nutzsignals \newline (Nutzsignalzeitreihen \bzw Nutzsignalmerkmale),
    \item der Zuordnung zu den Ausgangsklassen.
  \end{itemize}
  \item Zugriff auf die Datenverarbeitung und deren Parameter erlaubt,
  \item die Berechnung skalierbar macht,
  \item die Ergebnisse pr\"{a}sentiert und
  \item die Untersuchung und Ergebnisse archiviert.
\end{enumerate}

Im Folgenden wird zu Beginn die Umsetzung der GUI und die Implementierung der Algorithmen beschrieben. Daraufhin wird auf die M\"{o}glichkeit zur Skalierung und zum verteilten Rechnen, wie es zum schnellen Berechnen der Daten notwendig ist, eingegangen.

\section{Umsetzung und Methoden}

\begin{figure}[htbp]
        \centering
            \includegraphics[page=4]{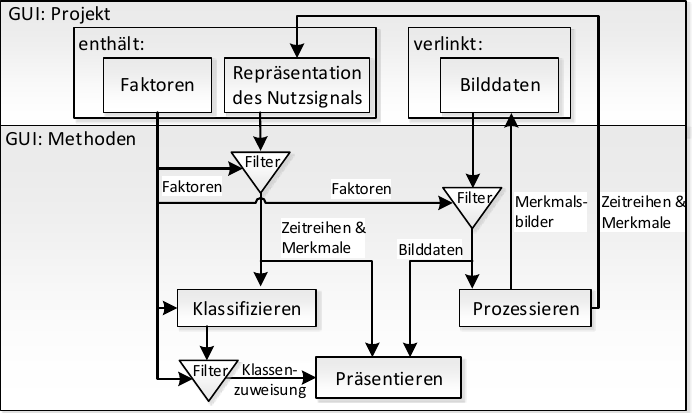}
        \caption[Prozessschritte und Signalfluss der grafischen Benutzeroberfl\"{a}che]{Prozessschritte und Signalfluss der grafischen Benutzeroberfl\"{a}che zur Datenverarbeitung bei \htsen}
        \label{fig:GUI_Uebersicht_4}

\end{figure}

Jede \hts wird als ein abgeschlossenes Projekt betrachtet. Ein Projekt enth\"{a}lt f\"{u}r jede Versuchseinheit eine Sammlung an Daten oder Verlinkungen (vgl. \abb \ref{fig:GUI_Uebersicht_4}). Eine Verlinkung ist eine Referenzierung auf den tats\"{a}chlichen Speicherort einer Datei. Das bedeutet, dass die Bilddaten nicht physisch im Projekt gespeichert werden, sondern lediglich auf die Datei verwiesen wird. Durch die Verweise bleiben die Projekte handlich, es entstehen keine mehrfachen Kopien der Bilddaten und es ist dennoch der direkte Zugriff auf die Daten \"{u}ber die Verlinkung gegeben.
Eine solche Sammlung wird f\"{u}r jede Versuchseinheit im Folgenden Datentupel genannt. Die Daten in einem Datentupel lassen sich in drei Gruppen einteilen: die Faktoren, die Bilddaten und die zu den Bilddaten geh\"{o}rige Repr\"{a}sentation des Nutzsignals. Mit der Repr\"{a}sentation ist die jeweilige Quantifizierung des \Nutzsigs aus dem Bildstrom bezeichnet, wie sie in Kapitel \ref{sec:BV_Module} durchgef\"{u}hrt wurde. Ein Projekt enth\"{a}lt zu Beginn nicht alle notwendigen Daten, sondern es wird durch die Methoden eine Reihe von Daten erzeugt, welche schlie{\ss}lich die Klassenzuweisung erm\"{o}glicht. Alle Daten lassen sich zu jedem Zeitpunkt pr\"{a}sentieren. Die Faktoren bestehen aus den St\"{o}r"~ und Planfaktoren sowie gesch\"{a}tzten Klassenzugeh\"{o}rigkeiten. Die Bilddaten sind der akquirierte Bildstrom und die Repr\"{a}sentation deren Quantifizierung.

Bei Projekterstellung werden sowohl die Bilddaten als auch die zugeh\"{o}rigen bekannten St\"{o}rfaktoren eingelesen. Die St\"{o}rfaktoren werden in einem solchen Zusammenhang auch  als Metadaten bezeichnet und sind \"{u}blicherweise entweder im Dateinamen und \bzw oder in einer zugeh\"{o}rigen Metadatei gespeichert.  Wie in \abb \ref{fig:GUI_Uebersicht_4} dargestellt, verlinkt das dargestellte Projekt Bilddaten und enth\"{a}lt Faktoren sowie Repr\"{a}sentationen der Bilddaten (Zeitreihen und Merkmale). Auf das Projekt lassen sich Methoden anwenden. Die zur Verf\"{u}gung stehenden Methoden sind entweder bereits in Gait-CAD (vgl. Abschnitt \ref{subsec:Gait-CAD}) integrierte Funktionalit\"{a}ten oder die Implementierung der Module aus Kapitel \ref{sec:BV_Module}. Die Methoden lassen sich in drei Gruppen einteilen:
\begin{enumerate}
  \item Methoden zum Prozessieren
  \item Methoden zum Klassifizieren
  \item Methoden zum Pr\"{a}sentieren.
\end{enumerate}
 Dabei  bezeichnet der Begriff Prozessieren alle Berechnungen zur Ermittlung der Merkmale, die das Nutzsignal repr\"{a}sentieren, somit also die Ermittlung von $\mathbf{X}$ aus dem Bildstrom $\mathbf{BS}$ (vgl. Abschnitt \ref{sec:Parameter_mathematisch} bzw. \abb \ref{fig:Signalmodell}). Die Methoden m\"{u}ssen dabei nicht zwingend auf die gesamten Daten angewandt werden, sondern es besteht die M\"{o}glichkeit, die Daten anhand der Faktoren zu filtern.

Nach dem Einlesen steht somit ein Projekt zur Verf\"{u}gung, welches bereits zum Erkunden des Datensatzes verwendet werden kann. So lassen sich (je nach aufgezeichneten Metadaten) beispielsweise alle Bilddaten eines bestimmten Aufnahmedatums oder Laboranten filtern und darstellen. Bei der Repr\"{a}sentation der Bilddaten handelt es sich immer um eine Quantifizierung. Die Quantifizierung ist Ergebnis der Bildverarbeitung und kann mit Hilfe der Benutzeroberfl\"{a}che auf alle Daten angewandt werden. Ebenso ist es m\"{o}glich, die Daten innerhalb der Oberfl\"{a}che mittels der Faktoren zu filtern. Nach dem Prozessieren steht die Quantifizierung in Form von \FischInfMerken zur Verf\"{u}gung. Hierbei werden auch weitere Faktoren erzeugt. Beispielsweise wird gespeichert, ob die Bildverarbeitung erfolgreich war oder nicht. Sollte die Bildverarbeitung \zB aufgrund von Fehlern oder Inhomogenit\"{a}ten in den Bilddaten gescheitert sein, so k\"{o}nnen solche Datentupel von der sp\"{a}teren Verarbeitung ausgeschlossen werden. Zwischenschritte des Prozessierens, wie etwa Zeitreihen oder Merkmalsbilder, k\"{o}nnen gespeichert und pr\"{a}sentiert werden. Bei Bildern als Ergebnis der Berechnung wird wiederum nur ein Link zu den Bildern im Projekt gespeichert und die Datei physisch auf dem Datenspeicher abgelegt, w\"{a}hrend die Werte der berechneten \FischInfMerke und \FischInfZeit vollst\"{a}ndig im Projekt abgelegt werden. Da es sich hierbei pro Bild, Merkmal und Abtastzeitpunkt im Normalfall um nur einen Wert handelt, bleibt die Gr\"{o}{\ss}e eines Projektes auch bei einer gro{\ss}en Anzahl an Datentupeln \"{u}berschaubar\footnote{Typische Projektgr\"{o}{\ss}e: \ca 9\,KB pro Datentupel}. Jeder Zwischenschritt l\"{a}sst sich pr\"{a}sentieren und mittels der Faktoren filtern \bzw gegen die Faktoren plotten (beispielsweise in Histogrammen). Das Filtern un Plotten ist sowohl f\"{u}r Merkmalsbilder als auch f\"{u}r Zeitreihen und Merkmale m\"{o}glich.

Da zur Erstellung eines Klassifikators ein Lerndatensatz ben\"{o}tigt wird, muss f\"{u}r dessen Erstellung zumindest f\"{u}r einen Teil der Datentupel die Klassenzugeh\"{o}rigkeit der Planfaktoren bekannt sein. Da die Klassenzuordnung von Experten anhand der Bilddaten erfolgt, wurde eine Funktionalit\"{a}t implementiert, die es erm\"{o}glicht, die zugeh\"{o}rigen Bilder oder Bildsequenzen von Datentupeln nacheinander anzeigen zu lassen und die Klassenzuordnung von Hand vorzunehmen. Der Vorgang wird als Labeln bezeichnet und kann in zuf\"{a}lliger Reihenfolge oder der Reihe nach erfolgen. Es besteht zus\"{a}tzlich die M\"{o}glichkeit, die Klassenzugeh\"{o}rigkeit aus einer Datei einzulesen, sollte die Information \"{u}ber die Daten von einem Experten \zB bereits in einer Excel-Datei abgelegt worden sein. Die Information wird als weiterer Faktor den Bildern zugeordnet und abgespeichert.  Anhand der bereits quantifizierten Bilddaten und der neuen Information durch das Lablen ist es nun m\"{o}glich, einen Klassifikator zu erstellen, der auf alle unbekannten Daten angewandt werden kann. Ist ein solcher Klassifikator \zB anhand von Kontrollen erstellt worden, l\"{a}sst sich der Klassifikator exportieren und in andere Projekte importieren. Ein solcher Export erm\"{o}glicht, unter der Voraussetzung konsistenter Datens\"{a}tze, die Klassifikation weiterer Projekte ohne die Notwendigkeit eines erneuten Labelns. Nach der Anwendung des Klassifikators sind alle Schritte abgeschlossen und k\"{o}nnen pr\"{a}sentiert werden. Es wurde zudem eine Funktionalit\"{a}t implementiert, die es gestattet, ein PDF"~Dokument zu generieren, das alle wichtigen Daten in Form eines Reports \"{u}bersichtlich zusammenstellt.

%

\section{Skalierung}

\begin{figure}[!htb]
        \centering
        \includegraphics[page=2]{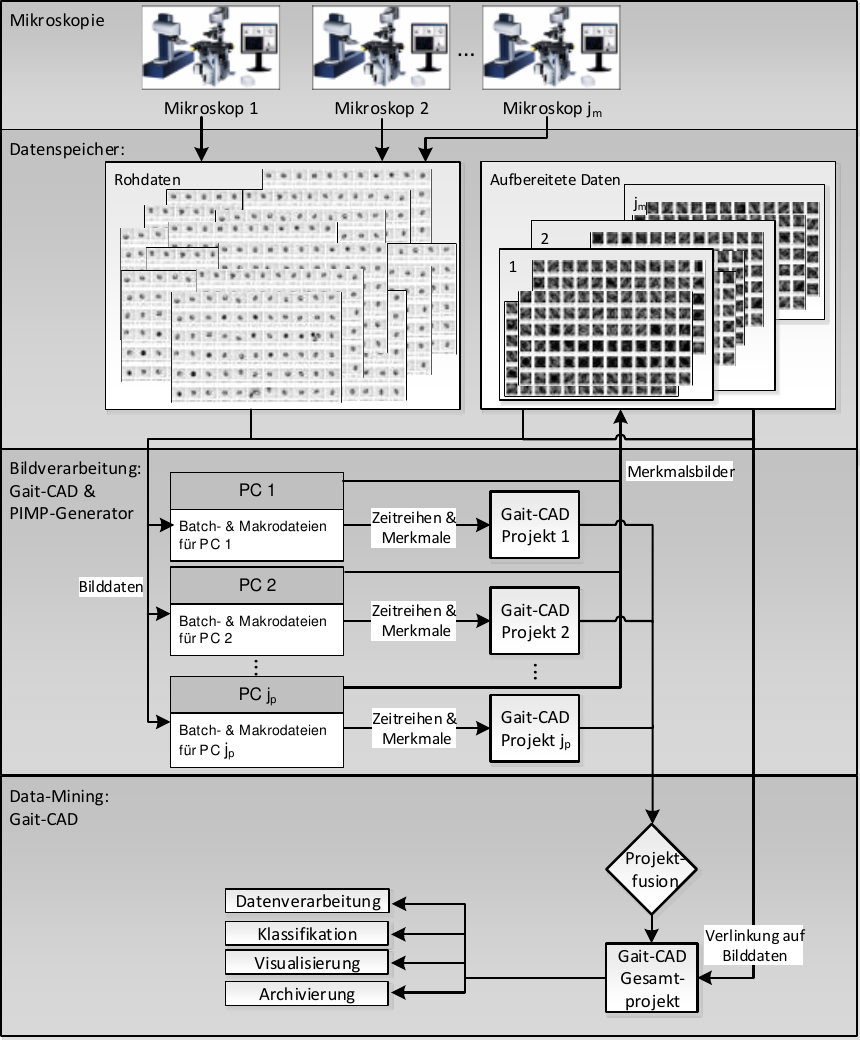}
        \caption[Skalierung f\"{u}r die Bildverarbeitung im Hochdurchsatzverfahren]{Skalierung f\"{u}r die Bildverarbeitung im Hochdurchsatzverfahren}
        \label{fig:GUI_Uebersicht_2}

\end{figure}

Die gro{\ss}e Anzahl $N$ an Einzelversuchen, welche in hohem Tempo in automatisierten Mikroskopen akquiriert werden, macht eine ebenso schnelle Datenverarbeitung notwendig. Die angewandte Methode zur Skalierung setzt ein schnelles lokales Datennetzwerk (LAN) voraus, ohne welches das einfache \"{U}bertragen der Daten mehr Zeit in Anspruch nimmt als sich durch verteiltes Rechnen einsparen l\"{a}sst. F\"{u}r die Daten\"{u}bertragung stand bei Durchf\"{u}hrung der vorliegenden Arbeit eine theoretische \"{U}bertragungsrate von 1000 Megabit/s (Gigabit-Ethernet) zur Verf\"{u}gung. Das Prinzip der hier angewandten verteilten Berechnung ist in \abb \ref{fig:GUI_Uebersicht_2} dargestellt.

Eine Anzahl an Mikroskopen zeichnet parallel die Bildstr\"{o}me auf und kopiert die Rohdaten \inkl der Metadaten auf einen zentralen Datenspeicher. Auf den Datenspeicher greifen einzelne Computer (Clients in \abb \ref{fig:GUI_Uebersicht_2} als PC 1 bis PC$j_p$ bezeichnet) zu. Auf den Clients l\"{a}uft jeweils eine Instanz der Bildverarbeitung ab. In der vorliegenden Arbeit ist die Instanz auf den Clienten entweder das Werkzeug Gait-CAD oder die Bildverarbeitungsoberfl\"{a}che \emph{PIMP} (vgl. Abschnitt \ref{sec:Werkzeuge}). Da beim Einlesen, Durchsehen und Berechnen der Bilddaten mehr als einmal auf die Bilddaten zugegriffen werden muss, hat es sich als sinnvoll erwiesen, auf jedem Client eine lokale Kopie der zu prozessierenden Daten abzuspeichern und nach erfolgreicher Berechnung lediglich die \"{A}nderungen zur\"{u}ck auf den Server zu speichern. Hierf\"{u}r wurde ein Programm der Firma Microsoft namens \emph{Synctoy ver. 2.1} eingesetzt, welches sich automatisiert mit Skripten steuern l\"{a}sst und f\"{u}r die Synchronisation \"{u}ber lokale Netzwerkstrukturen spezialisiert ist. Jeder Client erstellt in einer eigenen Instanz, wie im vorangegangenen Kapitel beschrieben, eine oder mehrere Projektdateien. In der Praxis hat es sich bew\"{a}hrt, je Mikroskopdatensatz eine Projektdatei anzulegen. Die aufbereiteten Daten (Merkmalsbilder \etc) werden von den Clients auf den Datenspeicher zur\"{u}ckgeschrieben und die Projektdateien werden nacheinander fusioniert. Wichtig ist hierbei, dass die Referenzierung auf die Daten nicht absolut, sondern relativ erfolgt und auf den Clients die gleiche Verzeichnisstruktur beibehalten wird     wie auf dem Datenspeicher. Ein anderes Vorgehen f\"{u}hrt zu fehlerhaften Referenzierungen der Bilddaten nach der Fusion. Nach Abschluss der Berechnung, Synchronisation und Fusion kann das finale Gesamtprojekt weiterverarbeitet werden. \"{U}blicherweise wird ein Klassifikator geladen oder erstellt, die Klassenzuweisung wird durchgef\"{u}hrt und die Ergebnisse werden visualisiert.

Um sicherzustellen, dass alle Clients die identischen Arbeitsschritte selbstst\"{a}ndig durchf\"{u}hren, wurde eine Kombination aus Makro- und Batchdateien erstellt.
Eine Makrodatei enth\"{a}lt eine feste Abfolge von Befehlen, welche an einem (Teil)Projekt durchgef\"{u}hrt werden. Eine Batchdatei ist in der Lage, eine definierte Anzahl an Projekten zu \"{o}ffnen und Makrodateien darauf anzuwenden. Die Makrodateien enthalten f\"{u}r jeden Mikroskopdatensatz die gleiche Vorgehensweise bei der Berechnung der Repr\"{a}sentationen f\"{u}r die Bilddaten. Der einzige Unterschied von Client zu Client sind die zugeordneten Bilddaten je Client. Jeder Client enth\"{a}lt somit f\"{u}r jeden Mikroskopdatensatz einen Satz Makrodateien, welche nacheinander abgearbeitet werden.
Die Aufspaltung in Makro- und Batchdateien erm\"{o}glicht eine erleichterte Fehlersuche. Es wird w\"{a}hrend der Ausf\"{u}hrung jeweils eine Ereignisprotokolldatei (Logdatei) erstellt, welche f\"{u}r jeden Makro- und jeden Batchbefehl einen Eintrag erh\"{a}lt, ob die Ausf\"{u}hrung erfolgreich war oder nicht.
Durch den vollst\"{a}ndig automatisierten Ablauf des \og Prozesses muss der Anwender in den Batchdateien nun lediglich die Speicherorte der Bilddateien anpassen. Die Erstellung der Projekte und des Gesamtprojektes erfolgt dann automatisch durch die Abfolge der beschriebenen Batch"~ und Makrodateien. Das Erstellen der Batchdateien ist im Gegensatz zur Projekterstellung nicht vollst\"{a}ndig grafisch implementiert, da in den Dateien lediglich die Referenzierungen zu den Bilddaten angepasst werden m\"{u}ssen und ein solch geringer Aufwand an Anpassung den zus\"{a}tzlichen Aufwand einer grafischen Umsetzung nicht rechtfertigt. In der Implementierung wurde das parallele Rechnen auf mehreren Kernen eines modernen Prozessors ausgenutzt, welches, unter Einhaltung diverser Restriktionen, direkt von MATLAB mittels des MATLAB-Befehls \emph{parfor} zur Verf\"{u}gung steht.
Zur Archivierung ist es ausreichend, die einzelnen Projektdateien und das Gesamtprojekt \inkl aller verwendeter Batch"~ und Makrodateien auf den Datenspeicher zu kopieren.

\section{Werkzeuge}\label{sec:Werkzeuge}
\subsection {Programmpaket Gait-CAD}\label{subsec:Gait-CAD}
Am Forschungszentrum Karlsruhe werden seit 1998 Algorithmen zur Klassifikation entwickelt und aus der Toolbox KAFKA (\textbf{Ka}rlsruher \textbf{F}uzzy Modellbildungs-, \textbf{K}lassifikations- und datengest\"{u}tzte \textbf{A}nalyse-Toolbox) entstand ab dem Jahr 2001  die auf MATLAB basierende Toolbox Gait-CAD (Gait: englisch f\"{u}r Gang und CAD: \textbf{C}omputer \textbf{A}ided \textbf{D}iagnosis)~\cite{Mikut07GaitcadEnglish,Mikut01}. Ab dem Jahr 2006 wurde Gait-CAD mit einer grafischen Benutzeroberfl\"{a}che als freie Software im Internet unter der GNU General Public License (GNU GPL) ver\"{o}ffentlicht \cite{Burmeister08,Mikut07ATP,Mikut08Biosig,Stegmaier12}. Die Anwendungsgebiete der Toolbox Gait-CAD haben sich von anf\"{a}nglich ausschlie{\ss}licher Ganganalyse in weitere Felder erstreckt, unter anderem f\"{u}r die vorliegende Arbeit auf das Feld der Bildverarbeitung mit zu Beginn Einzelbildern und schlie{\ss}lich auch Bildsequenzen.

\begin{figure}[htbp]
    \footnotesize
    \centering
        \centering
        \includegraphics[page=16,width=\linewidth]{Bilder/Diss_Zusammenhang}
        \caption[Die in Gait-CAD implementierten Module des Modulkatalogs]{Die in Gait-CAD implementierten Module des Modulkatalogs sind hervorgehoben.}
        \label{fig:TOX-Implementierte_Module}
        \centering
        \includegraphics[width=.725\linewidth]{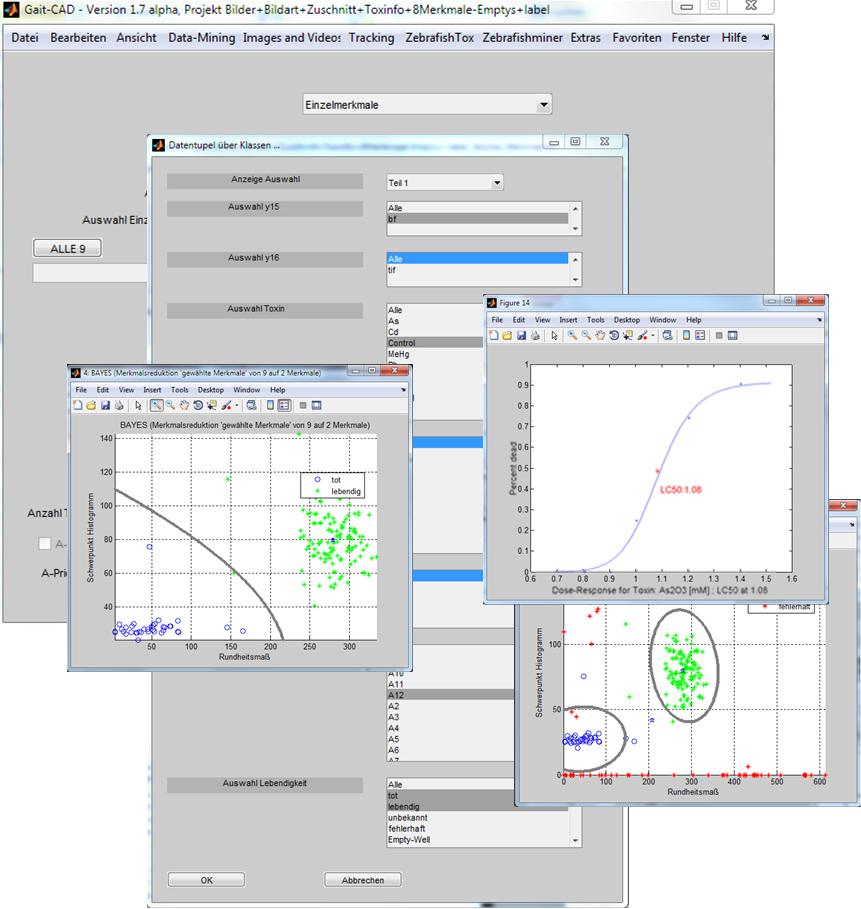}
        \caption[Gait-CAD: Die TOX-Erweiterung f\"{u}r den FET]{Gait-CAD: Die TOX-Erweiterung f\"{u}r den FET}
        \label{fig:Screenshot_Gait-CAD}

\end{figure}

Die Funktionalit\"{a}ten von Gait-CAD wurden im Rahmen der vorliegenden Arbeit um Bildverarbeitungsroutinen und f\"{u}r die Verarbeitung von Bildstr\"{o}men erweitert. Es stehen nun eine Reihe der vorgestellten Module in Form von Plugins zur Verf\"{u}gung, die sich auf die Datens\"{a}tze anwenden lassen. Die umgesetzten Module sind in \abb \ref{fig:TOX-Implementierte_Module} hervorgehobenen und wurden sowohl f\"{u}r Einzelbilder als auch f\"{u}r Bildsequenzen implementiert. Alle dargestellten Module sind entweder in Form von Plugins verf\"{u}gbar oder direkt von der GUI zu starten. Die bereits erw\"{a}hnte Funktionalit\"{a}t zum Labeln wurde ebenfalls erg\"{a}nzt. Die Auswertung des Anwendungsbeispiels \emph{Fisch Embryo Test} (FET), \vgl Kapitel \ref{chap:Anwendung}, wurde vollst\"{a}ndig in Gait-CAD integriert und steht "`auf Knopfdruck"' bis hin zur Erstellung eines automatischen Reports in Form eines PDF-Dokuments zur Verf\"{u}gung. Die \abb~\ref{fig:Screenshot_Gait-CAD} zeigt einige Screenshots der Funktionalit\"{a}ten. Im Vordergrund rechts ist eine EC$50$-Registrierung zu sehen. Im Vordergrund links ist ein Scatterplot \"{u}ber zwei Merkmale mit errechneter Diskriminanzfunktion dargestellt. Im Hintergrund ist eine beispielhafte Auswahl von Datentupeln sowie die Bedienoberfl\"{a}che von Gait-CAD zu sehen.

 F\"{u}r das Anwendungsbeispiel des \emph{Photomotor Response Screen} wurden Gait-CAD Projektdateien f\"{u}r alle Einzelversuche erstellt und durch Fusion in ein die gesamte Untersuchung umfassendes Projekt \"{u}berf\"{u}hrt. Innerhalb eines solchen Projektes ist es daraufhin m\"{o}glich, Klassifikationen, \zB bez\"{u}glich verschiedener Bewegungsmuster, durchzuf\"{u}hren. Die Bildverarbeitung erfolgt hier allerdings aus verschiedenen Gr\"{u}nden au{\ss}erhalb von Gait-CAD wie im folgenden Abschnitt beschrieben wird (\vgl \kap \ref{subsec:PIMP}).

\subsection {Grafische Oberfl\"{a}che PIMP} \label{subsec:PIMP}
\begin{figure}[htbp]
        \centering
        \includegraphics[width=\linewidth]{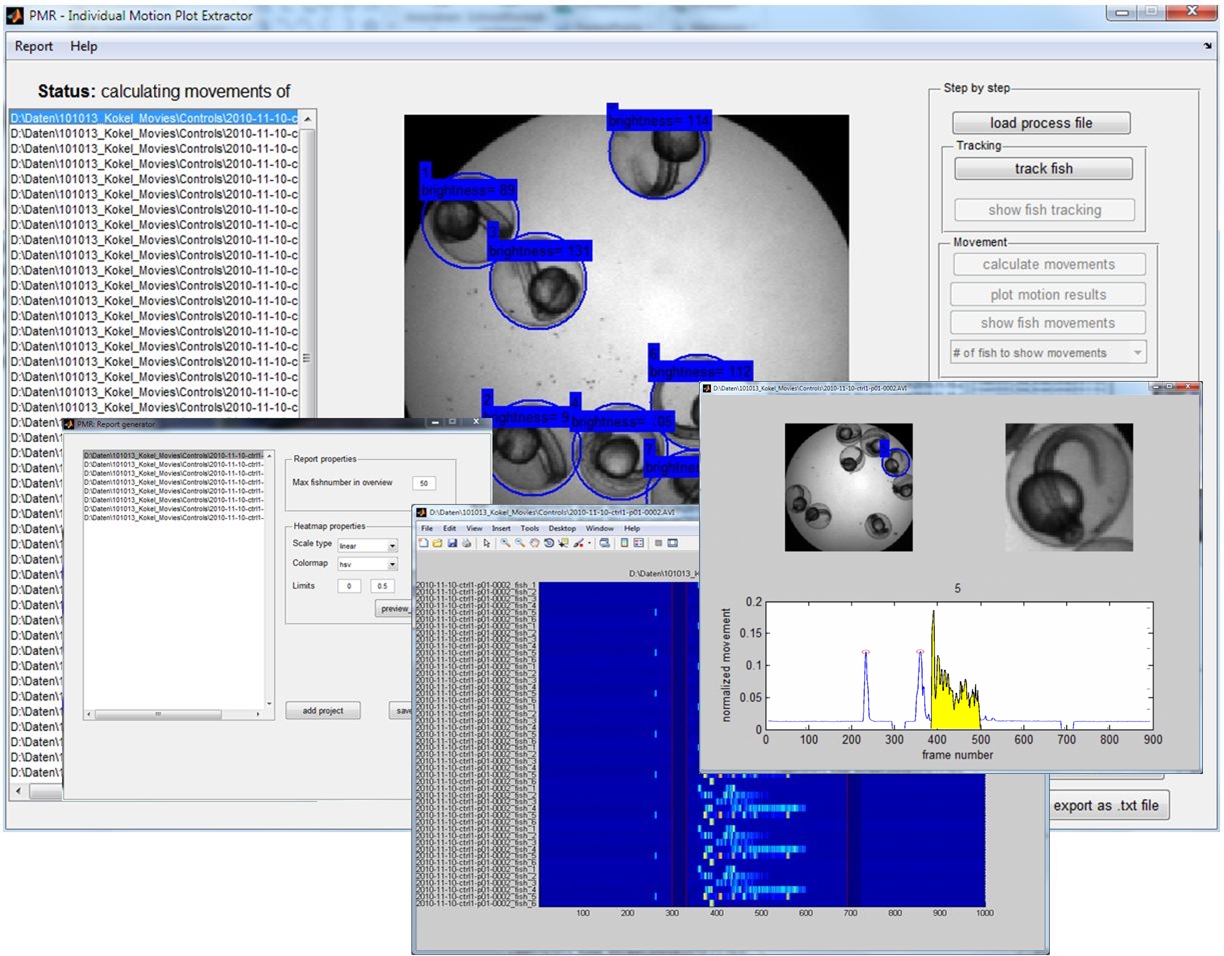}
        \caption[PIMP: Die grafische Oberfl\"{a}che zum Einlesen, Auswerten und Visualisieren]{PIMP: Die grafische Oberfl\"{a}che zum Einlesen, Auswerten und Visualisieren von PMR-Bildsequenzen\\}
        \label{fig:Screenshot_PIMP_1}

    \footnotesize
    \centering
        \centering
        \includegraphics[page=17,width=\linewidth]{Bilder/Diss_Zusammenhang}
        \caption[In \emph{PIMP} implementierte Module des Modulkatalogs]{In \emph{PIMP} implementierte Module des Modulkatalogs (Nicht implementierte Module sind ausgegraut.) }
        \label{fig:PIMP-Implementierte_Module}
\end{figure}

Die ebenfalls in MATLAB entworfene Auswertungseinheit \emph{PIMP} ist als grafische Be\-nutz\-er\-ober\-fl\"{a}che implementiert und enth\"{a}lt Routinen zum komfortablen Einlesen der Videosequenzen, der Auswahl und der manuellen Adaption von Auswerteparametern sowie dem Ausf\"{u}hren der Berechnung anhand der gew\"{a}hlten Parameter. Die Bildsequenzen lassen sich sowohl einzeln als auch in Gruppen einlesen und prozessieren. Vor der Durchf\"{u}hrung der Berechnung wird automatisch eine Vorschau erstellt, bei der das Ergebnis der Segmentierung direkt ersichtlich ist und anhand derer Parameter visuell, auch ohne Expertenwissen, angepasst werden k\"{o}nnen. Abbildung \ref{fig:Screenshot_PIMP_1} zeigt einige Screenshots der Oberfl\"{a}che. Im Vordergrund ist ein Einzelversuch mit den Ergebnissen des Trackings und der Bewegungsklassifikation (\vgl \kap \ref{subsec:Tracking}) dargestellt. Dahinter ist eine \"{U}bersicht mit \"{u}ber~50 solcher Einzelversuche in Form einer Heatmap dargestellt. Darauf folgt ein Screenshot des Report-Generators sowie schlie{\ss}lich ein Screenshot des Hauptteils der Oberfl\"{a}che, in welchem sich die Daten der Versuche ausw\"{a}hlen und alle Funktionen starten lassen.

Eine neue Oberfl\"{a}che wurde entworfen, da sie au{\ss}erhalb des Gesamtpakets Gait-CAD eine Reihe von Vorteilen bietet. Die Oberfl\"{a}che ist \"{u}bersichtlich und beschr\"{a}nkt sich auf die f\"{u}r den \emph{PMR} notwendigen Parameter. \emph{PIMP} wurde so konzipiert, dass es auch ohne eine vollst\"{a}ndige MATLAB-Installation (mit den entsprechenden MATLAB-Runtime-Paketen) lauff\"{a}hig ist. Zus\"{a}tzlich dazu ist es platt\-form\-un\-ab\-h\"{a}ngig und sowohl auf WINDOWS wie auch auf MAC/OS und UNIX Betriebssystemen lauff\"{a}hig. Das erm\"{o}glicht die Ausf\"{u}hrung auf beliebig vielen Computern, welche lokal, \zB direkt neben den aufzeichnenden Mikroskopen, platziert sind und spart das "`doppelte"' Kopieren wie bei der auf Gait-CAD basierten L\"{o}sung, bei welcher vom Mikroskop auf den Datenspeicher, von dem Datenspeicher auf den Client und schlie{\ss}lich wieder zur\"{u}ck kopiert werden muss. Lediglich die fertigen Daten werden zur Archivierung auf den Datenspeicher einmalig abgelegt. Damit wird die neue Softwarel\"{o}sung auch attraktiv f\"{u}r Labore, die weder \"{u}ber ein schnelles lokales Netzwerk noch \"{u}ber gro{\ss}e st\"{a}ndig erreichbare Speicherkapazit\"{a}ten verf\"{u}gen.
\begin{savenotes}
\begin{table}
\footnotesize
 \renewcommand{\arraystretch}{1.5}
\begin{tabular}
{p{.2\columnwidth}p{.4\columnwidth}p{.15\columnwidth}c}
\toprule
{\bf Programm}							&
{\bf Beschreibung}							&
{\bf Author}							&
{\bf Lizenz}\\
\midrule
\href{http://de.wikipedia.org/wiki/SyncToy}{Sync-Toy}	&freies Tool zur Synchronisierung von Dateien und Ordnern.	&Microsoft	&Freeware	&\\
\href{http://de.wikipedia.org/wiki/TortoiseSVN}{TortoiseSVN}	&Windows-Client für den Versionsverwaltungs-Dienst Subversion.	&The Tortoise\-SVN team	&GPL	&\\
\href{http://en.wikipedia.org/wiki/CombineZ}{CombineZ}	&Auf die Kombination mehrerer Fokusaufnahmen spezialisierte, batchfähige Software	&Alan Hadley	&GPL	&\\
\href{http://rsb.info.nih.gov/ij/}{ImageJ}	&Bildbearbeitungs- und Bildverarbeitungsprogramm	&Wayne\tabularnewline Rasband	&Public Domain	&\\
\href{http://www.irfanview.de}{IrfanView}	&Bildbetrachter mit umfangreichen Funktionen	&Irfan \^Skiljan	&Freeware	&\\
\href{http://www.mathworks.de/matlabcentral/fileexchange/25500-peakfinder}{PeakFinder\footnote{www.mathworks.de/matlabcentral/fileexchange/25500-peakfinder}}	&Implementierung zur Minima- und Maximadetektion innerhalb verrauschter Daten	&Nate Yoder	&BSD	&\\
\href{http://www.mathworks.de/matlabcentral/fileexchange/26978-hough-transform-for-circles}{Hough transform for circles\footnote{www.mathworks.de/matlabcentral/fileexchange/26978-hough-transform-for-circles}}	&Implementierung des Hough-Algorithmus zur Kreisdetektion	&David Young	&BSD	&\\
\href{http://www.mathworks.de/matlabcentral/fileexchange/28590-template-matching-using-correlation-coefficients}{Template Matching using Correlation Coefficients\footnote{www.mathworks.de/matlabcentral/fileexchange/28590-template-matching-using-correlation-coefficients}}	&Implementierung des Template-Matchings mittels Korrelations	&Yue Wu	&BSD	&\\
\href{http://www.mathworks.de/products/image/}{MATLAB Image Processing Toolbox}	&Umfangreiche Sammlung von Bildverarbeitungs- und Bildanalysefunktionen	&Mathworks	&Proprietär	&\\
\href{http://www.ni.com/analysis/lvaddon_vision.htm}{LabVIEW add-on: Vision/Image Processing}	&Bildverarbeitungserweiterung für LabVIEW	&National\tabularnewline  Instruments	&proprietär	&\\

\bottomrule
\end{tabular}
\caption[\"{U}bersicht der Fremdimplementierungen]{\"{U}bersicht der Fremdimplementierungen und verschiedener Tools, welche in Kombination mit der in Kapitel \ref{chap:Implementierung} vorgestellten Software zum Einsatz kommen.}\label{tab:Fremdimplementierungen}
\end{table}
\end{savenotes}

Zudem wurde die Option implementiert, bereits prozessierte Daten erneut einzulesen und so einzelne Bewegungsmuster einer jeden Versuchseinheit zu untersuchen und gegen andere zu plotten. Die Option erm\"{o}glicht das schnelle Inspizieren auff\"{a}lliger Ergebnisse. Der zus\"{a}tzlich integrierte Reportgenerator  erm\"{o}glicht das Erstellen eines ausf\"{u}hrlichen Berichtes mit einer Zusammenfassung sowie Zwischenergebnissen im PDF- oder wahlweise im HTML-Format. Die erzeugten Ergebnisse werden zudem f\"{u}r die weitere Verarbeitung und Klassifikation in Form von Gait-CAD Projekten abgespeichert. Die innerhalb von \emph{PIMP} umgesetzten Module des Modulkatalogs sind in \abb \ref{fig:PIMP-Implementierte_Module} hervorgehoben.  Tabelle \ref{tab:Fremdimplementierungen} bietet eine \"{U}bersicht verschiedener, bei der Umsetzung verwendeter Fremdimplementierungen sowie einige Tools, welche sich w\"{a}hrend der Arbeit bew\"{a}hrt haben. Beispielsweise das auf der Programmiersprache JAVA aufbauende \emph{ImageJ}, welches durch den offengelegten Quellcode leicht erweiterbar ist oder auch die au{\ss}erordentlich leistungsf\"{a}hige, wenn auch propriet\"{a}re, Bildverarbeitungs-Erweiterung in der Software MATLAB.

\emph{PIMP} wird im Rahmen der vorliegenden Arbeit zur Auswertung der Datens\"{a}tze in \kap \ref{sec:Anwendung_PMR} verwendet und ist dar\"{u}berhinaus derzeit als Auswerteeinheit der am KIT entwickelten Plattform f\"{u}r Hochdurchsatz-Untersuchungen am Zebrafisch im Einsatz\cite{marcato15}. Desweiteren befindet sich die Software derzeit im Labor von \emph{Randall T. Peterson}\footnote{\href{http://www.rtplab.org/}{http://www.rtplab.org/}} sowie im Labor von \emph{David Kokel}\footnote{\href{http://kokellab.com/}{http://kokellab.com/}} der Harvard Medical School und des Massachusetts General Hospital (Boston, USA) im Routineeinsatz.

\section{Bewertung}
Die vorgestellte Implementierung, die grafische Oberfl\"{a}che und das Skalierungskonzept lassen sich f\"{u}r die Auswertung beliebiger Daten von \htsen anwenden und falls notwendig adaptieren. W\"{a}hrend das Werkzeug PIMP f\"{u}r den Photomotor Response Screen spezialisiert ist, l\"{a}sst sich das Programmpaket Gait-CAD durch Plugins leicht auf neue Problemstellungen anpassen \bzw erweitern. Der grundlegende Ablauf vom Einlesen der Bilddaten bis hin zur Klassifikation bleibt hierbei identisch. Es muss f\"{u}r weitere Problemstellungen lediglich die Segmentierung ausgetauscht und anhand gelabelter Daten ein individueller Klassifikator entworfen werden. Somit steht nun, mit den vorgestellten Werkzeugen, eine zwar prototypische, jedoch vollst\"{a}ndige Implementierung f\"{u}r \htsen zur Verf\"{u}gung. Eine erste Anwendung ist in \cite{kokel13} zu finden. 
Im folgenden Kapitel werden die konzeptionellen und theoretischen Methoden an zwei praktischen Problemstellungen erprobt.

\chapter{Anwendungen}\label{chap:Anwendung}
\section {\"{U}bersicht}

Um die Leistungsf\"{a}higkeit des neuen Konzeptes nachzuweisen, wird es im Folgenden auf zwei Probleme, die sich mit \htsen l\"{o}sen lassen, angewandt. Dabei werden die Problemstellung und damit der zu untersuchende Bilddatensatz jeweils komplexer. Die vorgestellten Projekte sind:
\begin{itemize}
  \item eine toxikologische Untersuchung
  \begin{itemize}
\item  anhand von Einzelaufnahmen von vereinzelten Zebrab\"{a}rblingen (Abschnitt~\ref{subsec:FET_Koagulation}) und
  \item anhand von Bildsequenzen von vereinzelten Zebrab\"{a}rblingen (Abschnitt \ref{subsec:FET_Heartbeat}).
      \end{itemize}
  \item eine \hts anhand von Bildsequenzen von mehreren Zebrab\"{a}rblingen (Abschnitt \ref{sec:Anwendung_PMR}).
\end{itemize}

iAlle Projekte werden mittels des eingef\"{u}hrten Konzeptes entworfen. Es folgt die Erstellung einer Auswertungskette durch Auswahl geeigneter Module aus Kapitel \ref{sec:BV_Module}. Die Ergebnisse werden anschlie{\ss}end pr\"{a}sentiert und diskutiert. Weitere, nicht im Rahmen des Kapitels vorgestellte Anwendungen, auf die entweder direkt Einfluss genommen wurde oder die auf die Ergebnisse der vorliegenden Arbeit aufbauen, finden sich in \cite{marcato15,Pfriem11,pylatiuk14,Pylatiuk11,Sanchez2012}. Die Anwendungen wurden aus verschiedenen Gr\"{u}nden nicht in das Kapitel \ref{chap:Anwendung}  aufgenommen: \cite{Pfriem11,Pylatiuk11} haben ihren Schwerpunkt weniger in der Bildverarbeitung und -analyse als in der Automatisierungstechnik. In der Anwendung von \cite{pylatiuk14,Sanchez2012} werden einige Aspekte des hier vorgestellten Konzeptes umgesetzt und eine Bildverarbeitungsroutine zur Erkennung des Herzschlags und zur Extraktion der Herzfrequenz erarbeitet. Obwohl die Ergebnisse vielversprechend sind, war deren Umfang zum Zeitpunkt des Entstehens der vorliegenden Arbeit noch nicht ausreichend f\"{u}r eine Aufnahme in das folgende Kapitel.

\section{Fisch Embryo Test (FET)}\label{sec:Anwendung_FET}

Der FET stellt ein geeignetes Einsatzszenario f\"{u}r das erarbeitete Konzept dar, da alle Teilschritte ein hohes Potenzial zur Automatisierung bieten und in Teilbereichen wie \zB der Pr\"{a}paration bereits Automatisierungstechnik zur Verf\"{u}gung steht. Die bisher manuell erfolgende Klassifikation von \sog \emph{Endpunkten,} wie sie in den OECD-Richtlinien definiert sind \cite{OECD06,OECD06a}, zu automatisieren, ist der offensichtlich n\"{a}chste Schritt zur Vervollst\"{a}ndigung der Auswertungskette. Im \og Standard sind mehrere dieser \emph{Endpunkte} definiert. Hierbei handelt es sich jeweils um dichotome Pr\"{u}fungen, \dhe jede Larve erreicht einen der definierten Endpunkte oder nicht. Die Endpunkte und deren Erscheinungsformen sind:
\begin{itemize}
\item Koagulation

 Die Koagulation ist die einsetzende Gerinnung der Proteine im Ei. Bei unbefruchteten oder sich nicht entwickelnden Organismen tritt Koagulation ein. Sie ist in den Mikroskopaufnahmen durch einen dunklen Fleck im Inneren des Eies erkennbar. Koaguliertheit wird klassifiziert, wenn keine Larve sondern der beschriebene Fleck  im Bild vorhanden ist.
\item Herzschlag

Der ab \ca 24\,hpf einsetzende Herzschlag ist in der gesunden Larve, je nach Lage des Herzens, unterschiedlich deutlich sichtbar. Auch der Blutfluss in der Larve ist durch die Transparenz des Gewebes \zT erkennbar und ein indirektes Ma{\ss} f\"{u}r den Herzschlag. Der Herzschlag wird als vorhanden klassifiziert, wenn die Kontraktion des Herzens oder das Flie{\ss}en der Blutk\"{o}rperchen in der Bildsequenz erkennbar ist.

\item Spontanbewegung

Die ebenfalls nach \ca 24\,hpf eintretenden spontanen Bewegungen der Larve innerhalb des Chorions sind in Bildsequenzen durch ein Verwinden der Larve im Ei erkennbar. Die Spontanbewegung wird als vorhanden klassifiziert, wenn die Larve ihre Position im Bild deutlich ver\"{a}ndert, \dhe eine klar andere Position eingenommen hat, wie z.B. eine 180° Drehung.

\item Abl\"{o}sung des Schwanzes vom Dottersack

Nach \ca 24\,hpf l\"{o}st sich w\"{a}hrend der Entwicklung der Larve der Schwanz vom Dottersack. Das Abl\"{o}sen des Schwanzes wird als vorhanden klassifiziert, wenn ein Freiraum zwischen dem Schwanz und dem Dottersack in den Mikroskopaufnahmen erkennbar ist, \dhe der Schwanz der Kr\"{u}mmung des Dottersacks nicht unmittelbar folgt.
\end{itemize}

\begin{minipage}{\linewidth}
\begin{itemize}
\item Somiten

Ab \ca 16\,hpf werden im Schwanz und R\"{u}cken der Zebrab\"{a}rblingslarve \sog Somiten sichtbar. Somiten stellen sich auf den Bildern als regelm\"{a}{\ss}ige Linien im Schwanz dar und k\"{o}nnen je nach Position des Schwanzes \uU verdeckt oder auch unscharf abgebildet sein. Somiten werden als vorhanden klassifiziert, wenn die Struktur im R\"{u}cken der Larve erkennbar ist.
\end{itemize}
\end{minipage}

F\"{u}r jeden der Endpunkte soll gem\"{a}{\ss} des Standards eine ebenfalls dichotome Klassenzuweisung erfolgen. Nach Einsch\"{a}tzung von Experten ist jedoch sowohl die Abl\"{o}sung des Schwanzes vom Dottersack als auch die Ausbildung von Somiten im R\"{u}cken selbst vom Menschen nicht immer scharf einer der Klassen $\{$\emph{ja, nein}$\}$ zuzuordnen und eignet sich somit weniger f\"{u}r eine automatische Klassifikation. Es wird im Folgenden daher nur auf die ersten drei der genannten Endpunkte eingegangen.

\subsection{Auswirkungen der Orientierung der Larve im Ei}\label{subsec:Anwendung_Datensatzanalyse}
Die Larven k\"{o}nnen sich ab einem Alter von ca. 48\,hpf innerhalb des Eies frei bewegen. Damit ist die Pr\"{a}senz der biologisch relevanten Information in den Daten lageabh\"{a}ngig. Bei einer entwickelten Fischlarve sind je nach Position mehr oder weniger Organe sichtbar. Der Dottersack, mindestens ein Auge und der R\"{u}cken sind fast immer zu identifizieren. Feinere Strukturen wie das Herz, die Leber usw. sind nur schwer auszumachen. Schl\"{a}gt die Entwicklung komplett fehl, zersetzt sich das Ei und zeigt Koagulation. Um in der Lage zu sein, eine \hts erfolgreich durchzuf\"{u}hren, muss f\"{u}r einen akquirierten Datensatz untersucht werden, inwieweit sich der Datensatz zur automatisierten Klassifikation der \og Endpunkte eignet. Das Ziel der folgenden Analyse ist somit, die akquirierten Daten bez\"{u}glich den spezifischen Anforderungen der genannten Endpunkte zu untersuchen.

Die theoretisch m\"{o}gliche Anzahl von Orientierungen, die eine sich frei bewegende Larve innerhalb des Chorions annehmen kann, ist durch die kontinuierliche Positionierung unendlich gro{\ss}. Es l\"{a}sst sich jedoch eine Anzahl von typischen F\"{a}llen mit geometrischen Eigenschaften in Gruppen zusammenfassen, welche potenziell f\"{u}r die spezifische Untersuchung bez\"{u}glich der vorgestellten Endpunkte herangezogen werden k\"{o}nnen. Die Gruppen werden im Folgenden herausgearbeitet.

Ein typischer Datensatz wird bei \ca 48\,hpf akquiriert, da nach OECD-Norm sich hier alle der \og Endpunkte pr\"{u}fen lassen \cite{OECD06,OECD06a}. Bei einem Entwicklungszeitpunkt von 48\,hpf sind verschiedene Bereiche und Details in der Larve bereits unterscheidbar. Je nach Gr\"{o}{\ss}e sind die Details jedoch zum Teil nur schwer zu erkennen oder werden bei ung\"{u}nstiger Lage der Larven von verschiedenen K\"{o}rperteilen verdeckt. Die wichtigsten Bereiche, die sich in nahezu jedem Bild abgrenzen lassen, sind nach aufsteigendem Schwierigkeitsgrad der Auffindbarkeit geordnet: das Auge, der Dottersack, der R\"{u}cken und der Schwanz.

Die Vorgehensweise bei der Einteilung in Klassen orientiert sich vornehmlich daran, welche Bereiche der Larven gut zu erkennen sind. Hierf\"{u}r werden zur leichteren Klassifikation mehrere Merkmale zur Positionsbestimmung definiert. Markante Bereiche liegen \"{u}blicherweise im Vordergrund und werden nicht durch mehr oder minder transparente andere K\"{o}rperteile verdeckt. Das erste Merkmal ist somit das markanteste Merkmal am Kopf: die Augen, welche einen sehr starken Kontrast aufweisen und zumeist als dunkle Fl\"{a}chen oder "`Scheiben"' erscheinen. Das zweite Merkmal ist der Dottersack bzw. wie viel Prozent des Dottersacks ohne \"{U}berdeckung, \dhe im Vordergrund des Bildes, erscheinen.
Das dritte Merkmal zur Klasseneinteilung stellt der R\"{u}cken dar. Gemessen wird der Abstand der Konturlinie der Larve, also das Ende des R\"{u}ckens, bis zum Dottersack. Der Wert gibt an, wie viel Prozent des Durchmessers sichtbar ist, bezogen auf den maximalen Durchmesser einer "`normal"' entwickelten Larve bei einem Alter von 48\,hpf.
Das letzte Merkmal ist der Schwanz. Da keiner der ausgew\"{a}hlten Endpunkte Details innerhalb des Schwanzes auswertet, ist das Auftauchen des Schwanzes im Vordergrund eher hinderlich, weil es dann zumeist andere K\"{o}rperteile verdeckt. Das Merkmal misst die Menge an vom Schwanz verdeckten anderen K\"{o}rperteilen im Verh\"{a}ltnis zum Gesamtvolumen.

F\"{u}r die Einordnung der Lage der Larven werden in der vorliegenden Arbeit vier Klassen definiert.  Die Einteilung der Klassen erfolgt anhand der oben definierten Merkmale durch Anwendung folgender Regeln, welche aus Erfahrungswerten abgeleitet sind:


\begin{description}
  \item[\textbf{seitlich}] Die weitaus gr\"{o}{\ss}te Anzahl der Larven liegt auf der Seite. Bezogen auf die eingef\"{u}hrten vier Merkmale liegt eine Larve seitlich, wenn ein Auge deutlich im Vordergrund zu sehen ist, w\"{a}hrend das andere nicht im Vordergrund ist und sich im optimalen Fall sogar deckungsgleich auf der "`R\"{u}ckseite"' der Larve befindet. Beim Merkmal der Augen wird das Auge, welches einen gr\"{o}{\ss}eren Anteil an Pixeln im Vordergrund aufweist, als Auge~1 und das andere als Auge~2 bezeichnet. Somit ist f\"{u}r das Merkmal \emph{Auge} bei der Klasse \emph{Seitlich} Auge 1 zu mehr als 90\% im Vordergrund festgelegt, w\"{a}hrend von Auge 2 nicht mehr als 5\% im Vordergrund erkennbar sein darf.
      Der Dottersack ist immer im Bild sichtbar und kann lediglich durch den Schwanz oder den R\"{u}cken teilweise \"{u}berdeckt werden. Da bei seitlicher Lage der R\"{u}cken zumeist \"{u}ber der Larve liegt, sollte der Dottersack nicht oder nur unwesentlich verdeckt sein. Als zus\"{a}tzliches Kriterium ist somit eine  Sichtbarkeit des Dottersacks von 80\% festgelegt.
      Eine weitere Voraussetzung f\"{u}r die seitliche Lage ist, dass der R\"{u}cken zu mehr als 50\% \"{u}ber dem Dottersack sichtbar ist und sich weder vor noch hinter anderen K\"{o}rperteilen befindet.
      Da bei einer \"{U}berdeckung der Organe der Larve durch deren Schwanz solche wichtigen Bereiche im vorderen Teil der Larve nicht erkennbar werden, wird gefordert, dass die \"{U}berdeckung des Schwanzes 5\% des Volumens der Larve nicht \"{u}berschreiten darf.

\item[\textbf{von unten}] Die Klasse beinhaltet alle Larven, bei denen ein Teil des R\"{u}ckens durch den Dottersack verdeckt ist und beide Augen zumindest teilweise im Vordergrund erkennbar sind.
      Bezogen auf die Klasse wird gefordert, dass Auge 1 zu mehr als 90\% sichtbar und Auge~2 zu mehr als 60\% sichtbar ist. Da der Dottersack den R\"{u}cken verdeckt, ist er fast vollst\"{a}ndig im Vordergrund und es werden mehr als 80\% im Vordergrund f\"{u}r die Klasseneinteilung gefordert. Der R\"{u}cken muss verdeckt sein und befindet sich somit zu h\"{o}chstens 20\% im Vordergrund. Eine \"{U}berdeckung durch den Schwanz soll m\"{o}glichst ausgeschlossen werden. Die \"{U}berdeckung des Schwanzes soll daher 5\% des Volumens der Larve nicht \"{u}berschreiten.

  \item[\textbf{von oben}] Hierzu werden alle Larven gez\"{a}hlt, bei denen der Dottersack durch den R\"{u}cken verdeckt bzw. in zwei Bereiche geteilt wird und beide Augen zumindest teilweise im Vordergrund sichtbar sind. Die Augen im Vordergrund sind identisch zur Lage \emph{von unten} definiert mit mehr als 90\% f\"{u}r Auge 1 und mindestens 60\% f\"{u}r Auge 2. Da jedoch der Dottersack in der Lage \emph{von oben} durch den R\"{u}cken verdeckt ist, sind die Voraussetzungen gegen\"{u}ber \emph{von unten} umgekehrt. Der Dottersack muss zu maximal 60\% und der R\"{u}cken zu mindestens 50\% im Vordergrund sichtbar sein. Die \"{U}berdeckung des Schwanzes soll daher 5\% des Volumens der Larve nicht \"{u}berschreiten.

  \item[\textbf{hinten}] Zur Klasse \emph{hinten} werden die Larven gez\"{a}hlt, bei denen kein Auge im Vordergrund sichtbar ist, da der Kopf vom Dottersack verdeckt wird. Auge 1 soll f\"{u}r die Klasse \emph{hinten} zu maximal 10\% und Auge 2 zu maximal 5\% sichtbar sein. Der Dottersack befindet sich im Vordergrund zu mehr als 50\%. Eine \"{U}berdeckung des R\"{u}ckens oder des Schwanzes ist f\"{u}r die Klasse unerheblich und es werden alle Werte akzeptiert, \dhe 0\% - 100\%.

  \item[\textbf{undefiniert}] Larven, die sich keiner der \"{u}brigen vier Klassen zuordnen lassen, wurden als \emph{undefiniert} klassifiziert. Die klare Zuordnung scheitert in vielen F\"{a}llen dann, wenn die Larve schr\"{a}g oder verdreht im Ei liegt.

\end{description}
\begin{figure}[htb]\captionsetup[subfloat]{format=hang}
\centering
\subfloat[seitlich (h\"{a}ufigste Lage)]{
\includegraphics[width=.23\linewidth, height=.23\linewidth] {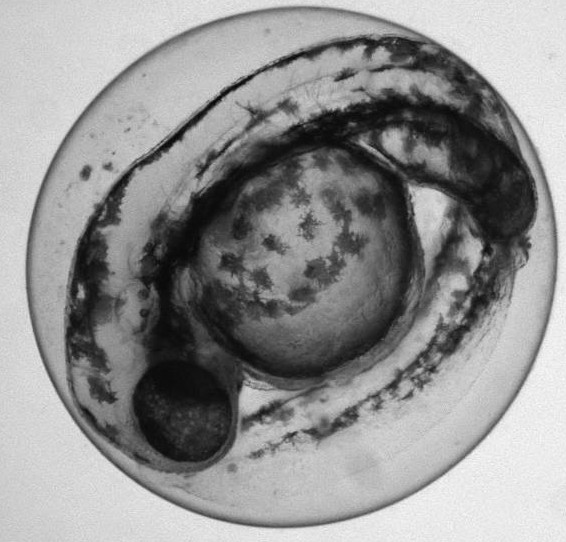}
}
\subfloat[von oben]{
\includegraphics[width=.23\linewidth, height=.23\linewidth] {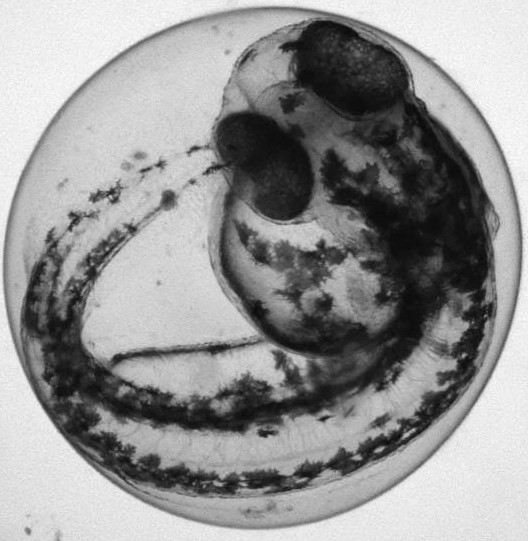}
\label{fig:Prozesskette_Mat_Meth_6}
}
\subfloat[von unten]{
\includegraphics[width=.23\linewidth, height=.23\linewidth] {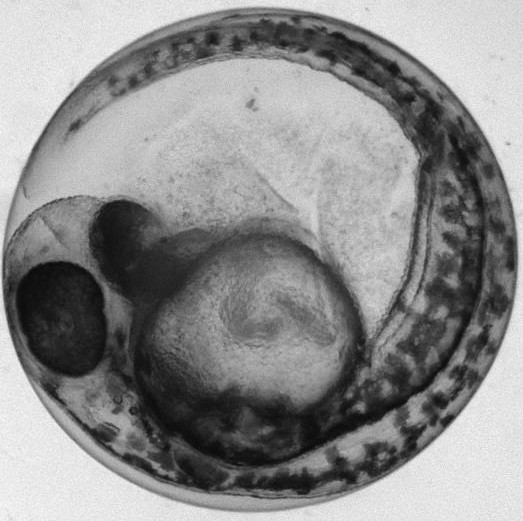}
}
\subfloat[von hinten]{
\includegraphics[width=.23\linewidth, height=.23\linewidth] {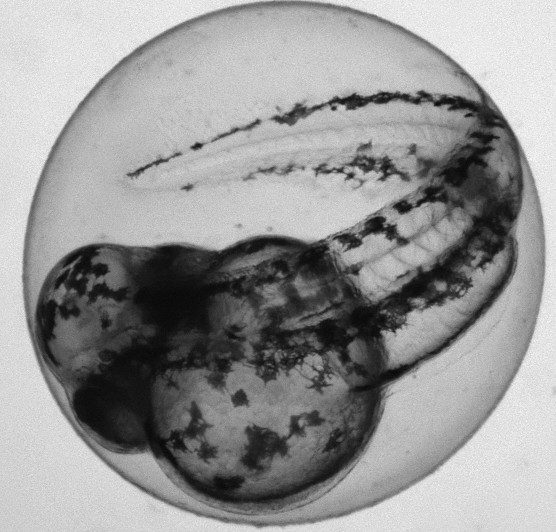}
}
\caption[Beispiele von Zebrab\"{a}rblingslarven in unterschiedlichen Orientierungen]{Beispiele von Zebrab\"{a}rblingslarven in unterschiedlichen Orientierungen im Ei}
\label{fig:Lage_Embryonen}
\end{figure}

In Tabelle \ref{tab:Lage_Embryonen} sind die Regeln der Einteilung zusammengefasst. Ein Beispiel f\"{u}r jede Position ist in \abb \ref{fig:Lage_Embryonen} gegeben. Ebenso sind hier die Ergebnisse einer manuellen Zuordnung von \"{u}ber 240 Zebrab\"{a}rblingslarven angegeben. Die Einteilung zeigt, dass nur wenig mehr als die H\"{a}lfte aller Larven in einer gleichen Position (der seitlichen Position) liegen. 25\% der Larven passen in keine der definierten Klassen. Eine Bildauswertung, die \Nutzsige in der Larve auswertet, muss daher von der Lage der Larven unabh\"{a}ngig sein, wenn keine Larven aufgrund der Lage verworfen werden sollen. Bei einer automatischen Erkennung der vier beschriebenen Klassen k\"{o}nnen zwar etwa 75\% der Larven ausgewertet werden, allerdings setzt dies voraus, dass das \Nutzsig ebenfalls in jeder Lage sichtbar ist. Des Weiteren muss f\"{u}r jede auszuwertende Klasse eine gesonderte Bildauswertung entwickelt werden. Eine Beschr\"{a}nkung auf die h\"{a}ufigste Klasse h\"{a}tte einen sehr hohen Ausschuss an Einzeluntersuchungen zur Folge, da etwa die H\"{a}lfte aller Versuche nicht ausgewertet werden k\"{o}nnen, was den Durchsatz halbiert. Beide Optionen eignen sich somit wegen des hohen Entwicklungs- \bzw Kostenaufwands nicht f\"{u}r den Praxiseinsatz.
\begin{table}[htbp]
\footnotesize
\centering
\begin{tabular}
{lD{=}{=}{-1}D{=}{=}{-1}D{=}{=}{-1}D{=}{=}{-1}D{=}{=}{-1}D{=}{=}{-1}D{=}{=}{-1}D{=}{=}{-1}}
\toprule
\multicolumn{1}{c}{\textbf{Klasse}} & \multicolumn {2}{c}{\textbf{Augen}} & \multicolumn{1}{c}{\textbf{Dottersack}} & \multicolumn{1}{c}{\textbf{R\"{u}cken}} & \multicolumn{1}{c}{\textbf{Schwanz}} & \multicolumn{1}{c}{\textbf{Anzahl}} \\
\cmidrule{2-3}
&  \multicolumn{1}{c}{\textbf{Auge 1}} &  \multicolumn{1}{c}{\textbf{Auge 2}} \\
\midrule
seitlich        &>= 90\,\% & <=\;\;\,5\,\% & >= 80\,\% & >= 50\,\% & <= 5\,\%& \multicolumn{1}{r}{$133$} \\
von oben        &>= 90\,\% & >= 60\,\% & <= 60\,\% & >= 50\,\% & <= 5\,\%& \multicolumn{1}{r}{$15$} \\
von unten       &>= 90\,\% & >= 60\,\% & >= 80\,\% & >= 20\,\% & <= 5\,\% & \multicolumn{1}{r}{$13$}\\
hinten      &<= 10\,\% & <= \;\;\,5\% & >= 50\% & \multicolumn{1}{r}{$0\,\text{-}100\%$} &  \multicolumn{1}{r}{$0\,\text{-}100\%$} & \multicolumn{1}{r}{$18$} \\
undefiniert & & & & & & \multicolumn{1}{r}{$61$}\\
\midrule
\midrule
\textbf{Summe} & & & & & & \multicolumn{1}{r}{$\mathbf{240}$}\\
\bottomrule
\end{tabular}

\caption[Zusammenfassung der Regeln zur Klassenzuordnung]{Zusammenfassung der Regeln zur Klassenzuordnung und manuelle Zuordnung von 240 unbehandelten Zebrab\"{a}rblingslarven; Augen, Dottersack: Prozent Pixel sichtbar im Vordergrund; R\"{u}cken: Prozent Pixel des R\"{u}cken\-durch\-mes\-sers sichtbar; Schwanz: Verdeckung anderer Teile der Larve in Prozent.}
\label{tab:Lage_Embryonen}
\end{table}
%

%

Zusammenfassend zeigt die Untersuchung, dass Larven im Ei sich f\"{u}r \htsen nur eignen, wenn deren \Nutzsig unabh\"{a}ngig von der Lage der Larve sichtbar \bzw quantifizierbar ist. F\"{u}r alle anderen Untersuchungen in der Larve muss auf einen zus\"{a}tzlichen Pr\"{a}parationsschritt zur\"{u}ckgegriffen werden, um eine definierte Position sicherzustellen.
Das Ergebnis ist somit, dass unter Verwendung des vorliegenden Datensatzes eine Auswertung nur Erfolg verspricht, wenn lageunabh\"{a}ngige Merkmale f\"{u}r jeden Endpunkt gefunden werden. Ist das nicht m\"{o}glich, muss die Biologie und Bildakquise erneut mit einem h\"{o}heren Informationsgehalt, beispielsweise im 3D-Raum, durchgef\"{u}hrt oder unter Verwendung von fluoreszierenden Markern erfolgen, welche weniger empfindlich auf Verdrehungen der Probe reagieren, wie \zB in \cite{Carvalho11,Gehrig09} gezeigt   .

\subsection{Fisch Embryo Test (FET) -- Koagulation \& EC$_{50}$ Regression}\label{subsec:FET_Koagulation}

Der FET soll in einer \hts zur Ermittlung der Sterberate bei unterschiedlichen Konzentrationen f\"{u}r mehrere Toxine eingesetzt werden. Zebrab\"{a}rblingseier sollen hierzu einzeln auf Koagulation untersucht werden und der Einfluss der Toxine bei steigender Konzentration ist zu bestimmen. Im Allgemeinen werden sich bei einer sehr niedrigen Konzentration fast alle Larven normal entwickeln (Sterberate bei wenigen Prozent), w\"{a}hrend bei einer sehr hohen Konzentration keiner der untersuchten Modellorganismen \"{u}berlebensf\"{a}hig ist (Sterberate 100\%). Die zu untersuchenden Toxine und die Verd\"{u}nnungsreihen je Konzentration sind in Tabelle \ref{tab:Datensatz_Tox_LC50} gegeben. Die Untersuchungen wurden am KIT durchgef\"{u}hrt.
\begin{table}[htbp]
\centering
\begin{tabular}
{p{1.3cm}rrrrrrrrrrr}
  \toprule
  \textbf{Toxin} &  \textbf{Einheit} & & & &\textbf{Dosis}\\
  \midrule
As2O3& [mM] &$ 0.80 $&$ 0.90 $&$ 1.00 $&$ 1.20 $&$ 1.40  $&$   $&$  $&      \\
CdCl& [mg/l] &$20$&$25$&$35$&$40$&$50$&$60$&$80$&$100$&$120$&$150$\\
Ethanol& [\%] &$ 1.5 $&$ 2.0 $&$ 2.25 $&$ 2.5 $&$ 3  $&$   $&$  $&\\
Methanol& [\%] &$ 3.00 $&$ 4.00 $&$ 4.50 $&$ 5.00 $&$ 6.00  $&$   $&$  $&  \\
PbCl2& [mg/l] &$ 80 $&$ 100 $&$ 120 $&$ 140 $&$ 160  $&$ 180  $&$ 200 $&$ 220$\\
  \bottomrule
\end{tabular}
\caption{ Toxine und Verd\"{u}nnungsreihen des untersuchten Datensatzes}\label{tab:Datensatz_Tox_LC50}
\end{table}

F\"{u}r die erfolgreiche Auswertung der \hts und zur Anwendung des erarbeiteten Konzeptes wird das in \abb \ref{fig:Ablaufdiagramm_Screendesign} vorgestellte Flussdiagramm zur anforderungskonformen Auswertung durchlaufen. Es werden die \"{u}bergeordneten Bl\"{o}cke der Versuchsplanung "`Biologie und Bildakquise"' sowie "`Analyse und Interpretation"' abgearbeitet. Jede Kategorie des Modulkataloges wird betrachtet und passende Module werden ausgew\"{a}hlt und angewandt. Schlie{\ss}lich wird die \hts durchgef\"{u}hrt und die Ergebnisse werden pr\"{a}sentiert. Die Anforderungen werden einzeln \"{u}berpr\"{u}ft. Ist eine Anforderung erf\"{u}llt, wird deren Kurzbezeichnung als Verweis auf Abschnitt \ref{sec:HDU_Anforderungen} in Klammern angegeben.

\subsubsection{Versuchsplanung: Biologie und Bildakquise} \label{subsubsec:Versuchsplanung-ToxI}
\begin{figure}[!htb]
               \centering
         \subfloat[
        ]{\reflectbox{\includegraphics[height=0.416\linewidth]{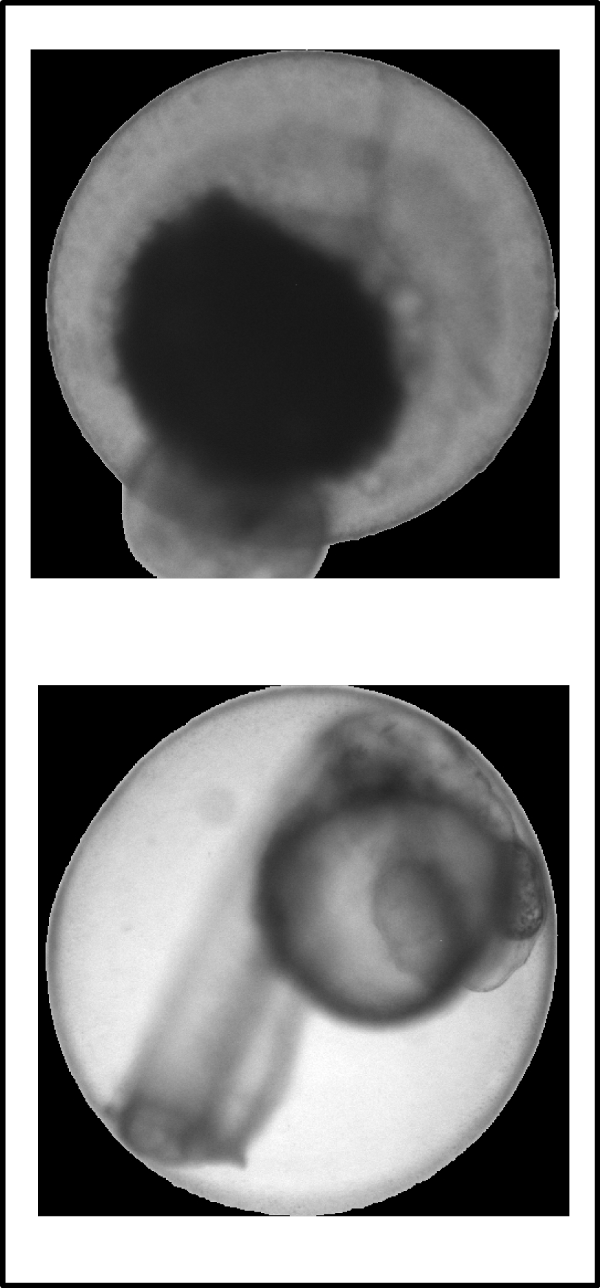}}\label{fig:Panel2_a}}
        \hfill
         \subfloat[
        ]{\includegraphics[page=4]{Bilder/03_Anwendung/Panel2/Panel}\label{fig:Panel2_b}}
        \hfill
         \subfloat[
        ]{\includegraphics[page=1]{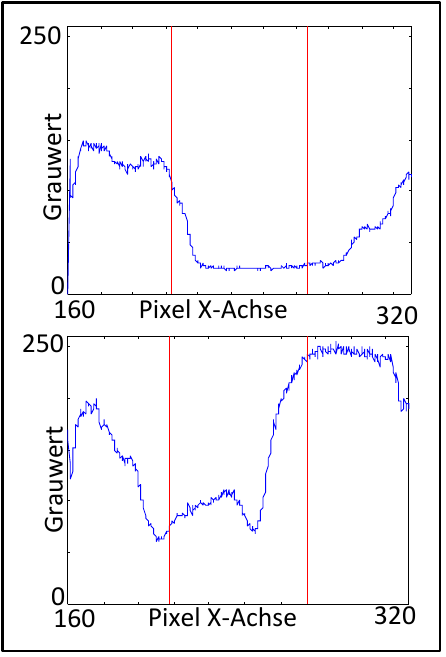}\label{fig:Panel2_c}}\hfill
         \subfloat[
        ]{\includegraphics[page=2]{Bilder/03_Anwendung/Panel2/Panel}\label{fig:Panel2_d}} \\
         \subfloat[
        ]{\includegraphics[page=3]{Bilder/03_Anwendung/Panel2/Panel}\label{fig:Panel2_e}}\hfill
         \subfloat[
        ]{\includegraphics[height=0.38\linewidth]{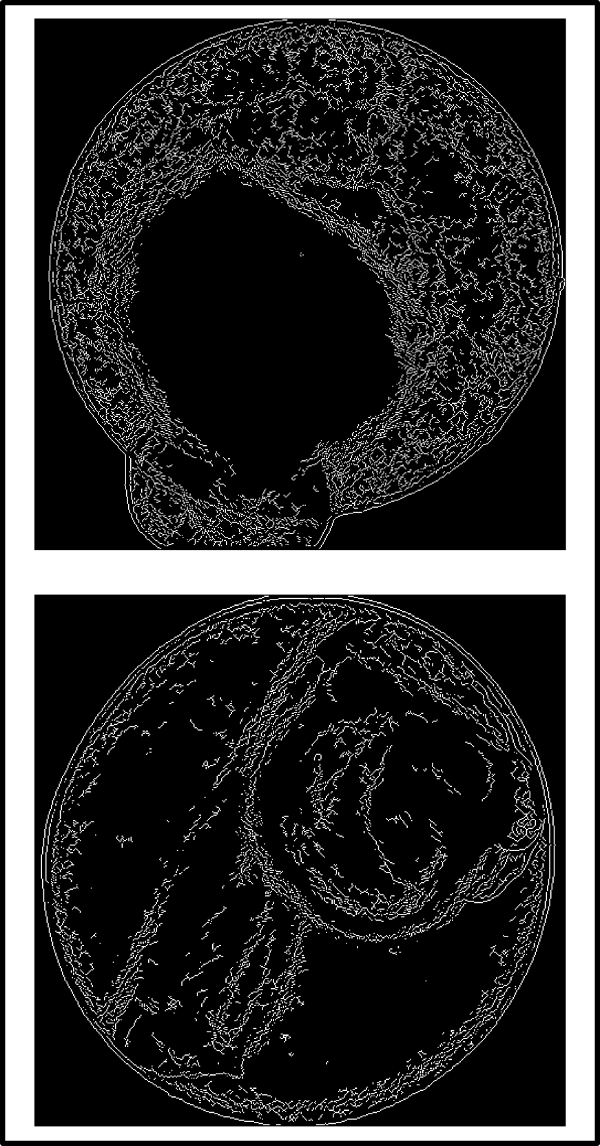}\label{fig:Panel2_f}}\hfill
         \subfloat[
        ]{\includegraphics[height=0.38\linewidth]{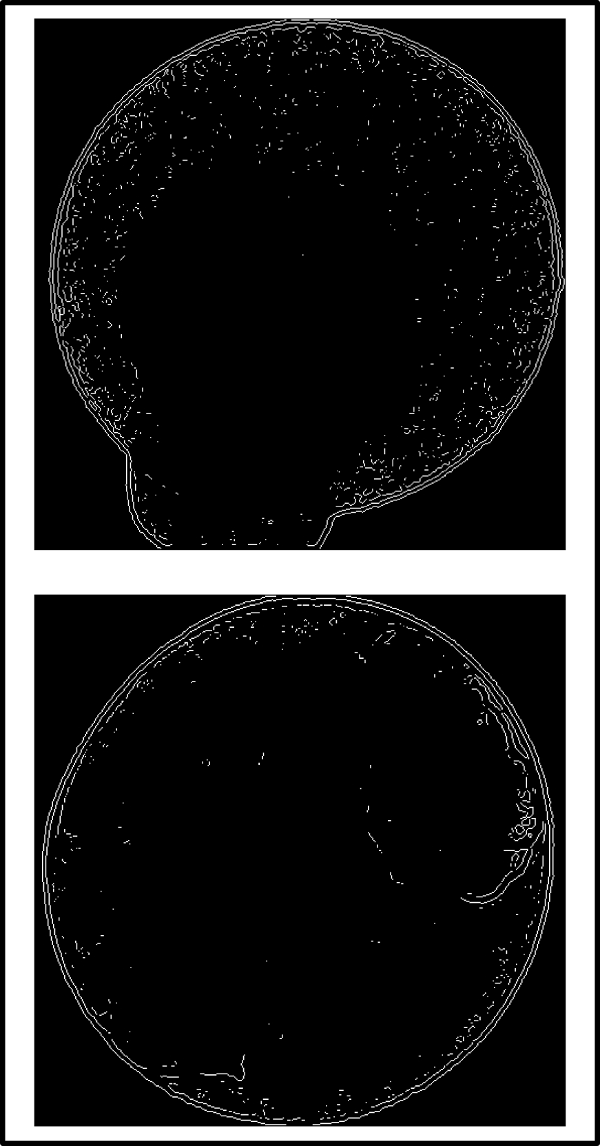}\label{fig:Panel2_g}}\hfill
         \subfloat[
        ]{
        \begin{minipage}[b]{0.18\linewidth}
        \includegraphics[width=\linewidth]{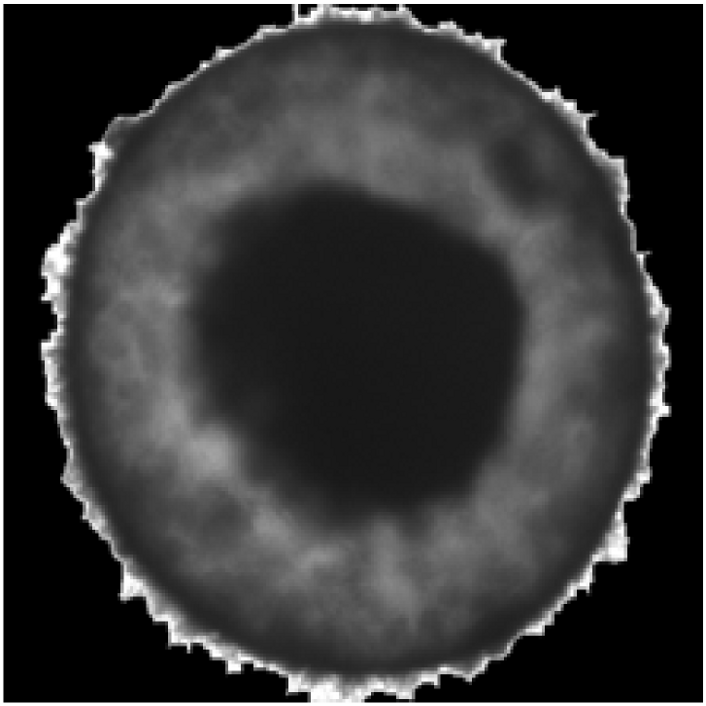}\\
        \includegraphics[width=\linewidth]{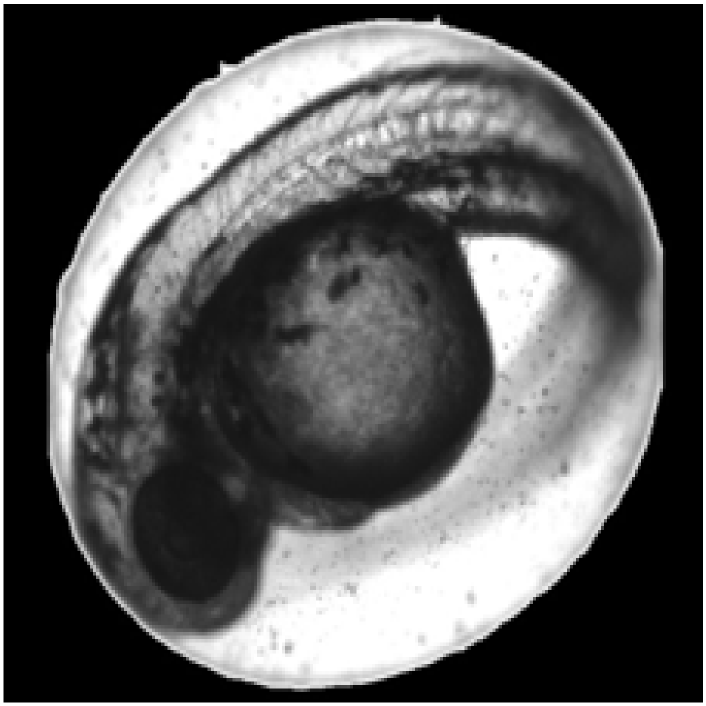}
        \tiny\\

        \end{minipage}
        \begin{minipage}[b]{0.18\linewidth}
        \includegraphics[width=\linewidth]{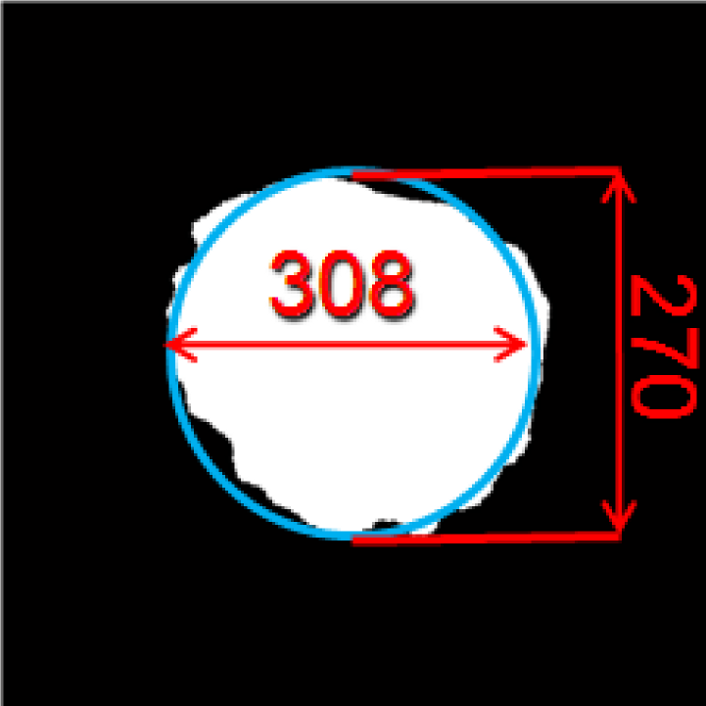}\\
        \includegraphics[width=\linewidth]{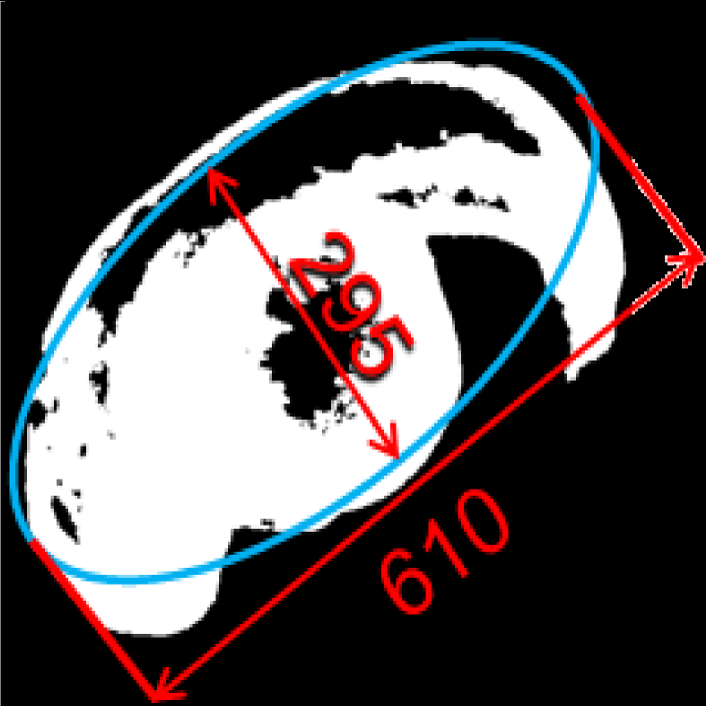}
        \tiny\\

        \end{minipage}\label{fig:Panel2_h}}
                \caption[Beispiele und Merkmale eines entwickelten sowie eines koaguliertes Eies eines Zebrab\"{a}rblings]{Beispiel f\"{u}r ein entwickeltes (\abb \ref{fig:Panel2_a} unten) sowie ein koaguliertes Ei eines Zebrab\"{a}rblings (\abb \ref{fig:Panel2_a} oben) und Illustration der extrahierten Merkmale $x_1 - x_7$ (\abb ~\ref{fig:Panel2_b}-\ref{fig:Panel2_h}; oben f\"{u}r die Klasse \emph{koaguliert,} unten f\"{u}r die Klasse \emph{entwickelt}) \cite{Alshut08}.}
                \label{fig:Panel2}
        \end{figure}

Die Bestimmung des biologischen Effekts, \dhe des \Nutzsigs, ist nach der eingef\"{u}hrten Vorgehensweise der erste Schritt und beim FET durch die OECD-Norm bereits durchgef\"{u}hrt. Ein Beispiel der zu unterscheidenden F\"{a}lle ist in \abb \ref{fig:Panel2_a} gegeben. Der markanteste Unterschied zwischen einer sich entwickelnden Larve und dem Eintreten der Koagulation ist der schwarze Punkt im Zentrum des Eies. Die Anforderungen (\vgl \kap \ref{sec:HDU_Anforderungen}) f\"{u}r die Durchf\"{u}hrung k\"{o}nnen als erf\"{u}llt betrachtet werden, da die Versuche der OECD-Norm den Gesetzm\"{a}{\ss}igkeiten entsprechen (Legitimit\"{a}t), die Larven in ausreichender Menge  zur Verf\"{u}gung stehen (logistische Machbarkeit), die Endpunkte reproduziert werden k\"{o}nnen (Reproduzierbarkeit) und auch die notwendigen Ressourcen zur Verf\"{u}gung stehen (Finanzielle Realisierbarkeit).

F\"{u}r die \"{U}berpr\"{u}fung der Anforderungen an die Messung m\"{u}ssen deren Randbedingungen abgesteckt werden. Die Messung erfolgt wie folgt: Die Fischlarven werden im Alter von 4\,hpf den Toxinen ausgesetzt, und ab 48\,hpf werden die Bilddaten akquiriert. Hierf\"{u}r wird in jedes N\"{a}pfchen von sieben Mikrotiterplatten jeweils eine Larve mittels einer Pipette transportiert. In jede Reihe wird eine h\"{o}here Konzentration eines Toxins gegeben. In jeder Mikrotiterplatte werden f\"{u}nf unterschiedliche Konzentrationen untersucht, wobei eine Reihe von Larven unbehandelt ist und die Negativ-Kontrolle darstellt. Jeweils zwei Reihen der Mikrotiterplatte blieben leer, somit sind 72 N\"{a}pfchen pro Platte besetzt. Jede Untersuchung wird doppelt ausgef\"{u}hrt (Replika \vgl Abschnitt \ref{subsec:Grundlagen_Bio}). Insgesamt werden f\"{u}r die Untersuchung somit 1008 Eier von Zebrab\"{a}rblingslarven verwendet (zweimal sieben Platten mit je 72 Larven).  Das Mikroskop erfasst zu jedem Zeitpunkt immer genau ein N\"{a}pfchen. Da die Steuer-Software des Mikroskops es nur erlaubt, die gesamte Platte zu akquirieren, werden auch Aufnahmen von den leeren N\"{a}pfchen aufgezeichnet, was sich auf die Akquise-Dauer niederschl\"{a}gt. Daher ist es ratsam, die Platten immer m\"{o}glichst voll zu besetzen.

Zur Messung wurde das Hochdurchsatz-Mikroskop Olympus-Scan$^{\mbox{\scriptsize{R}}}$ von Olympus Biosystems, ausgestattet mit einem Hamilton Micro-Lab-SWAP-Plattenwechsler, einem 2.5x Objektiv und einer Olympus Biosystems DB-1 (1344x1024 Pixel) CCD-Kamera verwendet. Die Mikroskop-Parameter (Fokus, Belichtungszeit usw.) wurden zu Beginn der Bildakquise einmalig per Hand bestimmt und mittels der mikroskopeigenen Software im XML-Format gespeichert. Es lassen sich jedoch nicht alle Parameter auf automatische Weise speichern. Wichtig ist, dass auch nicht automatisch protokollierte Parameter wie etwa ein Vorhang oder St\"{o}rungen wie \zB das Ansto{\ss}en an den Mikroskoptisch die Vergleichbarkeit von unterschiedlichen Durchl\"{a}ufen beeinflussen k\"{o}nnen und daher m\"{o}glichst zu vermeiden sind.

Nach der Akquise eines Bildstroms wird \"{u}berpr\"{u}ft, ob die Anforderungen der Messung erf\"{u}llt sind: In Abschnitt \ref{subsec:Anwendung_Datensatzanalyse} wurde gezeigt, dass sich die akquirierten Daten f\"{u}r die \hts eignen, unter der Voraussetzung, dass sich signifikante Merkmale extrahieren lassen, welche von der Lage der Zebrab\"{a}rblinge unabh\"{a}ngig sind (Eindeutige Pr\"{a}senz der biologischen Information). Da keine fluoreszierenden Marker eingesetzt werden und die Lichtmikroskopie keine bekannten toxischen Auswirkungen auf die Fische hat \cite{Pylatiuk11}, sind keine Effekte von der Bildakquise auf die Zebrab\"{a}rblinge zu erwarten (R\"{u}ckwirkungsfreiheit der Messung). Die Automatisierung der Mikroskopie mittels der Robotertechnik des \scanr soll eine hinreichend schnelle Messung sicherstellen. Die ben\"{o}tigte Zeit zur Messung eines Versuchs ist wie im Konzept beschrieben (\vgl Abschnitt \ref{sec:Parameter_mathematisch} Gleichung (\ref{eqn:RV_Dauer})) die Summe aus manuellen Pr\"{a}parationsschritten und der Dauer der Bild-Akquise. Es hat sich als zweckm\"{a}{\ss}ig erwiesen, die Pr\"{a}parationszeit f\"{u}r eine komplette Mikrotiterplatte zu bestimmen, denn typischerweise wird eine bestimmte Anzahl solcher Platten pro Tag verarbeitet und der Tagesablauf bei der Versuchsdurchf\"{u}hrung hiernach ausgerichtet. Da bei der durchgef\"{u}hrten \hts keine manuelle Ausrichtung der Larven erfolgt, sind, bez\"{u}glich der Pr\"{a}parationsschritte, lediglich die Dauer f\"{u}r die manuelle Vereinzelung ($t_{p_1}$ = \ca 6s pro N\"{a}pfchen) und f\"{u}r das manuelle Hinzuf\"{u}gen der Chemikalien ($t_{p_2}$= \ca 3s pro N\"{a}pfchen) zu ber\"{u}cksichtigen. Die Dauer der Bildakquise ist abh\"{a}ngig von der Anzahl zu akquirierender N\"{a}pfchen pro Mikrotiterplatte ($n_w$), der Anzahl von Aufnahmen pro N\"{a}pfchen hintereinander ($n_z$) und der Anzahl von Wiederholungen\footnote{Mit  "`Wiederholung der gesamten Platte"' ist hier die Wiederholung der Messung zu einem sp\"{a}teren Zeitpunkt zur Bestimmung von Bewegungen oder Entwicklungen gemeint und nicht zu verwechseln mit den Replika, welche Wiederholungen des gesamten Versuchs (inkl. Pr\"{a}paration und Akquise) sind (\vgl Abschnitt \ref{subsubsec:\"{U}bersichtHochdurschsatz}).} der gesamten Platte ($n_l$). Wird nun noch die Zeit ermittelt, welche das \scanr-Mikroskop zur Aufnahme eines Bildes ($t_{m_1}$), zum Bewegen zum n\"{a}chsten N\"{a}pfchen ($t_{m_3}$) sowie zum Bewegen an den Anfang der Mikrotiterplatte f\"{u}r eine Wiederholungsmessung ben\"{o}tigt ($t_{m_2}$), so berechnet sich die Akquise-Dauer mittels der oben aufgelisteten Parameter n\"{a}herungsweise\footnote{Abweichungen ergeben sich \zB durch Schwankungen in automatisch bestimmten Mikroskop-Parametern oder schwankende Belastung des PCs, \zB durch Datentransfer.}  mittels:
\begin{equation}\label{eq:Aquisezeit_scanr}
  t_{aq}=n_w\cdot n_z \cdot n_l \cdot t_{m_1}+ (n_w-1)\cdot n_l \cdot t_{m_2} + (n_l-1)\cdot t_{m_3}
\end{equation}
In \tab \ref{tab:Bildaquisezeit} sind die Parameter und die errechnete Dauer der Akquise f\"{u}r eine Mikrotiterplatte aufgef\"{u}hrt. Damit kann nach der bereits in Abschnitt \ref{sec:Parameter_mathematisch} eingef\"{u}hrten Formel~(\ref{eqn:RV_Dauer}) die Zeit $t_{RV}$, die f\"{u}r die Durchf\"{u}hrung eines Versuchs \inkl Akquise ben\"{o}tigt wird, bestimmt werden. Sie bel\"{a}uft sich hier auf \ca 11s und somit f\"{u}r eine Mikrotiterplatte auf weniger als 20min. Da das Zeitfenster eines vergleichbaren Alters der Larven bei \ca 4h liegt, k\"{o}nnen pro Tag und Mikroskop \ca 12 Platten oder 1152 Larven akquiriert werden. Die ermittelten maximal erreichbaren Werte sind f\"{u}r die geplante \hts ausreichend (Schnelligkeit der Messung), da f\"{u}r sieben Platten theoretisch ca. 2,5 bis 3 Stunden ben\"{o}tigt werden.
\begin{table}[htbp]
\begin{tabularx}{\linewidth}
{
>{\raggedright\arraybackslash}p{2cm}
>{\raggedleft\arraybackslash}p{.8cm}
X
}
\toprule
{\bf Parameter}							&
{\bf Wert } & {\textbf{Bemerkung}}							\\
\midrule
$n_l$&1& Wiederholungen innerhalb der Platte (nicht Replika!)\\
$n_w$&96& Standard-Mikrotiterplatte mit 96 N\"{a}pfchen\\
$n_z$&3& Anzahl der Fokusebenen\\
$t_{m_1}$&0.2&Zeit zur Aufnahme eines Bildes [s]\\
$t_{m_2}$&1.2&Zeit f\"{u}r Bewegung an den Anfang der Mikrotiterplatte [s]\\
$t_{m_3}$&4&Zeit f\"{u}r Bewegung von N\"{a}pfchen zu N\"{a}pfchen [s]\\
$t_{p_1}$&6.2& Zeit zum Vereinzeln einer Larve [s]\\
$t_{p_2}$&3.1& Zeit zum Exponieren einer Larve [s]\\
\midrule
\midrule
$t_{aq}$&1.8&Mittels Formel (\ref{eq:Aquisezeit_scanr}) ermittelte Gesamtzeit [s] f\"{u}r eine Larve\\
$t_{aq_{96}}$&172&Mittels Formel (\ref{eq:Aquisezeit_scanr}) ermittelte Gesamtzeit [s] f\"{u}r eine 96er-Platte\\
$t_{RV}$&11&Gesamter Zeitaufwand pro Larve [s]\\
$t_{RV_{96}}$&18&Gesamter Zeitaufwand pro 96er-Platte [min]\\

\bottomrule
\end{tabularx}
\caption[Ermittlung der notwendigen Zeit der Versuchsdurchf\"{u}hrung]{Ermittlung der notwendigen Zeit der Versuchsdurchf\"{u}hrung f\"{u}r eine Mikrotiterplatte}\label{tab:Bildaquisezeit}
\end{table}

\subsubsection{Versuchsplanung: Analyse und Interpretation}
F\"{u}r die Bildakquise nach \og Spezifikationen k\"{o}nnen nun die Parameter der \hts gem\"{a}{\ss} Abschnitt \ref{sec:Parameter_mathematisch} bestimmt werden.
Der biologische Effekt, die Koagulation, ist der Planfaktor $y$ und soll automatisch, anhand von aus dem Bildstrom~$\mathbf{BS}$ extrahierten Merkmalen~$\mathbf{x}$, gesch\"{a}tzt werden. Der Planfaktor geh\"{o}rt einer der Klassen $\{$\emph{ja, nein, unbekannt}$\}$ an. F\"{u}r einen Teil des Bildstroms wurden die Mikroskopaufnahmen der Eier manuell den Klassen $\{$\emph{ja, nein}$\}$ zugeordnet (gelabelt). Alle anderen Aufnahmen geh\"{o}ren der Klasse $\{$\emph{unbekannt}$\}$ an, \dhe es ist nicht bekannt, ob die Eier der Zebrab\"{a}rblinge koaguliert sind oder nicht. Die gelabelten Daten bilden die Lerndaten und werden zur Ermittlung der besten Merkmalskombination sowie zur Berechnung des Klassifikators herangezogen. Mit den Ergebnissen wird die Klassenzugeh\"{o}rigkeit  der unbekannten Daten $\hat{y}$ gesch\"{a}tzt  und einer der Klassen $\{$\emph{ja,nein}$\}$ zugeordnet.  Die Konzentration des Toxins ist der Parameter ${p}_y$ des Planfaktors und soll systematisch variiert werden. Es ist ebenso m\"{o}glich, den Datensatz nach den bekannten St\"{o}rfaktoren $z$ zu untersuchen. Diese sind das Toxin, die Position der untersuchten Larve innerhalb einer Konzentration auf der Mikrotiterplatte (hier immer in einer Reihe der Platte, \dhe Position 1 bis 12), das verwendete Mikroskop, die Wiederholung (Replika), das Aufnahmedatum und der Aufnahmezeitpunkt. Das Toxin ist deshalb als bekannter St\"{o}rfaktor zu betrachten, da zur Bestimmung des $LC_{\text{50}}$-Wertes lediglich die Konzentration des Toxins ge\"{a}ndert wird, nicht jedoch die \emph{Art} des Toxins. Eine Klassenzuweisung bez\"{u}glich der St\"{o}rfaktoren ist hier jedoch f\"{u}r die Bestimmung des $LC_{\text{50}}$-Wertes nicht weiter von Interesse. Im Gegenteil soll es bei gleichwertigen Bedingungen der \hts f\"{u}r einen optimalen Versuch unm\"{o}glich sein, eine Unterscheidung anhand der St\"{o}rfaktoren zu treffen (vgl. Abschnitt \ref{sec:Konzept_Versuchsauslegung}). Als Parameter der bekannten St\"{o}rfaktoren $p_z$ bleibt lediglich der Typ der verwendeten Mikrotiterplatte, da die Konzentration bereits Parameter des Planfaktors ist. Der Bildstrom wird in Farbe (Modalit\"{a}t $m_\mathbf{BS}=3$) bei einer Aufl\"{o}sung von 1334x1024 auf drei Fokusebenen (${u}_\mathbf{BS}=1334x1024x3$) zu lediglich einem Zeitpunkt (ein einziger Frame $t_\mathbf{BS}$=1) aufgezeichnet. Tabelle \ref{tab:Anw_Parameter_TOXI} fasst die Parameter des Versuchs zusammen.
\begin{table}[htb]
\footnotesize
\begin{tabularx}{\linewidth}{C{2cm}Zp{2.2cm}Zccc}
\toprule
$y/\hat{y}$	&$p_y$	&\multicolumn{1}{c}{$z/\hat{z}$}	&$p_z$ &\multicolumn{3}{c}{$p_\mathbf{BS}$}  \\
\cmidrule(l){5-7}
	&	&	&	&$m_\mathbf{BS}$	&${u}_\mathbf{BS}$	&$t_\mathbf{BS}$		 \\
\midrule
Koagu\-la\-tion	&Konzen\-tration	&Toxin\newline Position\newline Mikroskop\newline Wiederholung\newline Aufnahmedatum\newline Aufnahmezeit	 &$96er$-Mikro\-titer\-platte	 &$3$	 &$1334x1024x3$	&$1$		 \\
\bottomrule
\end{tabularx}
\caption[Parameter der \hts f\"{u}r einen Endpunkt]{Parameter der \hts f\"{u}r einen Endpunkt nach Abschnitt~\ref{sec:Parameter_mathematisch}; Mit $m_{\mathbf{BS}}$ Anzahl Kan\"{a}le, ${u}_\mathbf{BS}$ Anzahl Pixel und $t_\mathbf{BS}$ Anzahl Frames. }\label{tab:Anw_Parameter_TOXI}
\end{table}

\begin{figure}[htbp]
\centering
        \centering
               \includegraphics[page=13,
              width=\linewidth
               ]{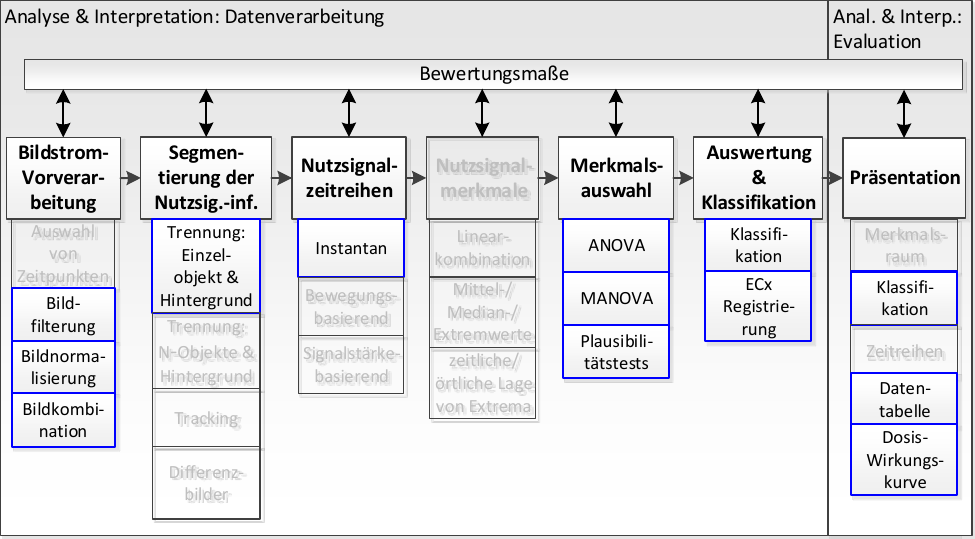}

        \caption[Module zur Auswertung und Bestimmung der EC$_{50}$ Konzentration]{Module, die zur Auswertung und Bestimmung der EC$_{50}$ Konzentration ausgew\"{a}hlt wurden}
        \label{fig:Anw_Module_LC50}
\end{figure}

F\"{u}r die Auswertung, den dritten Punkt im Ablaufdiagramm (\abb \ref{fig:Ablaufdiagramm_Screendesign}), werden geeignete Module identifiziert und zu einer Auswertungskette zusammengeschaltet. \abb \ref{fig:Anw_Module_LC50} hebt die ausgew\"{a}hlten Module im Modulkatalog hervor. Die Beschreibung der Module findet sich in Kapitel \ref{sec:BV_Module}, daher wird im Folgenden nur auf die Auswahl der Module und die damit gebildete Auswertungskette sowie deren Spezifizierung auf den FET eingegangen.

\textbf{Bildstrom-Vorverarbeitung:}
F\"{u}r die Bildstrom-Vorverarbeitung wurden die folgenden drei Module eingesetzt: In einer Bildkombination wird aus den Aufnahmen \"{u}ber drei Fokusebenen ein Bild mit erweitertem Fokus erstellt. Hierdurch sind die Details im Fischei deutlicher zu unterscheiden. Die Aufnahmen des Mikroskops \scanr nutzen lediglich einen kleinen Teil des zur Verf\"{u}gung stehenden Wertebereichs. Daher wird mittels Bildnormalisierung eine Spreizung des Histogramms hinzugef\"{u}gt, welche sich der Bildfilterung bedient. Als Ergebnis steht ein verbesserter Bildstrom $\mathbf{BS}^*$ zur Verf\"{u}gung.
      Da der Bildstrom f\"{u}r das \Nutzsig \emph{Koagulation} lediglich zu einem Zeitpunkt und an einem einzigen Mikroskop aufgezeichnet wurde, k\"{o}nnen eine aufwendigere Bildnormalisierung und die Auswahl von Zeitpunkten entfallen.

\begin{figure}[htbp]
        \centering
        \includegraphics[page=1]{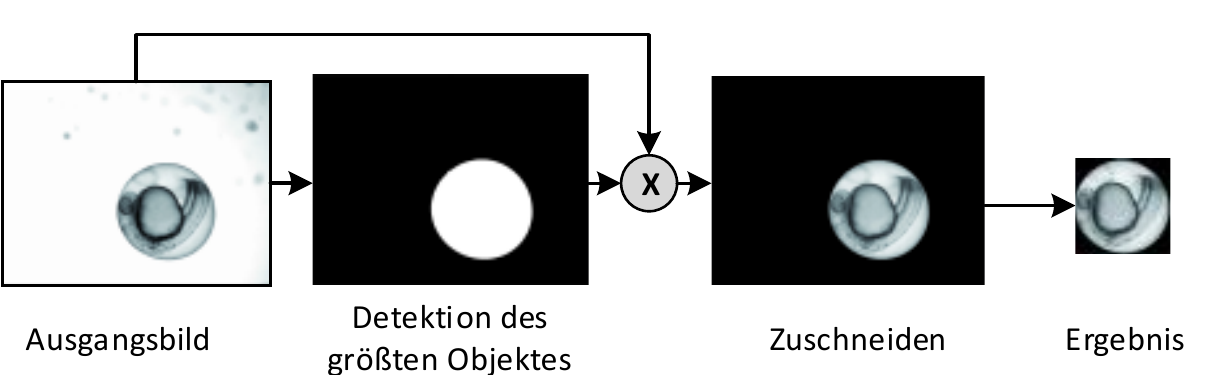}
        \caption{Veranschaulichung der Anwendung des Moduls \emph{Trennung von Einzelobjekt $\emph{\&}$ Hintergrund.}}
        \label{fig:Tox1_Trennung_Einzelobjekt}
      \end{figure}

\textbf{Segmentierung der Nutzsignalinformation:}
      Das Ei des Zebrab\"{a}rblings wird bei der Pr\"{a}paration vereinzelt, so dass lediglich ein Ei in jedem N\"{a}pfchen der Mikrotiterplatte und somit in jedem Bildstrom vorhanden ist. Hierf\"{u}r wird das Modul zum Trennen eines Einzelobjektes vom Hintergrund angewandt. Das Ergebnis ist eine Bildmaske, welche das Ei vom Hintergrund trennt und mit dem Originalbild multipliziert wird.  Fehlerhafte Aufnahmen werden wie im Modul beschrieben anhand der Gr\"{o}{\ss}e und "`Rundheit"' des Eies ausgeschlossen. \abb \ref{fig:Tox1_Trennung_Einzelobjekt} veranschaulicht die Anwendung der Segmentierung.

      Die Segmentierung l\"{a}sst sich robust auf den akquirierten Datensatz anwenden. Von 1008 Bildern werden sieben automatisch anhand der genannten Kriterien aussortiert. Die niedrige Fehlerrate erlaubt es, auf zus\"{a}tzliche Validierungspr\"{u}fungen zu verzichten. Die Fehler beruhen zudem ausschlie{\ss}lich auf qualitativen Inhomogenit\"{a}ten in den Bilddaten und k\"{o}nnen durch genauere Pr\"{a}paration abgestellt werden. Die Bilddaten nach der Segmentierung werden zur sp\"{a}teren Analyse gespeichert. Der vollst\"{a}ndige Prozess (\inkl Einlesen und Speichern) ben\"{o}tigt pro Bild 0.46s und f\"{u}r den gesamten Datensatz 7:42\,min auf einem normalen PC\footnote{CPU: Intel$^{\circledR}$ Core$^{\mbox{\scriptsize TM}}$ i5 750 @ 2.67GHz, Arbeitsspeicher: 4.00GB; OS: Windows 7 Enterprise 64Bit } (Segmentierbarkeit des biologischen Effekts im Bild).

\textbf{Nutzsignalzeitreihen:}
    Der Bildstrom wurde an lediglich einem Zeitpunkt aufgezeichnet. Somit kann auch nur f\"{u}r den Zeitpunkt der Aufzeichnung ein Merkmal extrahiert werden, was zu Einzelwerten \bzw einer Zeitreihe der L\"{a}nge $k=1$ f\"{u}hrt. Das Fehlen von Bildsequenzen beschr\"{a}nkt die Menge der extrahierbaren Merkmale auf die instantanen Merkmale (\vgl Abschnitt \ref{subsec:Module_Instantane_Merkmale}). Voraussetzung f\"{u}r die Merkmale ist, dass sie unabh\"{a}ngig von der Lage der Larve eine Klassifikation erlauben. Ein Beispiel beider Klassen des zu unterscheidenden Nutzsignals ist in \abb \ref{fig:Panel2_a} gegeben.

    Potenziell eignen sich alle instantanen Merkmale aus Abschnitt \ref{subsec:Module_Instantane_Merkmale} zur quantitativen  Beschreibung des Eiinhaltes bez\"{u}glich Koagulation. Es wurden daher auch alle Merkmale $x_1$ bis $x_8$ extrahiert. \abb \ref{fig:Panel2_b}-\ref{fig:Panel2_h} illustriert Beispiele f\"{u}r einige der Merkmale. Die obere Reihe der Abbildungen ist jeweils f\"{u}r ein koaguliertes Ei, die untere f\"{u}r eine normal entwickelte Larve. In \abb \ref{fig:Panel2_b} liegt der Wert von $x_1$ (Schwerpunkt des Histogramms) f\"{u}r koagulierte Eier bei deutlich niedrigeren Werten (hier 94 (koaguliert) zu 164 (entwickelt)), ebenso verh\"{a}lt sich das nicht dargestellte Merkmal $x_2$, welches sich auf die mittlere Helligkeit des gesamten Bildes bezieht. In \abb \ref{fig:Panel2_c} ist die mittlere Bildzeile des Bildes dargestellt. Von der Zeile wird der Mittelwert der mittleren 30\% als Merkmal $x_3$ extrahiert und ist in \abb \ref{fig:Panel2_d} durch die gr\"{u}ne Linie gekennzeichnet. Der Mittelwert f\"{u}r ein koaguliertes Ei liegt deutlich niedriger.  \abb \ref{fig:Panel2_e} illustriert das Merkmal $x_4$, welches die Schwankung der Grauwerte um den Mittelwert von $x_3$ quantifiziert. Das Merkmal ist die blau eingef\"{a}rbte Fl\"{a}che unter der Kurve.
     \abb \ref{fig:Panel2_f}-\ref{fig:Panel2_g} zeigt ein Beispiel f\"{u}r die Kantenbilder, deren Anzahl durch die Merkmale $x_5$ und $x_6$ extrahiert wird. \abb \ref{fig:Panel2_h} schlie{\ss}lich illustriert die l\"{a}ngste und k\"{u}rzeste Halbachse der umschlie{\ss}enden Ellipse, deren Differenz das Merkmal $x_7$ darstellt. Das Merkmal $x_8$ wurde nicht dargestellt, da es lediglich die Bildgr\"{o}{\ss}e enth\"{a}lt und als Validit\"{a}tspr\"{u}fung genutzt wird (vgl. hierzu Abschnitt \ref{Trennung von Einzelobjekt und Hintergrund} und Abschnitt \ref{subsec:Module_Instantane_Merkmale}).

\textbf{Dimensionsreduktion:}
Die Dimensionsreduktion der Zeitreihen zu Einzelmerkmalen entf\"{a}llt, da lediglich Einzelmerkmale extrahiert wurden.

\begin{figure}[htbp]
  \begin{minipage}[b]{.45\linewidth}
  \footnotesize
    \centering
\begin{tabularx}{\linewidth}{C{1.5cm}VV}
\toprule
\textbf{Merkmal} & \textbf{G\"{u}te-MANOVA} &\textbf{G\"{u}te- ANOVA}\\
\midrule
  $x_{7}$                                                               & {---\,\;} & 0.872  \\
 $x_{3}$                                                               & 0.898 & 0.693 \\
$x_{4}$                                                               & 0.886  & 0.556\\
$x_{6}$                                                               & 0.880  &0.430\\
$x_{5}$                                                               & 0.879  &0.389\\
$x_{1}$                                                               & 0.875  & 0.580\\
 $x_{8}$                                                               & 0.872  & 0.056\\
$x_{2}$                                                               & 0.872  & 0.249\\
\bottomrule
\end{tabularx}
      \vspace{1cm}
    \captionof{table}{Merkmalsrelevanzen -- Univariate und  Multivariate Analyse (bei Auswahl von 2 Merkmalen)\label{tab:Merkmalsrelevanzen_ToxI}}

  \end{minipage}\hfill
  \begin{minipage}[b]{.5\linewidth}
    \centering
  \includegraphics[page=1]{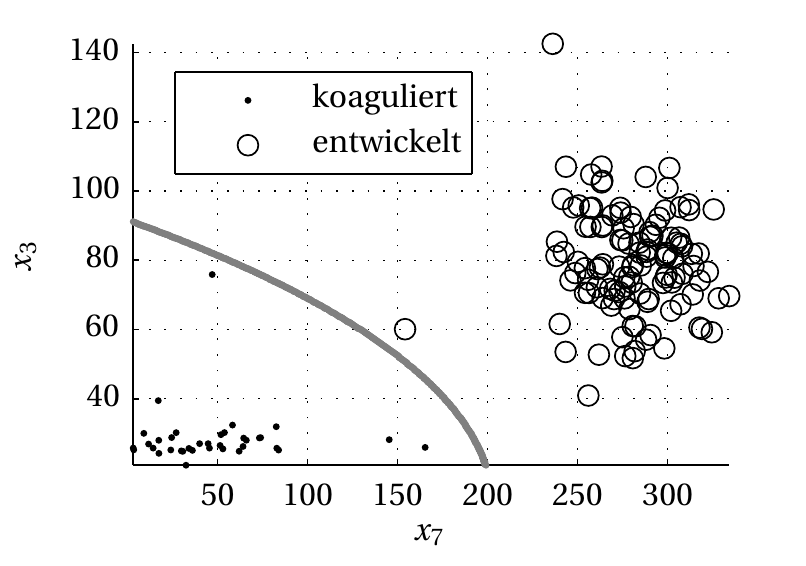}
\caption{Auf Basis der Lerndaten berechnete Diskriminanzfunktion\\}
\label{fig:Trainings_Daten}
  \end{minipage}
\end{figure}

\textbf{Merkmalsauswahl:}
Von einem Experten wurde ein Trainingsdatensatz aus einem Teil des aufgenommenen Datensatzes erstellt. Dabei wurden 35 Larven der Klasse \emph{koaguliert} und 120 Larven der Klasse \emph{entwickelt} zugeordnet. Anhand der Trainingsdaten wurde das trennst\"{a}rkste Merkmal mittels ANOVA ermittelt und mittels MANOVA die beste 2er-Merkmalskombination bestimmt. Es zeigt sich, dass das Merkmal $x_7$ mit einer G\"{u}te von $0.872$ in der ANOVA das beste Merkmal zur Klassifikation ist. Die beste Zweierkombination ergibt sich unter Hinzuziehung des Merkmals $x_3$ mit einer erreichten G\"{u}te von $0.898$ (vgl. \tab \ref{tab:Merkmalsrelevanzen_ToxI}). Somit ist eine Merkmalskombination extrahiert, die die \NutzsigInf beschreibt und eine gute Trennsch\"{a}rfe aufweist (Quantifizierbarkeit des biologischen Effekts im Bild).

\textbf{Auswertung \& Klassifikation:}
Anhand der identifizierten trennstarken Merkmalskombination und der Trainingsdaten wurde ein Bayes"~Klas\-sifi\-ka\-tor mit Ber\"{u}cksichtigung der klassenspezifischen Kovarianzmatrizen angelernt \cite{Mikut08}. Die Klassifikation zeigt auf den Testdaten in einer 5-fach-Crossvalidierung einen Klassifikationsfehler von $1.2\pm0.2\%$ (niedrige Fehlerrate bei der Detektion des biologischen Effekts). In \abb \ref{fig:Trainings_Daten} ist ein Scatterplot der Trainingsdaten \"{u}ber die Merkmale $x_7$ und $x_3$ sowie die Diskriminanzfunktion des Klassifikators abgebildet. Die der Klasse \emph{koaguliert} zugeordneten Larven zeigen f\"{u}r $x_3$ und $x_7$ deutlich geringere Werte als die der Klasse \emph{entwickelt} zugeordneten. Die Diskriminanzfunktion des Bayesklassifikators ist ebenfalls abgebildet.  Der Klassifikator kann nun bei Durchf\"{u}hrung der \hts auf Testdaten angewandt werden. Die Ergebnisse der Klassifikation erm\"{o}glichen die Zuordnung einer relativen Sterberate zu jeder Konzentration eines jeden Toxins.

Durch Anwendung des Moduls "'EC$_\text{x}$-Registrierung"' zur Ermittlung der Dosis-Wirkungskurve aus den Ergebnissen der Sterberaten der Konzentrationsreihen eines jeden Toxins l\"{a}sst sich bei geeigneten Werten der EC$_{50}$-Wert ermitteln. Im vorliegenden Datensatz lie{\ss}en die erzielten Effekte der Konzentrationen die Ermittlung des Wertes bei drei Toxinen zu, wobei keine Aufnahme so schlecht war, dass sie durch die Validierung ausgeschlossen werden musste. Es konnten somit alle aufgezeichneten Larven klassifiziert werden. Wenn es nicht m\"{o}glich war, ein EC$_{50}$"~Wert zu ermitteln war die gew\"{a}hlte Verd\"{u}nnungsreihe nicht passend (\vgl \kap \ref{subsec:Pr\"{a}s_EC50}). Die Anzahl der untersuchten Larven, Konzentrationen sowie die getroffenen Klassenzuordnungen sind in \tab \ref{tab:Ergebnisse_LC50} aufgelistet. \tab \ref{tab:Klassifikation_LC50} zeigt die Aufschl\"{u}sselung auf die einzelnen Konzentrationen. F\"{u}r das Toxin \emph{AS2O3} wurden beispielsweise 120 Larven untersucht (Zus\"{a}tzlich noch 24 unbehandelte Larven als Kontrolle), wovon sich jeweils 24 auf jede der 5 Konzentrationen verteilen (12 je Replika). Bei Konzentration \textbf{K1} von \emph{AS2O3} (0.80\,mM \vgl \tab \ref{tab:Datensatz_Tox_LC50}) wurden bei der ersten Wiederholung 0 von 12 Larven als koaguliert eingestuft (\dhe alle 12 Larven sind als  \emph{nicht koaguliert} eingestuft), beim zweiten Durchgang wurden 3 von 12 Larven als koaguliert gesch\"{a}tzt. Bei der Konzentration \textbf{K2} wurde keine der Larven als koaguliert eingestuft usw. Insgesamt wurden f\"{u}r das Toxin 49 Larven als koaguliert und 71 als entwickelt also nicht koaguliert eingesch\"{a}tzt. Der ermittelte EC$_{50}$ liegt bei einer Konzentration von 1.1\,mM.

\begin{table}[!htbp]
\begin{centering}
\begin{tabular}{
lrR{2cm}R{2cm}R{2cm}r
}
\toprule
\textbf{Toxine} & \textbf{Larven} & \multicolumn{1}{C{2.5cm}}{\textbf{Konzentrationen}} & \multicolumn{1}{C{2.5cm}}{\textbf{Sch\"{a}tzung koaguliert}} & \multicolumn{1}{C{2.5cm}}{\textbf{Sch\"{a}tzung entwickelt}} & \textbf{EC$_{50}$} \\
\midrule
As2O3        & 120 &  K1-K5 &  49 &  71 &  1.1\\
CdCl& 240 &  K1-K10 & 31 &209 &  -\\
Ethanol&120&K1-K5&27&93&2.5\\
Methanol&120&K1-K5&74&46&2.5\\
PbCl&240&K1-K8&47&193&-\\
Kontrolle&168&-&-&-&-\\
\bottomrule
\end{tabular}
\caption[Anzahl von untersuchten Larven und Konzentrationen]{Anzahl von untersuchten Larven und Konzentrationen sowie Klassifikationsergebnisse und errechnete EC$_{50}$-Werte \"{u}ber den Toxinen.}
\label{tab:Ergebnisse_LC50}
\end{centering}
\end{table}

\begin{table}[!htbp]
\begin{centering}
\begin{tabular}{
lrrrrrrrrrrrrrrrrrrrrrrr
}
\toprule
\textbf{Toxin} & \textbf{K1} & \textbf{K2} & \textbf{K3} & \textbf{K4} & \textbf{K5}&\textbf{K6}&\textbf{K7}&\textbf{K8}&\textbf{K9}& \textbf{K10} &\textbf{Kontrolle} \\
\midrule
AS2O3-1     &0	&0	&3	&9	&11	&	&	&	&	&	&1\\
AS2O3-2     &3	&0	&4	&9	&10	&	&	&	&	&	&0\\
CdCl-1      &1	&1	&5	&2	&0	&0	&2	&4	&2	&1	&0\\
CdCl-2      &1	&0	&0	&2	&1	&0	&0	&3	&4	&2	&1\\
Ethanol-1   &1	&0	&0	&6	&7	&	&	&	&	&	&0\\
Ethanol-2   &0	&0	&0	&4	&9	&	&	&	&	&	&0\\
Methanol-1  &0	&6	&10	&11	&12	&	&	&	&	&	&0\\
Methanol-2  &0	&2	&10	&12	&11	&	&	&	&	&	&0\\
PbCl-1      &3	&3	&4	&0	&6	&1	&5	&4	&	&	&0\\
PbCl-2      &5	&6	&1	&1	&3	&0	&2	&4	&	&	&0\\
\bottomrule
\end{tabular}
\caption[Als koaguliert eingestufte Larven von jeweils 12 Larven der Konzentration \"{u}ber den Toxinen]{Als koaguliert eingestufte Larven von jeweils 12 untersuchten Larven je Konzentration \"{u}ber den Toxinen der Mikrotiterplatten aus Tabelle \ref{tab:Ergebnisse_LC50}. Jede Mikrotiterplatte wurde jeweils einmal zur Kontrolle wiederholt.
(Bei~ CdCl-1(Kontrolle), CdCl-2(Kontrolle), PbCl-1(K1), PbCl-2(K1), PbCl-1(K2), PbCl-2(K2), PbCl"~1\-(Kontrolle) und PbCl"~2(Kontrolle) sind anstelle von 12 jeweils 24 Larven getestet worden.)}
\label{tab:Klassifikation_LC50}
\end{centering}
\end{table}

\begin{figure}[htbp]
\centering
        \centering
        \includegraphics[width=\linewidth]{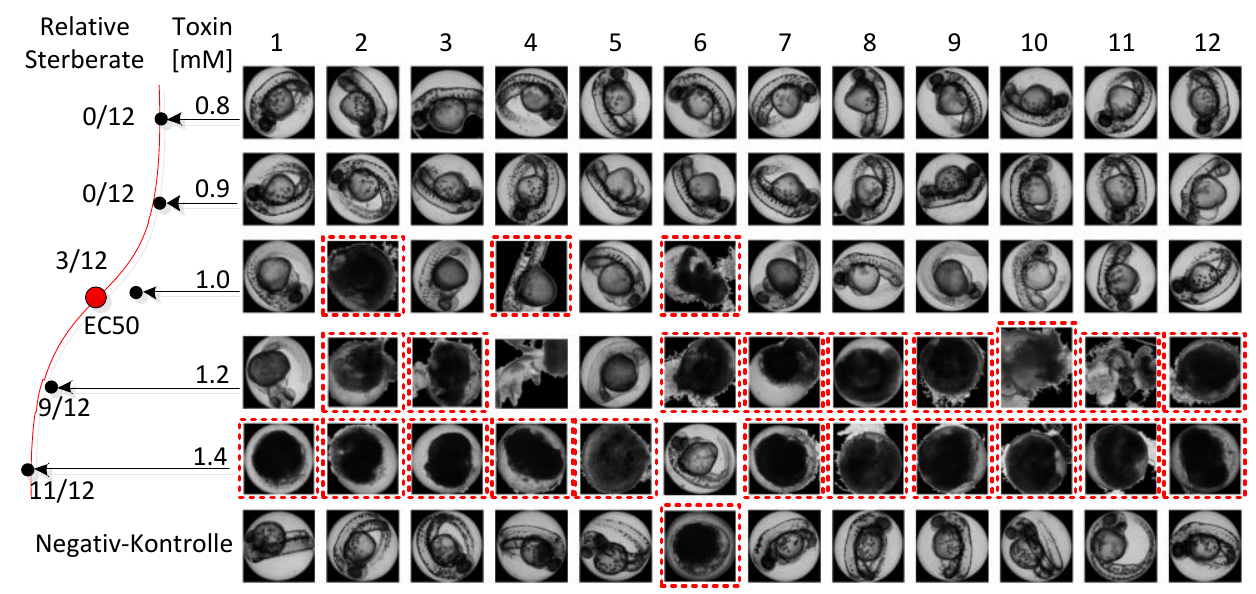}
        \caption[Visualisierung der Klassifikationsergebnisse der \hts]{Visualisierung der Klassifikationsergebnisse der \hts, hier f\"{u}r das Toxin As2O3; gestrichelt markierte Larven sind der Klasse \emph{Koaguliert} zugeordnet \cite{Alshut10}}
        \label{fig:Klassifikationsergebnisse}
\end{figure}

\begin{figure}[htbp]
\begin{minipage}[b]{0.45\linewidth}
\centering
\includegraphics[page=1,width=\linewidth]{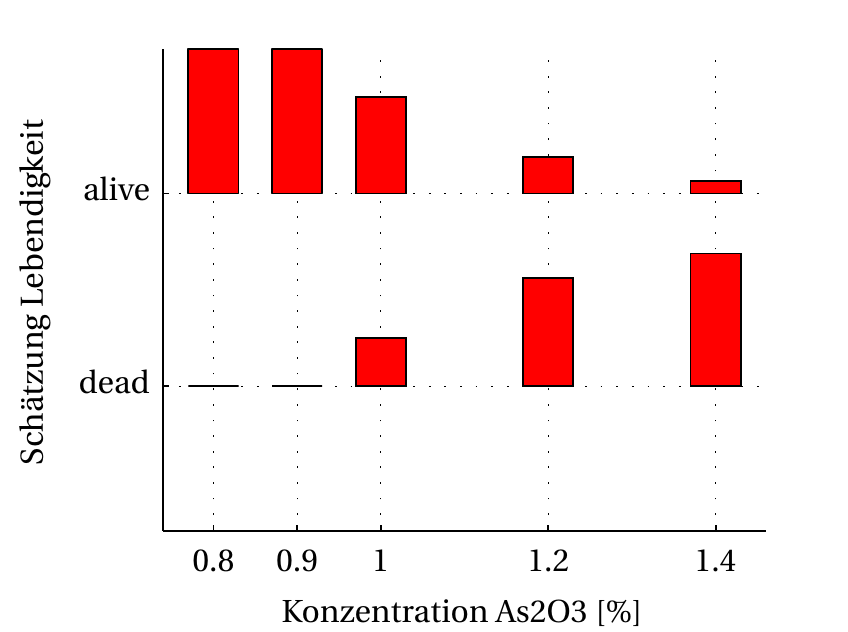}
\caption{2D-Histogramm koagulierter und entwickelter Larven}
\label{fig:Tox1_2DHistogram}
\end{minipage}
\hfill
\begin{minipage}[b]{0.5\linewidth}
\centering
\includegraphics[page=1,width=\linewidth]{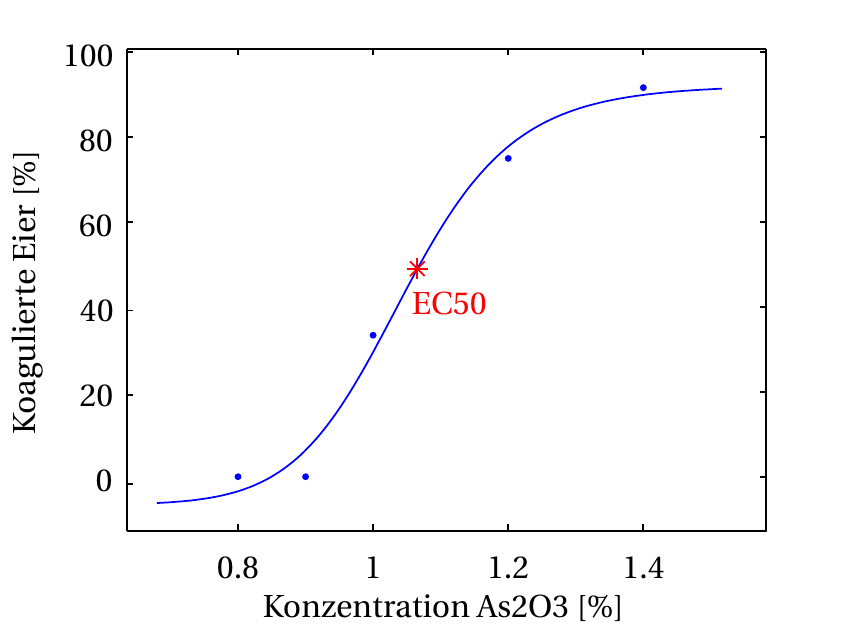}
\caption{Automatische Regression und EC$_{50}$-Berechnung}
\label{fig:Tox1_Regression}
\end{minipage}
\end{figure}

\textbf{Pr\"{a}sentation:}
Die Ergebnisse werden zur biologischen Auswertung in Form einer 22 Seiten starken Reportdatei im PDF-Format erstellt. Das Dokument enth\"{a}lt f\"{u}r jedes Toxin eine Reihe von Darstellungen der Ergebnisse, die jeweils auf einer DIN-A4 Seite das Ergebnis bereitstellen. Die Darstellungen sind: eine \"{U}berlagerung der Bilddaten mit den Klassifikationsergebnissen (vgl. \abb \ref{fig:Klassifikationsergebnisse}), eine Auswertung der Klassifikationsergebnisse in Tabellenform (\vgl \tab \ref{tab:Ergebnisse_LC50} und \tab \ref{tab:Klassifikation_LC50}), ein klassenspezifisches 2-D-Histogramm \"{u}ber die Konzentration (\vgl \abb \ref{fig:Tox1_2DHistogram}), die Regression auf die Dosis-Wirkungs-Kurve mit \"{U}berlagerung der ermittelten Konzentration EC$_{50}$ (\vgl \abb \ref{fig:Tox1_Regression}, dargestellt ist das Toxin Methanol), ein Scatterplot \"{u}ber Klassifikationsergebnisse im Merkmalsraum \"{u}ber alle Daten sowie f\"{u}r jede Konzentration. Zus\"{a}tzlich werden Metainformationen wie etwa der Name des Toxins und der Speicherort der Daten abgelegt. \abb \ref{fig:Klassifikationsergebnisse} zeigt die automatisch zugeschnittenen und tabellarisch in der Ordnung der Mikrotiterplatten angeordneten Bilder der Zebrafischlarven, wie sie f\"{u}r jedes Toxin in der Reportdatei abgelegt werden. Die Darstellung ergibt somit "`einen virtuellen Blick"' in die gesamte Mikrotiterplatte. Die als koaguliert klassifizierten Larven sind rot-gestrichelt umrandet. \abb \ref{fig:Tox1_2DHistogram} zeigt das klassenspezifische 2-D-Histogramm. Die H\"{o}he des roten Balkens gibt den Prozentwert an Larven an, die der jeweiligen Klasse koaguliert oder entwickelt (hier englisch \emph{dead }und \emph{alive}) zugeordnet wurden. Die beiden roten Balken ergeben in der Summe somit immer 100\%. Mit Hilfe der in der Reportdatei gesammelten Informationen steht dem Anwender das Endergebnis \"{u}bersichtlich zur Verf\"{u}gung und es lassen sich anhand der \"{U}berlagerung der Ergebnisse aus Segmentierung und Klassifikation alle Verarbeitungsschritte innerhalb weniger Minuten auf grunds\"{a}tzliche Konsistenz pr\"{u}fen (Pr\"{a}sentierbarkeit der Auswertung).

\subsubsection{Durchf\"{u}hrung und Resultat}

Nach dem Erstellen des Klassifikators l\"{a}sst sich die erarbeitete Auswertungskette durch die kurze ben\"{o}tigte Zeit zur Auswertung auch auf gro{\ss}e Datens\"{a}tze skalieren. Der gesamte Auswerteprozess ben\"{o}tigt (\inkl Einlesen und Speichern von Bildern, Projektdateien, Ergebnissen) pro Platte 73.5s, \dhe f\"{u}r die gesamte Untersuchung \ca 8:30\,min (Schnelligkeit der Auswertung). Ein erneutes Durchlaufen des Flussdiagramms ist nur notwendig, wenn eine \"{A}nderung der St\"{o}rfaktoren (\zB Anzahl der Mikroskope, Anzahl der Fischeier pro N\"{a}pfchen) vorgenommen werden muss. Alle Bilddaten, Gait-CAD-Projekte, Zwischenergebnisse und Reportdateien wurden auf einem Archiv des KIT namens LSDF (engl. Large Scale Data Facility) \cite{Stotzka11}, welches f\"{u}r die Speicherung gro{\ss}er Datenmengen \"{u}ber eine lange Zeit erstellt wurde, gespeichert (Wissenschaftliche Archivierung). Das Durchlaufen des vorgeschlagenen Flussdiagramms aus Abschnitt \ref{sec:Anforderungsgerechte_Anwendung} f\"{u}hrt erwartungsgem\"{a}{\ss} zu einer aussagekr\"{a}ftigen \hts , welche die Beantwortung der Fragestellung unter Einhaltung aller Anforderungen erlaubt. Es ist nun erstmals m\"{o}glich, den FET bildbasierend f\"{u}r den Endpunkt \emph{Koagulation} mit hoher Erfolgsrate automatisch auszuwerten \cite{Alshut09,Alshut11AT}.

Eine manuelle \"{U}berpr\"{u}fung der Ergebnisse zeigt jedoch, dass die Klassifikation durch die Beschr\"{a}nkung auf lediglich einen Endpunkt, aus biologischer Sicht, \dhe f\"{u}r die Beurteilung der Lebendigkeit der Zebrab\"{a}rblinge, zu einer hohen Falsch-Negativ-Rate kommt. Viele Larven weisen zwar keine Koagulation auf, k\"{o}nnen jedoch nicht als lebendig eingestuft werden, was sich dadurch zeigt, dass sie weder Herzschlag besitzen noch spontane Bewegungen ausf\"{u}hren. Das Ergebnis legt die Erweiterung der bildbasierten Auswertung bez\"{u}glich der beiden Endpunkte \emph{Herzschlag }und \emph{Spontanbewegung} nahe, welche im folgenden Abschnitt \ref{subsec:FET_Heartbeat} vorgestellt wird.

\subsection{Fisch Embryo Test (FET) -- Bewegungen} \label{subsec:FET_Heartbeat}

Die automatisierte Auswertung des FET aus Abschnitt \ref{subsec:FET_Koagulation} soll um die Endpunkte \emph{Herzschlag} und \emph{Spontanbewegung} erweitert werden, da so die Sensitivit\"{a}t der Untersuchung verbessert und eine geringere Falsch-Negativ-Rate erm\"{o}glicht wird.
Das Ziel der vorliegenden Versuche besteht darin, einen Standardablauf f\"{u}r das systematische Testen mit den genannten drei Endpunkten zu entwickeln. Dazu wurden drei Toxine mit bekannten Wirkungen gew\"{a}hlt, um typische Effekte sicher abzubilden. Methanol und Ethanol f\"{u}hren zu koagulierten Eiern, w\"{a}hrend das Arzneimittel Valproat in hohen Konzentrationen Muskeln sowie Nervensystem und folglich Herzschlag sowie Spontanbewegung beeintr\"{a}chtigt. F\"{u}r die erfolgreiche Erarbeitung einer \hts wird wiederum das in \abb \ref{fig:Ablaufdiagramm_Screendesign} vorgestellte Flussdiagramm zur anforderungskonformen Auswertung durchlaufen. Es kann hierbei auf einige Ergebnisse aus Abschnitt \ref{subsec:FET_Koagulation} zur\"{u}ckgegriffen werden. Alle Larven wurden von Experten auf die drei Endpunkte untersucht.
Pro Toxin wurde eine Mikrotiterplatte mit 96 N\"{a}pfchen verwendet, von denen 72 mit Fischlarven belegt wurden. F\"{u}nf Reihen mit je zw\"{o}lf Fischlarven wurde je einer bestimmten Toxinkonzentration ausgesetzt, w\"{a}hrend eine Reihe mit zw\"{o}lf unbehandelten Fischlarven als Negativ-Kontrolle ohne Toxinbehandlung diente. Damit wurde eine Gesamtzahl von 288 Larven untersucht. Der aufgenommene Datensatz und die manuelle Klassifikation bez\"{u}glich der drei Endpunkte sind in Tabelle \ref{tab:Datensatz_Tox_Bewegungen} gegeben.
\begin{table}[htbp]
\footnotesize
\centering
\begin{tabular}
{lrrrrrrccc}
  \toprule
  \multirow{2}{*}{\textbf{\footnotesize Toxin}} &
  \multirow{2}{*}{\textbf{\textbf{Einheit}}} &
  \multicolumn{5}{c}{\multirow{2}{*}{\textbf{Dosis}}}&
  \multicolumn{1}{C{1.5cm}}{\textbf{pr\"{a}sente Koagu\-lation}}&
  \multicolumn{1}{C{1.9cm}}{\textbf{pr\"{a}sente Spontan\-bewegung}}&
\multicolumn{1}{C{2cm}}{ \textbf{pr\"{a}senter Herzschlag}}\\
  \midrule
Methanol& [\%] & 1.67 & 2.27 & 2.5 & 2.75 & 3.75 & 9 &63 & 29\\
Ethanol& [\%] &1.33&1.82&2&2.2&3&16&31&8\\
Valproat I& [mg/l] & 186.7&254.5&280&308&420&6&21&10\\
Valproat II& [mg/l] & 186.7&254.5&280&308&420&17&25&18 \\
  \bottomrule
\end{tabular}
\caption[Toxine, Verd\"{u}nnungsreihen und manuell ermittelte Klassenzugeh\"{o}rigkeiten]{Toxine, Verd\"{u}nnungsreihen und manuell ermittelte Klassenzugeh\"{o}rigkeiten des untersuchten Datensatzes.}\label{tab:Datensatz_Tox_Bewegungen}
\end{table}

\subsubsection{Versuchsplanung: Biologie und Bildakquise}

Der biologische Effekt, \dhe die Endpunkte, sind bereits in der OECD-Norm definiert \cite{OECD06,OECD06a} und die Anforderungen zur Durchf\"{u}hrbarkeit sind wie bei der Klassifikation zur Koagulation erf\"{u}llt (Legitimit\"{a}t, logistische Machbarkeit, biologische Realisierbarkeit, Reproduzierbarkeit, finanzielle Realisierbarkeit).

Zur \"{U}berpr\"{u}fung der Anforderungen an die Messung werden lediglich die Unterschiede zum vorangegangenen Abschnitt gepr\"{u}ft. Das Mikroskop sowie die Pr\"{a}paration und die Zeitpunkte, an denen die Larven dem jeweiligen Toxin exponiert werden, sind identisch. 
Von jedem Ei sind, um die Messung und Detektion von Herzschlag und Spontanbewegung  zu realisieren, Bildzeitreihen notwendig. Der Herzschlag der Zebrab\"{a}rblingslarve hat eine Frequenz von ca. 120 Schl\"{a}gen pro Minute. Da der Herzschlag eine schnelle Bewegung darstellt, m\"{u}ssen hierf\"{u}r Bildzeitreihen mit einer hohen Abtastfrequenz (f $>$ 4\,\hertz) aufgenommen werden. Nach Voruntersuchungen wurde festgestellt, dass die maximale Akquise-Geschwindigkeit des \scanr-Mikroskops von zwei Bildern pro Sekunde (engl. fps - frames per second) ausreicht, um die Pr\"{a}senz des Herzschlags in einer Bildsequenz \"{u}ber zwei Sekunden zu identifizieren. Hierbei auftretende Aliasing-Effekte k\"{o}nnen vernachl\"{a}ssigt werden, weil nur eine dichotome Klassifikation, nicht aber die Bestimmung der Herzfrequenz Ziel der Untersuchung ist.
Die Spontanbewegung der Fischlarven erfolgt nicht konstant wie der Herzschlag. Zuweilen bewegt sich die Larve auch f\"{u}r mehrere Minuten nicht. Daher sind zu deren Analyse Bildzeitreihen mit niedrigen Frequenzen (f~$\ll$~1\,\hertz) notwendig. Eine Analyse der Klassifikationsergebnisse mit Mikroskopaufnahmen, die sich auf lediglich eine Fokusebene beziehen (\vgl \abb \ref{fig:extendedfokus}), zeigt keine signifikante Verschlechterung der Fehlerrate, wenn die ausgewertete Fokusebene weitgehend in der Mitte des Eies der untersuchten Zebrab\"{a}rblingslarve ist. Der Vorteil ist jedoch, dass sich der Datensatz um den Faktor 3 verringert und auf die Bildkombination verzichtet werden kann.

\begin{figure}[!htb]
\centering
        \centering
        \includegraphics[width=\linewidth]{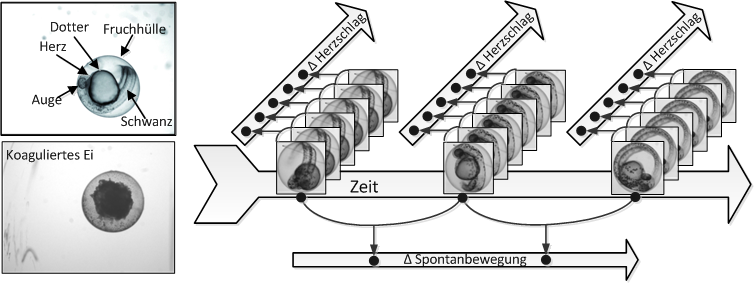}
        \caption[Typische Zebrab\"{a}rblingslarven und Bildzeitreihen f\"{u}r den Herzschlag]{Links: Typische Zebrab\"{a}rblingslarven (oben: entwickelt mit einzelnen Organen, unten: koaguliert), rechts: Bildzeitreihen f\"{u}r den Herzschlag wurden mit 2 Frames pro Sekunde akquiriert; Bildzeitreihen f\"{u}r die Spontanbewegung mit 0.25 Frames pro Minute.}
        \label{fig:Zeitstrahl_Bildaufnahme_AT}
\end{figure}

\begin{figure}[!htb]
\centering
\begin{minipage}[b]{0.45\linewidth}
        \centering
        \includegraphics[width=\linewidth]{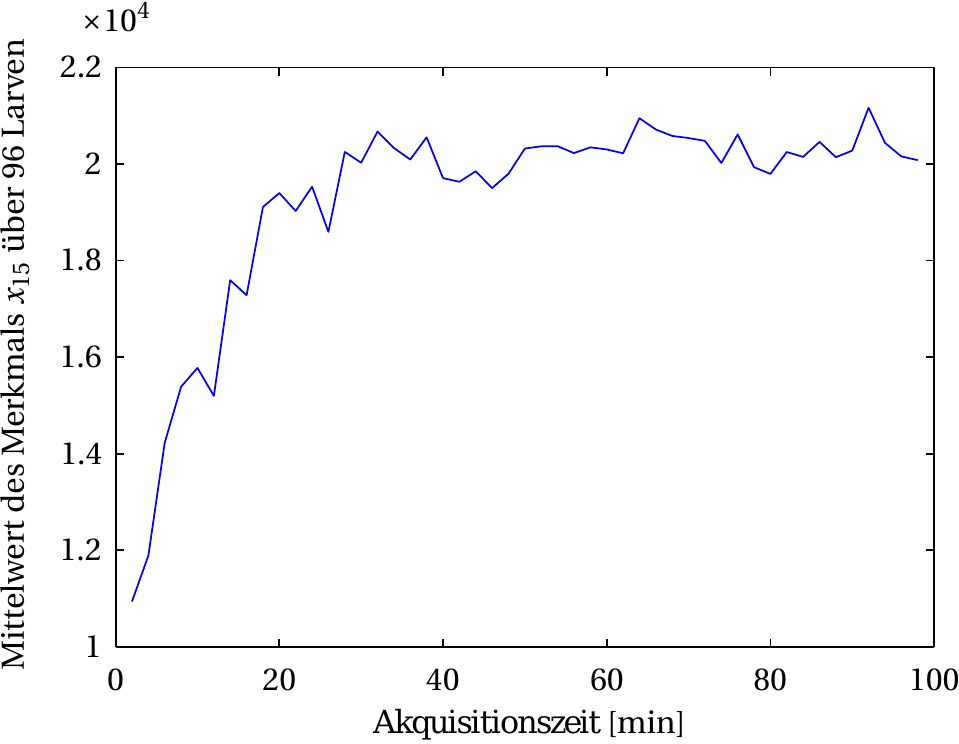}
        \caption[Mittelwert des Merkmals $x_{15}$ ]{Mittelwert des Merkmals $x_{15}$ (\vgl \kap \ref{subsec:Merkmale_Bewegung}) \"{u}ber 96 Larven einer Platte unbehandelter Larven (Kontrollen). Nach etwa $30$\,\minute~erreichen die Larven eine gleichbleibende Bewegungsmenge.}
        \label{fig:Eingewoehnung_Larven}
\end{minipage}
\hfill
   \begin{minipage}[b]{0.45\linewidth}
        \centering
        \includegraphics[width=.9\linewidth]{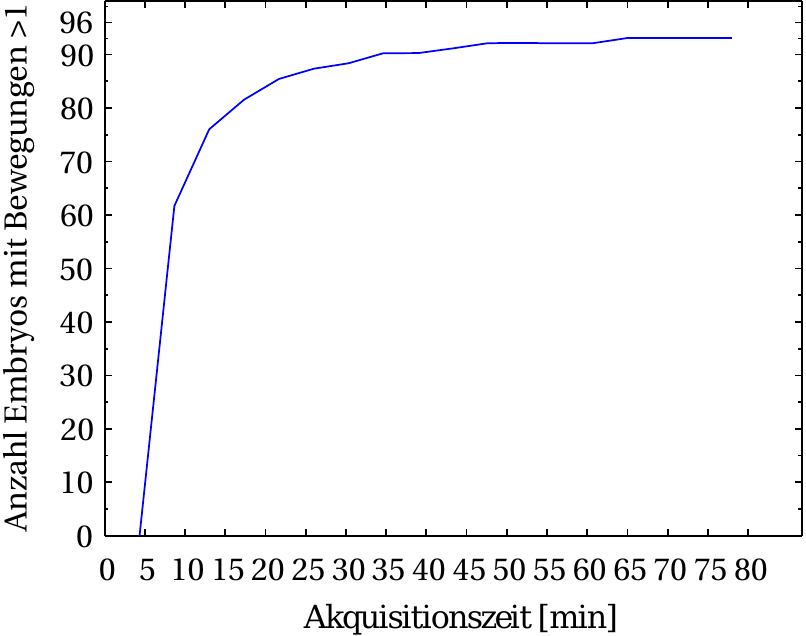}
        \caption[Zeitdauer bis zur ersten Bewegung einer Larve]{Anzahl von insgesamt 96 Larven aus der gleichen Platte wie \abb \ref{fig:Eingewoehnung_Larven}, bei welchen mindestens eine Spontanbewegung auf Basis des Merkmals $x_{15}$ (\vgl \kap \ref{subsec:Merkmale_Bewegung}) identifiziert wurde.}
        \label{fig:Spont_over_Time_Plot}
   \end{minipage}
\end{figure}

F\"{u}r die automatische Bildakquise wurde f\"{u}r die vorliegende Arbeit, um in der Lage zu sein, in nur einem Durchgang sowohl Herzschlag als auch Spontanbewegung aufzuzeichnen, ein zyklischer Akquiseablauf in die Steuerungssoftware des \scanr-Mikroskops programmiert. Damit wird ein Bildstrom nach Formel~(\ref{eq:Bildstrom}) aufgezeichnet. \abb \ref{fig:Zeitstrahl_Bildaufnahme_AT} illustriert die jeweils 5 schnell hintereinander aufgenommenen Bilder zur Herzschlagdetektion sowie die l\"{a}ngeren Pausen zwischen den einzelnen Sequenzen \cite{Alshut11AT}. Der Bildstrom $BS$ vereinfacht sich hier entsprechend, da nur eine Modalit\"{a}t ($I_c=1$) und auch nur eine Fokusebene ($I_z=1$) aufgezeichnet wurde. Dazu nimmt das Mikroskop f\"{u}r jede Larve eine zwei Sekunden lange Bildzeitreihe f\"{u}r den Herzschlag mit maximaler Akquise-Geschwindigkeit auf, was f\"{u}nf Aufnahmen ($n_z=5$) entspricht. Daraufhin setzt das Mikroskop die Aufnahme unmittelbar mit der n\"{a}chsten Larve fort, bis alle N\"{a}pfchen der 96er-Mikrotiterplatte ($n_w=96$) akquiriert wurden. Das Vorgehen wird 30 mal wiederholt ($n_l=30$). Die gro{\ss}e Anzahl von Wiederholungen ist notwendig, da die Larven nach dem Einsetzen in das Mikroskop, m\"{o}glicherweise aufgrund der Ersch\"{u}tterung, sich f\"{u}r eine bestimmte Zeitdauer wenig bewegen. Eine nur kurze Akquisitionszeit hat somit falsch-negativ (engl. \emph{false-negative }) Klassifikationen zur Folge. In \abb \ref{fig:Eingewoehnung_Larven} sind die anf\"{a}nglich verminderten Bewegungen der Larven anhand der mittleren Bewegungsmenge nach Merkmal $x_{15}$ (\vgl \kap \ref{subsec:Merkmale_Bewegung}) von 96 gleichzeitig ausgewerteten Larven unmittelbar nach dem Einsetzen in das Mikroskop gezeigt. Erst nach $20-30$\,\minute~erreicht das Merkmal einen etwa gleichbleibenden Wert, d.h. bis zu diesem Zeitpunkt sind deutlich weniger Bewegungen der Larven vorhanden. In \abb \ref{fig:Spont_over_Time_Plot} ist die Anzahl von Larven derselben Platte, welche mindestens eine Bewegung vollzogen haben \"{u}ber der Akquisitionszeit dargestellt.

Es l\"{a}sst sich ablesen, dass erst nach etwa einer Stunde alle Larven mindestens einmal eine Spontanbewegung durchgef\"{u}hrt haben. Aufgrund dieser Ergebnisse wurden die 30 Wiederholungen ($n_l=30$) festgelegt, was in einer Akquisezeit von \ca zwei Stunden resultiert. Somit ist sichergestellt, dass mindestens eine Bewegung von gesunden Larven erfasst wird. Damit kann die Akquise-Dauer mittels Formel (\ref{eq:Aquisezeit_scanr}) ermittelt werden. \abb~\ref{fig:Zeitstrahl_Bildaufnahme_AT} veranschaulicht den Datensatz anhand der Beschreibung des Bildstroms $BS$ (\vgl Formel (\ref{eq:Bildstrom}))   aus Abschnitt~\ref{sec:Parameter_mathematisch}. Die Zeit f\"{u}r die Akquise eines Versuchs bel\"{a}uft sich auf $t_{RV}$=76\,s. Dabei handelt es sich jedoch nur um einen theoretischen Wert, da die Wartezeiten zwischen den einzelnen Aufnahmen nicht gek\"{u}rzt werden k\"{o}nnen, um sicherzustellen, dass eine entwickelte Larve w\"{a}hrend der Akquise auch mindestens einmal eine Spontanbewegung durchgef\"{u}hrt hat (Eindeutige Pr\"{a}senz der biologisch relevanten Information). Auch durch die l\"{a}ngere Dauer der Akquise sind keine Auswirkungen auf die Larven zu erwarten (R\"{u}ckwirkungsfreiheit der Messung). Jede Mikrotiterplatte ben\"{o}tigt bei der Akquise mittels den in \tab \ref{tab:Bildaquisezeit_TOXII} angegebenen Parametern 123\,min.  Der zu verarbeitende Datensatz w\"{a}chst auf eine Gr\"{o}{\ss}e von fast 400\,MB Speicherplatz pro Modellorganismus und umfasst f\"{u}r den gesamten Datensatz \ca 1\,TB. Der Gr\"{o}{\ss}enunterschied im Vergleich zum Datensatz zur Ermittlung der Koagulation kommt durch den Einfluss der Bildsequenzen und der erw\"{a}hnten langen Akquisezeit zustande.
\begin{table}[htbp]
\begin{tabularx}{\linewidth}
{
>{\raggedright\arraybackslash}p{2cm}
>{\raggedleft\arraybackslash}p{.8cm}
X
}
\toprule
{\bf Parameter}							&
{\bf Wert } & {\textbf{Bemerkung}}							\\
\midrule
$n_l$&30& Wiederholungen\\
$n_w$&96& Standard Mikrotiterplatte mit 96 N\"{a}pfchen\\
$n_z$&5& Anzahl der Frames\\
$t_{m_1}$&0.2&Zeit zur Aufnahme eines Bildes [s]\\
$t_{m_2}$&1.2&Zeit f\"{u}r Bewegung an den Anfang der Mikrotiterplatte [s]\\
$t_{m_3}$&4&Zeit f\"{u}r Bewegung von N\"{a}pfchen zu N\"{a}pfchen [s]\\
$t_{p_1}$&6.2& Zeit zum Vereinzeln einer Larve [s] (manuell)\\
$t_{p_2}$&3.1& Zeit zum Exponieren einer Larve [s] (manuell)\\
\midrule
\midrule
$t_{aq}$&67&Mittels Gleichung (\ref{eq:Aquisezeit_scanr}) ermittelte Gesamtzeit [s] f\"{u}r eine Larve\\
$t_{aq_{96}}$&107&Mittels Gleichung (\ref{eq:Aquisezeit_scanr}) ermittelte Gesamtzeit [min] f\"{u}r eine 96er"~Platte\\
$t_{RV}$&76&Gesamter Zeitaufwand pro Larve [s]\\
$t_{RV_{96}}$&123&Gesamter Zeitaufwand pro 96er"~Platte [min]\\

\bottomrule
\end{tabularx}
\caption{Ermittlung der notwendigen Zeit der Versuchsdurchf\"{u}hrung f\"{u}r eine Mikrotiterplatte}\label{tab:Bildaquisezeit_TOXII}
\end{table}

Bei dem Zeitfenster zur Akquise von \ca 4h pro Tag lassen sich somit lediglich 2 Platten oder 192 Larven an einem Mikroskop akquirieren. Bei einer gro{\ss}en Anzahl zu untersuchender Substanzen m\"{u}ssen somit mehrere Mikroskope parallel betrieben werden. Der vorliegende Datensatz wurde auf nur einem Mikroskop aufgezeichnet, da die Erfahrung zeigt, dass die Verwendung mehrerer Mikroskope zu inkonsistenten Datens\"{a}tzen f\"{u}hren kann. Um die Vergleichbarkeit inhomogener Datens\"{a}tze sicherzustellen, muss der Einfluss der Mikroskope mittels spezieller Normalisierung beseitigt werden. F\"{u}r solche F\"{a}lle wurde das Modul in Abschnitt \ref{subsec:Modulkatalog_Normalisierung} entwickelt, welches auch dazu dienen kann, Datens\"{a}tze zwischen einzelnen Laboren vergleichbar zu machen. Die Leistungsf\"{a}higkeit wurde in \cite{Reischl10} gezeigt. Auch wenn das Modul in den Datens\"{a}tzen der vorliegenden Arbeit nicht zum Einsatz kommt, ist es f\"{u}r die Skalierbarkeit der Untersuchung wichtig und daher in den Modulkatalog aufgenommen. Die vier Mikrotiterplatten des betrachteten Datensatzes lassen sich an einem Mikroskop innerhalb von minimal zwei Tagen akquirieren (Schnelligkeit der Messung).

\subsubsection{Versuchsplanung: Analyse und Interpretation}

Die Parameter der \hts sind bis auf die Planfaktoren $\mathbf{Y}$, die Anzahl von akquirierten Frames ($t_{\mathbf{BS}}$) und der Akquise-Frequenz ($f_\mathbf{BS}$) identisch mit den Parametern der \hts f\"{u}r Koagulation. Die beiden neuen Planfaktoren \emph{Herzschlag} und \emph{Spontanbewegung} sollen automatisch, anhand aus dem Bildstrom $\mathbf{BS}$ extrahierter Merkmale $\mathbf{x}$, gesch\"{a}tzt werden. Zum Nachweis der Leistungsf\"{a}higkeit der automatischen Evaluation sind alle Aufnahmen der Larven von Experten einer der Klassen \emph{\{ja, nein, fehlerhaft\}} zugeordnet worden. Zwei Mikrotiterplatten (Methanol und Valproat~I) wurden  zum Anlernen der Algorithmen und die verbleibenden (Ethanol und Valproat~II) zum Validieren der Algorithmen verwendet. Die Parameter der \hts sind in \tab \ref{tab:Anw_Parameter_TOXII} zusammengefasst. Der Unterschied zu Tabelle \ref{tab:Anw_Parameter_TOXI} besteht in den Parametern zur Aufnahme der beschriebenen Bildsequenz mit entsprechenden Ab\-tast\-raten und Wiederholungen.
\begin{table}[htb]
\footnotesize
\begin{tabularx}{\linewidth}{p{2cm}Zp{2.2cm}Zcccc}
\toprule
$y/\hat{y}$	&$p_y$	&\multicolumn{1}{c}{$z/\hat{z}$}	&$p_z$	 &\multicolumn{4}{c}{$p_\mathbf{BS}$}  \\
\cmidrule(l){5-8}
	&	&	&	&$m_\mathbf{BS}$	&${u}_\mathbf{BS}$	&$t_\mathbf{BS}$	 &$f_\mathbf{BS}$	 \\
\midrule
Koagu\-la\-tion Herzschlag Spontanbew.	&Konzen\-tration	&Toxin\newline Position\newline Mikroskop\newline Wiederholung\newline Aufnahmedatum\newline Aufnahmezeit	 &$96$er-Mikro\-titerplatte	 &$3$	 &$1334x1024x3$	&$5$&$0.0042$		 \\
\bottomrule
\end{tabularx}
\caption[Parameter der \hts f\"{u}r drei Endpunkte nach Abschnitt \ref{sec:Parameter_mathematisch}]{Parameter der \hts f\"{u}r drei Endpunkte nach Abschnitt~\ref{sec:Parameter_mathematisch}; mit $m_{\mathbf{BS}}$ Anzahl Kan\"{a}le, ${u}_\mathbf{BS}$ Anzahl Pixel, $t_\mathbf{BS}$ Anzahl Frames und $f_\mathbf{BS}$ Akquise-Frequenz~[Hz]}\label{tab:Anw_Parameter_TOXII}
\end{table}

\begin{figure}[!htbp]
    \footnotesize
    \centering
        \centering
        \includegraphics[page=14,width=\linewidth]{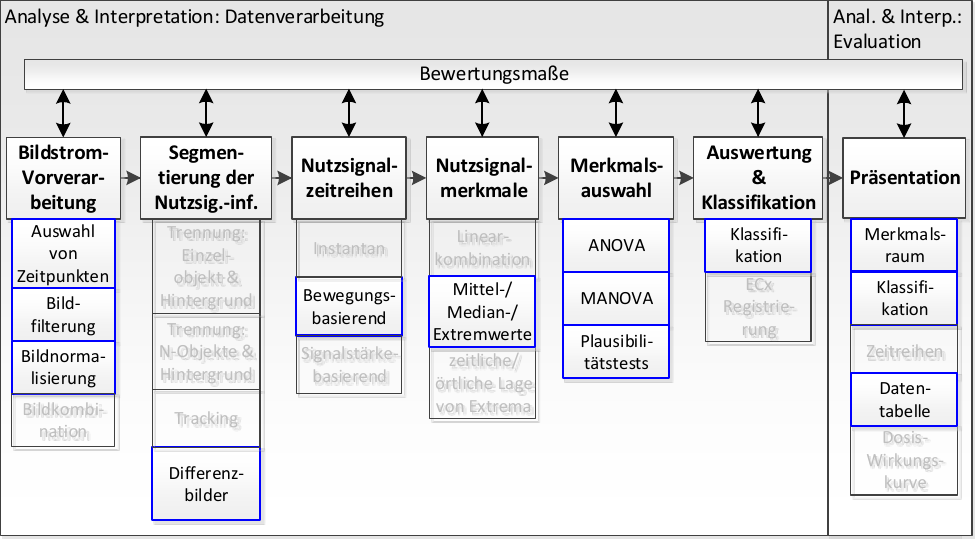}
        \caption[Module zur Auswertung von Herzschlag und Spontanbewegung]{Ausgew\"{a}hlte Module zur Auswertung von Herzschlag und Spontanbewegung}
        \label{fig:Module_TOXII}
\end{figure}

Die f\"{u}r die Auswertung der Endpunkte \emph{Herzschlag} und \emph{Spontanbewegung} zusammengeschalteten Module sind in \abb \ref{fig:Module_TOXII} hervorgehoben. Die Auswertung des Endpunktes \emph{Koagulation} erfolgt analog zum vorangegangenen Abschnitt, daher wird im Folgenden hierauf nicht weiter eingegangen.

\textbf{Bildstrom-Vorverarbeitung:}
Die Bildstrom-Vorverarbeitung verl\"{a}uft identisch wie bei der \hts zur Koagulation, jedoch kann, da nur eine Fokusebene akquiriert wurde, auf die Bildkombination f\"{u}r den Extended Focus verzichtet werden. F\"{u}r die Auswertung der Spontanbewegung ist es ausreichend, lediglich ein Bild jeder Kurzzeitreihe zu verwenden. Daher wurde das Modul zur zeitlichen Unterscheidung hinzugef\"{u}gt, was die notwendige Rechenzeit um den Faktor f\"{u}nf verringert.

\begin{figure}[!htbp]
    \footnotesize
    \centering
\subfloat[Originalbild
        ]{\includegraphics[width=.25\linewidth]{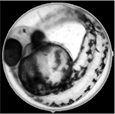}
        \label{fig:Herzschlag_Segmentierung_AT_a}}
\subfloat[Differenzbild
        ]{\includegraphics[width=.25\linewidth]{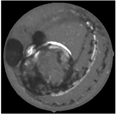}
        \label{fig:Herzschlag_Segmentierung_AT_b}}
\subfloat[Bin\"{a}rbild (Schwellenwert \"{u}ber Differenzbild)
        ]{\includegraphics[width=.25\linewidth]{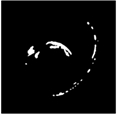}
        \label{fig:Herzschlag_Segmentierung_AT_c}}
\subfloat[\"{U}berlagerung (Bin\"{a}rbild mit Originalbild)
        ]{\includegraphics[width=.25\linewidth]{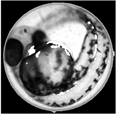}
        \label{fig:Herzschlag_Segmentierung_AT_d}}
        \caption[Quantifizierung der Bewegung im Bild]{Quantifizierung der Bewegung im Bild, von links nach rechts: Originalbild; Differenzbild aus der Bildzeitreihe f\"{u}r Herzschlag; Differenzbild nach Schwellenwertoperation; \"{U}berlagerung Originalbild und Differenzbild f\"{u}r Herzschlag. }
        \label{fig:Herzschlag_Segmentierung_AT}
\end{figure}

\textbf{Segmentierung der Nutzsignalinformation:}
F\"{u}r beide Endpunkte ist lediglich die Bewegung im Bild entscheidend. Da nur eine Larve pro Bild akquiriert wird und davon ausgegangen werden kann, dass sich Hintergrundobjekte und Fremdk\"{o}rper nicht bewegen, kann auf eine Trennung von Einzelobjekt und Hintergrund sowie ein Zuschneiden verzichtet werden. Zudem w\"{u}rde beides eine Registrierung der zugeschnittenen Bilder erforderlich machen, was rechenintensiv ist. Weiter kann gezeigt werden, dass sich sowohl die Spontanbewegung als auch der Herzschlag durch die Ver\"{a}nderung der Pixel \"{u}ber die Zeit widerspiegelt, was in \abb \ref{fig:Herzschlag_Segmentierung_AT} dargestellt ist. Die Abbildung illustriert den Einfluss der Bewegungen des Herzschlags und des Blutflusses. Bildbereiche mit einem hohen Anteil an Bewegung erscheinen im Differenzbild (\abb~\ref{fig:Herzschlag_Segmentierung_AT_b}) als helle Regionen, welche sich durch einen Schwellenwert vom Hintergrund trennen lassen (\abb~\ref{fig:Herzschlag_Segmentierung_AT_c}). \"{U}berlagert man schlie{\ss}lich die segmentierten Bildbereiche mit dem Originalbild ergibt sich \abb~\ref{fig:Herzschlag_Segmentierung_AT_d}.

Zur Segmentierung des Bildstroms und f\"{u}r die Merkmalsextraktion von Spontanbewegung und Herzschlag werden lediglich die Merkmalsbilder \"{u}ber absolute Differenzen im Bild errechnet und abgespeichert. Der segmentierte Bildstrom besteht somit aus Kurzzeitreihen von je vier Differenzbildern f\"{u}r den Herzschlag und 29 Differenzbildern f\"{u}r die Spontanbewegung. Die Differenzbilder der Kurzzeitreihen wurden \"{u}ber die mit hoher Frequenz aufgezeichneten Bilder errechnet und die f\"{u}r Spontanbewegung \"{u}ber die mit niedriger Frequenz aufgezeichneten Bilder.

\textbf{Nutzsignalzeitreihen:}
Aus jedem Differenzbild werden Merkmale extrahiert. Hierdurch entsteht eine Zeitreihe f\"{u}r den Herzschlag und eine Zeitreihe f\"{u}r die Spontanbewegung. Es wurden die Merkmale $x_9-x_{15}$ extrahiert, wie sie im Modulkatalog im Abschnitt \ref{subsec:Merkmale_Bewegung} beschrieben sind.

%
%
%
%
%

\textbf{Dimensionsreduktion:}
Der Merkmalssatz wird durch Anwendung von Maximum- (MAX), Median- (MEDIAN) und Mittelwertoperatoren (MEAN) \"{u}ber die Abtastzeitpunkte gebildet. Die Operatoren MAX extrahieren die maximale Bewegung \"{u}ber der Zeit, was starke Bewegungs\"{a}nderungen der Larve widerspiegelt. Der Median und der Mittelwert geben Auskunft \"{u}ber die Menge der Bewegung in der gesamten Zeitreihe.

\textbf{Merkmalsauswahl:}
Anhand der Trainingsdaten wurde das trennst\"{a}rkste Merkmal mittels ANOVA ermittelt und mittels MANOVA die beste 2er-Kombination bestimmt. In Tabelle \ref{tab:Merkmalsrelevanzen_ToxII_spont} und Tabelle \ref{tab:Merkmalsrelevanzen_ToxII_herz} sind die ermittelten Merkmalsrelevanzen f\"{u}r jedes alle Merkmale aufgelistet.

  Die Kombination der Merkmale \emph{MEAN ZR $x_{15}$ }und\emph{ MAX ZR $x_{12}$ }liefert f\"{u}r die Klassifikation der Spontanbewegung die beste G\"{u}te (\vgl \tab \ref{tab:Merkmalsrelevanzen_ToxII_spont}). Das erstgenannte Merkmal quantisiert den Durchschnitt der maximalen Bewegung. Das Merkmal weist, f\"{u}r starke Bewegungen, die \"{u}ber die gesamte Zeitreihe auftreten, hohe Werte auf. Besonders bei gesunden, aktiven Larven tendiert das Merkmal daher zu hohen Werten. Das zweite Merkmal quantifiziert das Maximum signifikanter Bewegungen. Da der Maximalwert extrahiert wird, toleriert es somit auch Ruhephasen ohne Bewegungen, wenn die Larven einige Zeit still liegen bleiben. Eine tote oder beeintr\"{a}chtigte Fischlarve bewegt sich gar nicht oder nur schwach, was in beiden Merkmalen durch niedrige Werte angezeigt wird.

Die Merkmalsrelevanzen der ANOVA und MANOVA, welche den Herzschlag beschreiben, sind hingegen relativ niedrig (vgl. \tab \ref{tab:Merkmalsrelevanzen_ToxII_herz}). Das beste Ergebnis ergibt sich bei Merkmalen, welche nicht eine einzelne Bewegung im Gesamtbild, sondern eine \"{A}nderung beschreiben, welche \"{u}ber die gesamte Zeitreihe konstant hoch ist. Das ist verst\"{a}ndlich, da der Herzschlag im Vergleich zur Spontanbewegung ein eher lokales Ph\"{a}nomen ist, das wenig globale Pixel\"{a}nderungen hervorruft jedoch ohne Pause auftritt. Die extrahierten Merkmale beschreiben somit die \NutzsigInf und weisen eine ausreichende Trennsch\"{a}rfe auf (Quantifizierbarkeit des biologischen Effekts im Bild).
\begin{table}[!htbp]

    \begin{minipage}[b]{0.48\linewidth}
        \footnotesize
\begin{tabularx}{\linewidth}{lVV}
\toprule
\textbf{Merkmal} & \textbf{G\"{u}te-MANOVA} &\textbf{G\"{u}te- ANOVA}\\
\midrule
 MEAN ZR $x_{15}    $& ----- &0.879   \\
 MAX ZR $x_{12}     $& 0.895& 0.789 \\
 MAX ZR $x_{13}     $& 0.894& 0.801 \\
MAX ZR $x_{9}     $& 0.893& 0.859 \\
 MEAN ZR $x_{13}    $& 0.887& 0.834 \\
 MEAN ZR $x_{12}    $& 0.887& 0.829 \\
MAX ZR $x_{10}     $& 0.886& 0.771 \\
 MEDIAN ZR $x_{13}  $& 0.885& 0.819\\
 MEDIAN ZR $x_{12}  $& 0.885& 0.813 \\
MAX ZR $x_{14}     $& 0.883  & 0.721   \\
 MAX ZR $x_{15}     $& 0.882  & 0.730 \\
 MEAN ZR $x_{9}    $& 0.882& 0.820  \\
 MEDIAN ZR $x_{9}  $& 0.881 & 0.801    \\
 MEAN ZR $x_{14}    $& 0.881& 0.878  \\
 MEAN ZR $x_{10}    $& 0.880& 0.769    \\
 MEDIAN ZR $x_{10}  $& 0.880 & 0.746      \\
 MEDIAN ZR $x_{14}  $& 0.879& 0.873  \\
 MEDIAN ZR $x_{15}  $& 0.879& 0.874 \\
 MAX ZR $x_{11}     $& 0.879& 0.777\\
 MEDIAN ZR $x_{11}  $& 0.879& 0.749  \\
 MEAN ZR $x_{11}    $& 0.879  & 0.772       \\
\bottomrule
\end{tabularx}
      \vspace{1cm}
    \captionof{table}{Merkmalsrelevanzen f\"{u}r Spontanbewegung -- Univariate und  Multivariate Analyse (bei Auswahl von 2 Merkmalen)\label{tab:Merkmalsrelevanzen_ToxII_spont}}
    \end{minipage}
    \hfill
        \begin{minipage}[b]{0.48\linewidth}
    \footnotesize
        \centering
\begin{tabularx}{\linewidth}{lVV}
\toprule
\textbf{Merkmal} & \textbf{G\"{u}te-MANOVA} &\textbf{G\"{u}te- ANOVA}\\
\midrule
 MEAN ZR $x_{14}     $& ----- &0.342  \\
MEDIAN ZR $x_{9}   $& 0.577 & 0.016   \\
MEAN ZR $x_{9}     $& 0.542 & 0.032 \\
 MEAN ZR $x_{13}     $& 0.489 & 0.015    \\
 MEAN ZR $x_{12}     $& 0.478 & 0.033    \\
 MEDIAN ZR $x_{13}   $& 0.476 & 0.004 \\
 MEDIAN ZR $x_{12}   $& 0.471  & 0.057 \\
MEAN ZR $x_{10}     $& 0.464& 0.017  \\
MEDIAN ZR $x_{10}   $& 0.461& 0.007     \\
 MAX ZR $x_{12}      $& 0.447& 0.000   \\
 MAX ZR $x_{13}      $& 0.440 & 0.033 \\
 MAX ZR $x_{10}      $& 0.432 & 0.031     \\
 MAX ZR $x_{9}      $& 0.423 &  0.036 \\
 MAX ZR $x_{14}      $& 0.344 & 0.261 \\
 MAX ZR $x_{15}      $& 0.344 & 0.261 \\
 MEDIAN ZR $x_{14}   $& 0.343& 0.297   \\
 MEDIAN ZR $x_{15}   $& 0.343 & 0.297 \\
 MEAN ZR $x_{15}     $& 0.342 & 0.342  \\
 MAX ZR $x_{11}      $& 0.342 &  0.033 \\
 MEDIAN ZR $x_{11}   $& 0.342& 0.009   \\
 MEAN ZR $x_{11}     $& 0.342 & 0.020 \\
\bottomrule
\end{tabularx}
      \vspace{1cm}
    \captionof{table}{Merkmalsrelevanzen f\"{u}r Herzschlag -- Univariate und  Multivariate Analyse (bei Auswahl von 2 Merkmalen)\label{tab:Merkmalsrelevanzen_ToxII_herz}}
  \end{minipage}

\end{table}

\begin{figure}[!htbp]

        \centering
        \includegraphics[width=\linewidth]{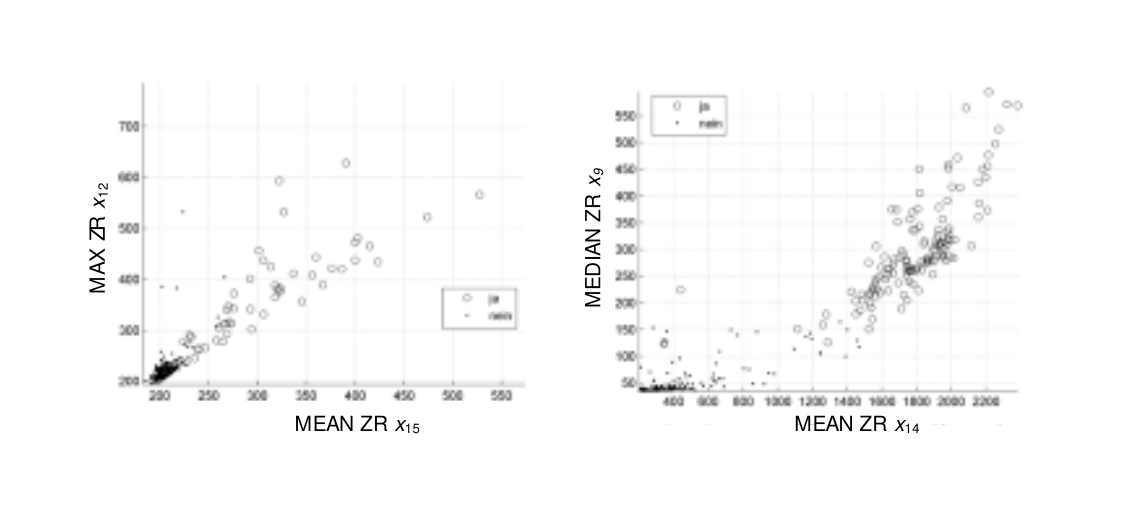}
        \caption[Scatterplot der besten Merkmale nach MANOVA]{Scatterplot der besten Merkmale nach MANOVA; Herzschlag (links) und Spontanbewegung (rechts) \cite{Alshut11AT} }
        \label{fig:Fig8}
\end{figure}

\textbf{Auswertung \& Klassifikation:}
Der Klassifikator der Spontanbewegung zeigte 1.92\% Fehler auf den Lerndaten (156 Beispiele) und 5.52\% Fehler auf den Testdaten (145 Beispiele). Wie in \abb \ref{fig:Fig8} zu erkennen ist, \"{u}berlappen sich die Klassen, was zu weiteren Klassifikationsfehlern f\"{u}hrt. Die \"{U}berlappung ist auf den Einfluss der Toxine zur\"{u}ckzuf\"{u}hren, welche die Nerven und Muskelaktivit\"{a}ten nicht schlagartig, sondern stufenweise ausschalten. Eine Diskussion mit dem Experten ergab, dass bereits bei der manuellen Klassifikation viele Zuordnungen nicht scharf vornehmbar waren. Der Grund hierf\"{u}r ist, dass es im \"{U}bergangsbereich stark vom subjektiven Eindruck des Experten abh\"{a}ngig ist, ob eine leichte Bewegung noch als Herzschlag gewertet wird oder nicht.

Die Klassifikation des Herzschlags weist 5.0\% Fehler \"{u}ber den Lerndaten (140 Beispiele) und 9.85\% Fehler \"{u}ber den Testdaten (132 Beispiele) auf. Die Herzschlagdetektion ist sensibler f\"{u}r Fehlklassifikationen, da die bewegte Stelle sehr klein ist. Ebenso ist der \"{U}bergang von einem sichtbaren Herzen zu einem verdeckten flie{\ss}end, was wiederum zu Fehlklassifikationen f\"{u}hren kann. Des Weiteren bewirken die Toxine oftmals eine stark herabgesetzte Funktion oder Frequenz des Herzens, was ebenfalls zu Fehlklassifikationen f\"{u}hrt, da ein Experte \"{u}blicherweise nach einem kontinuierlich schlagenden Herzen sucht, w\"{a}hrend der Klassifikator selbst einen einzelnen Schlag abbildet.
Die Scatterplots in \abb \ref{fig:Fig8} zeigen die Verteilung aller Datenpunkte. Wenige Bewegungen in einem Bild f\"{u}hren zu niedrigen Merkmalswerten; daher befinden sich die Datenpunkte toter oder beeintr\"{a}chtigter Fischlarven bei niedrigen Werten. Hohe Bewegung, wie es \zB bei vorhandenem Herzschlag bzw. Spontanbewegung der Fall ist, hat demnach gr\"{o}{\ss}ere Werte und h\"{o}here Varianzen  zur Folge. Deutlich zu erkennen ist der "`flie{\ss}ende"' \"{U}bergang zwischen den Einzelversuchen der Klassen (in  Abbildung \ref{fig:Fig8} durch Punkte bzw. Kreise dargestellt) vor allem beim Herzschlag. Insgesamt liefert die Klassifikationsroutine eine ausreichende Genauigkeit, da sich alle toxikologischen Gr\"{o}{\ss}en von Interesse aus den Ergebnissen berechnen lassen (niedrige Fehlerrate bei der Detektion des biologischen Effekts).

\begin{figure}[htbp]
        \centering
        \includegraphics[width=.85\linewidth]{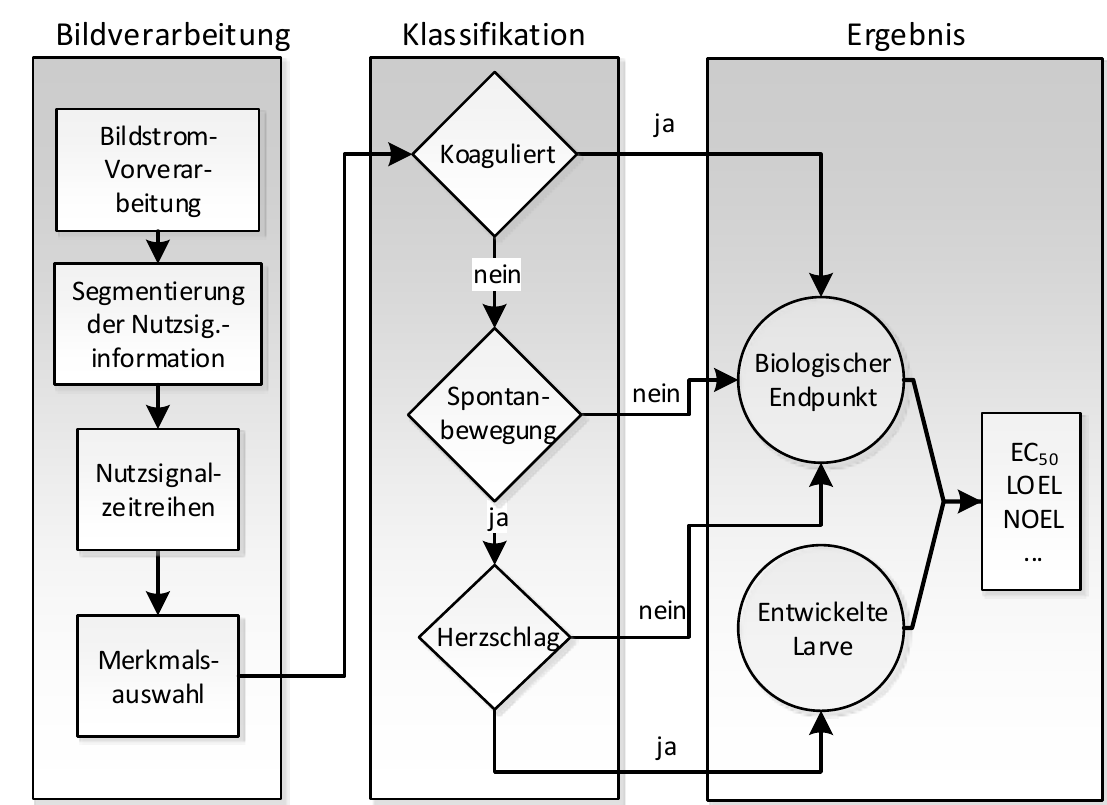}
        \caption[Klassifikatorkaskade \"{u}ber drei Endpunkte]{Klassifikatorkaskade \"{u}ber drei Endpunkte \cite{Alshut11AT}}
        \label{fig:Klassifikator_Kaskade}
\end{figure}

Alle Klassifikatoren wurden im Anschluss zu einer Kaskade kombiniert (vgl.  \abb~\ref{fig:Klassifikator_Kaskade}). Im ersten Schritt erfolgt die Bildverarbeitung, wie sie in den vorangegangenen Abschnitten beschrieben ist, bis hin zur Merkmalsauswahl. Mithilfe der Merkmale werden die Klassifikatoren angelernt. Jede unbekannte Larve, \dhe nicht gelabelte Versuchseinheit, durchl\"{a}uft nun die abgebildete Kaskade. Im ersten Schritt der Klassifikation wird, mit Hilfe der ermittelten Merkmalswerte, auf Koaguliertheit gepr\"{u}ft. Alle nicht koagulierten Larven werden dann bez\"{u}glich Spontanbewegung und im Anschluss daran auf Herzschlag gepr\"{u}ft. Nur wenn keiner der Endpunkte erreicht wird, wird die Larve als \emph{entwickelt} eingestuft. Alle anderen Larven werden einem biologischen Endpunkt zugeordnet. Aus den Zuordnungen lassen sich daraufhin alle statistischen Werte von Interesse berechnen.

\textbf{Pr\"{a}sentation:} Die Ergebnisse werden identisch zum Endpunkt \emph{Koagulation} dargestellt (Pr\"{a}sentierbarkeit der Auswertung) und archiviert (Wissenschaftliche Archivierung).

\subsubsection{Durchf\"{u}hrung und Resultat}
Die vorgestellte neue Strategie zur Bildakquise gestattet es, alle drei Endpunkte anhand eines einzigen Datensatzes zu pr\"{u}fen. Die durch den nur sporadisch auftretenden biologischen Effekt verursachte lange Akquise-Dauer und der demzufolge gro{\ss}e Datensatz von \ca 1\,TB Rohdaten l\"{a}sst die Auswertezeit pro Platte ansteigen. Um bei Verwendung mehrerer Mikroskope die Hochdurchsatzf\"{a}higkeit zu sichern, wurden die Algorithmen f\"{u}r die vorliegende Arbeit so implementiert (\vgl Kapitel \ref{chap:Implementierung}), dass ein verteiltes Rechnen m\"{o}glich ist (Schnelligkeit der Auswertung).

Zum Nachweis der Leistungsf\"{a}higkeit der Klassifikatorkaskade wurden die drei Endpunkte mit den manuell gelabelten Daten bestimmt und mit den gesch\"{a}tzten Ergebnissen f\"{u}r einen Endpunkt und der Kaskade \"{u}ber  alle drei Endpunkte verglichen (vgl. \tab \ref{tab:vgl_1_3_Endpkt}).
\begin{table}[!htbp]
\begin{centering}
\begin{tabular}{
cR{4cm}c
}
\toprule
&\textbf{Nur Koaguliertheit}&\textbf{3 Endpunkte}\\
\midrule
Datensatz I (\tab \ref{tab:Datensatz_Tox_LC50}) & $99$\,\% ($98,8$\,\%)& --- \\
Datensatz II (\tab \ref{tab:Datensatz_Tox_Bewegungen}) & $42,3$\,\% ($41,1$\,\%) & $95,8$\,\% ($82,2$\,\%)  \\
\bottomrule
\end{tabular}
\caption[Vergleich der Klassifikationsergebnisse auf Basis von einem und von drei Endpunkten.]{Vergleich der Klassifikationsergebnisse auf Basis von einem und von drei Endpunkten. Jeweils Ergebnisse \"{u}ber Lerndaten und in Klammern \"{u}ber Testdaten \bzw bei Datensatz I das Ergebnis aus der 5-fach Crossvalidierung. (Die manuell gelabelten Werte in Datensatz II entsprechen dem Ergebnis der gesamten Klassifikatorkaskade)}
\label{tab:vgl_1_3_Endpkt}
\end{centering}
\end{table}

Wirken die Toxine vor allem auf die Bewegungen \bzw Muskeln der Larven, ist die Falsch-Positiv-Rate, trotz der guten Klassifikationsg\"{u}te f\"{u}r den Endpunkt Koagulation, hoch, da sich dies im Endpunkt \emph{Koagulation} nicht widerspiegelt. Solche Larven weisen keinerlei Merkmale der Koagulation auf. Die kombinierte Klassifikation der drei Endpunkte hingegen zeigt eine gute \"{U}bereinstimmung (< 80\% auf Testdaten) mit den manuell gelabelten Daten. Die Nachteile sind jedoch der deutlich gr\"{o}{\ss}ere Datensatz sowie die lange Dauer der Akquise und der Auswertung. Die Zeit zur Aufnahme einer einzigen Platte $t_{RV_{96}}$ beispielsweise  betr\"{a}gt 123\,min im Vergleich zu lediglich 18\,min in Abschnitt \ref{subsec:FET_Koagulation} (vgl. auch \tab \ref{tab:Bildaquisezeit} und \tab \ref{tab:Bildaquisezeit_TOXII}).

\section{Photomotor Response (\pmr)}\label{sec:Anwendung_PMR}

\subsection{\"{U}bersicht}
Der Photomotor Response (\pmr) ist ein biologischer Versuch mit Zebrab\"{a}rblingslarven, welcher im Hochdurchsatz zur Anwendung kommt und erstmals von \emph{Kokel et al.} vorgestellt wurde \cite{Kokel10}. Die Arbeit stellt in Umfang und Durchsatz die bisher gr\"{o}{\ss}te ver\"{o}ffentlichte \hts mit Zebrab\"{a}rblingen dar. Dennoch wird in der Arbeit erhebliches Potenzial, gerade in Bezug auf die Datenextraktion und Bildverarbeitung, nicht genutzt. Auch ist das Verh\"{a}ltnis von \Nutzsig zu Hintergrundrauschen aufgrund verschiedener Fehlereinfl\"{u}sse wie Beleuchtung und Kontrast hoch. Die Anwendung bietet somit ein optimales Szenario, die in Kapitel \ref{chap:Neues_Konzept} vorgestellte Methodik anzuwenden und deren Potenzial an einem realen Beispiel zu belegen. In einer Projektarbeit zwischen dem Institut f\"{u}r Angewandte Informatik und den \og Autoren von der Harvard Medical School wurde das ausgearbeitete Konzept auf die Auswertung des Bildstroms angewandt und implementiert. Die Vorgehensweise und Ergebnisse sind im Folgenden dargestellt. Zu Beginn wird der Hintergrund des Versuchs zum besseren Verst\"{a}ndnis skizziert, daraufhin die vorgestellte Methodik angewandt und schlie{\ss}lich deren Ergebnisse vorgestellt.

\subsection{\pmr: Hintergrund, Rahmenbedingungen und Schwachpunkte}
\begin{figure}[h!tbp]
\centering
    \begin{minipage}[t]{0.52\linewidth}
        \centering
        \includegraphics[height=6cm]{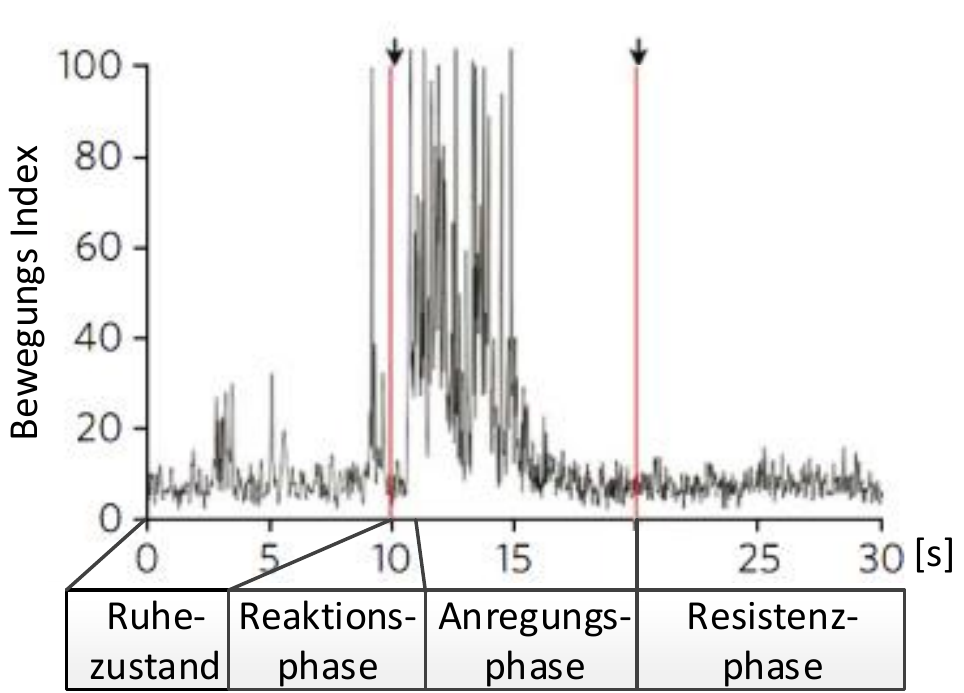}
        \caption[Repr\"{a}sentatives Beispiel f\"{u}r die aggregierte  Bewegung w\"{a}hrend eines \mbox{\emph{PMR}"~Ver\-suchs} \"{u}ber der Zeit.]{Repr\"{a}sentatives Beispiel f\"{u}r die aggregierte  Bewegung w\"{a}hrend eines \mbox{\emph{PMR}"~Ver\-suchs} \"{u}ber der Zeit. Die Pfeile markieren die Zeitpunkte der Lichtstimuli (in Anlehnung an \cite{Kokel10})}
        \label{fig:PMR_Urspruenglicher_Plot}
    \end{minipage}
\hfill
    \begin{minipage}[t]{0.4\linewidth}
        \centering
        \includegraphics[page=1,height=5.5cm]{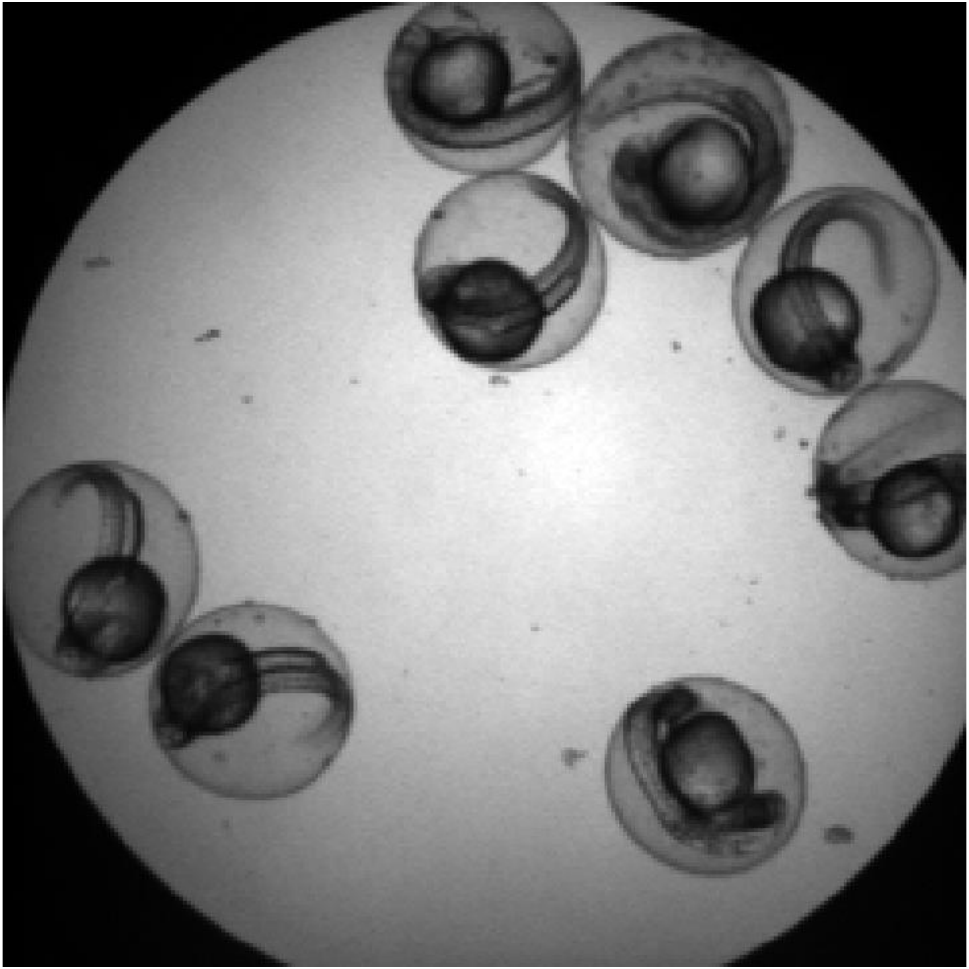}
        \caption[Einzelframe einer Bildsequenz aus einem \emph{PMR}"~Versuch.]{Einzelframe einer Bildsequenz aus einem \emph{PMR}"~Versuch. Acht Larven befinden sich in einem N\"{a}pfchen einer 96er-Mikro\-titer\-platte~\cite{Kokel10}.}
        \label{fig:PMR_Urspruenglicher_Plot_2}
        \end{minipage}
\end{figure}

Der \pmr nutzt ein stereotypes Verhalten von gesunden Zebrab\"{a}rblingslarven, welches durch einen intensiven Lichtstimulus ausgel\"{o}st werden kann. Zebrab\"{a}rblingslarven k\"{o}nnen sich bereits in einem fr\"{u}hen Entwicklungsstadium, mit einem Alter von ca.~28\,hpf, innerhalb der Eih\"{u}lle bewegen. Werden die Larven nun einem starken Lichtimpuls ausgesetzt, so antworten sie mit einem typischen \emph{Zappeln}. Alle Bewegungen der Larven werden in \cite{Kokel10} durch Vergleich von f\"{u}nf Bildzeilen eines jeden Frames gemessen. Ein repr\"{a}sentatives Beispiel einer solchen Bewegungsaufzeichnung \"{u}ber alle Larven der Bildsequenz ist in \abb \ref{fig:PMR_Urspruenglicher_Plot} dargestellt. Der gesamte Versuch wird als Photomotor Response (\pmr) bezeichnet und l\"{a}sst sich grob in folgende vier Phasen einteilen: Die erste Phase (Ruhezustand) des \pmr\ ist das nat\"{u}rliche Verhalten bzw. die \sog \emph{Spontanbewegungen} der Larven ohne jeglichen Einfluss von au{\ss}en. Entsprechend bewegen sich die Larven mit der typischen H\"{a}ufigkeit wie es bei ca. 28\,hpf \"{u}blich ist~\cite{Westerfield93}. Die Bewegungen treten zuf\"{a}llig auf und werden in \cite{Kokel10} als Hintergrundsignal aufgefasst. Die erste Phase ist in \abb \ref{fig:PMR_Urspruenglicher_Plot} der Bereich von 0 bis 10 Sekunden des Versuchs. Die typischen Spontanbewegungen sind die Ausschl\"{a}ge im Bewegungsindex. Die Pfeile markieren jeweils den Zeitpunkt eines Lichtstimulus. In einem typischen \emph{PMR}"~Versuch werden zwei Lichtstimuli verwendet. Die zweite Phase kann als Reaktionszeit oder Latenzzeit bezeichnet werden und wird durch den Einsatz des Lichtstimulus nach 10 Sekunden sowie dem Beginn starker Bewegungen der Larven bei \ca 11 Sekunden abgegrenzt. Es ist somit die Ruhephase zwischen dem ersten Stimulus und der Reaktion. Die dritte Phase ist die Bewegungsphase oder Anregungszeit. Die Phase zeichnet sich durch starke Zappelbewegungen der Larven aus und endet mit dem Abklingen der starken Bewegungen bei \ca 20 Sekunden. Es werden zwei typische Bewegungsereignisse unterschieden: Das \sog "`Coiling"' und das "`Swimming"'.
Die Begriffe stehen f\"{u}r unterschiedliche charakteristische Bewegungen, welche die Larve typischerweise nach einem Lichtimpuls durchf\"{u}hrt. Coilings sind starke Drehungen der Larve und werden als ein instinktives "`Umdrehen"' der Fische in eine zuf\"{a}llige Richtung interpretiert. Swimmings sind Bewegungen, bei denen die Larve, ohne gro{\ss}e Drehungen auszuf\"{u}hren, sich auf der Stelle mit geringen Amplituden bewegt. Ein Swimming kann auch als eine Art "`Zittern"' beschrieben werden und wird als ein fiktives "`Fl\"{u}chten"' oder Wegschwimmen der Larve interpretiert. Nach der Anregungszeit erfolgt ein weiterer Lichtstimulus, auf welchen die Larven typischerweise nicht reagieren und der die vierte und letzte Phase einleitet, die \sog Refractory-Phase oder Resistenzzeit. In der letzten Phase sind die Larven nicht nur unempfindlich gegen\"{u}ber dem zweiten Lichtstimulus, sondern die Anzahl der nat\"{u}rlichen, zuf\"{a}llig auftretenden Spontanbewegungen ist auch deutlich geringer, verglichen mit dem \emph{Ruhezustand} zwischen 0 und 5 Sekunden.

Das Verhalten der Zebrab\"{a}rblingslarven ist hochgradig reproduzierbar und der Versuch l\"{a}sst sich derart skalieren, dass er sich automatisiert in 96er"~Mikrotiterplatten durchf\"{u}hren l\"{a}sst. Beim vorliegenden Hochdurchsatzversuch wird eine gro{\ss}e Anzahl von Eiern (bis zu 5000 Larven pro Tag) gewonnen. Die Eier werden bis zu einem Alter von 28\,hpf in einem Inkubator gehalten und daraufhin in Mikrotiterplatten in Gruppen zu je 8-10 Larven  pipettiert (vgl. \abb \ref{fig:PMR_Urspruenglicher_Plot_2}). Durch den Einfluss von chemischen Substanzen kann das Bewegungsmuster der Larven erheblich von dem typischen oben beschriebenen Muster (vgl. \abb \ref{fig:PMR_Urspruenglicher_Plot}) unbehandelter Larven abweichen. Die Abweichungen lassen sich systematisch zu verschiedenen Einflussgruppen zusammenfassen, aus denen charakteristische Fingerabdr\"{u}cke der Chemikalien erstellt werden k\"{o}nnen \cite{kokel12}.

 Die in \cite{Kokel10} beschriebene Vorgehensweise und Bildverarbeitung hat sich als durchaus robust erwiesen, jedoch sind verschiedene Probleme, gerade im Bereich der Datenanalyse, nicht optimal gel\"{o}st. Beispielsweise erfolgt jegliche Datenakquise bzw. Zeitreihenextraktion aus den beschriebenen f\"{u}nf Linien, welche \"{a}quidistant \"{u}ber das Bild verteilt sind, was den  Nachteil hat, dass keine Information \"{u}ber das einzelne Individuum und dessen Bewegung generiert werden kann. Es wird lediglich der Mittelwert \"{u}ber alle Bewegungen gebildet und so die aggregierte Bewegung berechnet. Des Weiteren wird in \cite{kokel12} die Anzahl von Larven im N\"{a}pfchen nicht exakt ber\"{u}cksichtigt. Da die untersuchte Anzahl jedoch zwischen 8 und 10 Larven variiert \cite{kokel12}, unterliegt somit auch die Auswertung der gleichen Variation. Zudem ist nicht sichergestellt, ob die ausgewerteten Linien die Eier der Larven \"{u}berhaupt \bzw nahe des Zentrums schneiden, was zur Folge hat, dass die Auswertung je nach Verteilung der Larven im N\"{a}pfchen schwankt.  Deshalb ist zu keinem Zeitpunkt klar, wie viel relevante Information ausgewertet wird und wie hoch der Rauschanteil ist. Die als Merkmal zur Charakterisierung der Bewegungsmuster herangezogene Messung unterliegt somit bei gleichen Bedingungen deutlichen Schwankungen.


Durch Anwendung des in der vorliegenden Arbeit vorgestellten Konzeptes lassen sich die genannten Fehler\-einfl\"{u}sse erheblich verringern, \zT beseitigen und somit die Aussagekraft des Versuchs deutlich verbessern, ohne Einbu{\ss}en im Durchsatz in Kauf nehmen zu m\"{u}ssen. Die folgenden Untersuchungen beschr\"{a}nken sich auf unbehandelte Larven, da es notwendig ist, die nat\"{u}rliche Reaktion der Larven auf Lichtstimuli vollst\"{a}ndig zu untersuchen und zu beschreiben, bevor Vergleiche mit behandelten Larven vorgenommen werden k\"{o}nnen. Das Ziel der Anwendung des Konzeptes ist es,
\begin{itemize}
  \item charakteristische Bewegungsereignisse von einzelnen Larven zu klassifizieren sowie
  \item Auftrittsh\"{a}ufigkeit und Dauer der Bewegungsereignisse einzelner Larven zu bestimmen.
\end{itemize}

Hierzu ist es notwendig,
\begin{itemize}
  \item einzelne Larven in den Bildsequenzen zu finden und
  \item Merkmale zur Quantifizierung der Bewegung zu extrahieren.
\end{itemize}


\subsection{Versuchsplanung: Biologie und Bildakquise}

Da in \cite{Kokel10} die Untersuchung bereits im Hochdurchsatz durchgef\"{u}hrt wurde, sind auch die Anforderungen an die Durchf\"{u}hrung und Messung erf\"{u}llt. Somit k\"{o}nnen diese Punkte im Flussdiagramm \abb \ref{fig:Ablaufdiagramm_Screendesign} \"{u}bersprungen werden und direkt mit der Erstellung der Auswertungskette fortgefahren werden. Um die Vergleichbarkeit der Auswertung mit den urspr\"{u}nglichen Ergebnissen von \cite{Kokel10} zu gew\"{a}hrleisten, wurde nur das Merkmal $x_{16}$ aus \kap \ref{subsec:Merkmale_Bewegung}  extrahiert, welches identisch mit dem in \cite{Kokel10} verwendeten ist. Da die Aussagekr\"{a}ftigkeit des Merkmales von \cite{Kokel10} bereits gezeigt wurde, wurde auf einen Vergleich mit den anderen in der vorliegenden Arbeit eingef\"{u}hrten Merkmalen zur Beschreibung von Bewegungen ($x_9-x_{17}$) verzichtet. Die Klassifikation der Bewegungsereignisse muss zudem direkt an der extrahierten Zeitreihe erfolgen. Aus diesen Gr\"{u}nden kann auf die Kategorien \emph{Dimensionsreduktion} und \emph{Merkmalsauswahl} des Modulkatalogs verzichtet werden.

\subsection{Versuchsplanung: Analyse und Interpretation}

Die Parameter der \hts f\"{u}r die Auswertung werden bestimmt. Der biologische Effekt, die Bewegungsereignisse {Coiling} und {Swimming}, sind die Planfaktoren $\mathbf{Y}$ und deren Auftreten soll automatisch, anhand von aus dem Bildstrom $\mathbf{BS}$ extrahierten Zeitreihen $\mathbf{y_{ZR}}$ gesch\"{a}tzt werden. Die Planfaktoren geh\"{o}ren den Ereignissen \emph{\{ja,nein,unbekannt\} }an. Im Hochdurchsatz soll der Versuch zum Ermitteln des Einflusses und zum Kategorisieren von Chemikalien $p_y$ angewandt werden. Als St\"{o}rfaktor $z$ wurde die Position der Larven auf einer 96er-Mikrotiterplatte aufgezeichnet. Der Typ der Mikrotiterplatte ist der Parameter $p_z$ des St\"{o}rfaktors. Die Bildsequenzen wurden in Farbe $m_\mathbf{BS}=3$ bei einer Aufl\"{o}sung von 250x250 Pixeln auf einer Fokusebene f\"{u}r 1000 Frames bei 30fps akquiriert. Eine Bildsequenz, eines Wells, belegt einen Speicherplatz von \ca 200MB. F\"{u}r die Auswertung lag ein Datensatz von 60 Bildsequenzen vor. Tabelle \ref{tab:Anw_Parameter_PMR} fasst die Parameter der \hts zusammen.
\begin{table}[htb]
\begin{tabularx}{\linewidth}{ZZZZccc}
\toprule
$y/\hat{y}$	&$p_y$	&$z/\hat{z}$	&$p_z$	 &\multicolumn{3}{c}{$p_\mathbf{BS}$}  \\
\cmidrule(l){5-7}
	&	&	&	&$m_\mathbf{BS}$	&${u}_\mathbf{BS}$	&$t_\mathbf{BS}$		 \\
\midrule
Coiling Swimming	&Chemikalie &Position	&$96$-Mikro\-titer\-platte	 &$3$	 &$250x250x1$	 &$1000$		\\
\bottomrule
\end{tabularx}
\caption[Parameter der \hts nach Abschnitt \ref{sec:Parameter_mathematisch} f\"{u}r \emph{PMR}"~Versuch]{Parameter der \hts nach Abschnitt \ref{sec:Parameter_mathematisch} f\"{u}r \emph{PMR}"~Versuch; Mit $m_{\mathbf{BS}}$ Anzahl Kan\"{a}le, ${u}_\mathbf{BS}$ Anzahl Pixel und $t_\mathbf{BS}$ Anzahl Frames. }\label{tab:Anw_Parameter_PMR}
\end{table}

\begin{figure}[htbp]
\centering
        \centering
       \includegraphics[page=15, width=\linewidth]{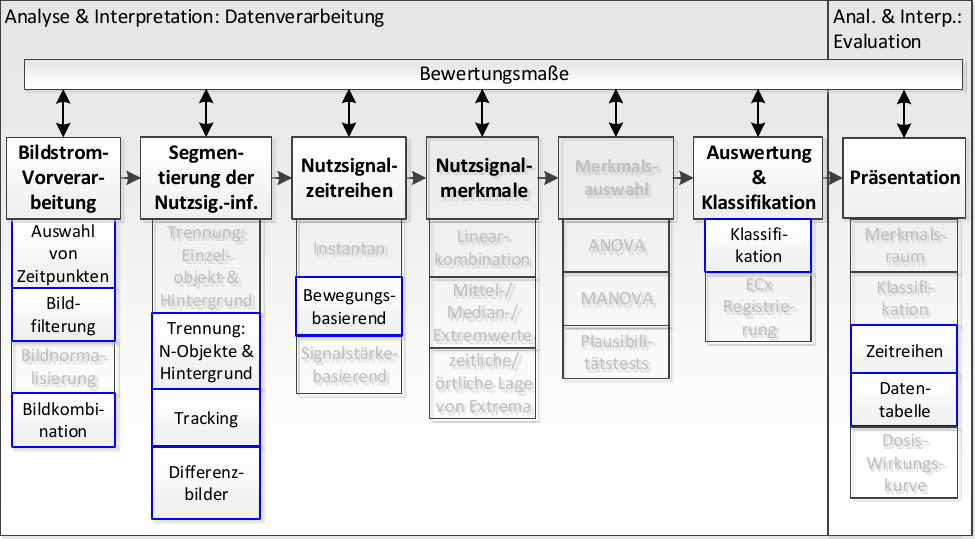}

        \caption{Module zur Auswertung der Bewegungsereignisse {Coiling} und {Swimming}.}
        \label{fig:Module_PMR}
\end{figure}

Die f\"{u}r die Auswertung der Bewegungsereignisse {Coiling} und {Swimming} ausgew\"{a}hlten Module sind in \abb \ref{fig:Module_PMR} hervorgehoben und werden im Folgenden an den jeweiligen Abschnitten erkl\"{a}rt.

\textbf{Bildstrom-Vorverarbeitung:}
Da keine Farbinformationen von Interesse sind, k\"{o}nnen alle Farbebenen durch Mittelung der Farbkan\"{a}le zu einem Grauwertbild kombiniert werden, was den Speicherbedarf auf ein Drittel der urspr\"{u}nglichen Gr\"{o}{\ss}e reduziert. Zur Verminderung von Bildrauschen wird der Bildstrom mittels eines Gau{\ss}filters gegl\"{a}ttet. Die Auswahl von Zeitpunkten kann vorgenommen werden, da w\"{a}hrend der beiden Lichtstimuli die Kamera lediglich wei{\ss}e (ges\"{a}ttigte) Bilder zeigt und die ersten drei Aufnahmen meist noch unterbelichtet sind. Alle Bilder, die w\"{a}hrend der beiden Lichtstimuli (jeweils w\"{a}hrend der Frames oder Abtastzeitpunkte $k=250\dots300,\;k=650\dots700$) aufgezeichnet werden sowie die ersten drei Aufnahmen des Bildstroms ($k=1\dots3$) werden daher verworfen (Auswahl von Zeitpunkten).

\begin{figure}[htbp]
\centering
\captionsetup{font=footnotesize}
    \begin{minipage}[t]{0.32\linewidth}
        \centering
        \includegraphics[page=2,height=4cm]{Bilder/03_Anwendung/PMR_Konzeptumsetzung_Figures}
        \caption[Kantenbild des Beispiel-Screenshots aus \abb \ref{fig:PMR_Urspruenglicher_Plot_2}.]{
        Kantenbild des Beispiel-Screenshots aus \abb \ref{fig:PMR_Urspruenglicher_Plot_2}. Die Eih\"{u}llen sind deutlich als z.T. unterbrochene Kreise zu erkennen. 
        }
        \label{fig:PMR_Konzeptumsetzung_Figures_Kantenbild}
    \end{minipage}
\hfill
    \begin{minipage}[t]{0.32\linewidth}
        \centering
        \includegraphics[page=3,height=4cm]{Bilder/03_Anwendung/PMR_Konzeptumsetzung_Figures}
        \caption[Akkumula\-tions\-matrix (heatmap) der Hough-\-Kreis\-detektion]{
        Akkumula\-tions\-matrix (heatmap) der Hough-\-Kreis\-detektion f\"{u}r Radien zwischen 20 und 30 Pixel}
        \label{fig:PMR_Konzeptumsetzung_Figures_Akkumulationsmatrix}
    \end{minipage}
    \hfill
    \begin{minipage}[t]{0.325\linewidth}
        \centering
        \includegraphics[height=4cm]{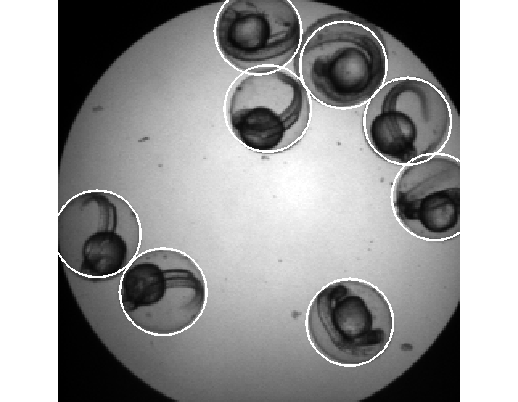}
        \caption[Identifizierte Kreise aus Mittelpunkten aller Maxima]{
        Identifizierte Kreise aus Mittelpunkten aller Maxima der Akkumulationsmatrix \"{u}ber definiertem Schwellenwert}
        \label{fig:PMR_Konzeptumsetzung_Figures_Hough_Ergebniss}
    \end{minipage}
\end{figure}

\textbf{Segmentierung der Nutzsignalinformation:}
Nach der Bildstrom-Vorverarbeitung m\"{u}ssen alle Objekte von Interesse im Bild gefunden, vom Hintergrund getrennt und voneinander unterschieden werden. Dabei handelt es sich bei den Aufnahmen um eine unbekannte Anzahl von Eiern. Die Anzahl schwankt  \"{u}blicherweise zwischen 8 und 10, kann jedoch bei einem versehentlich nicht gef\"{u}llten N\"{a}pfchen auf 0 sinken oder auch deutlich h\"{o}her liegen, \zB durch einen Pipettier-Fehler. Zur Anwendung kommt das Modul \emph{Trennung mehrerer Objekte vom Hintergrund}. Es wird auf das erste betrachtete Frame ($k=4$) angewandt. Im bin\"{a}ren Kantenbild, welches bei der Kreissuche initial erstellt wird, sind die zu suchenden Kreise deutlich zu erkennen, wie  \abb \ref{fig:PMR_Konzeptumsetzung_Figures_Kantenbild} exemplarisch zeigt.

In der Abbildung ist weiter gut zu erkennen, dass die Kreise nur bei ausreichendem Kontrast durch das Kantenbild vollst\"{a}ndig wiedergegeben werden. Es zeigte sich, dass \zB ein niedriger Wasserstand im N\"{a}pfchen das Licht der Durchlichtbeleuchtung derart streut, dass sich ein starker Helligkeitsverlauf zum Rand hin einstellt. In bestimmten  Toleranzgrenzen kann der Helligkeitsverlauf von dem Algorithmus kompensiert werden. Im betrachteten Beispiel ist der Grauwert des Hintergrundes im Zentrum ges\"{a}ttigt, \dhe vollst\"{a}ndig wei{\ss} bei einem Pixelwert von 255, w\"{a}hrend der Wert der Pixel zum Rand hin auf 80 abf\"{a}llt. Je dichter die Eier am Rand liegen und je dunkler der Hintergrund ist, desto geringer ist der Kontrast und desto schlechter k\"{o}nnen die Eier erkannt werden. Die zur Kreisdetektion im n\"{a}chsten Schritt berechnete Akkumulationsmatrix der Hough-Kreisdetektion des Moduls ist in \abb \ref{fig:PMR_Konzeptumsetzung_Figures_Akkumulationsmatrix} dargestellt. Es wurde nach Kreisen mit einem Radius zwischen 20 und 30 Pixel gesucht. Bei der Darstellung wird jedes Pixel im Bild als potentieller Kreismittelpunkt betrachtet. F\"{u}r jedes wei{\ss}e Pixel im Kantenbild werden alle theoretisch m\"{o}glichen Kreismittelpunkte in der Akkumulationsmatrix markiert. Tats\"{a}chliche Kreismittelpunkte erhalten so einen hohen Score in der Akkumulationsmatrix. Alle Punkte \"{u}ber einen Score von 500 in der Akkumulationsmatrix werden schlie{\ss}lich als tats\"{a}chlicher Kreismittelpunkt aufgefasst.
Der Beleuchtungseffekt ist deutlich an schw\"{a}cheren Maxima in den Randregionen der Akkumulationsmatrix  zu erkennen. In \abb \ref{fig:PMR_Konzeptumsetzung_Figures_Hough_Ergebniss} sind die segmentierten Eier markiert.
Ist der Beleuchtungseffekt zu gro{\ss}, werden Eier fehlerhaft segmentiert, was eine Validit\"{a}tspr\"{u}fung erfordert. Die tats\"{a}chlich vorhandenen Kreise m\"{u}ssen von den f\"{a}lschlicherweise detektierten unterschieden werden. Es wird der Abstand aller Kreismittelpunkte ermittelt und alle diejenigen identifiziert, welche n\"{a}her beieinander liegen als 90\% des typischen Eidurchmessers. Daraufhin werden nacheinander Kreismittelpunkte verworfen, welche potenziell die Zebrab\"{a}rblingslarve nicht vollst\"{a}ndig erfassen, wie es \zB in \abb \ref{fig:PMR_Konzeptumsetzung_Figures_Hough_Ergebniss} oben zu sehen ist. Ein typisches Merkmal f\"{u}r eine vollst\"{a}ndig erfasste Larve ist der mittlere Grauwert innerhalb des Kreises. Da Vordergrundpixel, also Pixel, welche die Larve abbilden, dunkel erscheinen, ist der mittlere Grauwert niedriger, wenn die Larve korrekt gefunden wurde und daher mehr Vordergrundpixel enthalten sind. Es werden nacheinander Kreise mit hohen Mittelwerten verworfen, bis alle \"{u}brigen einen Abstand von mehr als 90\% des typischen Eidurchmessers zueinander haben.
Zwei weitere Plausibilit\"{a}tskriterien schlie{\ss}en gefundene Eier aus, welche entweder einen deutlich zu hohen oder niedrigen Mittelwert aufweisen. Ein solcher Fall kann eintreten, wenn \zB ein unbefruchtetes Ei, also eine leere Eih\"{u}lle, in der Bildsequenz vorkommt. Manchmal sind auch Luftbl\"{a}schen auf den Bildern, welche f\"{a}lschlicherweise als Eier erkannt werden. Durch die Plausibilit\"{a}tskriterien k\"{o}nnen solche F\"{a}lle abgefangen werden.

\begin{table}[!htbp]
  \begin{tabularx}{\linewidth}{z{3cm}X}
\toprule
\textbf{Kriterium [\#]}&\textbf{Beschreibung} \\
\midrule
    1 & Frames, die 25\% heller sind als der zuvor bestimmte Mittelwert korrekter Frames, werden verworfen.  \\
    2 & Teilweise oder vollst\"{a}ndig au{\ss}erhalb der Bildmatrix detektierte Eier werden verworfen.\\
    3 & Detektierte Eier mit einer mittleren Helligkeit von \"{u}ber 200 werden verworfen.\\
    4 & Detektierte Eier mit einer mittleren Helligkeit unter 50 werden verworfen.\\
    5 & Eier mit einem Durchmesser gr\"{o}{\ss}er oder kleiner als die Suchradien  (hier zwischen 20 und 30 Pixel) werden verworfen.\\
    6 & Au{\ss}erhalb des Suchraums f\"{u}r das Tracking (hier 4 Pixel in X- und Y-Richtung) detektierte Eier werden verworfen.\\
    7 & Eier, deren Mittelpunkte n\"{a}her als 90\% des zuvor ermittelten typischen Eidurchmessers sind, werden verworfen.\\
    8 & Bewegungen au{\ss}erhalb des detektierten Kreises der detektierten Eier werden verworfen.\\
\bottomrule
\end{tabularx}
\caption{\"{U}bersicht der Validit\"{a}tskriterien beim PMR }\label{tab:PMR_Validit\"{a}t}
\end{table}

\begin{figure}[!htbp]
            \centering
                    \centering
                   \includegraphics[page=5, width=.8\linewidth]{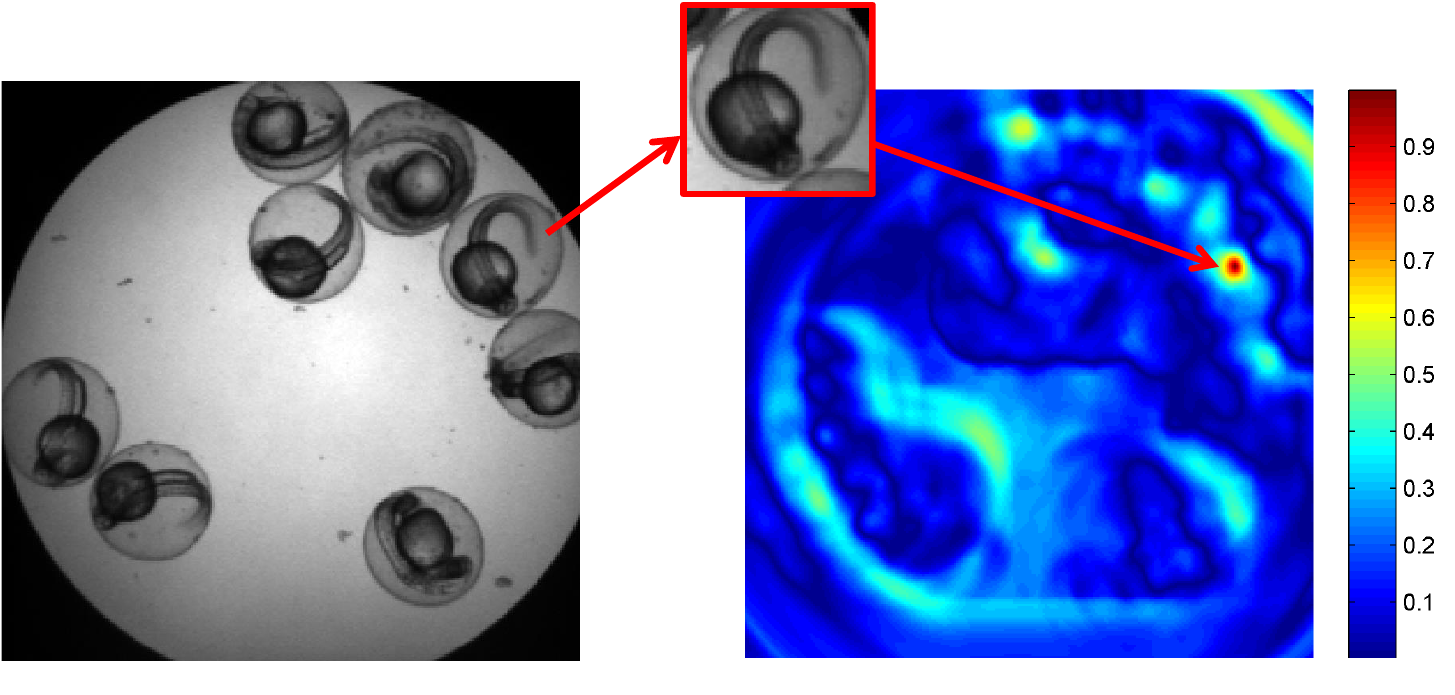}

                    \caption[Kreuzkorrelation des Quell- oder Objektbildes]{Kreuzkorrelation des Quell- oder Objektbildes (Templates) mit dem Zielbild oder Referenzbild aus dem darauf folgenden Frame. Links: Originalbild; Mitte: Template; rechts: Erzielte \"{U}bereinstimmung des Templates f\"{u}r die jeweilige Position im Zielbild. Hohe Werte bedeuten eine hohe \"{U}bereinstimmung.}
                    \label{fig:PMR_Konzeptumsetzung_Figures_Kreuzkorr}
            \end{figure}

Werden alle Eier im Bild fehlerfrei gefunden, so ist das initiale Segmentieren aller Zebrab\"{a}rblingslarven abgeschlossen und die Position im Bild sowie die Abma{\ss}e des umspannenden Rechtecks werden gespeichert. Der n\"{a}chste Schritt ist das Verfolgen der Eier, hierf\"{u}r wird das Modul "`\emph{Tracking}"' angewandt. Exemplarisch ist die Anwendung des Moduls in \abb \ref{fig:PMR_Konzeptumsetzung_Figures_Kreuzkorr} abgebildet. Zur besseren Visualisierung wurde in der Abbildung der Suchraum nicht durch Vorwissen eingegrenzt, wie es bei der Implementierung der Fall ist. Bei dem Verfahren wird das in der Abbildung rot eingerahmte Suchbild (engl. Template) \"{u}ber das Zielbild geschoben und mittels Kreuzkorrelation die \"{U}bereinstimmung ermittelt. Es lassen sich sieben Maxima erkennen, von denen sich eines deutlich abhebt. Da sich zwischen einzelnen Frames das Ei nur wenige Pixel bewegt haben kann, ist der Suchraum auf jeweils vier Pixel in X- und Y-Richtung eingeschr\"{a}nkt. Damit ist eine Verwechslung der Maxima ausgeschlossen. Das Modul ermittelt die wahrscheinlichste Position des Suchbildes im Zielbild.

Die Bilder der Eier zu jedem Abtastzeitpunkt werden nicht zugeschnitten und erneut abgespeichert, sondern es wird lediglich die Position und Ausdehnung des Eies aufgezeichnet. Das Ergebnis ist ein Datensatz, der zu den Rohdaten an jedem Abtastzeitpunkt charakteristische Gr\"{o}{\ss}en f\"{u}r jedes Ei enth\"{a}lt. Die Gr\"{o}{\ss}en sind Eigr\"{o}{\ss}e und Position sowie der mittlere Grauwert. Das Tracking aller Eier einer Bildsequenz ben\"{o}tigt \ca 20\,s (inkl. Einlesen der Bildsequenz und Speichern der Ergebnisse) auf einem zum Zeitpunkt des Entstehens der vorliegenden Arbeit \"{u}blichen Desktop-PC (Segmentierbarkeit des biologischen Effekts/Schnelligkeit der Auswertung).  In der Benutzer\-ober\-fl\"{a}che (vgl. Kapitel \ref{chap:Implementierung}) werden als Parameter Eidurchmesser zwischen 20 und 30 Pixel gew\"{a}hlt (bei einer Aufl\"{o}sung der Sequenzen von 250 x 250 Pixel) und detektierte Eier mit einem mittleren Grauwert unter 50 und \"{u}ber 200 werden verworfen. Tabelle \ref{tab:PMR_Validit\"{a}t} listet alle Validit\"{a}tskriterien auf.

\textbf{Nutzsignalzeitreihen:}
Ist die Position der Eier in jedem Frame ermittelt, m\"{u}ssen f\"{u}r jeden Zeitpunkt Merkmale f\"{u}r die Bewegung extrahiert werden. Bewegungen im Bild f\"{u}hren direkt zu Ver\"{a}nderungen in Pixelwerten, welche nun individuell f\"{u}r jedes Ei berechnet werden k\"{o}nnen. Jede Bewegung der Larve ist mit einer Ver\"{a}nderung der Pixelwerte in der zugeordneten Region verbunden, daher wird f\"{u}r die Bewegung die Summe aller Absolutwerte der Pixeldifferenzen der Region gebildet. Das Ergebnis ist eine Zeitreihe, welche die Bewegung der jeweiligen Larve quantifiziert und im Folgenden \emph{Bewegungsindex} genannt wird. Da das Merkmal nicht mehr wie zuvor die Bewegung mittels Bildzeilen \"{u}ber die gesamte Breite der Rohdaten quantifiziert, sondern der Bewegungsindex f\"{u}r jede Larve individuell bestimmt wird, schwankt die Dynamik und damit der Bewegungsindex im Bildausschnitt, je nachdem ob sich die Larve an einer gut oder schlecht ausgeleuchteten Position im Well befindet. Um den Beleuchtungseffekt zu minimieren, wird der Bewegungsindex mit der Standardabweichung des jeweiligen Bildausschnitts normalisiert. Der Bewegungsindex setzt sich aus den Merkmalswerten des Merkmals $x_{16}$ des Modulkatalogs zu jedem Abtastzeitpunkt zusammen.

\begin{figure}[h!tbp]
\centering
    \begin{minipage}[t]{0.48\linewidth}
        \centering
        \includegraphics[page=8, width=\linewidth]{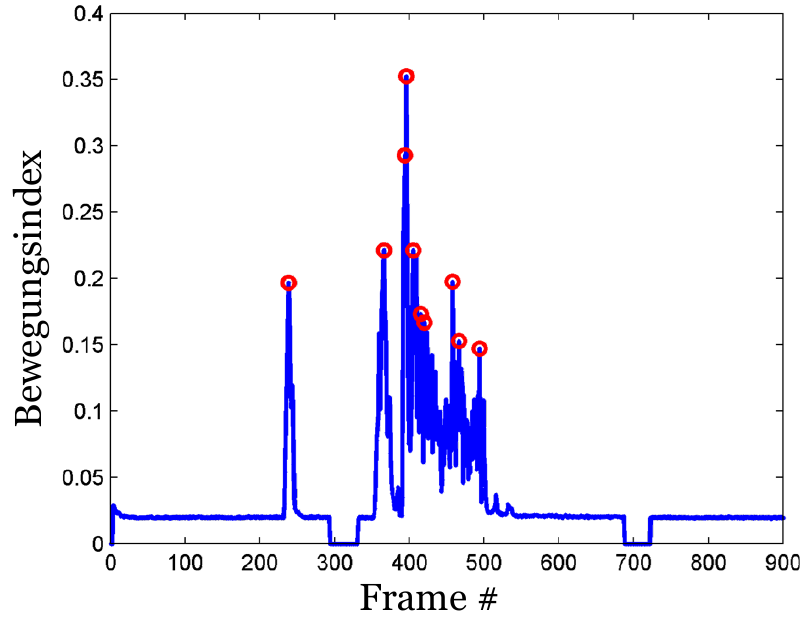}
        \caption[Klassifikation der Coiling-Ereignisse]{Klassifikation der Coiling-Ereignisse anhand von Peak-Erkennung in der Bewegungszeitreihe.}
        \label{fig:PMR_Konzeptumsetzung_Figures_Klassifikation_coiling}
    \end{minipage}
\hfill
    \begin{minipage}[t]{0.48\linewidth}
        \centering
        \includegraphics[page=9, width=\linewidth]{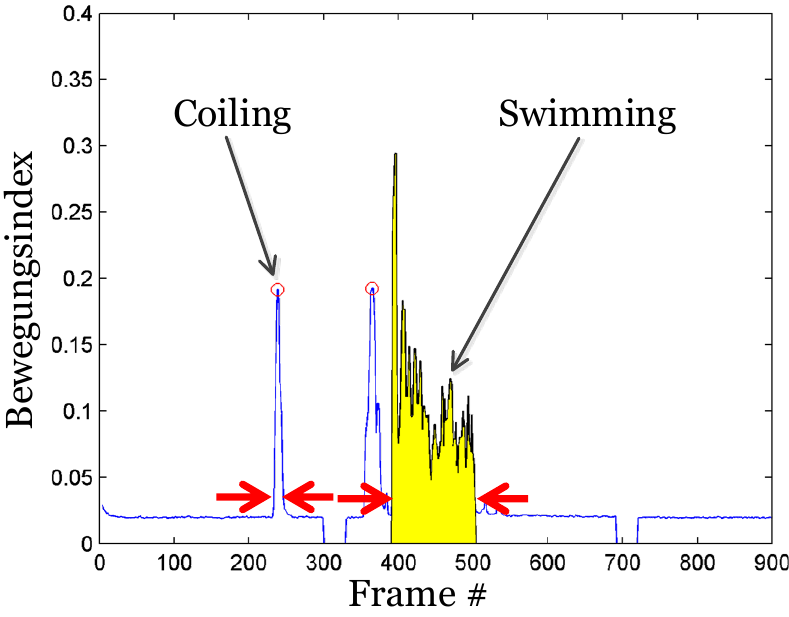}
        \caption[Klassifikation der Swimming-Ereignisse]{Klassifikation der Swimming-Ereignisse anhand von zwei Schwellenwerten.}
        \label{fig:PMR_Konzeptumsetzung_Figures_Klassifikation_Swimming}
        \end{minipage}
\end{figure}

Das Ermitteln der Eiposition mithilfe des Moduls zum Tracken auf Basis eines einfachen Kreises ist zwar \"{a}u{\ss}erst robust gegen\"{u}ber dem beschriebenen Wegdriften der Ei\-posi\-tion, jedoch weicht die gefundene Position in manchen F\"{a}llen um wenige Pixel von der optimalen Position ab. In solchen F\"{a}llen werden Ver\"{a}nderungen der Pixelwerte gemessen, welche nicht auf die Bewegung der Larve, sondern auf die Bewegung der Bildausschnitte zueinander zur\"{u}ckzuf\"{u}hren ist. Um lediglich Bewegungen der Larven zu verfolgen, wurde eine Feinjustierung durchgef\"{u}hrt. Hierzu wird als Suchbild anstatt des einfachen Kreises der Ausschnitt des Originalbildes verwendet, welches auf alle Nachbarpositionen verschoben wird und daraufhin erneut \og Merkmal berechnet wird. Die Position, an der der Wert des Merkmals minimal wird, ist identisch mit der optimalen Position des Suchbildes und enth\"{a}lt die reine Bewegungsinformation. Zur weiteren Verbesserung der Robustheit wurde das beschriebene Verfahren f\"{u}r eine Abweichung von bis zu zwei Pixeln durchgef\"{u}hrt. Eine weitere Verbesserung der Merkmalswerte wird durch Hinzuf\"{u}gen einer zus\"{a}tzlichen Region von Interesse (ROI) innerhalb des segmentierten Bereichs erzielt. Da die Bildausschnitte in Matlab nur als rechteckige Matrizen gespeichert werden k\"{o}nnen, erscheinen Teile von Nachbareiern, welche sich diagonal zu dem aktuellen Ei befinden, im ausgewerteten Bildausschnitt. Bewegt sich nun die Nachbarlarve, erzeugt diese Bewegung bei Ber\"{u}cksichtigung des kompletten Bildausschnitts eine leichte Bewegung in der Zeitreihe, die jedoch nur der Nachbarlarve zuzuordnen ist. Die ROI wurde durch die Multiplikation der Bildausschnitte mit einer kreisrunden Maske durchgef\"{u}hrt, welche den typischen Eidurchmesser aufweist und alle Pixel au{\ss}erhalb des auszuwertenden Eies auf den Wert 0 setzt. Das Ergebnis ist eine Zeitreihe der Bewegung einer einzelnen Larve und exemplarisch in \abb \ref{fig:PMR_Konzeptumsetzung_Figures_Klassifikation_coiling} dargestellt. Es wurden 425 Bewegungszeitreihen von Larven, also ca. sieben Larven pro Bildsequenz, gefunden und extrahiert (Quantifizierbarkeit des biologischen Effekts im Bild).

\textbf{Auswertung \& Klassifikation:}
Da f\"{u}r jede Larve eine detaillierte Information \"{u}ber die Reaktion auf den Lichtstimulus extrahiert wurde, kann nun eine genaue Untersuchung der typischen Reaktionsformen und Bewegungsereignisse folgen. Die Zeitreihen werden als Gait-CAD-Projekt gespeichert und dort bez\"{u}glich der Bewegungsereignisse klassifiziert. Im Bewegungsprofil sind Coiling Bewegungen als deutlich separierte hohe Ausschl\"{a}ge erkennbar, w\"{a}hrend das Swimming ein etwas l\"{a}ngerer Abschnitt mittlerer Amplitude ist.
Zur Ermittlung der Anzahl und zum Unterscheiden zwischen beiden Ereignissen werden zuerst alle hohen Ausschl\"{a}ge im Bewegungsindex identifiziert (\vgl \abb \ref{fig:PMR_Konzeptumsetzung_Figures_Klassifikation_coiling}).

Daraufhin wird die L\"{a}nge des Intervalls bestimmt, in dem sich die Larve bewegt. Ein manueller Vergleich von Videosequenzen und den Werten des Bewegungsindex ergibt, dass typische Coiling-Ereignisse weniger als 40\,Frames andauern (oder 1,33\,s bei 30,03\,fps). In \abb \ref{fig:PMR_Konzeptumsetzung_Figures_Klassifikation_Swimming} ist die L\"{a}nge zweier Intervalle anhand roter Pfeile gekennzeichnet. Das linke Intervall ist ein Coiling-Ereignis, w\"{a}hrend das rechte, deutlich l\"{a}ngere Intervall ein Swimming-Ereignis ist (gelb gekennzeichnet). Mittels der so bestimmten L\"{a}nge der Intervalle um die detektierten Ausschl\"{a}ge werden alle Ereignisse mit einer Intervalll\"{a}nge von weniger als 40\,Frames als Coiling-Ereignisse und alle anderen als Swimming-Ereignisse klassifiziert.

\begin{figure}[h!tbp]
\centering
    \begin{minipage}[t]{0.48\linewidth}
        \centering
        \includegraphics[page=10, width=\linewidth]{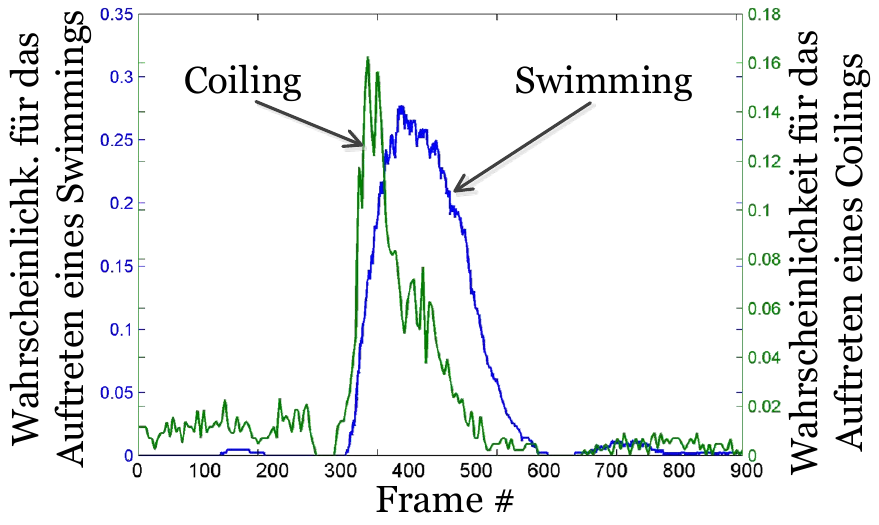}
        \caption[Wahrscheinlichkeit f\"{u}r das Auftreten von genau einem Swimming- oder Coiling-Ereignis]{Wahrscheinlichkeit f\"{u}r das Auftreten von genau einem Swimming- oder Coiling-Ereignis f\"{u}r jedes Frame}
        \label{fig:PMR_Konzeptumsetzung_Figures_Wahrscheinlichkeiten}
    \end{minipage}
\hfill
    \begin{minipage}[t]{0.48\linewidth}
        \centering
        \includegraphics[page=11, width=.83\linewidth]{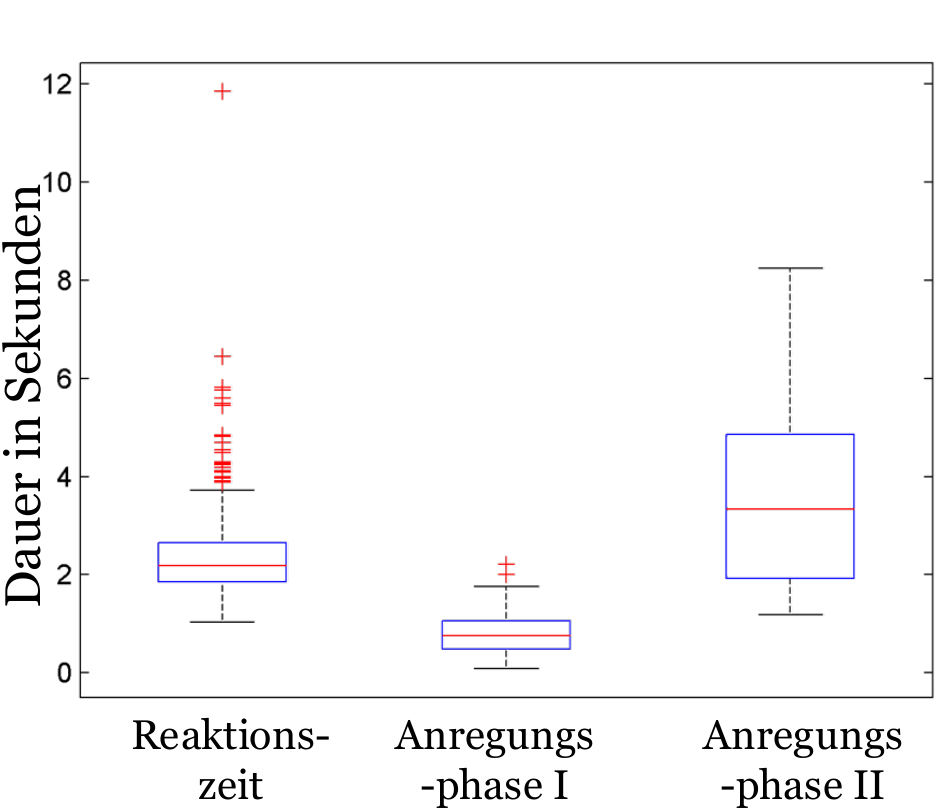}
        \caption{Dauer der Phasen, ermittelt anhand der Klassifikationen}
        \label{fig:PMR_Konzeptumsetzung_Figures_Zeitdauer}
        \end{minipage}
\end{figure}

Die Anwendung der Klassifikation auf alle extrahierten 425 Zeitreihen erm\"{o}glicht es, typische Bewegungsmuster und die Zeitdauer verschiedener Phasen zu bestimmen. In \abb \ref{fig:PMR_Konzeptumsetzung_Figures_Wahrscheinlichkeiten} ist die Wahrscheinlichkeit f\"{u}r ein Auftreten genau eines Coiling-Events (gr\"{u}n) \bzw eines Swimming-Events (blau) an jedem Frame aufgetragen. Deutlich zu sehen ist der Peak der Wahrscheinlichkeit direkt nach dem ersten Lichtstimulus (Frame 300) sowie die Abnahme der Wahrscheinlichkeit f\"{u}r einen Coil in der Resistenzphase (Frame 700-900) im Vergleich mit dem Ruhezustand (Frame 0-300). Die Wahrscheinlichkeit f\"{u}r ein Swimming hat ihren Peak kurz nach dem Peak Coiling-Wahrscheinlichkeit.

 In \abb \ref{fig:PMR_Konzeptumsetzung_Figures_Zeitdauer} ist auf der Grundlage der gemessenen Bewegungen die jeweilige Dauer der Phasen bestimmt und \"{u}ber alle Larven als Boxplot dargestellt. Die Reaktionszeit ist die Dauer in Sekunden vom Lichtstimulus bis zur ersten messbaren Reaktion der Larve (Median bei etwa 2 Sekunden Dauer). Die in \abb \ref{fig:PMR_Urspruenglicher_Plot} eingef\"{u}hrte Anregungsphase l\"{a}sst sich mit Hilfe der Klassifikationen nun in zwei Teile aufspalten. Anregungsphase I ist die Dauer vom Ende der Reaktionszeit bis zum Einsetzen der Swimming-Ereignisse (Median bei etwa einer Sekunde). Die Dauer der Swimming-Ereignisse ist Anregungsphase II (Median bei etwa drei Sekunden). 
 Die gesamte Auswertung einer Bildsequenz ben\"{o}tigt \ca 2\,min. Eine 96er-Mikrotiterplatte mit \ca 770 Larven kann an einem einzigen Computer somit unter 3:20\,h ausgewertet werden. Die Art der Implementierung (siehe Kapitel \ref{chap:Implementierung}) erm\"{o}glicht zudem die Beschleunigung der Auswertung durch das verteilte Rechnen auf mehreren Computern (Schnelligkeit der Auswertung).

\begin{figure}[h!tbp]
\centering
    \begin{minipage}[t]{0.48\linewidth}
        \centering
        \includegraphics[page=6, width=.8\linewidth]{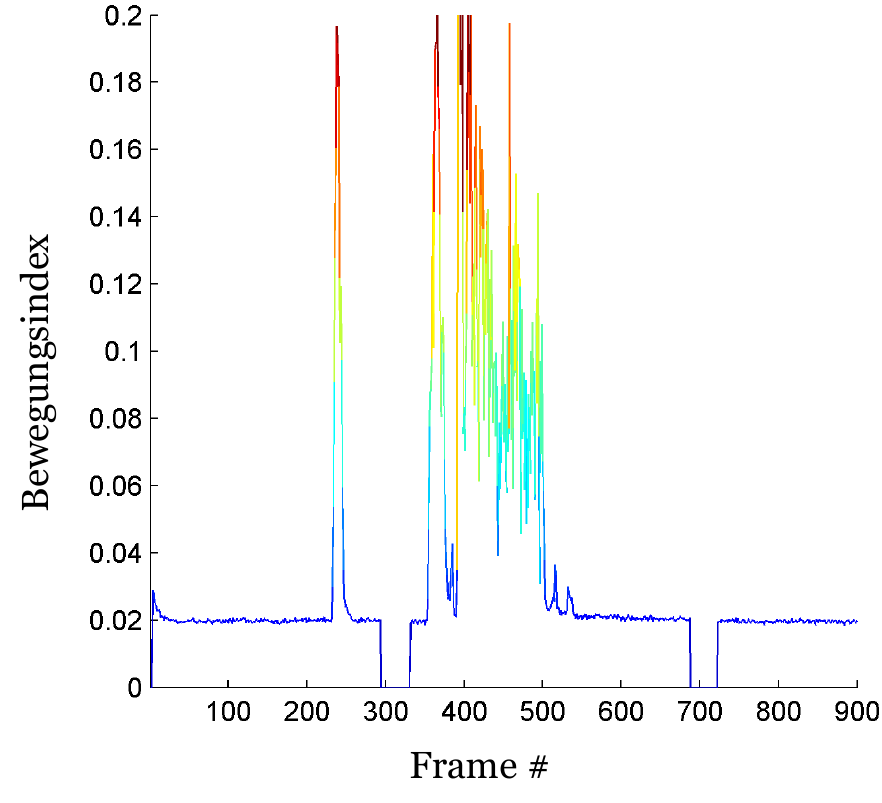}
        \caption[Extrahierte Bewegungszeitreihe]{Extrahierte Bewegungszeitreihe. Hohe Werte sind rot, niedrige blau eingef\"{a}rbt.}
        \label{fig:PMR_Konzeptumsetzung_Figures_Bewegungszeitreihe}
    \end{minipage}
\hfill
    \begin{minipage}[t]{0.48\linewidth}
        \centering
        \includegraphics[page=7, width=\linewidth]{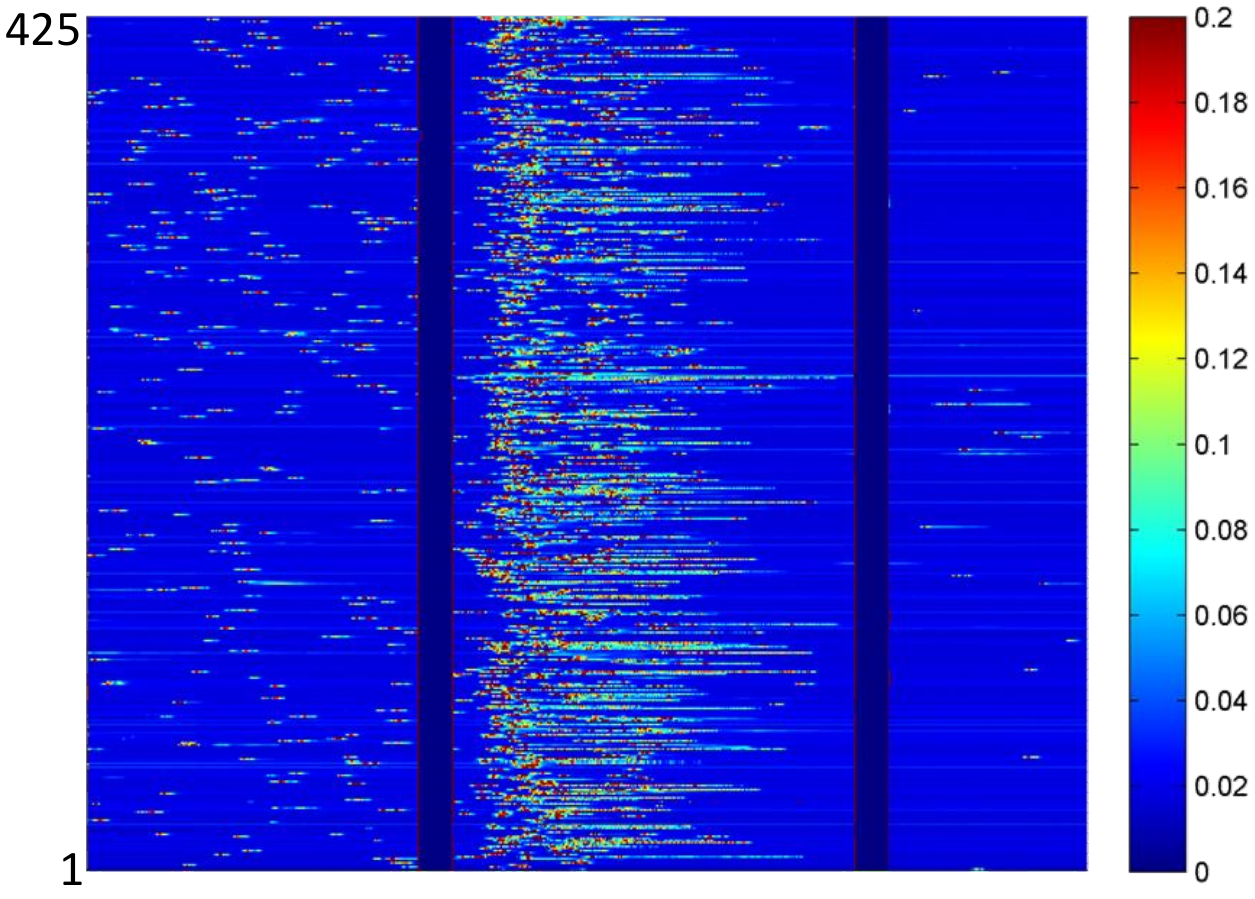}
        \caption[Zusammenfassung aller 425 Fischlarven als Heatmap.]{Zusammenfassung aller 425 Fischlarven als Heatmap. Die einzelnen Bewegungsphasen der Larven sind deutlich zu erkennen.}
        \label{fig:PMR_Konzeptumsetzung_Figures_Heatmap}
        \end{minipage}
\end{figure}

\textbf{Pr\"{a}sentation:}
Ziel der Visualisierung ist es, die gro{\ss}e Menge an Bewegungszeitreihen und Klassifikationen derart zu veranschaulichen, dass Bewegungsmuster erkannt werden k\"{o}nnen. Wird die Bewegungszeitreihe je nach H\"{o}he des Ausschlages mit einem Farbcode eingef\"{a}rbt und wird die Kurve  "`von oben"' betrachtet, so ergibt sich ein Farbprofil, in welchem blau f\"{u}r keine Bewegung und rot f\"{u}r starke Bewegungen definiert wurde (\vgl \abb \ref{fig:PMR_Konzeptumsetzung_Figures_Bewegungszeitreihe}). Wird der Farbcode f\"{u}r jede Larve in jeweils einer Zeile aufgetragen, ergibt sich eine Heatmap-\"{U}bersicht wie sie in \abb \ref{fig:PMR_Konzeptumsetzung_Figures_Heatmap} dargestellt ist.
Deutlich sind die vereinzelten Ausschl\"{a}ge im Ruhezustand, die Ruhepause w\"{a}hrend der Reaktionszeit, die Anregungsphase sowie die Resistenzzeit zu erkennen. Die \"{U}bersicht erm\"{o}glicht einen schnellen \"{U}berblick \"{u}ber alle Larven und l\"{a}sst sich bis \ca 500 Larven gut visualisieren. Die Anzahl von Ereignissen wiederum l\"{a}sst sich zu \emph{Datentabellen} zusammenfassen. Die Zeitdauer der Phasen l\"{a}sst sich entsprechend der \abb \ref{fig:PMR_Konzeptumsetzung_Figures_Zeitdauer} \zB mittels Boxplots darstellen.

Zur biologischen Auswertung und Archivierung werden die Ergebnisse in Form eines PDF-Dokumentes automatisch generiert. Das Dokument enth\"{a}lt einen Screenshot des ersten und letzten Frames. Hierin sind die Ergebnisse der Segmentierung \"{u}berlagert und die Larven durchnummeriert (\"{a}hnlich zu \abb \ref{fig:PMR_Konzeptumsetzung_Figures_Hough_Ergebniss}). So kann mit einem Blick gepr\"{u}ft werden, ob die Segmentierung und das Tracking erfolgreich sind. Ebenso werden die Bewegungszeitreihen der Larven in \og Heatmaps dargestellt (Pr\"{a}sentierbarkeit der Auswertung). Alle Daten wurden identisch zu den anderen Untersuchungen archiviert (Wissenschaftliche Archivierung).

\subsubsection{Durchf\"{u}hrung und Resultat}
Mit den Zeitreihen ist es nun erstmals m\"{o}glich, die individuellen Bewegungen zu klassifizieren. Es wurden Frames mit 1.540 Coiling-Ereignissen und 19.775 Frames mit Swimming-Ereignissen identifiziert (Aus einer Gesamtzahl von 382.500 Frames). Zeitlich ereignet sich das Maximum der Swimming-Ereignisse nach dem Maximum der Coiling-Ereignisse. Dies l\"{a}sst sich auch in \abb \ref{fig:PMR_Konzeptumsetzung_Figures_Wahrscheinlichkeiten} ablesen. In der Abbildung ist die Wahrscheinlichkeit f\"{u}r das Auftreten eines Coilings (blaue Kurve) und das Auftreten eines Swimmings (gr\"{u}ne Kurve) zu jedem Zeitpunkt angegeben. Die Verteilung zeigt ein starkes Ansteigen der Bewegungen nach dem Stimulus. Nach der Reaktionszeit von 2.2\,s (66\,Frames) auf den Stimulus steigt das Auftreten der Coiling-Frequenz um etwa das 10\,fache von 0.07 Coilings/s auf 0.82 Coilings/s f\"{u}r die mittlere Dauer von \ca 0.8\,s (\abb \ref{fig:PMR_Konzeptumsetzung_Figures_Zeitdauer}). W\"{a}hrend vor dem Stimulus nur bei 2 Larven ein Swimming klassifiziert wurde, zeigen 161 Larven nach dem Stimulus ein Swimming-Ereignis, mit einer mittleren Dauer von 3.3\,s. Die Werte legen den Schluss nahe, dass die Anregungsphase sich in eine Anregungs-Phase I und Anregungs-Phase II aufspalten l\"{a}sst. Nach der Anregungsphase f\"{a}llt die Rate der Coilings auf einen Wert von 0.02 Coilings/s ab und gesunde Larven reagieren nicht auf einen weiteren Lichtstimulus (\abb \ref{fig:PMR_Konzeptumsetzung_Figures_Heatmap}). Die Ergebnisse trugen schlie{\ss}lich zur Identifikation von biologischen Wirkungsketten und zu einem tieferen Verst\"{a}ndnis der Reaktionen und Bewegungsmuster der Larven bei \cite{kokel13}. In \cite{kokel13} findet sich zudem eine genauere biologische Interpretation der o.g. Werte.

\section{Bewertung}
In Kapitel \ref{chap:Anwendung} wurde die Funktionsweise des in der vorliegenden Arbeit neu entwickelten Konzeptes aus Kapitel~\ref{chap:Neues_Konzept} anhand von zwei realen Problemstellungen mit jeweils steigender Komplexit\"{a}t belegt. F\"{u}r die Untersuchungen steht nun eine automatische und aussagekr\"{a}ftige Auswerteroutine zur Verf\"{u}gung.  Auf die Ergebnisse beider Projekte wird im Folgenden gesondert eingegangen.
\subsection{Fisch Embryo Test}
In Abschnitt \ref{sec:Anwendung_FET} wurde das neue Konzept auf zwei toxikologische Untersuchungen angewandt. Die Auswertung des Fisch Embryo Tests (FET) wurde f\"{u}r drei verschiedene biologische Endpunkte bildbasierend und automatisiert durchgef\"{u}hrt. Anhand von Einzelaufnahmen gelingt es, den Endpunkt \emph{Koagulation} zu klassifizieren. Mittels Bildsequenzen wird die Auswertung auf die Endpunkte \emph{Herzschlag} und \emph{Spontanbewegung} ausgeweitet.

In einer Voruntersuchung wurden die Versuchs- und Auswerteparameter nach dem neuen Konzept gem\"{a}{\ss} \abb \ref{fig:Neues_Konzept_Prinzip_b} festgelegt. Zur Sicherstellung der Ber\"{u}cksichtigung aller in Abschnitt \ref{sec:Anforderungsgerechte_Anwendung} ermittelten Anforderungen an eine \hts wurde bei der Auslegung das Flussdiagramm \abb \ref{fig:Ablaufdiagramm_Screendesign} durchlaufen. Ein wesentliches Ergebnis der Voruntersuchung ist, dass f\"{u}r eine erfolgreiche Untersuchung alle extrahierten Merkmale zur Klassifikation lageunabh\"{a}ngig sein m\"{u}ssen.

F\"{u}r beide Untersuchungen wurde auf Basis der Voruntersuchung eine Auswertungskette durch Auswahl geeigneter Module aus Kapitel \ref{sec:BV_Module} erstellt. Durch die neuen Methoden zur Segmentierung und zur Vorbereitung des Bild- und Datenstroms gelingt f\"{u}r beide F\"{a}lle die Segmentierung und Merkmalsextraktion. Die f\"{u}r die vorliegende Arbeit entwickelten instantanen Merkmale $x_1$ bis $x_8$ des Modulkatalogs (\vgl \kap \ref{subsec:Module_Instantane_Merkmale}) erm\"{o}glichen die Klassifikation des Endpunktes \emph{Koagulation}. Die Merkmale $x_9$ bis $x_{15}$ zur Klassifikation von Bewegungen erm\"{o}glichen die Erweiterung auf die Endpunkte \emph{Herzschlag} und \emph{Spontanbewegung}. Die Ergebnisse der Klassifikation wurden durch geeignete Methoden des Modulkatalogs pr\"{a}sentiert.

Abschlie{\ss}end werden die Ergebnisse beider Untersuchungen zu einer Klassifikatorkaskade kombiniert (\vgl \abb \ref{fig:Klassifikator_Kaskade}), mit der nun erstmals die bildbasierte automatisierte Auswertung des FET anhand von drei Endpunkten zur Verf\"{u}gung steht.
\subsection{Photomoter Response}
In Abschnitt \ref{sec:Anwendung_PMR} wurde das in der vorliegenden Arbeit erarbeitete Konzept aus Kapitel \ref{chap:Neues_Konzept} erneut erfolgreich angewandt. Nach der Anwendung auf Einzelaufnahmen und der anschlie{\ss}enden Erweiterung auf Bildsequenzen f\"{u}r vereinzelte Zebrab\"{a}rblingslarven innerhalb der Versuche des FET in Abschnitt \ref{sec:Anwendung_FET} ist es nun erstmals m\"{o}glich, einzelne Larven auch ohne vorherige manuelle Vereinzelung zu untersuchen. F\"{u}r den PMR bedeutet das, bei unver\"{a}ndertem Durchsatz, eine erhebliche Verringerung der Fehlereinfl\"{u}sse und eine deutliche Verbesserung der Aussagekraft verglichen mit dem in \cite{Kokel10} vorgestellten Verfahren. Die verbesserte Aussagekraft des in der vorliegenden Arbeit vorgestellten Verfahrens trug direkt zur Identifikation sog. \emph{Response Cells} im Rautenhirn (engl. Hindbrain) bei, welche f\"{u}r die Ausl\"{o}sung der stereotypen Antwort der Zebrab\"{a}rblinge auf Lichtimpulse verantwortlich sind \cite{kokel13}.

Die Versuchs- und Auswerteparameter wurden wieder nach dem neuen Konzept gem\"{a}{\ss} \abb \ref{fig:Neues_Konzept_Prinzip_b} festgelegt und die Erf\"{u}llung aller in Abschnitt \ref{sec:Anforderungsgerechte_Anwendung} ermittelten Anforderungen sichergestellt.  Durch Auswahl geeigneter Module aus dem Modulkatalog wurde eine Auswertungskette erstellt. Die in der vorliegenden Arbeit neu entwickelten Methoden zur Segmentierung und zum Tracking der Fischeier erm\"{o}glichen, nach Extraktion von Merkmalen, das Ermitteln eines charakteristischen Bewegungsindex f\"{u}r jede Fischlarve einzeln. Auf Basis des Bewegungsindex wird eine Klassifikation der beiden Ereignisse \emph{Coiling} und \emph{Swimming} m\"{o}glich. Schlie{\ss}lich wird die Leistungsf\"{a}higkeit der Auswertungskette auf 425 Zebrab\"{a}rblingslarven belegt und die Dauer charakteristischer Phasen (Anregungsphase I und II, Reaktionszeit) ermittelt.

In Form der Auswertungskette steht nun ein leistungsf\"{a}higes Verfahren zur Analyse und Klassifikation einzelner Zebrab\"{a}rblinge, ohne die Notwendigkeit einer vorherigen manuellen Vereinzelung, zur Verf\"{u}gung.

\chapter{Zusammenfassung und Ausblick}\label{chap:Zusammenfassung}

Bildbasierte \htsen am Zebrab\"{a}rbling setzen hohe Anforderungen sowohl an den Versuchsentwurf als auch die automatische Analyse und Interpretation. Die gro{\ss}e Anzahl an Einzeluntersuchungen und die dementsprechend gro{\ss}e Datenmenge nach der Bildakquise machen ein umfassendes Konzept zur Versuchsauslegung sowie neue Auswerteverfahren notwendig.

Zur optimierten Versuchsauslegung schl\"{a}gt die vorliegende Arbeit zum einen ein strukturiertes Verfahren vor, mit dem es m\"{o}glich wird, die Versuchsparameter, angepasst f\"{u}r die automatisierte Bildauswertung, zu bestimmen. Zum anderen werden neue Bildverarbeitungsmodule, die sowohl f\"{u}r den Zebrab\"{a}rbling als auch den Hochdurchsatz entwickelt wurden, vorgestellt. Die Kombination beider Teile bietet dem Anwender, \ia dem Biologen, einen L\"{o}sungsweg f\"{u}r \htsen am Zebrab\"{a}rbling, der nicht nur neue Nutzsignale auswertbar macht, sondern auch die Auslegung optimiert und somit Datenmenge, redundante Informationen und Arbeitsaufwand sowie Klassifikationsfehler reduziert.

Die Kernpunkte der vorliegenden Arbeit umfassen:

\begin{itemize}
  \item Die Entwicklung eines Konzeptes zur strukturierten Versuchsauslegung f\"{u}r die \hts am Zebrab\"{a}rbling (Kapitel \ref{chap:Neues_Konzept}).

  \item Die Entwicklung neuer robuster, hochdurchsatzf\"{a}higer Bildverarbeitungsmodule f\"{u}r Untersuchungen am Zebrab\"{a}rbling (Kapitel \ref{sec:BV_Module}).

  \item Die Implementierung der Algorithmen zur skalierbaren Anwendung auf \htsen (Kapitel \ref{chap:Implementierung}) sowie
  \item die Anwendung und der Nachweis der Leistungsf\"{a}higkeit des Konzeptes und der Module auf zwei exemplarische biologische Problemstellungen (Kapitel \ref{chap:Anwendung}).
\end{itemize}

Die wichtigsten Ergebnisse der vorliegenden Arbeit sind:

  \begin{enumerate}
\setlength{\itemsep}{-8pt}
    \item 	Entwicklung eines neuen Konzeptes zum systematischen Entwurf einer Hochdurchsatz-Prozesskette aus Versuchsplanung, Versuchsvorbereitung, Datenerfassung, Datenverarbeitung und Evaluation.
    \item 	Ableitung einer Entscheidungslogik f\"{u}r die Auslegung einer geeigneten Hochdurchsatz-Datenakquise.
    \item 	Herleitung neuer mathematischer Beschreibungen zur Bewertung von Versuchsauslegungen von Hochdurchsatzversuchen, die speziell auf die Anforderungen zur Automatisierbarkeit des Gesamtprozesses eingehen.
    \item 	Entwicklung einer neuen Methode zur Minimierung des Einflusses von St\"{o}rgr\"{o}{\ss}en auf die Klassifikation.
    \item 	Ableitung einer neuen dichotomen Klassifikatorkaskade zur Evaluation von Herzschlag und Spontanbewegung von Zebrab\"{a}rblingslarven f\"{u}r den Hochdurchsatz.
    \item 	Entwicklung eines neuen Verfahrens zur robusten Segmentierung und Tracking von Zebrab\"{a}rblingseiern.
    \item 	Entwicklung eines neuen Verfahrens zur Extraktion von Bewegungen und Klassifikation der Bewegungsmuster von Zebrab\"{a}rblingslarven.
    \item 	Ableitung einer Methodik zur Realisierung eines automatischen Reportgenerators sowie Nachweis der Funktionalit\"{a}t.
    \item 	Entwicklung der neuen Benutzeroberfl\"{a}che \emph{PIMP} zum benutzerfreundlichen Import, zur Adaption,  Ausf\"{u}hrung, Klassifikation, Visualisierung und Archivierung von \htsen am Zebrab\"{a}rbling.
    \item 	Nachweis der Funktionalit\"{a}t des Konzeptes durch Versuchsentwurf und "~durchf\"{u}hrung zweier hochdurchsatzf\"{a}higer Untersuchungen.
    \item 	Nachweis der Leistungsf\"{a}higkeit der vorgestellten Module durch Anwendung auf zwei reale Problemstellungen.
    \item 	Implementierung der Algorithmen in die Matlab-Toolbox Gait-CAD in Form eines Plug-ins.

  \end{enumerate}

  In Bezug auf die automatisierte Bildauswertung m\"{u}ssen sich weiterf\"{u}hrende Arbeiten mit der vollst\"{a}ndigen Automatisierung der \zT manuell durchgef\"{u}hrten Vorbereitungsschritte befassen.

\begin{figure}[htbp]
        \centering
                \includegraphics[page=11,
                width=\linewidth
                ]{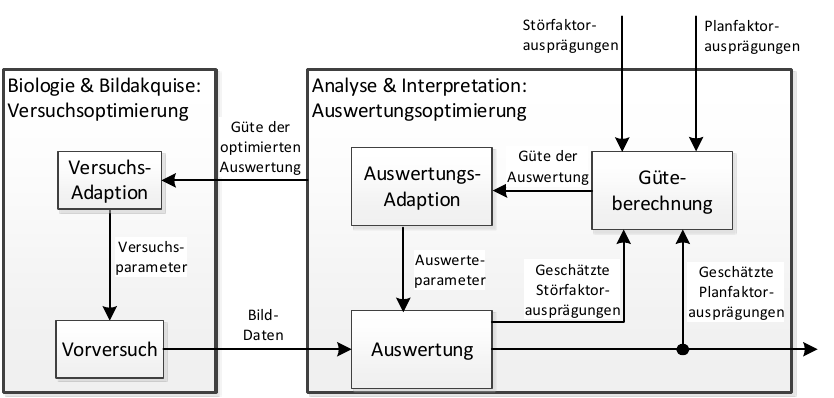}
\caption[Optimierung sowohl der Auswerte- als auch der Versuchsparameter ]{Ausblick: Anhand der erzielten G\"{u}te einer \hts werden sowohl Auswerte- als auch Versuchsparameter angepasst und verbessert.}
\label{fig:Ausblick}
\end{figure}

  Eine zuk\"{u}nftige Weiterentwicklung erfordert, die Bildauswertung, Bildakquise sowie Versuchsdurchf\"{u}hrung weiter zu verkn\"{u}pfen. Hierbei wird der Ansatz der computergest\"{u}tzten Bildakquise weitergef\"{u}hrt zum computergest\"{u}tzten Experiment. Der Verlauf derartiger Experimente ist von der Reaktion der Proben abh\"{a}ngig. Eine solche R\"{u}ckkopplung der Bildauswertung auf den Versuchsverlauf w\"{u}rde \zB das Feld des Behaviorismus \cite{skinner1953} erstmals f\"{u}r den Hochdurchsatz er\"{o}ffnen. Weiteres Potenzial f\"{u}r methodische Entwicklungen bietet z.B. die dreidimensionale Registrierung von Zebrab\"{a}rblingseiern nach Spontanbewegungen im Chorion. Eine solche Methode erm\"{o}glicht eine \hts, bei der sich der Entwicklungsprozess von einzelnen Organen st\"{a}ndig und ohne Pr\"{a}paration verfolgen l\"{a}sst. Weiteres Potenzial bietet sich im Bereich der Auswertung und R\"{u}ckf\"{u}hrung von Klassifikationsergebnissen. Eine M\"{o}glichkeit ist es, die Optimierung der erzielten G\"{u}te der Klassifikation nicht nur, wie in der vorliegenden Arbeit vorgestellt, zur Merkmalsauswahl und Normierung zu verwenden, sondern auf die Auswahl aller Parameter der Verarbeitungskette auszuweiten (siehe erste Ans\"{a}tze in \cite{Khan12smps}).  \abb \ref{fig:Ausblick} zeigt einen m\"{o}glichen Aufbau einer solchen r\"{u}ckgekoppelten Parameterauswahl wie sie bei Weiterf\"{u}hrung der hier vorgestellten Ergebnisse aussehen k\"{o}nnte.

\addcontentsline{toc}{chapter}{Symbolverzeichnis}

\chapter*{Symbolverzeichnis}
\markboth{Symbolverzeichnis}{Symbolverzeichnis}

\section*{Notationsvereinbarung}
\begin{tabular}{@{}ll@{}}
Skalare & kursiv: $a, b, c,\dots$\\
Vektoren& fett: $\mathbf{a, b, c,}\dots$\\
Matrizen& fett, gro{\ss}: $\mathbf{A, B, C, \dots}$\\
Mengen& kalligraphisch, gro{\ss}: $\mathcal{A, B, C}$\\
Konstanten, Bezeichner& Kleinbuchstaben: $\mathrm{a, b, c}$\\

\end{tabular}

\section*{Symbole}
 \begin{longtable}{p{.27\textwidth}p{.67\textwidth}}

\toprule
{\bf Formelzeichen}							&
{\bf Bedeutung} \\
\midrule
$a$                             & Schwellenwert im Bewegungsindex\\
$B$                             &Bildanzahl\\
$\mathbf{BS}$                   & Bildstrom eines Einzelversuchs\\
$\mathbf{BS^*}$                   & Bildstrom eines Einzelversuchs nach der Bildvorverarbeitung\\

$c$                             &Quantil bei der Merkmalsberechnung\\
$CV$                            &Messgenauigkeitsparameter  \\
$d$                   &Hill-Koeffizient; Absolutwert der Steigung der Dosis-Wirkungs-Kurve\\
$EC_x$                            &Wendestelle der Dosis-Wirkungskurve\\

$f_{{BS}}$                         & Aufnahmeh\"{a}ufigkeit des Bildstroms\\
$f_{\Psi}$                         & Abtastfrequenz des Signals\\

${H}$                   &Histogramm eines Bildes\\
 $h(x,y)$                         & Filtermaske \\

$\mathbf{I}$                    &Bildmatrix\\
$\mathcal{I}$                    &Indexmenge ausgew\"{a}hlter Merkmale\\
$\mathcal{I}^*$                    &Durch Filterung adaptierte Indexmenge $\mathcal{I}$   \\
$\mathcal{I}^{**}$                    &Durch einen Wrapper adaptierte Indexmenge $\mathcal{I}$   \\
$\mathcal{I}_{BV}$              &Indexmenge zur zeitlichen Unterscheidung im Bildstrom\\
 $i_c$                          & Laufindex f\"{u}r Modalit\"{a}ten eines segmentierten Bildstroms\\
$I_c$                    &Anzahl an Modalit\"{a}ten eines Bildstroms\\
 $i_{c\Psi}$                          & Laufindex f\"{u}r Modalit\"{a}ten eines segmentierten Bildstroms\\
$I_{c\Psi}$                    &Anzahl an Modalit\"{a}ten eines segmentierten Bildstroms\\
$\mathbf{I}_d$                    &Differenzbild\\
 $I^*_d$                          & Pixelwert eines mit Rauschen \"{u}berlagerten Differenzbildes\\
 $\mathbf{I}_d,\text{smoothed}$                    &Gegl\"{a}ttetes Differenzbild\\
$I_\text{dyn}$                     & Dynamischer Schwellenwert bei der Merkmalsberechnung\\
$\mathbf{I}_E$                    &Einzelne Bildmatrix innerhalb eines Bildstroms\\
  $i_f$                          & Laufindex der Frames einer Bildsequenz\\
    $i_{f\Psi}$                          & Laufindex der Frames einer Bildsequenz im segmentierten Bildstrom\\
 $i_h$                            &Laufindex im Histogramm\\
$I_\text{hoch}$                            & Oberer Schwellenwert zur Bildnormalisierung\\
     ${I}_{i_xi_y}$                            & Pixel einer Bildmatrix\\
$ {I}_{i_xi_y}^*$                &Pixelwerte nach Bildnormalisierung\\
$\mathbf{I}_{\text{kante}}$      &Bin\"{a}res Kantenbild\\
$I_\text{max}$                     & Maximaler Intensit\"{a}tswert eines Bildes\\
  $i_p$                          & Laufindex Pr\"{a}parationsschritte\\
$  I_{\operatorname{pmr}}$& Indikator bei der Klassifikation des Bewegungsindex\\
$I_{\text{quant}_c}$&  Mittels eines Quantils $c$ ermittelter Schwellenwert \"{u}ber Pixelwerte\\
$I_{\text{quant}_{1-c}}$& Mittels eines Quantils $1-c$ ermittelter  Schwellenwert \"{u}ber Pixelwerte \\
$I_\text{schwell}$      &Schwellenwert in einem Bild\\
 $i_{s}$                            &Laufindex f\"{u}r Spitzen im Bewegungsindex\\
 $i_{sy}$                            &Laufindex Planfaktoren\\
 $i_{sz}$                            &Laufindex St\"{o}rfaktoren\\
 $I_\text{tief}$                            & Unterer Schwellenwert zur Bildnormalisierung\\
  $i_w$                          & Laufindex f\"{u}r Wiederholungen einer Abtastung\\
    $i_{w\Psi}$                          & Laufindex f\"{u}r Wiederholungen einer Abtastung im segmentierten Bildstrom\\
 $i_x$                          & Laufindex f\"{u}r Spalten in einem Bild \\
 $I_x$                          & Anzahl Spalten in einem Bild \\
   $i_{x\Psi}$                          & Laufindex f\"{u}r Spalten im segmentierten Bildstrom \\
   $i_y$                          & Laufindex f\"{u}r Zeilen in einem Bild \\
  $I_y$                          & Anzahl Zeilen in einem Bild \\
   $i_{y\Psi}$                          & Laufindex f\"{u}r Zeilen im segmentierten Bildstrom \\
 $i_z$                          & Laufindex f\"{u}r Schichten in einem dreidimensionalen Bild \\
   $I_z$                          & Anzahl Schichten in einem dreidimensionalen Bild \\
      $i_{z\Psi}$                          & Laufindex f\"{u}r Schichten im dreidimensionalen und segmentierten Bildstrom \\

$J$                             &Randpunkte des Bewegungsindex\\
$j$                             &Laufvariable\\
$j_m$                             &Anzahl an Mikroskopen\\
$j_p$                             &Anzahl an PCs \bzw Projekten\\
$K$                             &Anzahl Abtastzeitpunkte\\
$k$                            & Laufindex Abtastzeitpunkte\\



$M$                             &Intervall um eine Spitze im Bewegungsindex\\
$m_1$                             &Anzahl der Zeitreihen\\
$m_2$                             &Anzahl der Merkmale\\
$m_3$                             &Anzahl ausgew\"{a}hlter Merkmale\\
$m_{BS}$                        & Modalit\"{a}t des Bildstroms\\
$m_d$                             &Anzahl aggregierter Merkmale\\
$M_{\text{max}}$                             &L\"{a}ngste Halbachse einer umschlie{\ss}enden Ellipse\\
$M_{\text{min}}$                             &K\"{u}rzeste Halbachse einer umschlie{\ss}enden Ellipse\\
$MSR$                             &Minimum-Signifikanz-Verh\"{a}ltnis\\
$m_{y,i_{sy}}$                           & Anzahl an Klassen des $i_{sy}$-ten Planfaktors\\
$m_{z,i_{sz}}$                           & Anzahl an Klassen des $i_{sz}$-ten St\"{o}rfaktors\\
$m_\Psi$                        & Modalit\"{a}t des Nutzsignals\\

$n$                             &Laufindex Versuchseinheiten\\
$N$                             &Anzahl Einzelversuche in einer Hochdurchsatz-Untersuchung\\
$\mathbb{N}$                            & Raum der nat\"{u}rlichen Zahlen \\
$n_{C^+}$                       & Anzahl an Positiv-Kontrollen\\
$n_{C^-}$                       & Anzahl an Negativ-Kontrollen\\
$n_F$                           &Anzahl an Bildsequenzen eines Bildstroms\\
$n_l$                       & Anzahl an Wiederholungsdurchl\"{a}ufen innerhalb einer Platte\\
$n_p$                           &Anzahl an Pr\"{a}parationsschritten zur Realisierung eines Versuchs\\
$n_s$                           &Anzahl an pr\"{a}parierten Versuchseinheiten\\
$n_w$                           &Anzahl an N\"{a}pfchen (Wells)\\
$n_z$                           &Anzahl an Fokusebenen\\

$O$                            & Optimierungskriterium zur Normalisierung\\

$\mathbf{p}$                      & Vektor aller Versuchsparameter \\
$\mathbb{P}$                      &Sensor- oder Pixelwerte\\
$\mathbf{p^{*}}$                      & Durch Filterung adaptierte Versuchsparameter $\mathbf{p}$ \\
$\mathbf{p^{**}}$                      & Durch einen Wrapper adaptierte Versuchsparameter $\mathbf{p}$ \\
$\mathbf{p}_{{BS}}$        & Versuchsparameter des Bildstroms \\
$\mathbf{p}_{fn}$                  & Versuchsparameter der Bildstrom-Vorverarbeitung\\
$\mathbf{p}_{S\Psi}$
                                  &Versuchsparameter der Klassifikation  mit $\mathbf{p}_{S\Psi}=[\mathbf{p}_{S{\Psi_1}},\mathbf{p}_{S{\Psi_2}},\mathcal{I},\mathbf{p}_{S{\Psi_4}},\mathbf{p}_{S{\Psi_5}}]^T$\\
$\mathbf{p}_{{y}}$         & Versuchsparameter der Planfaktoren  \\
$\mathbf{p}_{{z}}$         & Versuchsparameter der St\"{o}rfaktoren  \\
$\mathbf{p}_{{\Psi}}$      & Versuchsparameter des \Nutzsigs \\

$q$                     &Parameter zur Zentrierung bei Normalisierung\\
$\mathbb{R},\mathbb{R}^s,\dots$            & Raum der reellen Zahlen\\
$\mathbf{RV}$                   & Reale Werte des Einzelversuchs\\
$\mathbb{RV}$            & Raum welcher den realen Einzelversuch beschreibt\\
$r_F$                   & Anzahl an Frames einer Bildsequenz\\
$r_{F\Psi}$                   & Anzahl an Frames einer Bildsequenz im segmentierten Bildstrom\\
$r_w$                   & Anzahl an Wiederholungen einer Abtastung\\
$r_{w\Psi}$                   & Anzahl an Wiederholungen einer Abtastung im segmentierten Bildstrom\\

$SF$                            &Signal-Fenster\\
 $SHV$                           & Signal-Hintergrund-Verh\"{a}ltnis\\
 $SNR$                          & Signal-Rausch-Verh\"{a}ltnis\\
$S_{RV_1}$                           &Abbildung bei der Bildakquise\\
$S_{RV_2}$                             &Abbildung zur Bildstrom-Vorverarbeitung\\
$S_{RV_3}$                             &Abbildung bei der Segmentierung\\
 $s_y$                            &Anzahl Planfaktoren\\
$S_y$                           &Gesamt-Abbildung zur Sch\"{a}tzung der Klassenzugeh\"{o}rigkeit des realen Versuchs bez\"{u}glich der Planfaktoren\\
 $s_z$                            &Anzahl St\"{o}rfaktoren\\
$S_z$                           &Gesamt-Abbildung zur Sch\"{a}tzung der Klassenzugeh\"{o}rigkeit des realen Versuchs bez\"{u}glich der St\"{o}rfaktoren\\
$S_{\Psi_1}$                           &Abbildung zur Bestimmung von Nutzsignalzeitreihen\\
$S_{\Psi_2}$                           &Abbildung zur Bestimmung von Nutzsignalmerkmalen\\
$S_{\Psi_3}$                           &Abbildung zur Merkmalsauswahl\\
$S_{\Psi_4}$                           &Abbildung zur Merkmalsaggregation\\
$S_{\Psi_5y}$                           &Abbildung zur Entscheidungsfindung bez\"{u}glich der Planfaktoren\\
$S_{\Psi_5z}$                           &Abbildung zur Entscheidungsfindung bez\"{u}glich der St\"{o}rfaktoren\\

$t_{aq}$                             &Erforderliche Zeit f\"{u}r die Akquise einer Standard-Platte mit 96 N\"{a}pfchen\\
$t_{{BS}}$                   &Anzahl an Frames des Bildstroms\\
$t_{\mathrm{hpf}}$                  & Alter der Zebrab\"{a}rblingslarven in einer \hts\\
$t_{p}$                             &Erforderliche Zeit f\"{u}r eine Bewegung des automatischen Mikroskops\\
$t_{p}$                             &Erforderliche Zeit f\"{u}r einen Pr\"{a}parationsschritt einer Probe in einer \hts\\
$t_{{RV}}$                   &Erforderliche Zeit f\"{u}r die Akquise einer Standard-Platte mit 96 N\"{a}pfchen\\
$t_{{\Psi}}$                   &Aufnahmedauer bei der Bildakquise\\

$\mathbf{u}_{{BS}}$& Indizes, die den Auftrittsort des Nutzsignals  innerhalb des Bildstroms beschreiben mit $\mathbf{{u}}_{BS}=[x_{BS_1},x_{BS_2},y_{BS_1},y_{BS_2},z_{BS_1},z_{BS_2}]^T$\\
$\mathbf{{u}}_{\Psi}$ & Indizes, die den Auftrittsort des Nutzsignals innerhalb des realen Versuchs beschreiben mit $\mathbf{{u}}_{\Psi}=[x_{\Psi_1},x_{\Psi_2},y_{\Psi_1},y_{\Psi_2},z_{\Psi_1},z_{\Psi_2}]^T$\\

$v$                             &Parameter der Verteilung (\zB Standardverteilung) zur Normalisierung\\
$VK$                            &Variationskoeffizient\\

$\mathbf{W}$                  & Matrix der aggregierten Merkmale\\
$x$                            & Merkmal (allgemein) \\
$\mathbf{X}$                   & Matrix der Merkmale ($N$ Zeilen, $s$ Spalten) \\
$x^*$                            & Hilfsvariable bei der Merkmalsberechnung\\
$x_\text{off}$                            & Konstante bei der Merkmalsberechnung\\
$x_{I}$                        & Abszissenachse eines Bildes \\

${y}$                    &Planfaktor\\
${y}$                    &Gesch\"{a}tzte Klasse eines Planfaktors\\
$\mathbf{\hat y}$                        &Gesch\"{a}tzte Klassenzugeh\"{o}rigkeit der Planfaktoren \\
$\mathbf{Y}$                    &Matrix der Planfaktoren\\
$\mathbf{y}_c$                        &Erwarteter Effekt nach Regression\\
$y_{I}$                        & Ordinatenachse eines Bildes \\
$\mathbf{Y}_{ZR}$                    &Zeitreihe von Merkmalswerten\\
${z}$                    &St\"{o}rfaktor\\
$\hat{\mathbf{z}}$              &Gesch\"{a}tzte Klassenzugeh\"{o}rigkeit der St\"{o}rfaktoren\\
$\mathbf{Z}$                    &Matrix der St\"{o}rfaktoren\\
$Z'$                            & Parameter zur Quantifizierung der Signifikanz einer \hts  \\
$Z''$                            & Parameter zur Quantifizierung der Signifikanz eines Einzelversuchs einer \hts  \\

$\mathbf{z}_c$                    &Konzentrationen einer Untersuchung\\
$\mathbf{z}_{RV}$                        &Ersatzgr\"{o}{\ss}e weiterer Faktoren einer \hts\\
$\alpha$                        &Parameter zur Wichtung zwischen Plan"~ und St\"{o}rfaktoren bei der Optimierung\\
$\delta$                        &Bildzeile\\
$\mu_{\mathrm{max}}$ & Maximaler Erwartungswert einer Versuchsreihe\\
$\mu_{\mathrm{min}}$ & Minimaler Erwartungswert einer Versuchsreihe\\
$\xi_y$                           &Erfolgsrate bei Sch\"{a}tzung der Planfaktoren\\
$\overline{\xi_y}$                &Klassifikationsg\"{u}te bei Sch\"{a}tzung der Planfaktoren\\
$\overline{\xi_y,\text{ziel}}$                &Optimale Klassifikationsg\"{u}te f\"{u}r $\xi_y$\\
$\xi_z$                           &Erfolgsrate bei Sch\"{a}tzung der St\"{o}rfaktoren\\
$\overline{\xi_z}$                &Klassifikationsg\"{u}te bei Sch\"{a}tzung der St\"{o}rfaktoren\\
$\overline{\xi_z,\text{ziel}}$                &Optimale Klassifikationsg\"{u}te f\"{u}r $\xi_z$\\

$\sigma$                        & Standardabweichung der Messwerte eines Versuchs\\
$\sigma_d$                      & Standardabweichung der Differenz von Versuchen \\
$\sigma_{\mathrm{max}/\mathrm{min}}$ &Versuch mit gr\"{o}{\ss}ter/kleinster Standardabweichung der Messwerte\\
$\phi$                         & Farbtiefe eines Bildes\\
$\boldsymbol{\Psi}$                & Segmentierter Bildstrom\\

\\
\bottomrule

\end{longtable}


\begin{thebibliography}{100}

\bibitem{Ackerman04}
\textsc{Ackerman, F.}; \textsc{Massey, R.}; \textsc{Ministerr{\aa}d, N.};
  \textsc{R{\aa}d, N.}: \emph{The true costs of REACH}. Nordic Council of
  Ministers, 2004.

\bibitem{adams94}
\textsc{Adams, R.}; \textsc{Bischof, L.}: Seeded region growing. \emph{Pattern
  Analysis and Machine Intelligence, IEEE Transactions on} 16 (1994) 6,
  S.~641--647.

\bibitem{Ahrens74}
\textsc{Ahrens, H.}; \textsc{L{\"a}uter, J.}: \emph{{Mehrdimensionale
  Varianzanalyse: Hypothesenpr\"ufung, Dimensionserniedrigung, Diskrimination
  bei multivariaten Beobachtungen}}. Berlin: Akademie-Verlag, 1974.

\bibitem{Alshut08}
\textsc{Alshut, R.}: \emph{{Entwicklung eines Bildverarbeitungsalgorithmus zur
  automatisierten Evaluierung eines Hochdurchsatzexperimentes mit toxikologisch
  behandelten Modellorganismen}}. Diplomarbeit, Karlsruher Institut f{\"u}r
  Technologie (KIT), 2008.

\bibitem{Alshut10}
\textsc{Alshut, R.}; \textsc{Legradi, J.}; \textsc{Liebel, U.}; \textsc{Yang,
  L.}; \textsc{van Wezel, J.}; \textsc{Str{\"a}hle, U.}; \textsc{Mikut, R.};
  \textsc{Reischl, M.}: Methods for automated high-throughput toxicity testing
  using zebrafish embryos. \emph{Lecture Notes in Artificial Intelligence} 6359
  (2010), S.~219--226.

\bibitem{Alshut09}
\textsc{Alshut, R.}; \textsc{Legradi, J.}; \textsc{Mikut, R.};
  \textsc{Str{\"a}hle, U.}; \textsc{Reischl, M.}: Robust identification of
  coagulated zebrafish eggs using image processing and classification
  techniques. In: \emph{Proc., 19. Workshop Computational Intelligence}, S.
  9--21, 2009.

\bibitem{Alshut11AT}
\textsc{Alshut, R.}; \textsc{Mikut, R.}; \textsc{Legradi, J.}; \textsc{Liebel,
  U.}; \textsc{Str{\"a}hle, U.}; \textsc{Bretthauer, G.}; \textsc{Reischl, M.}:
  {Automatische Klassifikation von Bildzeitreihen f{\"u}r toxikologische
  Hochdurchsatz-Untersuchungen}. \emph{at-Automatisierungstechnik} 59(5)
  (2011), S.~259--268.

\bibitem{Aquino11}
\textsc{Aquino, D.}; \textsc{Sch{\"o}nle, A.}; \textsc{Geisler, C.};
  \textsc{v~Middendorff, C.}; \textsc{Wurm, C.}; \textsc{Okamura, Y.};
  \textsc{Lang, T.}; \textsc{Hell, S.}; \textsc{Egner, A.}: Two-color nanoscopy
  of three-dimensional volumes by {4Pi} detection of stochastically switched
  fluorophores. \emph{Nature Methods} 8 (2011) 4, S.~353--359.

\bibitem{AssayGuidanceManualCommitteeMaryland05}
\textsc{{Assay Guidance Manual Committee Maryland}}: Assay guidance manual
  version. In: \emph{Eli Lilly and Company and NIH Chemical Genomics Center,
  Bethesda}, 2005.

\bibitem{Baker11}
\textsc{Baker, M.}: Screening: the age of fishes. \emph{Nature Methods} 8
  (2011) 1, S.~47--51.

\bibitem{Bandemer77}
\textsc{Bandemer, H.}: \emph{Theorie und Anwendung der optimalen
  Versuchsplanung}, Bd.~1. Akademie-Verlag, 1977.

\bibitem{Best08}
\textsc{Best, J.~D.}; \textsc{Alderton, W.~K.}: Zebrafish: An in vivo model for
  the study of neurological diseases. \emph{Neuropsychiatric Disease and
  Treatment} 4 (2008) 3, S.~567.

\bibitem{Best08a}
\textsc{Best, J.~D.}; \textsc{Berghmans, S.}; \textsc{Hunt, J.};
  \textsc{Clarke, S.~C.}; \textsc{Fleming, A.}; \textsc{Goldsmith, P.};
  \textsc{Roach, A.~G.}: Non-associative learning in larval zebrafish.
  \emph{Neuropsychopharmacology} 33 (2008), S.~1206--1215.

\bibitem{bhat09}
\textsc{Bhat, S.}; \textsc{Liebling, M.}: Cardiac tissue and erythrocyte
  separation in bright-field microscopy images of the embryonic zebrafish heart
  for motion estimation. In: \emph{Biomedical Imaging: From Nano to Macro,
  2009. ISBI'09. IEEE International Symposium on}, S. 746--749, IEEE, 2009.

\bibitem{bhat12}
\textsc{Bhat, S.}; \textsc{Ohn, J.}; \textsc{Liebling, M.}: Motion-based
  structure separation for label-free high-speed 3-D cardiac microscopy.
  \emph{Image Processing, IEEE Transactions on} 21 (2012) 8, S.~3638--3647.

\bibitem{Biometrica08}
\textsc{Biometrica, U.}: Complex object parametric analyzer and sorter (COPAS).
  In: \emph{http://www.unionbio.com}, 2008.

\bibitem{Blackburn11}
\textsc{Blackburn, J.~S.}; \textsc{Liu, S.}; \textsc{Raimondi, A.~R.};
  \textsc{Ignatius, M.~S.}; \textsc{Salthouse, C.~D.}; \textsc{Langenau,
  D.~M.}: High-throughput imaging of adult fluorescent zebrafish with an {LED}
  fluorescence macroscope. \emph{Nature Protocols} 6 (2011) 2, S.~229--241.

\bibitem{Bleicher03}
\textsc{Bleicher, K.~H.}; \textsc{Bohm, H.-J.}; \textsc{Muller, K.};
  \textsc{Alanine, A.~I.}: Hit and lead generation: beyond high-throughput
  screening. \emph{Nature Reviews Drug Discovery} 2 (2003) 5, S.~369--378.

\bibitem{Blow2009}
\textsc{Blow, N.}: High-throughput screening: designer screens. \emph{Nature
  Methods} 6 (2009) 1, S.~105--108.

\bibitem{Bowman10}
\textsc{Bowman, T.~V.}; \textsc{Zon, L.~I.}: Swimming into the future of drug
  discovery: In vivo chemical screens in zebrafish. \emph{{ACS} Chemical
  Biology} 5 (2010) 2, S.~159--161.

\bibitem{Braunbeck05}
\textsc{Braunbeck, T.}; \textsc{B{\"o}ttcher, M.}; \textsc{Hollert, H.};
  \textsc{Kosmehl, T.}; \textsc{Lammer, E.}; \textsc{Leist, E.};
  \textsc{Rudolf, M.}; \textsc{Seitz, N.}: Towards an alternative for the acute
  fish {LC50} test in chemical assessment: the fish embryo toxicity test goes
  multi-species -- an update. \emph{{ALTEX~- Alternativen zu Tierexperimenten}}
  22(2) (2005), S.~87--102.

\bibitem{Braunbeck98}
\textsc{Braunbeck, T.}; \textsc{Hinton, D.~E.}; \textsc{Streit, B.}: \emph{Fish
  ecotoxicology}. Birk{h\"a}user, 1998.

\bibitem{Broach96}
\textsc{Broach, J.~R.}; \textsc{Thorner, J.}: High-throughput screening for
  drug discovery. \emph{Nature} 384 (1996) 6604 Suppl, S.~14--16.

\bibitem{VandenBulck11}
\textsc{Van~den Bulck, K.}; \textsc{Hill, A.}; \textsc{Mesens, N.};
  \textsc{Diekman, H.}; \textsc{De~Schaepdrijver, L.}; \textsc{Lammens, L.}:
  Zebrafish developmental toxicity assay: A fishy solution to reproductive
  toxicity screening, or just a red herring? \emph{Reproductive Toxicology} 32
  (2011) 2, S.~213--219.

\bibitem{burger06}
\textsc{Burger, W.}; \textsc{Burge, M.~J.}: \emph{Digitale Bildverarbeitung}.
  Springer, 2006.

\bibitem{Burmeister08}
\textsc{Burmeister, O.}; \textsc{Reischl, M.}; \textsc{Bretthauer, G.};
  \textsc{Mikut, R.}: {Data-Mining-Analysen mit der MATLAB-Toolbox Gait-CAD}.
  \emph{at-Automatisierungstechnik} 56(7) (2008), S.~381--389.

\bibitem{burt1983}
\textsc{Burt, P.}; \textsc{Adelson, E.}: The laplacian pyramid as a compact
  image code. \emph{Communications, IEEE Transactions on} 31 (1983) 4,
  S.~532--540.

\bibitem{cachat11}
\textsc{Cachat, J.}; \textsc{Stewart, A.}; \textsc{Utterback, E.};
  \textsc{Hart, P.}; \textsc{Gaikwad, S.}; \textsc{Wong, K.}; \textsc{Kyzar,
  E.}; \textsc{Wu, N.}; \textsc{Kalueff, A.~V.}: Three-dimensional
  neurophenotyping of adult zebrafish behavior. \emph{PLoS One} 6 (2011) 3.

\bibitem{Canny83}
\textsc{Canny, J.}: Finding edges and lines in images. {Techn.\ Ber.},
  Massachusetts Institute of Technology, 1983.

\bibitem{Cao09}
\textsc{Cao, Y.}; \textsc{Semanchik, N.}; \textsc{Lee, S.~H.}; \textsc{Somlo,
  S.}; \textsc{Barbano, P.~E.}; \textsc{Coifman, R.}; \textsc{Sun, Z.}:
  Chemical modifier screen identifies {HDAC} inhibitors as suppressors of {PKD}
  models. \emph{Proceedings of the National Academy of Sciences} 106 (2009) 51,
  S.~21819 --21824.

\bibitem{Carolyn75}
\textsc{Carolyn, K.}; \textsc{Ballard, D.}; \textsc{Sklansky, J.}: Finding
  circles by an array of accumulators. \emph{Commun. {ACM}} 18 (1975) 2,
  S.~120--122.

\bibitem{Carpenter07}
\textsc{Carpenter, A.~E.}: Image-based chemical screening. \emph{Nature
  Chemical Biology} 3(8) (2007), S.~461.

\bibitem{Carradice08}
\textsc{Carradice, D.}; \textsc{Lieschke, G.}: Zebrafish in hematology: sushi
  or science? \emph{Blood} 111 (2008) 7, S.~3331--3342.

\bibitem{Carreira-Perpinan97}
\textsc{Carreira-Perpin{\'a}n, M.~A.}: A review of dimension reduction
  techniques. \emph{Department of Computer Science. University of Sheffield.
  Tech. Rep. CS-96-09} 9 (1997), S.~1--69.

\bibitem{Carvalho11}
\textsc{Carvalho, R.}; \textsc{de~Sonneville, J.}; \textsc{Stockhammer, O.};
  \textsc{Savage, N.}; \textsc{Veneman, W.}; \textsc{Ottenhoff, T.};
  \textsc{Dirks, R.}; \textsc{Meijer, A.}; \textsc{Spaink, H.}: A
  high-throughput screen for tuberculosis progression. \emph{PLoS One} 6 (2011)
  2.

\bibitem{Chan09}
\textsc{Chan, P.}; \textsc{Lin, C.}; \textsc{Cheng, S.}: Noninvasive technique
  for measurement of heartbeat regularity in zebrafish {(danio} rerio) embryos.
  \emph{{BMC} Biotechnology} 9 (2009) 1, S.~11.

\bibitem{cleland1963}
\textsc{Cleland, W.}: The kinetics of enzyme-catalyzed reactions with two or
  more substrates or products: I. Nomenclature and rate equations.
  \emph{Biochimica et Biophysica Acta (BBA)-Specialized Section on
  Enzymological Subjects} 67 (1963), S.~104--137.

\bibitem{Colowick06}
\textsc{Colowick, S.~P.}; \textsc{Kaplan, N.~O.}; \textsc{Abelson, J.~N.};
  \textsc{Simon, M.~I.}: \emph{Methods in enzymology: Measuring biological
  responses with automated microscopy}. Academic Press, 2006.

\bibitem{Cremers07}
\textsc{Cremers, D.}; \textsc{Rousson, M.}; \textsc{Deriche, R.}: A review of
  statistical approaches to level set segmentation: integrating color, texture,
  motion and shape. \emph{International Journal of Computer Vision} 72 (2007)
  2, S.~195--215.

\bibitem{Cristianni00}
\textsc{Cristianni, N.}; \textsc{Shawe-Taylor, J.}: \emph{{Support vector
  machines}}. Cambridge, United Kingdom: Cambridge University Press, 2000.

\bibitem{dAlencon10}
\textsc{{d'Alencon}, C.}; \textsc{Pena, O.}; \textsc{Wittmann, C.};
  \textsc{Gallardo, V.}; \textsc{Jones, R.}; \textsc{Loosli, F.};
  \textsc{Liebel, U.}; \textsc{Grabher, C.}; \textsc{Allende, M.}: A
  high-throughput chemically induced inflammation assay in zebrafish.
  \emph{{BMC} Biology} 8 (2010) 1, S.~151.

\bibitem{Eastwood06}
\textsc{Eastwood, B.~J.}; \textsc{Farmen, M.~W.}; \textsc{Iversen, P.~W.};
  \textsc{Craft, T.~J.}; \textsc{Smallwood, J.~K.}; \textsc{Garbison, K.~E.};
  \textsc{Delapp, N.~W.}; \textsc{Smith, G.~F.}: The minimum significant ratio:
  a statistical parameter to characterize the reproducibility of potency
  estimates from concentration-response assays and estimation by
  replicate-experiment studies. \emph{Journal of Biomolecular Screening} 11
  (2006) 3, S.~253--261.

\bibitem{Eaton74}
\textsc{Eaton, R.~C.}; \textsc{Farley, R.~D.}: Spawning cycle and egg
  production of zebrafish. \emph{Copeia}  (1974), S.~195--204.

\bibitem{Efron95}
\textsc{Efron, B.}; \textsc{Tibshirani, R.}: Cross-validation and the
  bootstrap: estimating the error rate of a prediction rule. {Techn.\ Ber.}
  TR-477, Dept. of Statistics, Stanford University, 1995.

\bibitem{Eimon09}
\textsc{Eimon, P.}; \textsc{Rubinstein, A.}: The use of in vivo zebrafish
  assays in drug toxicity screening. \emph{Expert Opinion on Drug Metabolism
  and Toxicology} 5 (2009) 4, S.~393--401.

\bibitem{Eisen96}
\textsc{Eisen, J.~S.}: Zebrafish make a big splash. \emph{Cell} 87 (1996) 6,
  S.~969--977.

\bibitem{Forster04}
\textsc{Forster, B.}; \textsc{Van De~Ville, D.}; \textsc{Berent, J.};
  \textsc{Sage, D.}; \textsc{Unser, M.}: Complex wavelets for extended
  depth-of-field: A new method for the fusion of multichannel microscopy
  images. \emph{Microscopy Research and technique} 65 (2004) 1-2, S.~33--42.

\bibitem{Gad05}
\textsc{Gad, S.}: \emph{Drug discovery handbook}, Bd.~10. Wiley Online Library,
  2005.

\bibitem{Gehrig09}
\textsc{Gehrig, J.}; \textsc{Reischl, M.}; \textsc{Kalmar, E.}; \textsc{Ferg,
  M.}; \textsc{Hadzhiev, Y.}; \textsc{Zaucker, A.}; \textsc{Song, C.};
  \textsc{Schindler, S.}; \textsc{Liebel, U.}; \textsc{M{\"u}ller, F.}:
  Automated high throughput mapping of promoter-enhancer interactions in
  zebrafish embryos. \emph{Nature Methods} 6 (2009) 12, S.~911--916.

\bibitem{Goldsmith04}
\textsc{Goldsmith, P.}: Zebrafish as a pharmacological tool: The how, why and
  when. \emph{Current opinion in pharmacology} 4 (2004) 5, S.~504--512.

\bibitem{Gonzalez08}
\textsc{Gonzalez, R.~C.}; \textsc{Woods, R.~E.}: \emph{Digital image
  processing}. Pearson Prentice Hall, 3. Aufl., 2008.

\bibitem{green12}
\textsc{Green, J.}; \textsc{Collins, C.}; \textsc{Kyzar, E.~J.}; \textsc{Pham,
  M.}; \textsc{Roth, A.}; \textsc{Gaikwad, S.}; \textsc{Cachat, J.};
  \textsc{Stewart, A.~M.}; \textsc{Landsman, S.}; \textsc{Grieco, F.};
  \textsc{et~al.}: Automated high-throughput neurophenotyping of zebrafish
  social behavior. \emph{Journal of Neuroscience Methods} 210 (2012) 2,
  S.~266--271.

\bibitem{Hall99}
\textsc{Hall, M.}: \emph{Correlation-based feature selection for machine
  learning}. Dissertation, University of Waikato, Hamilton, New Zealand, 1999.

\bibitem{Haralick92}
\textsc{Haralick, R.~M.}; \textsc{Shapiro, L.~G.}: \emph{Computer and robot
  vision}. {Addison-Wesley} Longman Publishing Co., Inc. Boston, {MA,} {USA},
  1992.

\bibitem{Hathaway87}
\textsc{Hathaway, R.}; \textsc{Bezdek, J.~C.}; \textsc{Tucker, W.}: An improved
  convergence theory for the fuzzy {ISODATA} clustering algorithms. In:
  \emph{Analysis of Fuzzy Information} (\textsc{Bezdek, J.}, Hg.), Bd.~3, S.
  123--132, Boca Rota: CRC Press, 1987.

\bibitem{Hell94}
\textsc{Hell, S.~W.}; \textsc{Wichmann, J.}: Breaking the diffraction
  resolution limit by stimulated emission: stimulated-emission-depletion
  fluorescence microscopy. \emph{Optics letters} 19 (1994) 11, S.~780--782.

\bibitem{Henn11}
\textsc{Henn, K.}; \textsc{Braunbeck, T.}: Dechorionation as a tool to improve
  the fish embryo toxicity test {(FET)} with the zebrafish {(Danio} rerio).
  \emph{Comparative Biochemistry and Physiology Part C: Toxicology \&
  Pharmacology} 153 (2011) 1, S.~91--98.

\bibitem{Hill05}
\textsc{Hill, A.~J.}; \textsc{Teraoka, H.}; \textsc{Heideman, W.};
  \textsc{Peterson, R.~E.}: Zebrafish as a model vertebrate for investigating
  chemical toxicity. \emph{Toxicological Sciences} 86 (2005) 1, S.~6--19.

\bibitem{Hogg08}
\textsc{Hogg, R.~C.}; \textsc{Bandelier, F.}; \textsc{Benoit, A.};
  \textsc{Dosch, R.}; \textsc{Bertrand, D.}: An automated system for
  intracellular and intranuclear injection. \emph{Journal of Neuroscience
  Methods} 169 (2008) 1, S.~65--75.

\bibitem{Hough62}
\textsc{Hough, P. V.~C.}: Method and means for recognizing complex patterns.
  1962.

\bibitem{Huang95}
\textsc{Huang, L.}; \textsc{Wang, M.}: Image thresholding by minimizing the
  measures of fuzziness. \emph{Pattern Recognition} 28 (1995) 1, S.~41--51.

\bibitem{Huertas86}
\textsc{Huertas, A.}; \textsc{Medioni, G.}: Detection of intensity changes with
  subpixel accuracy using Laplacian-Gaussian Masks. \emph{IEEE Transactions on
  Pattern Analysis and Machine Intelligence} 8(5) (1986), S.~651--664.

\bibitem{Huisken04}
\textsc{Huisken, J.}; \textsc{Swoger, J.}; \textsc{Del~Bene, F.};
  \textsc{Wittbrodt, J.}; \textsc{Stelzer, E.}: Optical sectioning deep inside
  live embryos by selective plane illumination microscopy. \emph{Science} 305
  (2004) 5686, S.~1007--1009.

\bibitem{Hueser06}
\textsc{H{\"u}ser, J.}: \emph{High-throughput screening in drug discovery}.
  {Wiley-VCH}, 2006.

\bibitem{Ingham97}
\textsc{Ingham, P.}: Zebrafish genetics and its implications for understanding
  vertebrate development. \emph{Human Molecular Genetics} 6 (1997),
  S.~1755--1760.

\bibitem{Inglese07}
\textsc{Inglese, J.}; \textsc{Johnson, R.~L.}; \textsc{Simeonov, A.};
  \textsc{Xia, M.}; \textsc{Zheng, W.}; \textsc{Austin, C.~P.}; \textsc{Auld,
  D.~S.}: High-throughput screening assays for the identification of chemical
  probes. \emph{Nature Chemical Biology} 3 (2007) 8, S.~466--479.

\bibitem{Irons10}
\textsc{Irons, T.}; \textsc{{MacPhail}, R.}; \textsc{Hunter, D.};
  \textsc{Padilla, S.}: Acute neuroactive drug exposures alter locomotor
  activity in larval zebrafish. \emph{Neurotoxicology and Teratology} 32 (2010)
  1, S.~84--90.

\bibitem{ISO07}
\textsc{{ISO}}: \emph{{ISO} 15088:2007 Water quality --determination of the
  acute toxicity of waste water to zebrafish eggs {(Danio} rerio)}. 2007.

\bibitem{Iversen06}
\textsc{Iversen, P.~W.}; \textsc{Eastwood, B.~J.}; \textsc{Sittampalam, G.~S.};
  \textsc{Cox, K.~L.}: A comparison of assay performance measures in screening
  assays: Signal window, z' factor, and assay variability ratio. \emph{Journal
  of Biomolecular Screening} 11 (2006) 3, S.~247 --252.

\bibitem{Jahne05}
\textsc{J{\"a}hne, B.}: \emph{Digitale Bildverarbeitung}. Springer, 6.,
  {\"u}berarb. und erw. aufl. Aufl., 2005.

\bibitem{Jain00}
\textsc{Jain, A.~K.}; \textsc{Duin, R. P.~W.}; \textsc{Mao, J.}: Statistical
  pattern recognition: a review. \emph{IEEE Transactions on Pattern Analysis
  and Machine Intelligence} 22(1) (2000), S.~4--36.

\bibitem{kautsky02}
\textsc{Kautsky, J.}; \textsc{Flusser, J.}; \textsc{Zitov{\'a}, B.};
  \textsc{{\v{S}}imberov{\'a}, S.}: A new wavelet-based measure of image focus.
  \emph{Pattern Recognition Letters} 23 (2002) 14, S.~1785--1794.

\bibitem{Keiter10}
\textsc{Keiter, S.}; \textsc{Peddinghaus, S.}; \textsc{Hollert, H.};
  \textsc{Feiler, U.}; \textsc{Reifferscheid, G.}; \textsc{v.~d. Goltz, B.};
  \textsc{Braunbeck, T.}; \textsc{Hafner, C.}; \textsc{Ottermanns, R.};
  \textsc{Hammers-Wirtz, M.}; \textsc{et~al.}: DanTox--ein BMBF-Verbundprojekt
  zur Ermittlung spezifischer Toxizit{\"a}t und molekularer Wirkungsmechanismen
  sedimentgebundener Umweltschadstoffe mit dem Zebrab{\"a}rbling (Danio rerio).
  \emph{Umweltwissenschaften und Schadstoff-Forschung} 22 (2010) 2, S.~94--98.

\bibitem{Keller08SC}
\textsc{Keller, P.}; \textsc{Schmidt, A.}; \textsc{Wittbrodt, J.};
  \textsc{Stelzer, E.}: {Reconstruction of zebrafish early embryonic
  development by scanned light sheet microscopy}. \emph{Science} 322 (2008)
  5904, S.~1065--1069.

\bibitem{Khan12smps}
\textsc{Khan, A.}; \textsc{Reischl, M.}; \textsc{Schweitzer, B.};
  \textsc{Weiss, C.}; \textsc{Mikut, R.}: Automatic tuning of image
  segmentation routines by means of fuzzy feature evaluation. \emph{Advances in
  Intelligent Systems and Computing} 190 (2013) 5, S.~459--467.

\bibitem{Kimmel95}
\textsc{Kimmel, C.}; \textsc{Ballard, W.~W.}; \textsc{Kimmel, S.~R.};
  \textsc{Ullmann, B.}; \textsc{Schilling, T.~F.}: Stages of embryonic
  development of the zebrafish. \emph{Developmental Dynamics} 203 (1995) 3,
  S.~253--310.

\bibitem{Klein07}
\textsc{Klein, B.}: \emph{Versuchsplanung-DoE: Einf{\"u}hrung in die
  Taguchi/Shainin-Methodik}. Oldenbourg Wissenschaftsverlag, 2007.

\bibitem{Kohavi95}
\textsc{Kohavi, R.}: A study of cross-validation and bootstrap for accuracy
  estimation and model selection. In: \emph{Proc., International Joint
  Conference on Artificial Intelligence}, 1995.

\bibitem{Kohavi97}
\textsc{Kohavi, R.}; \textsc{John, G.~H.}: Wrappers for feature subset
  selection. \emph{Artificial Intelligence} 97 (1-2) (1997), S.~273--324.

\bibitem{Kokel10}
\textsc{Kokel, D.}; \textsc{Bryan, J.}; \textsc{Laggner, C.}; \textsc{White,
  R.}; \textsc{Cheung, C. Y.~J.}; \textsc{Mateus, R.}; \textsc{Healey, D.};
  \textsc{Kim, S.}; \textsc{Werdich, A.~A.}; \textsc{Haggarty, S.~J.};
  \textsc{Macrae, C.~A.}; \textsc{Shoichet, B.}; \textsc{Peterson, R.~T.}:
  Rapid behavior-based identification of neuroactive small molecules in the
  zebrafish. \emph{Nature Chemical Biology} 6 (2010) 3, S.~231--237.

\bibitem{kokel13}
\textsc{Kokel, D.}; \textsc{Dunn, T.~W.}; \textsc{Ahrens, M.~B.};
  \textsc{Alshut, R.}; \textsc{Cheung, C. Y.~J.}; \textsc{Saint-Amant, L.};
  \textsc{Bruni, G.}; \textsc{Mateus, R.}; \textsc{van Ham, T.~J.};
  \textsc{Shiraki, T.}; \textsc{et~al.}: Identification of nonvisual photomotor
  response cells in the vertebrate hindbrain. \emph{The Journal of
  Neuroscience} 33 (2013) 9, S.~3834--3843.

\bibitem{kokel12}
\textsc{Kokel, D.}; \textsc{Rennekamp, A.}; \textsc{Shah, A.}; \textsc{Liebel,
  U.}; \textsc{Peterson, R.}: Behavioral barcoding in the cloud: embracing
  data-intensive digital phenotyping in neuropharmacology. \emph{Trends in
  Biotechnology}  (2012).

\bibitem{Lahl06}
\textsc{Lahl, U.}; \textsc{Hawxwell, K.~A.}: {REACH-The} new european chemicals
  law. \emph{Environmental science \& technology} 40 (2006) 23, S.~7115--7121.

\bibitem{Lammer09a}
\textsc{Lammer, E.}: \emph{Refinement of the fish embryo toxicity test (FET)
  with zebrafish (Danio rerio): Is it a real replacement of the acute fish
  toxicity test?} Dissertation, Combined faculty for mathematics and natural
  sciences, Ruprecht-Karl-University Heidelberg, 2009.

\bibitem{Lammer09}
\textsc{Lammer, E.}; \textsc{Carr, G.}; \textsc{Wendler, K.}; \textsc{Rawlings,
  J.}; \textsc{Belanger, S.}; \textsc{Braunbeck, T.}: Is the fish embryo
  toxicity test ({FET}) with the zebrafish (\emph{{Danio rerio}}) a potential
  alternative for the fish acute toxicity test? \emph{Comparative Biochemistry
  and Physiology Part C: Toxicology \& Pharmacology} 149 (2009), S.~196--209.

\bibitem{Langheinrich02}
\textsc{Langheinrich, U.}; \textsc{Hennen, E.}; \textsc{Stott, G.};
  \textsc{Vacun, G.}: Zebrafish as a model organism for the identification and
  characterization of drugs and genes affecting p53 signaling. \emph{Current
  Biology} 12 (2002) 23, S.~2023--2028.

\bibitem{lawson2011}
\textsc{Lawson, N.~D.}; \textsc{Wolfe, S.~A.}: Forward and reverse genetic
  approaches for the analysis of vertebrate development in the zebrafish.
  \emph{Developmental cell} 21 (2011) 1, S.~48--64.

\bibitem{Lehmann02}
\textsc{Lehmann, T.}; \textsc{{Meyer zu Bexten}, E.}: \emph{Handbuch der
  Medizinischen Informatik}. M\"unchen: Hanser-Verlag, 2002.

\bibitem{Lessman11}
\textsc{Lessman, C.~A.}: The developing zebrafish {(Danio} rerio): A vertebrate
  model for high-throughput screening of chemical libraries. \emph{Birth
  Defects Research Part C: Embryo Today: Reviews} 93 (2011) 3, S.~268--280.

\bibitem{Liebel03}
\textsc{Liebel, U.}; \textsc{Starkuviene, V.}; \textsc{Erfle, H.};
  \textsc{Simpson, J.}; \textsc{Poustka, A.}; \textsc{Wiemann, S.};
  \textsc{Pepperkok, R.}: A microscope-based screening platform for large-scale
  functional protein analysis in intact cells. \emph{FEBS Letters} 554 (2003)
  3, S.~394--398.

\bibitem{Lieschke07}
\textsc{Lieschke, G.~J.}; \textsc{Currie, P.~D.}: Animal models of human
  disease: Zebrafish swim into view. \emph{Nature Reviews Genetics} 8 (2007) 5,
  S.~353--367.

\bibitem{Lin00}
\textsc{Lin, S.}: Transgenic zebrafish. In: \emph{Developmental biology
  protocols: Volume II} (\textsc{Tuan, R.~S.}; \textsc{Lo, C.~W.}, Hg.), Bd.
  136 von \emph{Methods in Molecular Biology}, S. 375--383, Humana Press, 2000.

\bibitem{liu12}
\textsc{Liu, R.}; \textsc{Lin, S.}; \textsc{Rallo, R.}; \textsc{Zhao, Y.};
  \textsc{Damoiseaux, R.}; \textsc{Xia, T.}; \textsc{Lin, S.}; \textsc{Nel,
  A.}; \textsc{Cohen, Y.}: Automated phenotype recognition for zebrafish embryo
  based in vivo high throughput toxicity screening of engineered
  nano-materials. \emph{PloS One} 7 (2012) 4.

\bibitem{Liu07d}
\textsc{Liu, T.}: A quantitative zebrafish phenotyping tool for developmental
  biology and disease modeling. \emph{Signal Processing Magazine, IEEE} 24
  (2007) 1, S.~126--129.

\bibitem{Liu07c}
\textsc{Liu, T.}; \textsc{Nie, J.}; \textsc{Li, G.}; \textsc{Guo, L.};
  \textsc{Wong, S.}: ZFIQ: Zebrafish image quantitator. In: \emph{Proc.,
  IEEE/NIH Life Science Systems and Applications Workshop}, S. 59--62, 2007.

\bibitem{Liu08b}
\textsc{Liu, T.}; \textsc{Nie, J.}; \textsc{Li, G.}; \textsc{Guo, L.};
  \textsc{Wong, S. T.~C.}: ZFIQ: a software package for zebrafish biology.
  \emph{Bioinformatics} 24 (2008) 3, S.~438--439.

\bibitem{Love04}
\textsc{Love, D.~R.}; \textsc{Pichler, F.~B.}; \textsc{Dodd, A.}; \textsc{Copp,
  B.~R.}; \textsc{Greenwood, D.~R.}: Technology for high-throughput screens:
  the present and future using zebrafish. \emph{Current Opinion in
  Biotechnology} 15 (2004) 6, S.~564--571.

\bibitem{Lu07}
\textsc{Lu, Z.}; \textsc{Chen, P.~C.}; \textsc{Nam, J.}; \textsc{Ge, R.};
  \textsc{Lin, W.}: A micromanipulation system with dynamic force-feedback for
  automatic batch microinjection. \emph{Journal of Micromechanics and
  Microengineering} 17 (2007), S.~314--321.

\bibitem{MacRae12}
\textsc{MacRae, C.}: Zebrafish in toxicity testing. \emph{Annual Review of
  Pharmacology and Toxicology} 52 (2012) 1.

\bibitem{mahalanobis1936}
\textsc{Mahalanobis, P.~C.}: On the generalized distance in statistics.
  \emph{Proceedings of the National Institute of Sciences (Calcutta)} 2 (1936),
  S.~49--55.

\bibitem{Malo06}
\textsc{Malo, N.}; \textsc{Hanley, J.~A.}; \textsc{Cerquozzi, S.};
  \textsc{Pelletier, J.}; \textsc{Nadon, R.}: Statistical practice in
  high-throughput screening data analysis. \emph{Nature Biotechnology} 24
  (2006) 2, S.~167--175.

\bibitem{Mandrell12}
\textsc{Mandrell, D.}; \textsc{Truong, L.}; \textsc{Jephson, C.};
  \textsc{Sarker, M.~R.}; \textsc{Moore, A.}; \textsc{Lang, C.};
  \textsc{Simonich, M.~T.}; \textsc{Tanguay, R.~L.}: Automated zebrafish
  chorion removal and single embryo placement. \emph{Journal of Laboratory
  Automation} 17 (2012) 1, S.~66--74.

\bibitem{marcato15}
\textsc{Marcato, D.}; \textsc{Alshut, R.}; \textsc{Breitwieser, H.};
  \textsc{Mikut, R.}; \textsc{Str{\"a}hle, U.}; \textsc{Pylatiuk, C.};
  \textsc{Peravali, R.}: An automated and high-throughput Photomotor Response
  platform for chemical screens. In: \emph{Engineering in Medicine and Biology
  Society (EMBC), 2015 37th Annual International Conference of the IEEE}, S.
  7728--7731, IEEE, 2015.

\bibitem{Maruyama88}
\textsc{Maruyama, N.}; \textsc{Shibata, Y.}; \textsc{Sekine, T.}:
  \emph{Microtiter plate}. US Patent, 4,735,778, 1988.

\bibitem{Mayer86}
\textsc{Mayer, F.~L.}; \textsc{Ellersieck, M.~R.}; \textsc{Fish, U.~S.}:
  \emph{Manual of acute toxicity: Interpretation and data base for 410
  chemicals and 66 species of freshwater animals}. {US} Dept. of the Interior,
  Fish and Wildlife Service, 1986.

\bibitem{McGrath08}
\textsc{McGrath, P.}; \textsc{Li, C.}: {Zebrafish: a predictive model for
  assessing drug-induced toxicity}. \emph{Drug Discovery Today} 13 (2008) 9-10,
  S.~394--401.

\bibitem{Metscher99}
\textsc{Metscher, B.}; \textsc{Ahlberg, P.}: Zebrafish in context: uses of a
  laboratory model in comparative studies. \emph{Developmental Biology} 210
  (1999), S.~1--14.

\bibitem{Mikula08}
\textsc{Mikula, K.}; \textsc{Peyri{\'e}ras, N.};
  \textsc{Reme{\v{s}}{\i}kov{\'a}, M.}; \textsc{Sarti, A.}: 3D embryogenesis
  image segmentation by the generalized subjective surface method using the
  finite volume technique. \emph{Finite Volumes for Complex Applications V:
  Problems and Perspectives}  (2008), S.~585--592.

\bibitem{Mikula11}
\textsc{Mikula, K.}; \textsc{Peyri{\'e}ras, N.};
  \textsc{Reme{\v{s}}{\'\i}kov{\'a}, M.}; \textsc{Sm{\'\i}{\v{s}}ek, M.}: 4D
  Numerical schemes for cell image segmentation and tracking. \emph{Finite
  Volumes for Complex Applications VI: Problems and Perspectives}  (2011),
  S.~693--701.

\bibitem{Mikut08}
\textsc{Mikut, R.}: \emph{Data Mining in der Medizin und Medizintechnik}.
  Universit\"atsverlag Karlsruhe, 2008.

\bibitem{Mikut08Biosig}
\textsc{Mikut, R.}; \textsc{Burmeister, O.}; \textsc{Braun, S.};
  \textsc{Reischl, M.}: The open source Matlab toolbox Gait-CAD and its
  application to bioelectric signal processing. In: \emph{Proc., DGBMT-Workshop
  Biosignalverarbeitung, Potsdam}, S. 109--111, 2008.

\bibitem{Mikut07ATP}
\textsc{Mikut, R.}; \textsc{Burmeister, O.}; \textsc{Grube, M.};
  \textsc{Reischl, M.}; \textsc{Bretthauer, G.}: {Interaktive Auswertung von
  aufgezeichneten Zeitreihen f\"ur Fehlerdiagnosen und
  Mensch-Maschine-Interfaces}. \emph{atp - Automatisierungstechnische Praxis}
  49(8) (2007), S.~30--34.

\bibitem{Mikut13}
\textsc{Mikut, R.}; \textsc{Dickmeis, T.}; \textsc{Driever, W.};
  \textsc{Geurts, P.}; \textsc{Hamprecht, F.~A.}; \textsc{Kausler, B.~X.};
  \textsc{Ledesma-Carbayo, M.~J.}; \textsc{Mar{\'e}e, R.}; \textsc{Mikula, K.};
  \textsc{Pantazis, P.}; \textsc{et~al.}: Automated processing of zebrafish
  imaging data: A survey. \emph{Zebrafish} 10 (2013) 3, S.~401--421.

\bibitem{Mikut07GaitcadEnglish}
\textsc{Mikut, R.}; \textsc{Loose, T.}; \textsc{Burmeister, O.}; \textsc{Braun,
  S.}; \textsc{Reischl, M.}: Gait-{CAD}.
  http://sourceforge.net/projects/gait-cad/, 2007.

\bibitem{Mikut01}
\textsc{Mikut, R.}; \textsc{Peter, N.}; \textsc{Malberg, H.};
  \textsc{J{\"a}kel, J.}; \textsc{Gr{\"o}ll, L.}; \textsc{Bretthauer, G.};
  \textsc{Abel, R.}; \textsc{D{\"o}derlein, L.}; \textsc{Rupp, R.};
  \textsc{Schablowski, M.}; \textsc{Gerner, H.}: \emph{{Diagnoseunterst\"utzung
  f\"ur die instrumentelle Ganganalyse (Projekt GANDI)}}. Forschungszentrum
  Karlsruhe, 2001.

\bibitem{Montgomery05}
\textsc{Montgomery, D.~C.}: \emph{Design and analysis of experiments}. Wiley,
  6. Aufl., 2005.

\bibitem{Mueller06c}
\textsc{M{\"u}ller, W.~A.}; \textsc{Hassel, M.}: \emph{Entwicklung bedeutsamer
  Modellorganismen II: Wirbeltiere}, S. 115--170. Springer, Berlin, 2006.

\bibitem{Nagel02}
\textsc{Nagel, R.}: {DarT:} The embryo test with the Zebrafish (\emph{{Danio
  rerio}}) --a general model in ecotoxicology and toxicology. \emph{{ALTEX~-
  Alternativen zu Tierexperimenten}} 19 Suppl 1 (2002), S.~38--48.

\bibitem{niblack85}
\textsc{Niblack, W.}: \emph{An introduction to digital image processing}.
  Strandberg Publishing Company, 1985.

\bibitem{Nuesslein-Volhard80}
\textsc{N{\"u}sslein-Volhard, C.}; \textsc{Wieschaus, E.}: Mutations affecting
  segment number and polarity in Drosophila. \emph{Nature} 287 (1980),
  S.~795--801.

\bibitem{OECD06a}
\textsc{{OECD}}: \emph{Background paper on fish embryo toxicity assay}. 2006.

\bibitem{OECD06}
\textsc{{OECD}}: \emph{Fish embryo toxicity {(FET)} test. Draft {OECD}
  guideline for the testing of chemicals}. 2006.

\bibitem{Ohn11}
\textsc{Ohn, J.}; \textsc{Liebling, M.}: In vivo, high-throughput imaging for
  functional characterization of the embryonic zebrafish heart. In:
  \emph{Proc., IEEE International Symposium on Biomedical Imaging: From Nano to
  Macro}, S. 1549--1552, IEEE, 2011.

\bibitem{Otsu75}
\textsc{Otsu, N.}: A Threshold Selection Method from Gray-Level Histograms.
  \emph{Automatica} 11 (1975), S.~285--296.

\bibitem{Otsu79}
\textsc{Otsu, N.}: A threshold selection method from gray-level histograms.
  \emph{IEEE Transactions on Systems, Man and Cybernetics} 9 (1979), S.~62--66.

\bibitem{Pardo-Martin10}
\textsc{{Pardo-Martin}, C.}; \textsc{Chang, T.}; \textsc{Koo, B.~K.};
  \textsc{Gilleland, C.~L.}; \textsc{Wasserman, S.~C.}; \textsc{Yanik, M.~F.}:
  High-throughput in vivo vertebrate screening. \emph{Nature Methods} 7 (2010)
  8, S.~634--636.

\bibitem{Patton01}
\textsc{Patton, E.}; \textsc{Zon, L.}: {The art and design of genetic screens:
  Zebrafish}. \emph{Nature Reviews Genetics} 2(12) (2001), S.~956--966.

\bibitem{Pavlidis90}
\textsc{Pavlidis, T.}; \textsc{Liow, Y.}: Integrating region growing and edge
  detection. \emph{Pattern Analysis and Machine Intelligence, IEEE Transactions
  on} 12 (1990) 3, S.~225--233.

\bibitem{Pawley08}
\textsc{Pawley, J.}; \textsc{Masters, B.~R.}: Handbook of biological confocal
  microscopy. \emph{Optical Engineering} 35 (1996) 9, S.~2765--2766.

\bibitem{Peravali11}
\textsc{Peravali, R.}; \textsc{Gehrig, J.}; \textsc{Giselbrecht, S.};
  \textsc{L{\"u}tjohann, D.}; \textsc{Hadzhiev, Y.}; \textsc{M{\"u}ller, F.};
  \textsc{Liebel, U.}: Automated feature detection and imaging for
  high-resolution screening of zebrafish embryos. \emph{{BioTechniques}} 50
  (2011) 5, S.~319--324.

\bibitem{Perkel12}
\textsc{Perkel, J.}: LIFE SCIENCE TECHNOLOGIES: Animal-free toxicology:
  Sometimes, in vitro is better. \emph{Science} 335 (2012) 6072, S.~1122--1125.

\bibitem{Pfriem11}
\textsc{Pfriem, A.}; \textsc{Schulz, S.}; \textsc{Pylatiuk, C.};
  \textsc{Alshut, R.}; \textsc{Bretthauer, G.}: {Robotersysteme f{\"u}r
  Hochdurchsatzverfahren in der Bioanalysetechnik}.
  \emph{at-Automatisierungstechnik} 59 (2011) 2, S.~134--140.

\bibitem{Pieper83}
\textsc{Pieper, R.~J.}; \textsc{Korpel, A.}: Image processing for extended
  depth of field. \emph{Applied Optics} 22 (1983) 10, S.~1449--1453.

\bibitem{Prasher95}
\textsc{Prasher, D.~C.}: Using GFP to see the light. \emph{Trends in Genetics}
  11 (1995) 8, S.~320--323.

\bibitem{Pylatiuk11}
\textsc{Pylatiuk, C.}; \textsc{Pfriem, A.}; \textsc{Liebel, U.};
  \textsc{Schulz, S.}; \textsc{Bretthauer, G.}: Ingenieurtechnische
  Besonderheiten bei der automatischen Handhabung von biologischen Organismen.
  \emph{at-Automatisierungstechnik} 59 (2011) 11, S.~692--698.

\bibitem{pylatiuk14}
\textsc{Pylatiuk, C.}; \textsc{Sanchez, D.}; \textsc{Mikut, R.};
  \textsc{Alshut, R.}; \textsc{Reischl, M.}; \textsc{Hirth, S.};
  \textsc{Rottbauer, W.}; \textsc{Just, S.}: Automatic zebrafish heartbeat
  detection and analysis for zebrafish embryos. \emph{Zebrafish} 11 (2014) 4,
  S.~379--383.

\bibitem{Quinlan86}
\textsc{Quinlan, J.~R.}: Induction of decision trees. \emph{Machine Learning} 1
  (1986), S.~81--106.

\bibitem{Rasch07}
\textsc{Rasch, D.}; \textsc{Verdooren, L.~R.}; \textsc{Gowers, J.~I.}:
  \emph{Planung und Auswertung von Versuchen und Erhebungen}. Lehrbuch
  international, M{\"u}nchen: Oldenbourg, 2. Aufl., 2007.

\bibitem{Reischl10}
\textsc{Reischl, M.}; \textsc{Alshut, R.}; \textsc{Mikut, R.}: On robust
  feature extraction and classification of inhomogeneous data sets. In:
  \emph{{Proc., 20.~Workshop Computational Intelligence}}, S. 124--143, KIT
  Scientific Publishing, 2010.

\bibitem{Reischl03}
\textsc{Reischl, M.}; \textsc{Gr{\"o}ll, L.}; \textsc{Mikut, R.}: {Optimierte
  Klassifikation f\"ur Mehrklassenprobleme am Beispiel der Bewegungssteuerung
  von Handprothesen}. In: \emph{{Proc., 13.~Workshop Fuzzy Systeme}}, S.
  124--143, Forschungszentrum Karlsruhe, 2003.

\bibitem{Reischl04a}
\textsc{Reischl, M.}; \textsc{Gr{\"o}ll, L.}; \textsc{Mikut, R.}: Optimized
  classification of multiclass problems applied to {EMG}-control of hand
  prostheses. In: \emph{{Proc., IEEE International Joint Conference on Neural
  Networks}}, S. 1473--1478, 2004.

\bibitem{Stegmaier12a}
\textsc{Reischl, M.}; \textsc{Mikut, R.}: Challenges of Uncertainty Propagation
  in Image Analysis. In: \emph{Proceedings. 22. Workshop Computational
  Intelligence}, S.~55, KIT Scientific Publishing, 2014.

\bibitem{Rihel10}
\textsc{Rihel, J.}; \textsc{Prober, D.~A.}; \textsc{Arvanites, A.};
  \textsc{Lam, K.}; \textsc{Zimmerman, S.}; \textsc{Jang, S.};
  \textsc{Haggarty, S.~J.}; \textsc{Kokel, D.}; \textsc{Rubin, L.~L.};
  \textsc{Peterson, R.~T.}; \textsc{Schier, A.~F.}: Zebrafish behavioral
  profiling links drugs to biological targets and {rest/wake} regulation.
  \emph{Science} 327 (2010) 5963, S.~348--351.

\bibitem{Rinkwitz11}
\textsc{Rinkwitz, S.}; \textsc{Mourrain, P.}; \textsc{Becker, T.~S.}:
  Zebrafish: An integrative system for neurogenomics and neurosciences.
  \emph{Progress in Neurobiology} 93 (2011) 2, S.~231--243.

\bibitem{Russ02}
\textsc{Russ, J.}: \emph{The image processing handbook}. CRC Press, 2002.

\bibitem{Ruxton11}
\textsc{Ruxton, Graeme D. ;~Colegrave, N.}: \emph{Experimental design for the
  life sciences}. Oxford Univ. Press, 3. Aufl., 2011.

\bibitem{Sabaliauskas06}
\textsc{Sabaliauskas, N.}; \textsc{Foutz, C.}; \textsc{Mest, J.};
  \textsc{Budgeon, L.}; \textsc{Sidor, A.}; \textsc{Gershenson, J.};
  \textsc{Joshi, S.}; \textsc{Cheng, K.}: {High-throughput zebrafish
  histology}. \emph{Methods} 39 (2006), S.~246--254.

\bibitem{Sanchez2012}
\textsc{S\'{a}nchez, D.}: \emph{{Entwicklung einer Methode zur automatisierten
  Extraktion der Herzfrequenz aus Videosequenzen von Zebrafischlarven}}.
  Diplomarbeit, Karlsruher Institut f{\"u}r Technologie (KIT), 2012.

\bibitem{sauvola97}
\textsc{Sauvola, J.}; \textsc{Seppanen, T.}; \textsc{Haapakoski, S.};
  \textsc{Pietikainen, M.}: Adaptive document binarization. In: \emph{Document
  Analysis and Recognition, 1997., Proceedings of the Fourth International
  Conference on}, Bd.~1, S. 147--152, IEEE, 1997.

\bibitem{Scholkopf96}
\textsc{Sch{\"o}lkopf, B.}; \textsc{Smola, A.}; \textsc{M{\"u}ller, K.}:
  {Nonlinear component analysis using a kernel eigenvalue problem}. {Techn.\
  Ber.}~44, Max-Planck-Institut f\"ur biologische Kybernetik, 1996.

\bibitem{Schreiber07}
\textsc{Schreiber, S.}; \textsc{Kapoor, T.~M.}; \textsc{Wess, G.}:
  \emph{Chemical biology: From small molecules to systems biology and drug
  design}. {Wiley-VCH}, 2007.

\bibitem{Selderslaghs12}
\textsc{Selderslaghs, I.~W.}; \textsc{Blust, R.}; \textsc{Witters, H.~E.}:
  Feasibility study of the zebrafish assay as an alternative method to screen
  for developmental toxicity and embryotoxicity using a training set of 27
  compounds. \emph{Reproductive Toxicology} 33 (2012) 2, S.~142 -- 154.

\bibitem{Semmlow04}
\textsc{Semmlow, J.~L.}: \emph{Biosignal and biomedical image processing:
  {MATLAB-based} applications}. {CRC} Press, 2004.

\bibitem{Shannon98}
\textsc{Shannon, C.}: {Communication in the presence of noise}.
  \emph{Proceedings of the IEEE} 86 (1998) 2, S.~447--457.

\bibitem{Shariff10}
\textsc{Shariff, A.}; \textsc{Kangas, J.}; \textsc{Coelho, L.~P.};
  \textsc{Quinn, S.}; \textsc{Murphy, R.~F.}: Automated image analysis for
  {High-Content} screening and analysis. \emph{Journal of Biomolecular
  Screening} 15 (2010) 7, S.~726 --734.

\bibitem{Shimomura05}
\textsc{Shimomura, O.}: The discovery of aequorin and green fluorescent
  protein. \emph{Journal of microscopy} 217 (2005) 1, S.~3--15.

\bibitem{Shimomura62}
\textsc{Shimomura, O.}; \textsc{Johnson, F.~H.}; \textsc{Saiga, Y.}:
  Extraction, purification and properties of aequorin, a bioluminescent protein
  from the luminous hydromedusan, aequorea. \emph{Journal of Cellular and
  Comparative Physiology} 59 (1962) 3, S.~223--239.

\bibitem{skinner1953}
\textsc{Skinner, B.~F.}: \emph{Science and human behavior}. Simon and Schuster,
  1953.

\bibitem{Spector06}
\textsc{Spector, D.}; \textsc{Goldman, R.}: \emph{Basic methods in microscopy:
  Protocols and concepts from cells: A laboratory manual}. {CSHL} Press, 2006.

\bibitem{Spomer12}
\textsc{Spomer, W.}; \textsc{Pfriem, A.}; \textsc{Alshut, R.};
  \textsc{Christian, P.}: High-throughput screening of zebrafish embryos using
  automated heart detection and imaging. \emph{Journal of Laboratory
  Automation}  (2012).

\bibitem{Sprague06}
\textsc{Sprague, J.}; \textsc{Bayraktaroglu, L.}; \textsc{Clements, D.};
  \textsc{Conlin, T.}; \textsc{Fashena, D.}; \textsc{Frazer, K.};
  \textsc{Haendel, M.}; \textsc{Howe, D.~G.}; \textsc{Mani, P.};
  \textsc{Ramachandran, S.}; \textsc{et~al.}: The Zebrafish Information
  Network: The zebrafish model organism database. \emph{Nucleic acids research}
  34 (2006) suppl 1, S.~581--585.

\bibitem{Starkuviene07}
\textsc{Starkuviene, V.}; \textsc{Pepperkok, R.}: The potential of high-content
  high-throughput microscopy in drug discovery. \emph{British journal of
  pharmacology} 152 (2007) 1, S.~62--71.

\bibitem{Stegeman10}
\textsc{Stegeman, J.~J.}; \textsc{Goldstone, J.~V.}; \textsc{Hahn, M.~E.};
  \textsc{Dr. Steve F.~Perry, D. M.~E.}: Perspectives on zebrafish as a model
  in environmental toxicology. In: \emph{Zebrafish}, Bd. Volume 29, S.
  367--439, Academic Press, 2010.

\bibitem{Stegmaier12}
\textsc{Stegmaier, J.}; \textsc{Alshut, R.}; \textsc{Reischl, M.};
  \textsc{Mikut, R.}: Information fusion of image analysis, video object
  tracking, and data mining of biological images using the open source matlab
  toolbox gait-cad. \emph{Biomedizinische Technik (Biomedical Engineering)} 57
  (S1) (2012), S.~458--461.

\bibitem{Steinbrecher93}
\textsc{Steinbrecher, R.}: \emph{{Bildverarbeitung in der Praxis}}. Oldenbourg,
  1993.

\bibitem{Stotzka11}
\textsc{Stotzka, R.}; \textsc{Hartmann, V.}; \textsc{Jejkal, T.};
  \textsc{Sutter, M.}; \textsc{van Wezel, J.}; \textsc{Hardt, M.};
  \textsc{Garcia, A.}; \textsc{Kupsch, R.}; \textsc{Bourov, S.}: Perspective of
  the Large Scale Data Facility (LSDF) supporting nuclear fusion applications.
  In: \emph{Parallel, Distributed and Network-Based Processing (PDP), 2011 19th
  Euromicro International Conference on}, S. 373--379, IEEE, 2011.

\bibitem{Straehle11}
\textsc{Str{\"a}hle, U.}; \textsc{Scholz, S.}; \textsc{Geisler, R.};
  \textsc{Greiner, P.}; \textsc{Hollert, H.}; \textsc{Rastegar, S.};
  \textsc{Schumacher, A.}; \textsc{Selderslaghs, I.}; \textsc{Weiss, C.};
  \textsc{Witters, H.}; \textsc{et~al.}: Zebrafish embryos as an alternative to
  animal experiments -- A commentary on the definition of the onset of
  protected life stages in animal welfare regulations. \emph{Reproductive
  Toxicology} 33 (2012) 2, S.~128--132.

\bibitem{Tran07}
\textsc{Tran, T.~C.}; \textsc{Sneed, B.}; \textsc{Haider, J.}; \textsc{Blavo,
  D.}; \textsc{White, A.}; \textsc{Aiyejorun, T.}; \textsc{Baranowski, T.~C.};
  \textsc{Rubinstein, A.~L.}; \textsc{Doan, T.~N.}; \textsc{Dingledine, R.};
  \textsc{et~al.}: Automated, quantitative screening assay for antiangiogenic
  compounds using transgenic zebrafish. \emph{Cancer research} 67 (2007) 23,
  S.~11386--11392.

\bibitem{Traver03}
\textsc{Traver, D.}; \textsc{Herbomel, P.}; \textsc{Patton, E.~E.};
  \textsc{Murphey, R.~D.}; \textsc{Yoder, J.~A.}; \textsc{Litman, G.~W.};
  \textsc{Catic, A.}; \textsc{Amemiya, C.~T.}; \textsc{Zon, L.~I.};
  \textsc{Trede, N.~S.}: The zebrafish as a model organism to study development
  of the immune system. \emph{Advances in Immunology} 81 (2003), S.~253--330.

\bibitem{Valdecasas01}
\textsc{Valdecasas, A.}; \textsc{Marshall, D.}; \textsc{Becerra, J.};
  \textsc{Terrero, J.}: On the extended depth of focus algorithms for bright
  field microscopy. \emph{Micron} 32 (2001) 6, S.~559--569.

\bibitem{Vaughan10}
\textsc{Vaughan, M.}; \textsc{van Egmond, R.}: The use of the zebrafish
  {(danio} rerio) embryo for the acute toxicity testing of surfactants, as a
  possible alternative to the acute fish test. \emph{{ATLA.} Alternatives to
  laboratory animals} 38 (2010) 3, S.~231--238.

\bibitem{Veith05}
\textsc{Veith, D.}; \textsc{Veith, M.}: Biologie fluoreszierender Proteine: Ein
  Regenbogen aus dem Ozean. \emph{Biologie in unserer Zeit} 35 (2005) 6,
  S.~394--404.

\bibitem{Vogt09}
\textsc{Vogt, A.}; \textsc{Cholewinski, A.}; \textsc{Shen, X.}; \textsc{Nelson,
  S.}; \textsc{Lazo, J.}; \textsc{Tsang, M.}; \textsc{Hukriede, N.}: Automated
  image-based phenotypic analysis in zebrafish embryos. \emph{Developmental
  Dynamics} 238 (2009) 3, S.~656--663.

\bibitem{walker12}
\textsc{Walker, S.~L.}; \textsc{Ariga, J.}; \textsc{Mathias, J.~R.};
  \textsc{Coothankandaswamy, V.}; \textsc{Xie, X.}; \textsc{Distel, M.};
  \textsc{K{\"o}ster, R.~W.}; \textsc{Parsons, M.~J.}; \textsc{Bhalla, K.~N.};
  \textsc{Saxena, M.~T.}; \textsc{et~al.}: Automated reporter quantification in
  vivo: High-throughput screening method for reporter-based assays in
  zebrafish. \emph{PloS One} 7 (2012) 1.

\bibitem{Wang07}
\textsc{Wang, W.}; \textsc{Liu, X.}; \textsc{Sun, Y.}: Contact detection in
  microrobotic manipulation. \emph{The International Journal of Robotics
  Research} 26 (2007) 8, S.~821--828.

\bibitem{waters09}
\textsc{Waters, J.~C.}: Accuracy and precision in quantitative fluorescence
  microscopy. \emph{The Journal of Cell Biology} 185 (2009) 7, S.~1135--1148.

\bibitem{Westerfield93}
\textsc{Westerfield, M.}: \emph{{The zebrafish book: a guide for the laboratory
  use of zebrafish (Brachydanio rerio)}}. University of Oregon Press Eugene,
  OR, 1993.

\bibitem{Williams09}
\textsc{Williams, E.~S.}; \textsc{Panko, J.}; \textsc{Paustenbach, D.~J.}: The
  european union's {REACH} regulation: A review of its history and
  requirements. \emph{Crit Rev Toxicol} 39 (2009) 7, S.~553--575.

\bibitem{Wittmann12}
\textsc{Wittmann, C.}; \textsc{Reischl, M.}; \textsc{Shah, A.}; \textsc{Mikut,
  R.}; \textsc{Liebel, U.}; \textsc{Grabher, C.}: Facilitating drug discovery:
  An automated high-content inflammation assay in zebrafish. \emph{Journal of
  Visualized Experiments: JoVE} 65 (2012).

\bibitem{Woelcke01}
\textsc{W{\"o}lcke, J.}; \textsc{Ullmann, D.}: Miniaturized {HTS} technologies
  - {uHTS}. \emph{Drug Discovery Today} 6 (2001) 12, S.~637--646.

\bibitem{Xie11}
\textsc{Xie, Y.}; \textsc{Sun, D.}; \textsc{Tse, H.~Y.}; \textsc{Liu, C.};
  \textsc{Cheng, S.~H.}: Force sensing and manipulation strategy in
  {robot-assisted} microinjection on zebrafish embryos. \emph{{IEEE/ASME}
  Transactions on Mechatronics} 16 (2011) 6, S.~1002--1010.

\bibitem{Xu99}
\textsc{Xu, Q.}: Microinjection into zebrafish embryos. \emph{Methods in
  molecular biology (Clifton, NJ)} 127 (1999), S.~125--132.

\bibitem{Yang09}
\textsc{Yang, L.}; \textsc{Ho, N.}; \textsc{Alshut, R.}; \textsc{Legradi, J.};
  \textsc{Weiss, C.}; \textsc{Reischl, M.}; \textsc{Mikut, R.}; \textsc{Liebel,
  U.}; \textsc{M{\"u}ller, F.}; \textsc{Str{\"a}hle, U.}: Zebrafish embryos as
  models for embryotoxic and teratological effects of chemicals.
  \emph{Reproductive Toxicology} 28 (2009), S.~245--253.

\bibitem{Yee05}
\textsc{Yee, N.~S.}; \textsc{Pack, M.}: Zebrafish as a model for pancreatic
  cancer research. \emph{Pancreatic Cancer: Methods and Protocols}  (2005),
  S.~273--298.

\bibitem{zhang1995}
\textsc{Zhang, T.}; \textsc{Rawson, D.~M.}: Studies on chilling sensitivity of
  zebrafish (Brachydanio rerio) embryos. \emph{Cryobiology} 32 (1995) 3,
  S.~239--246.

\bibitem{Zhang11}
\textsc{Zhang, X.}; \textsc{Lu, Z.}; \textsc{Gelinas, D.}; \textsc{Ciruna, B.};
  \textsc{Sun, Y.}: Batch transfer of zebrafish embryos into multiwell plates.
  \emph{{IEEE} Transactions on Automation Science and Engineering} 8 (2011) 3,
  S.~625--632.

\bibitem{Ziou98}
\textsc{Ziou, D.}; \textsc{Tabbone, S.}: Edge detection techniques-an overview.
  In: \emph{International Journal of Pattern Recognition and Image Analysis},
  Citeseer, 1998.

\bibitem{Zon10}
\textsc{Zon, L.~I.}; \textsc{Peterson, R.}: The new age of chemical screening
  in zebrafish. \emph{Zebrafish} 7 (2010) 1, S.~1--1.

\bibitem{Zon05}
\textsc{Zon, L.~I.}; \textsc{Peterson, R.~T.}: In vivo drug discovery in the
  zebrafish. \emph{Nature Reviews Drug Discovery} 4(1) (2005), S.~35--44.

\end{thebibliography}

\listoffigures
\listoftables
\addcontentsline{toc}{chapter}{Tabellenverzeichnis}


\end{document}